%% file: neurips_for_arxiv.tex
\documentclass{article}
\usepackage{amsmath,amssymb,amsfonts,amsthm,mathtools}

\PassOptionsToPackage{numbers}{natbib}
\usepackage[main, final]{neurips_2026}
\usepackage[utf8]{inputenc}
\usepackage[T1]{fontenc}
\usepackage{url}
\usepackage{booktabs}
\usepackage{nicefrac}
\usepackage{microtype}
\usepackage{xcolor}
\usepackage{enumitem}
\usepackage{graphicx}

\usepackage{float}
\usepackage{algorithm}
\usepackage{algpseudocode}

\usepackage{comment}

\usepackage{enumitem}

\newtheorem{theorem}{Theorem}[section]
\newtheorem{lemma}[theorem]{Lemma}
\newtheorem{corollary}[theorem]{Corollary}
\newtheorem{proposition}[theorem]{Proposition}
\newtheorem{assumption}[theorem]{Assumption}
\theoremstyle{definition}

\newtheorem{remark}[theorem]{Remark}

\newcommand{\ignore}[1]{}
\newcommand{\R}{\mathbb{R}}
\newcommand{\E}{\mathbb{E}}

\newcommand{\distpm}{\operatorname{dist}_{\pm}}

\newcommand{\ot}{\otimes}

\newcommand{\norm}[1]{\left\lVert #1\right\rVert}

\usepackage{amsmath,amssymb}

\newcommand{\dist}{\operatorname{dist}}

\usepackage[utf8]{inputenc} 
\usepackage[T1]{fontenc}    
\usepackage{url}            
\usepackage{booktabs}       
\usepackage{amsfonts}       
\usepackage{nicefrac}       
\usepackage{microtype}      
\usepackage{xcolor}         
\usepackage[toc,page,header]{appendix}
\usepackage{minitoc}
\usepackage[hypertexnames=false]{hyperref}

\title{The Multiscale Single-Index Model: An MSE Analysis of Online SGD Dynamics}

\author{%
  Gordon Dai \\
  New York University\\
  \texttt{td2568@nyu.edu} \\
  \And
    Joan Bruna \\
  New York University\\
  \texttt{jb4496@nyu.edu} \\
}

\begin{document}
\doparttoc 
\faketableofcontents 

\part{} 

\maketitle

\begin{abstract}

We consider the Multiscale Single-Index Model (MSIM), first introduced in \cite{oymak2021learning}, as a stylized  model for hierarchical feature learning with scale separation. Each layer extracts a shared single-index feature at one physical scale and passes it to the next, giving a tractable setting in which to study how deep architectures learn local multiscale representations. Under non-degeneracy and delocalization assumptions on the link function and planted features respectively, for the two hidden-layer case ($K$=2), we analyze joint online spherical SGD under MSE loss, which is a canonical regression objective, in discrete time and show that it achieves weak recovery from random initialization in $O(d\log^2 d)$ samples, followed by strong recovery to accuracy $\varepsilon$ in $O(d\log(1/\varepsilon))$ additional samples. The main technical challenge arises from the non-Gaussian learned features observed by the second layer. We address this challenge through a quantitative Gaussian comparison using delocalization, combined with sharp trajectory-level control of the coupled SGD dynamics. This work is a companion to \cite{bruna2026multiscalesingleindexmodelstylized}, which studies recovery under a correlation objective for general $K.$
\end{abstract}

\section{Introduction}


Feature learning, informally defined as the ability to extract informative low-dimensional representations out of high-dimensional data, is the defining characteristic of Neural Networks trained with gradient descent ---and the basis for the modern AI paradigm of pretraining then fine-tuning. 
Over the recent years, a feature learning theory has emerged, focusing on a stylized model for high-dimensional learning based on multi-index models \cite{arous2021onlinestochasticgradientdescent, dudeja2018learning, damian2022neural, damian2024computational, dandi2023learning, barbier2019optimal, cui2024asymptotics}; see \cite{bruna2025surveyalgorithmsmultiindexmodels} for a recent survey, demonstrating the ability of shallow neural networks to perform efficient feature learning, and even identifying sharp phase transitions \cite{montanari2026phasetransitionsfeaturelearning}. 

A natural shortcoming for such theory is that shallow NNs are limited idealizations of modern NNs, where depth is a critical component. Towards addressing this gap, several works \cite{allen-zhu2019what, nichani2023provable, ren2026provablelearningrandomhierarchy, dandi2025computationaladvantagedepthlearning, wang2023learning} have studied feature learning in deeper architectures, raising novel questions such as (i) elucidating the hierarchy between features at different layers, or (ii) how to adapt gradient methods to facilitate learning across all layers.

In this work, motivated by applications in data domains such as images or text, we consider a stylized model for deep feature learning based on the principle of \emph{scale separation}, where each layer of the network can be associated with a different physical scale. In the context of computer vision, the model becomes a Convolutional Neural Network (CNN) with disjoint receptive fields, as originally introduced in \cite{oymak2021learning}; the first hidden-layer extracts local features at each patch, and subsequent layers capture local information at the next scale, progressively coarse-graining the domain \cite{bruna2013invariant, mallat2016understanding}.

We instantiate each feature learning stage using a single-index model. Specifically, we consider a setting with Gaussian inputs $X \in (\mathbb{R}^{d})^{\otimes K}$, capturing a multiscale structure with $K$ scales and `patches' of size $d$, and a target obtained by a $K$-layer network, whereby each layer $k \in [K]$ coarse-grains the feature map $Z^{(k-1)} \in (\R^{d})^{\otimes K-k+1}$ as $Z^{(k)} = \phi( Z^{(k-1)} \theta_{k}^*) \in (\R^{d})^{\otimes (K-k)}$; see Figure \ref{fig:schema}. The model $X \mapsto f_{\theta^*}(X):=Z^{(K)} \in \R$ is thus parametrized by hidden directions or features $\theta^*_1, \ldots, \theta^*_K \in \mathbb{S}^{d-1}$, and a non-linear activation function $\phi: \R \to \R$. 

By considering the Wiener chaos expansion of the target $f(X)$, this model can be viewed as a non-linear analogue of Tensor PCA \cite{montanari2014statistical}, where the planted directions $\theta_1^*, \ldots, \theta_K^*$ may be estimated from a spike in the associated chaos decomposition. 
This naturally leads to the study of two different algorithmic frameworks: (i) a spectral method, motivated by its efficiency in related problems, eg single- and multi-index models \cite{mondelli2018fundamental, damian2024computational, joshi2024complexity}, spiked tensor PCA, and tensor-decomposition methods for non-overlapping CNNs \cite{oymak2021learning}; and (ii) stochastic gradient-descent, inspired by the practical connection with deep learning. The latter setting is not addressed by these tensor-decomposition analyses and therefore is the main focus of our work.

Our analysis makes two key structural assumptions: we consider the setting where $\phi$ is \emph{non-degenerate}, in the sense that $\E_{G \sim N(0,1)}[\phi'(G)] :=\kappa \neq 0$ \footnote{This corresponds to information exponent $1$ in the language of \cite{arous2021onlinestochasticgradientdescent}.}, and further assume that the planted directions are \emph{delocalized}, unaligned with the basis on which the non-linear activations act. Exploiting these two assumptions, we analyze online SGD in the case (K=2), the immediate multiscale extension of the single-index setting, we establish a sample complexity of $n = O( d (\log(d)^2 + \log \varepsilon^{-1}))$ to achieve error $\varepsilon$, nearly matching the spectral estimator up to logarithmic factor. It is important to emphasize that the machinery of Ben Arous et al. \cite{arous2021onlinestochasticgradientdescent} does \emph{not} directly apply in this setting, due to the non-Gaussian features seen by the second hidden layer activations.

\paragraph{Related Works.}

Understanding the benefits of depth in Neural Networks is a longstanding question in the field, with initial progress centered on approximation questions \cite{telgarsky2016benefits,eldan2016power}. The efforts then moved towards incorporating 
optimization aspects, with early notable works such as \cite{pmlr-v178-safran22a,malach2021connection}, and later developed in \cite{wang2023learninghierarchicalpolynomialsthreelayer, nichani2023provable, allen-zhu2019what}. One aspect shared by all these works is the fact that the input high-dimensional distribution is rotation-invariant, ie with no particular physical structure. Early works that also exploited tensor representations of data to capture scale separation are \cite{cohen2016expressive}, which focused on expressivity. 

The closest works to ours are \cite{dandi2025computationaladvantagedepthlearning} and \cite{ren2026provablelearningrandomhierarchy}, which we discuss briefly.
Dandi et al. \cite{dandi2025computationaladvantagedepthlearning} study Single/Multi-Index Gaussian Hierarchical Targets $f^\star(x) = g^\star(a^{\star\top} P_k(W^\star x)/\sqrt{d^{\epsilon_1}})$ with $W^\star \in \mathbb{R}^{d^{\epsilon_1}\times d}$, where the hierarchy is one of \emph{dimensions} $d \to d^{\epsilon_1} \to 1$. They train a three-layer MLP \emph{layer-wise} 
and propagate Hermite decompositions across layers via the asymptotic Gaussianity of the inner statistic $h^\star$, which holds because $\mathrm{IE}(P_k) = 2$. Our setup differs in three essentials: (1) the hierarchy is one of \emph{physical scales} (within- vs.\ across-patch), with both layers acting in the same dimension $d$ on single-index teachers $\beta^\star,\theta^\star \in \mathbb{S}^{d-1}$; (2) both layers are trained \emph{jointly} by online SGD on the same timescale, so the second-layer student sees a moving, non-Gaussian feature map throughout; and (3) at $\mathrm{IE}(\phi) = 1$ the inner statistic is not asymptotically Gaussian, so we replace the CLT by a quantitative Stein/Lindeberg comparison driven by the delocalization. 
Ren et al. \cite{ren2026provablelearningrandomhierarchy} prove that $L$-layer convolutional networks trained layer-wise learn $L$-level Random Hierarchy Models with $O(m^{(1+o(1))L})$ samples, via a problem-independent ``shallow-to-deep chaining'' reduction: layer-wise training succeeds whenever the target correlates with lower-level latents, the signal to lower layers is clean, and the lower-level features are weakly identifiable. Their setting is complementary to ours: The data are discrete tokens drawn from a probabilistic context-free grammar rather than continuous Gaussian patches, and the targets are categorical class labels rather than scalar regressions, so the relevant geometry is a tree of synonym classes rather than the product sphere. Our companion work \cite{bruna2026multiscalesingleindexmodelstylized} shows that  under correlation loss between teacher model and student model $\mathbb{E}\bigl[f_\theta(X)f_{\theta^\star}(X)\bigr]$ 
when initialized in a favorable basin of constant-in-$d$ probability, online SGD achieves achieves weak recovery in $\widetilde{O}(d^{K-1})$ samples, and continues to $1-o_d(1)$ overlap after a lower-order continuation phase, for any fixed depth $K.$ Yet it remains unclear how the online SGD behaves under standard loss such as Mean Square Error (MSE) loss, which is the main focus of this work.



\paragraph{Notation:} All expectations \(\E_X\) are with respect to \(X\), with the directions held fixed, unless explicitly stated otherwise. For unit vectors \(u,v\), define the sign-invariant distance $ \distpm(u,v):=\min\{\norm{u-v}_2,\norm{u+v}_2\}.$ 

\section{The Multiscale Single-Index Model}

Let \(K\ge1\) be fixed. We model the input as a $K$-th order Gaussian tensor of dimension $d$, ie $X\in(\R^d)^{\ot K}$
have independent \(N(0,1)\) entries. Given unit vectors
\(\theta_1,\ldots,\theta_K\in S^{d-1}\), define $Z^{(0)}=X$, and recursively 
\begin{align}
  Y^{(r)}_{i_{r+1},\ldots,i_K}
  &=\sum_{j=1}^d\theta_r(j)
    Z^{(r-1)}_{j,i_{r+1},\ldots,i_K},
    \qquad 1\le r\le K,\label{eq:model-y}\\
  Z^{(r)}&=\phi(Y^{(r)})\quad\text{entrywise}.\nonumber
\end{align}
At the final layer \(Z^{(K)}\in\R\), and we write 
  $f_\theta(X):=Z^{(K)}.$
\begin{figure}
    \centering
    \includegraphics[width=0.5\linewidth]{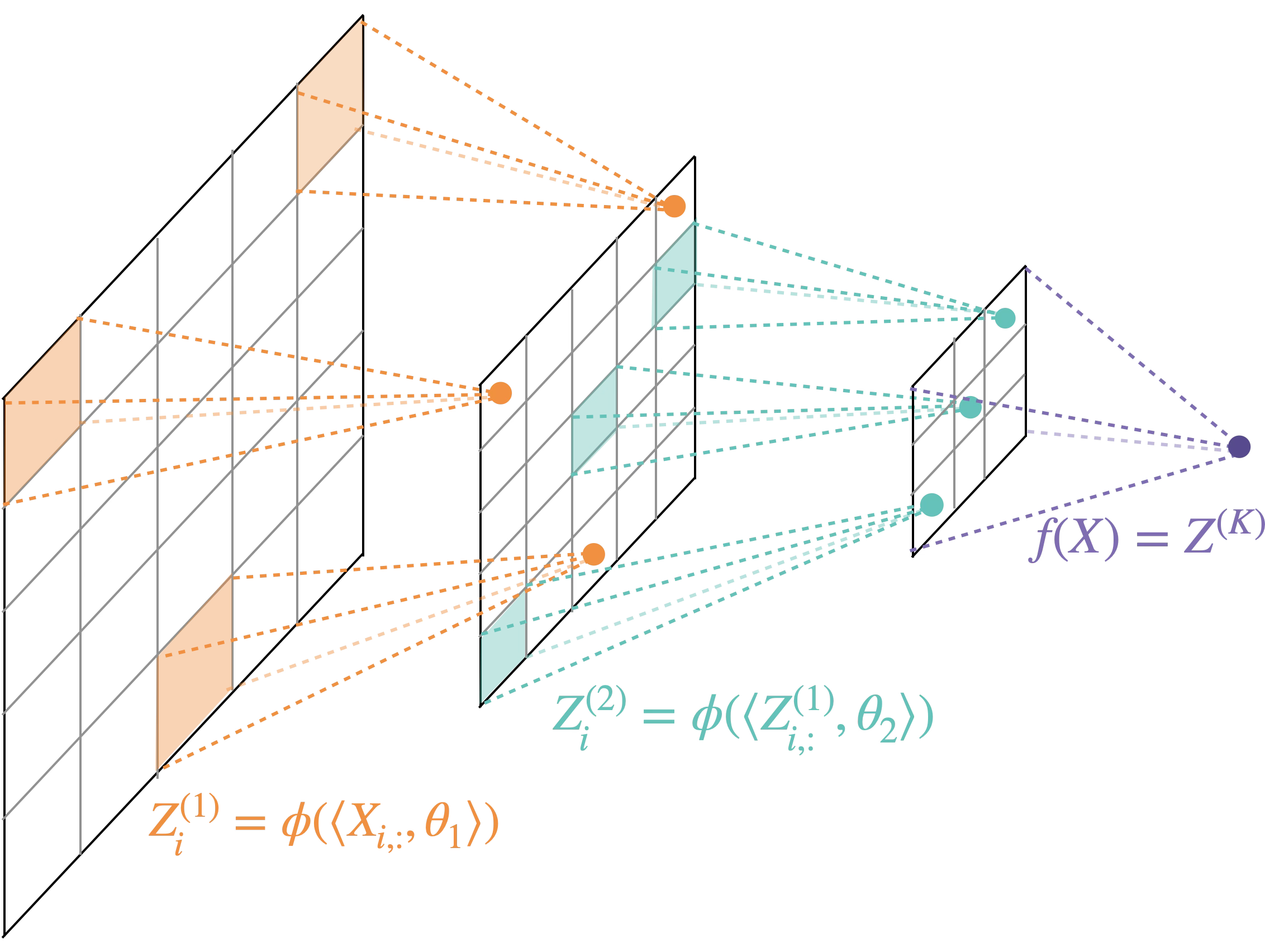}
    \caption{Schema of the Multiscale SIM model. Each patch within the same layer $k$ shares the same local feature, parametrized by a SIM of direction $\theta_k$. }
    \label{fig:schema}
\end{figure}
As illustrated in Figure \ref{fig:schema}, this is a stylized model of a NN with physical structure: each layer computes \emph{local} and \emph{translation-invariant} features at a fixed spatial \emph{scale}. In the case where $K=2$, for instance, 
 the first layer extracts a single shared `local' feature $\theta_1^*$ capturing internal correlations within each patch, while the second layer extracts a coarse feature $\theta^*_2$ capturing correlations across patches. 

Our algorithmic approaches will be based on two main assumptions: 
\begin{assumption}[Non-degeneracy and regularity, informal]
\label{ass:nondegen_informal}
Let \(G\sim N(0,1)\). We assume
\begin{equation}
  \E\phi(G)=0,
  \qquad
  \E\phi(G)^2=1,
  \qquad
  \kappa:=\E\phi'(G)\ne0.
  \tag{A1}\label{ass:normalization-informal}
\end{equation}
Further, $\phi$ satisfies certain standard regularity assumptions, which differ slightly on each setting (discussed in each section).  
\end{assumption}

In the language of Ben Arous et al. \cite{arous2021onlinestochasticgradientdescent}, non-degeneracy corresponds to the case where the information exponent of $\phi$ is $1$. In the shallow setting $K=1$, this assumption greatly simplifies the geometry of the model, ensuring a descent direction from random initializations. In our setting $K>1$, the non-degeneracy is instrumental for gradient methods to escape the mediocrity, as will be shown in later sections.

\begin{assumption}[Incoherence, informal]
\label{ass:incoherence_informal}
    The planted directions are \emph{delocalized}, in the sense that $\| \theta_k\|_\infty \ll 1$. 
\end{assumption}
Incoherence corresponds to the notion that no patch exerts an dominant influence on the output, and enables a mean-field approximation, or equivalently a \emph{Gaussian universality} in our SGD analysis, similarly as in \cite{gheissari2025universality}.

\section{Main Result}\label{sec:online-sgd}

We now turn to verify that, indeed, $K=2$ is a tractable setting for SGD dynamics to achieve strong recovery under the natural MSE loss rather than linear correlation in our companion paper \cite{bruna2026multiscalesingleindexmodelstylized}. For ease of presentation, we name $\theta_1^*= \theta^*$ and $\theta_2^* = \beta^*$ in the remainder of this section.
 \subsection{Setup}\label{sec:setup}
Given the teacher direction $(\theta^*,\beta^*)\in\mathbb{S}^{d-1}\times\mathbb{S}^{d-1}$ we want to study the ability of  \textit{online stochastic gradient descent} (online SGD) to minimize the population loss 
\begin{align}
\label{eq:bib}
    \mathcal{L}(\beta_t, \theta_t) &:= \|  f_{\beta_t, \theta_t} - f_{\beta^*, \theta^*} \|^2 = \mathbb{E}_{x \sim \gamma_D}  [f_{\beta_t, \theta_t}(x) - f_{\beta^*, \theta^*}(x)]^2,
\end{align}
where the label is generated by $z_i := \phi(X_i\cdot\beta^*), f_{\beta^*, \theta^*}(X) := \psi(z\cdot\theta^*), i\in[d].$ To recover the teacher, we run online SGD on $\mathbb{S}^{d-1}\times\mathbb{S}^{d-1}:$
\begin{align}\label{eq:eachlayerupdate}
\beta_{t+1}
= \frac{\beta_t - \delta \nabla_\beta^{\mathcal S} \ell(\beta_t,\theta_t;x_t)}
       {\big\|\beta_t - \delta \nabla_\beta^{\mathcal S} \ell(\beta_t,\theta_t;x_t)\big\|},
\qquad
\theta_{t+1}
= \frac{\theta_t - \delta \nabla_\theta^{\mathcal S} \ell(\beta_t,\theta_t;x_t)}
       {\big\|\theta_t - \delta \nabla_\theta^{\mathcal S} \ell(\beta_t,\theta_t;x_t)\big\|},
\end{align}
for $t=0,1,2,\dots$, where $(x_t \in \R^{d \times d})$ are fresh samples independent of the past, $\delta$ is the step-size, and the spherical gradient is defined as:
\[
\nabla_\beta^{\mathcal S}\ell(\beta,\theta;x):=(I-\beta\beta^\top)\nabla_\beta\ell(\beta,\theta;x),
\qquad
\nabla_\theta^{\mathcal S}\ell(\beta,\theta;x):=(I-\theta\theta^\top)\nabla_\theta\ell(\beta,\theta;x).
\]

Following \cite{arous2021onlinestochasticgradientdescent, bruna2023singleindexmodelsgaussian}, throughout the learning process, we track the evolution of summary statistics, defined as overlaps $\rho_t:=\beta_t\cdot \beta^*$ and $\chi_t:=\theta_t\cdot \theta^*$ for the first and second layer's parameter respectively.



\begin{theorem}[Informal]\label{thm:informal}
    With our assumptions (A1)-(A4) (see Section \ref{sec:assumptions}), for any $\varepsilon >0,$ under online SGD training (\ref{eq:eachlayerupdate}), we have:
    \begin{itemize}
        \item \emph{Weak Recovery}: starting from initialization, with probability at least $1-K\varepsilon,$ choosing step size $\delta\asymp 1/(d\log d),$ SGD achieves weak recovery:  
        $\min\{\rho_t,\chi_t\}=\Theta(1),$ using $O(d\log^2 d)$ samples.\label{thm:weak-informal}
        \item \emph{Strong Recovery}: conditioned on the weak recovery event, with probability at least $1-O\left(\log(1/\varepsilon)/d\right),$ choosing step size $\delta \asymp 1/d,$ SGD achieves strong recovery: 
        $\min\{\rho_t,\chi_t\}\geq1-\varepsilon,$ using $O\left(d\log(1/\varepsilon)\right)$  additional samples.\label{thm:strong-informal}
    \end{itemize}
\end{theorem}


We build our analysis on top of the analytical framework introduced in Ben Arous et al. \cite{arous2021onlinestochasticgradientdescent} by modeling the learning process as drifts with fluctuations. The main technical point is to leverage delocalization  (which, as remarked in \ref{eeee}, will be true w.h.p. under \ref{ass:favorable-initialization}) to control the gaussian approximation $\mathcal{L}_G$ of the true loss $\mathcal{L}$ to derive the continuous-population level on the gradient flow dynamics: Based on  Stein's Coupling (see Theorem \ref{thm:stein-product}), for each layer, in Lemma~\ref{lem:uniform-first-order-gradient-comparison}, we bound the distance between the spherical gradient of the true loss $\nabla_{\mathbb S,\beta}\mathcal{L}$ and its Gaussian proxy $\nabla_{\mathbb S,\beta}\mathcal{L}_G$ by some constant factor of the incoherence $\eta_t:=\|\theta_t\|_{\infty}+\|\theta^*\|_{\infty}$ that holds true uniformly in time:
\begin{align}\label{eq:eta-control-of-gaussian-non-gaussian}
\left|
\left\langle
\nabla_{\mathbb S,\beta}(L-L_G)(\theta_t,\beta_t),
\beta^*
\right\rangle
\right|
\le C_\beta^{(1)}\eta_t, \qquad
\left|
\left\langle
\nabla_{\mathbb S,\theta}(L-L_G)(\theta_t,\beta_t),
\theta^*
\right\rangle
\right|
\le C_\theta^{(1)}\eta_t.
\end{align}

\paragraph{Continuous-time Analysis}\label{subsec:cts-time-weak}

In continuous time,  we consider the population-level spherical gradient flow defined as \begin{align}\label{eq:popgradientflow}
    \dot\beta_t=-\nabla_{\mathbb S,\beta}L(\theta_t,\beta_t), \quad \dot\theta_t=-\nabla_{\mathbb S,\theta}L(\theta_t,\beta_t).
\end{align}
Based on \eqref{eq:eta-control-of-gaussian-non-gaussian}, in Lemma~\ref{lem:uniform-positive-overlap-drift} we derive a system of coupled ODE inequalities that captures the evolution of overlaps as
\begin{align}\label{eq:two-drift-inequalities}
\dot\rho_t
\ge
a_\rho\chi_t(1-\rho_t^2)-b_\rho\eta_t,
\qquad
\dot\chi_t
\ge
a_\chi\rho_t(1-\chi_t^2)-b_\chi\eta_t,
\end{align}
where \(a_\rho,a_\chi,b_\rho,b_\chi>0\) are independent of \(d\) and $(\theta_t, \beta_t)$. From here, we can see that for the RHS to be non-vacuous and provide enough signal for the growth of $(\rho_t,\theta_t)$, terms $b_\rho\eta_t$ and $b_\chi\eta_t$ must not exceed $a_\rho\chi_t(1-\rho_t^2)$ and $a_\chi\rho_t(1-\chi_t^2),$ respectively. This requires us to provide a quantitative control of $\eta_t$, derived in Lemma~\ref{lemma:secondlayercontrol}:
\begin{align}\label{eq:secondlayercontrol}
            \dot{\eta}_t & \leq C_1 \rho_t \| \theta^* \|_\infty + C_2 \chi_t \rho_t  \eta_t + O(\eta_t^2)~. 
\end{align}

We make the following remarks on our setting:
\begin{itemize}
    \item \emph{On the Delocalization hypothesis}: our proof crucially relies on the Gaussian comparison in order to obtain an overlap growth differential inequality. Delocalization is not necessary for efficient recovery, e.g. if $\| \theta^*\|_0 = O_d(1)$ the target is a multi-index model, for which gradient-based guarantees exist \cite{damian2022neural,bietti2023learning}. A natural follow-up question is to understand whether our SGD can obtain recovery guarantees uniform on $\theta^*$. 
    \item \emph{On non-degeneracy} The key mechanism for learning is described in \eqref{eq:two-drift-inequalities}. Whenever the information exponent of $\phi$ is $s>1$, one would instead obtain 
    \begin{align}\label{eq:two-drift-inequalities_bis}
\dot\rho_t
\ge
a_\rho\chi_t \rho_t^{s-1} (1-\rho_t^2)-b_\rho\eta_t,
\qquad
\dot\chi_t
\ge
a_\chi\rho_t \chi_t^{s-1} (1-\chi_t^2)-b_\chi\eta_t,
\end{align}
in which the perturbative $\eta_t$ term dominates the positive drift at initialization. This difficulty is not merely technical; our numerical simulations in Section \ref{sec:experiments} show that online SGD does not manage to learn in this setting. 

    \item \emph{On Gaussian inputs} For simplicity we state our result in the Gaussian setting, but using delocalization in the first layer feature, our arguments naturally extend to the class of well-behaved product measures for which Stein's CLT coupling applies.
\end{itemize}

\subsection{Main Elements of the Proof}\label{sec:main-element-sgd}

Below, we present the high-level roadmap of proofs in getting Theorem~\ref{thm:informal}. 

\subsubsection{Weak Recovery}\label{sec:weak-recovery} The proof has two dynamical regimes, Phase I and Phase II, separated by
\[
\tau_A:=\inf\{t:Z_t^2\ge A\eta_t\}, \quad Z_t:=\rho_t+\chi_t.
\]
In Phase I, $Z_t$ is still below the incoherence scale. In this phase, the crude \(O(\eta_t)\) comparison error is too large under spherical initialization; in Lemma~\ref{lem:refined-first-order-comparison-small} we bound the deviation of true population drifts from its Gaussian proxy by a refined third-moment quantity
\begin{align}\label{eq:omegat}
    \omega_t:=\sum_{i=1}^d\bigl(|\theta_{t,i}|+|\theta_i^*|\bigr)^3,
\end{align}
which itself is controlled in Lemma~\ref{lem:refined-small-overlap-drift} and Lemma~\ref{lem:population-third-moment-control}. Under the small-overlap regime of weak recovery, the dynamics is well approximated by the coupled drift inequalities
\begin{equation}\label{eq:small-overlap-approximation-pair}
\dot\rho_t
\ge
a\chi_t-b\omega_t,
\qquad
\dot\chi_t
\ge
a\rho_t-b\omega_t.
\end{equation}
In Lemma~\ref{lem:popescape}, we showed that (i) starting from initialization, \eqref{eq:small-overlap-approximation-pair} pulls \(Z_t\) from \(d^{-1/2}\) to the signal-dominant scale \(Z_t^2\sim\eta_t\) in \(O(\log d)\) time; (ii) after \(\tau_A\), the incoherence error is of a lower order compared with the positive \(Z\)-drift. The sum \(Z_t\) then grows from order \(d^{-1/4}\) to a fixed small constant, while the imbalance \(D_t=\rho_t-\chi_t\) contracts. The entry estimate \(D_{\tau_A}=o(Z_{\tau_A})\) ensures that both overlaps become positive constants simultaneously under another $O(\log d)$ time, obtaining weak recovery.

\paragraph{Extension to Discrete-Time SGD}\label{sec:discrete-time-weak-recovery}
The discrete-time weak-recovery proof mirrors the same two-regime mechanism, with additional martingale and discretization controls: We start from the one-step expansion in Lemma~\ref{lem:sgd-one-step},
\begin{align}
       \rho_{t+1}=\rho_t + \delta F_t^{(\rho)}+\delta M_t^{(\rho)} +R_t^{(\rho)}\qquad
    \chi_{t+1}=\chi_t + \delta F_t^{(\chi)}+\delta M_t^{(\chi)} +R_t^{(\chi)},\label{eq:rho-chi-one-step}
\end{align}
where we take the step size $\delta\asymp 1/d\log d.$ 
During Phase I, plugging the $\omega$-controlled population drift in Lemma~\ref{lem:refined-small-overlap-drift} into \eqref{eq:rho-chi-one-step}, we obtained the aggregate one-step update expression
\begin{align}\label{eq:Z-iterate-discrete}
Z_{t+1}
\ge
(1+\beta\delta)Z_t-C\delta\omega_t+\varepsilon_t^{(Z)},
\end{align}
where \(\varepsilon_t^{(Z)}\) as the one-step increment of martingale and discretization errors, with their time-accumulation is controlled by Abel summation from the event \(E_T^{(Z)}\). Besides, for the third-moment term $w_t$ in \eqref{eq:Z-iterate-discrete}, under Lemma~\ref{lem:sgd-third-moment-control} as the discretized version of third-moment control made in Lemma~\ref{lem:population-third-moment-control} we showed that for Phase I, \(\omega_t\lesssim (1+t\delta)/\sqrt d\), making the weighted forcing remains of order \(d^{-1/2}\). Therefore \(Z_t\) grows from its initialization scale \(d^{-1/2}\) to the signal-dominant scale \(Z_{\tau_A}\gtrsim d^{-1/4}\) in \(O(\delta^{-1}\log d)\) iterations, while \(\eta_t\lesssim (\log d)/\sqrt d\). Same as continuous case, we arrived at an entry-balance estimate $D_{\tau_A}=o(Z_{\tau_A})$ using the small-overlap \(D\)-drift derived in \ref{lem:D-drift}. 

Entering Phase II, as shown in Lemma~\ref{lem:sgd-phase2}, once \(Z_t^2\gtrsim \eta_t\), the incoherence error is negligible compared with the positive \(Z\)-drift, so \(Z_t\) grows from order \(d^{-1/4}\) to a fixed constant \(c_Z\) within another \(O(\delta^{-1}\log d)\) iterations. At the same time, the \(D\)-recursion is contractive up to lower-order forcing, which keeps \(|D_t|\le Z_t/2\). Hence, as shown in Theorem~\ref{thm:sgd-escape-mediocrity-pullup}, at the hitting time \(t_{\rm esc}:=\inf\{t:Z_t\ge c_Z\}\), both overlaps are bounded below with high probability:
\[
\min\{\rho_{t_{\rm esc}},\chi_{t_{\rm esc}}\}
=
\frac{Z_{t_{\rm esc}}-|D_{t_{\rm esc}}|}{2}
\ge
\frac{c_Z}{4}.
\]
\ignore{Theorem~\ref{thm:sgd-escape-mediocrity-pullup} formalizes this weak recovery for discrete-time SGD with high probability on the weak good event, \(\mathcal G_T^{\rm weak},\) from the concentration inequalities in Lemma~\ref{lem:sgd-martingale-Z}, Lemma~\ref{lem:sgd-discretization}, Lemma~\ref{lem:uniform-taylor-weak}, Lemma~\ref{lem:freedman-bound}, and Lemma~\ref{lem:sgd-third-moment-control}.}

\subsubsection{Strong Recovery}\label{sec:strong-recovery}

\paragraph{Continuous-time Strong Recovery Results}\label{subsec:cts-time-strong}

After weak recovery, we show in Lemma~\ref{lem:entry-local-cap-after-corrected-weak} that under additional $O(1)$ time, the population dynamics drives \((\theta_t,\beta_t)\) into a fixed local cap of arbitrary level around $y^*:=(\theta^*, \beta^*),$ the global minimizer of the true population loss. While the small-overlap estimate from weak recovery no longer holds, we employ the uniform bounds in Lemma~\ref{lem:uniform-positive-overlap-drift} and Lemma~\ref{lemma:secondlayercontrol} to show $\eta_t=o_d(1),$ which guarantees both overlaps enter the claimed local cap.

Once inside this local cap, the proof switches from overlap drift to local
geometry. We analyze the Hessian of the loss function by decomposing it as:
\begin{align}\label{eq:hessian-decomposition}
    H_L(y) = \underbrace{\bigl(H_L(y)-H_L(y^*)\bigr)}_{\rm (A)} + \underbrace{\bigl(H_L(y^*)-H_{L_G}(y^*)\bigr)}_{\rm (B)} + \underbrace{H_{L_G}(y^*)}_{\rm (C)}.    
\end{align}
From \eqref{eq:hessian-decomposition}, we derive Lemma~\ref{lem:teacher-full-hessian-comparison} to lower bound term $\rm (B)$ by some constant factor of $-\|\theta^*\|_{\infty},$ and Lemma~\ref{lem:uniform-hessian-continuity} to lower bound term $\rm (A)$ by some constant factor of $-\dist_{\mathcal{M}}(y,y^*):=-\sqrt{\arccos(\rho)^2+\arccos(\chi)^2},$ taken arbitrarily small. Finally, as $\rm (C)$ is lower bounded by a Hermite-coefficient quantity that dominates the other constants, in Lemma~\ref{lem:true-local-product-strong-convexity}, we obtain the positivity of the Hessian for the true loss $H_L(y)$ throughout the local cap. 

Hence, under this implied local strong convexity, in Lemma~\ref{lem:product-contraction}, by exploiting the the log-drift inequality
\begin{align}\label{eq:log-drift-inequality}
\left\langle
\nabla_{\mathcal M}L(y_t),
\operatorname{Log}_{y_t}(y^*)
\right\rangle
\le
-2\mu_{\rm loc}Q_t,\quad Q_t:=\frac12\arccos(\chi_t)^2+\frac12\arccos(\rho_t)^2,
\end{align}
we make a standard Riemannian gradient-flow argument to obtain the exponential contraction of $\dist_{\mathcal{M}}(y,y^*),$ in the sense that 
\begin{align}
\operatorname{dist}_{\mathcal M}(y_t,y^*)^2
\le
e^{-2\mu_{\rm loc}(t-t_\mathrm{esc})}
\operatorname{dist}_{\mathcal M}(y_{t_\mathrm{esc}},y^*)^2.
\end{align}
Converting the product distance back to overlap gives the strong recovery of 
\begin{align}\label{eq:strong-cts}
    \min\{\rho_t,\chi_t\}\ge 1-\varepsilon
\end{align}
after an additional \(O(\log(1/\varepsilon))\) time from weak recovery. We formalize this result in Theorem~\ref{thm:continuous-weak-to-strong-recovery}.

\paragraph{Extension to Discrete-Time SGD}\label{sec:discrete-time-strong-recovery} Same as the extension to discrete-time SGD in Section~\ref{sec:weak-recovery}, here we must control stochastic fluctuations of SGD inside the strongly-convex basin to ensure the local contraction. Finishing the discrete-time weak recovery gave by Theorem~\ref{thm:sgd-escape-mediocrity-pullup}, we switch the SGD step size to $\delta_{\rm loc}=\frac{c_{\rm loc}}{d},$ that is optimal to strong recovery as remarked in Remark\ref{rmk:optimal-delta-strong}. 

Mirroring  Lemma~\ref{lem:entry-local-cap-after-corrected-weak}, Proposition~\ref{prop:discrete-entry-local-basin} shows that after an additional \(O(d)\) iterations beyond weak recovery, the trajectory enters the local strongly-convex cap in discrete-time. Inside this cap, we track the \textit{local
distance}
\begin{align}
    Q_t:=\frac12\arccos(\chi_t)^2+\frac12\arccos(\rho_t)^2. 
\end{align}
From its one-step recursion in Lemma~\ref{lem:one-step-distance-recursion}:
\begin{align}\label{eq:one-step-q}
Q_{t+1} \le Q_t + \delta_{\rm loc} \langle \bar g_t,\mathcal L_t\rangle + \delta_{\rm loc}\Xi_t + C_{\rm ret}\delta_{\rm loc}^2G_t, 
\end{align}
we bound the population drift $\langle \bar g_t,\mathcal L_t\rangle$ by the log-drift inequality \eqref{eq:log-drift-inequality} in continuous time, and we bound both the martingale fluctuation term \(\Xi_t:=\langle g_t-\bar g_t,\mathcal L_t\rangle\) and the quadratic gradient term \(G_t:=\|g_t\|^2\) in Lemma~\ref{lem:sharp-local-self-quenching-sgd} by the sharp local self-quenching estimate: 
\begin{align}\label{eq:self-quenching}
    \mathbb E[\Xi_t^2\mid\mathcal F_t]\lesssim Q_t^2 \text{, and } \mathbb E[G_t\mid\mathcal F_t]\lesssim dQ_t,
\end{align}
which show that both the gradient energy and the martingale variance vanish proportionally to the current distance to the teacher. 
Thus the SGD noise becomes weaker as the iterates approach the teacher: Proposition~\ref{prop:one-local-epoch} proves that one local SGD epoch
of length \(O(\delta_{\rm loc}^{-1})\) contracts the distance scale by a
factor of \(1/2\) with probability \(1-O(\delta_{\rm loc})\), while the
trajectory remains inside the local cap. Iterating dyadic epochs in
Proposition~\ref{prop:iterated-local-contraction} yields $Q_t\le \varepsilon$ after $O\!\left(d\log\frac1\varepsilon\right)$ additional samples. We conclude our discrete-time result in Theorem~\ref{thm:discrete-weak-to-strong-recovery}, where we obtained  \eqref{eq:strong-cts}, namely the strong recovery under discrete-time SGD, with a total sample complexity of $O(d\log^2 d)+O\!\left(d\log\frac1\varepsilon\right).$


\section{Conclusions and Future Directions}
In this work, we studied the Multiscale Single-Index Model, in which a $d^K$-dimensional input with multiscale structure is iteratively coarse-grained via local, translation-invariant single-index models, and where the number of scales $K$ corresponds directly to the depth of the associated neural network. Provided the non-linearities are non-degenerate and the planted directions are delocalized, we proved  weak and strong recovery of online SGD in the three-layer setting $K=2$ using $O(d (\log(d)^2)$ samples. 

Our work leaves several open questions ahead of us: the first, most pressing one is to extend the SGD recoverability to general depth $K>2$. While \cite{bruna2026multiscalesingleindexmodelstylized} provides $d$-asymptotic strong recovery for general $K$ under the correlation loss, it remains unclear how a general $K$ model behaves under canonical regression objectives such as MSE loss. In our current SGD analysis, the restriction to $K=2$ is tied to the weak-recovery phase near random initialization. Indeed, the MSE objective decomposes as
\begin{align}
\mathcal{L}_{\mathrm{MSE}}(\Theta)
=
\mathbb{E}\bigl[f_\Theta(X)^2\bigr]
-2\mathbb{E}\bigl[f_\Theta(X)f_{\Theta^\star}(X)\bigr]
+\mathbb{E}\bigl[f_{\Theta^\star}(X)^2\bigr].
\end{align}

When $m_j=\langle\theta_j,\theta_j^\star\rangle\asymp d^{-1/2}$, the teacher-dependent correlation drift in mode $i$ has size
$\prod_{\ell\neq i}m_\ell\asymp d^{-(K-1)/2}$.
In contrast, our current first-order bounds allow teacher-independent finite-$d$ self-energy corrections at scale $d^{-1/2}$.
For $K=2$, these terms occur at the same scale, and our refined third-moment and incoherence estimates allow us to prove that the resulting two-overlap dynamics still has a positive cooperative drift to escape mediocrity.
For $K>2$, however, the planted drift is of strictly smaller order, and a first-order Gaussian or norm comparison no longer determines the sign of the overlap evolution.
Moreover, unlike the perturbations in the correlation objective, the self-energy corrections are not naturally accompanied by planted-overlap factors. Figure \ref{fig:k2_vs_k3}, Section \ref{sec:experiments} shows the empirical evidence of hardness of MSE loss for $K>2.$

The next natural question is to consider the regime of fixed $d$ (the so-called `receptive field') and diverging depth $K$; this will also require substantial technical innovation, since our CLT/universality arguments do not directly apply. 

Finally, an important limitation of our work is that it imposes a form of scale separation: each layer lives on its dedicated scale. While this is a valid assumption for certain physical data (e.g images), most popular deep architectures do not have scale separation; for instance in ResNets \cite{he2015deepresiduallearningimage}/ NeuralODEs \cite{chen2019neuralordinarydifferentialequations} the features remain at the same scale throughout several layers. Building toy models that explain feature learning in these settings remains an outstanding challenge. 

\section{Acknowledgments}

GD was partially supported by the Dean's Undergradaute Research Fund from NYU's College of Arts and Science. This work was completed during the JB sabbatical visit at the Flatiron Institute, which also provided the computing resources used in the project. The authors would also like to thank Hengzhi Li for helping them running experiments, and  Dr. Denny Wu, Yunwei Ren, Hanzhe Wu for reviewing our draft and providing helpful feedback.

\bibliographystyle{abbrv}
\bibliography{refs}

\newpage
\appendix
\addcontentsline{toc}{section}{Appendix} 
\part{Appendix} 
\parttoc 

\section{Numerical Experiments}
\label{sec:experiments}




We illustrate the performance of online SGD on several representative experiments. 
We use minibatch online SGD. The figure captions and experiment descriptions specify the batch sizes and learning rates.
We initialize the parameters $(\theta^*, \beta^*)$ drawn i.i.d. from the uniform measure, and we initialize $(\theta_0, \beta_0)$ uniformly at random from the non-negative-overlap hemispheres $\{ \theta; \theta \cdot \theta^* \geq 0\}, \{ \beta; \beta \cdot \beta^* \geq 0\}$.

\begin{figure}
    \centering
    \includegraphics[width=0.75\linewidth,trim=0 0 0 21bp,clip]{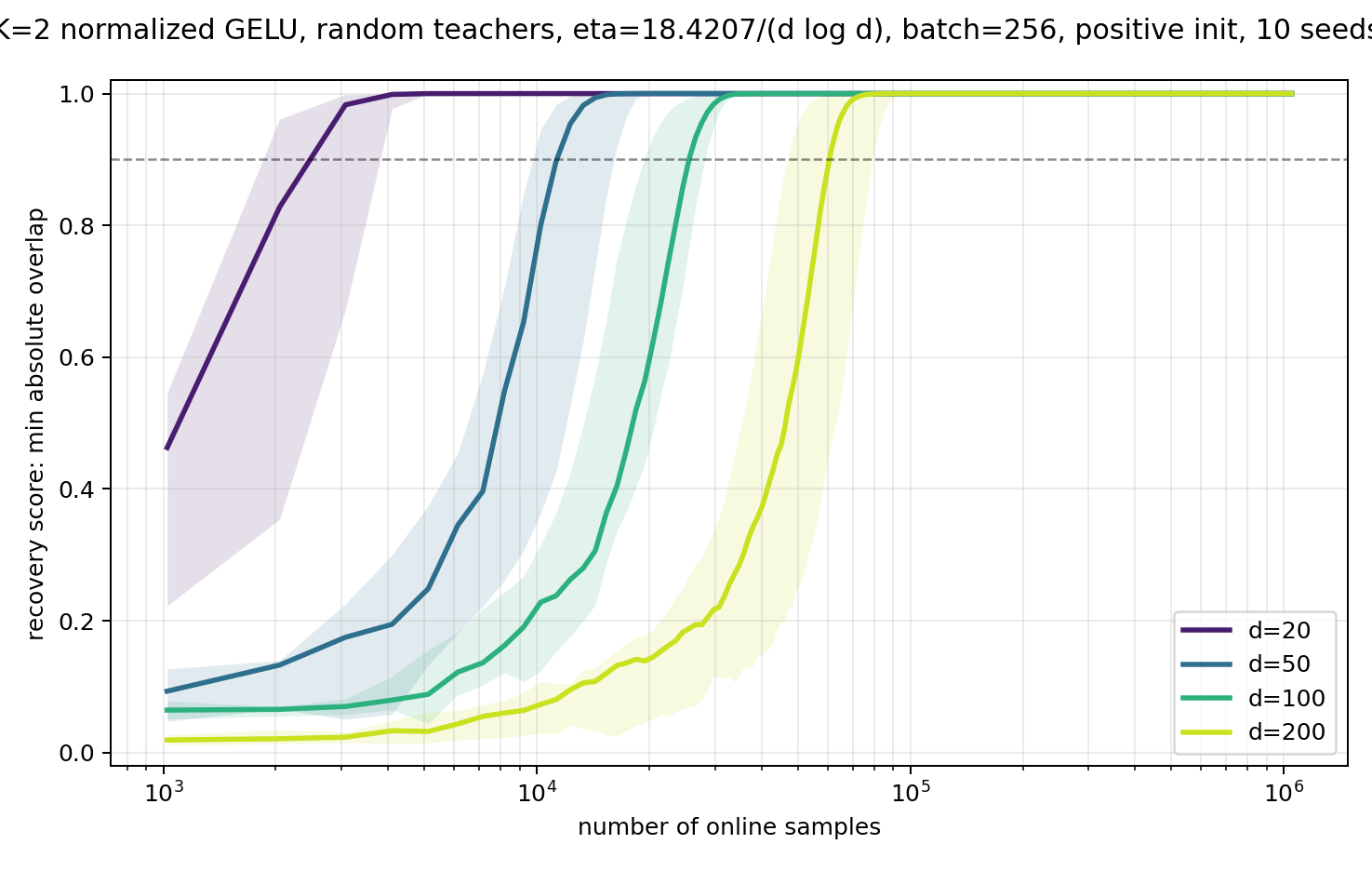}
    \caption{Minimum absolute overlap for $K=2$ with normalized GELU at $d\in\{20,50,100,200\}$. We use $\eta=18.4207/(d\log d)$, batch size $256$, and $10$ seeds.}
    \label{fig:dimensionsweep}
\end{figure}

\paragraph{Sample complexity} In Figure \ref{fig:dimensionsweep} we plot the recovery performance of online SGD for $d\in \{20,50,100,200\}$, using normalized GELU. We track $\min(|\chi_t|,|\rho_t|)$ over $10$ independent seeds and show variability across seeds.  As predicted by our theory, the algorithm successfully recovers the planted parameters, at a rate that is roughly $n \simeq d \log d$.

\paragraph{Comparing different $K,d$ pair choices} To empirically see the potential of online SGD under MSE loss extends beyond $K>2$ and compare its performance with correlation loss as studied in our companion work \cite{bruna2026multiscalesingleindexmodelstylized}, we ran a set of experiments with varying $K.$ Specifically, based on our model set-up, we consider a fixed dimension of model input, $d^K=16777216,$ and tested pair choices $(K,d)\in \{(2,4096),(3,256),(4,64),(6,16)\}$ for both correlation and MSE loss. All experiments use normalized $\phi = \mathrm{GeLU}$ activations, a learning rate of \(\eta=0.000391\), minibatches of size 16, and \(2^{27}\) training examples across 10 random seeds. As the figure shows, only  for $K=2$ do we see reliable recovery for both losses. Their performance diverges substantially as depth increases, with correlation loss exhibits meaningful recovery among all remaining trials.

\begin{figure}
    \centering\includegraphics[width=1\linewidth]{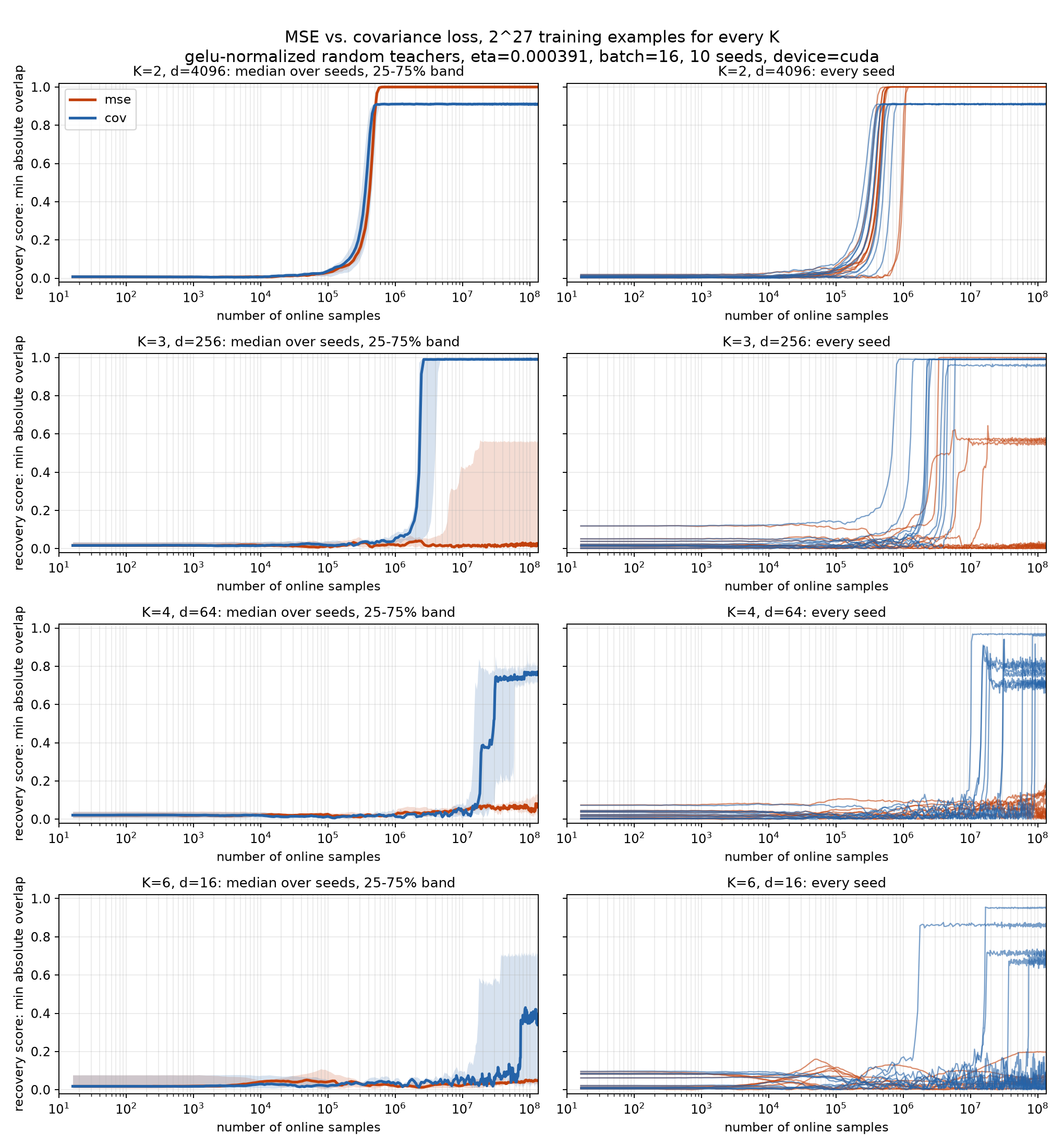}
    \caption{Comparison between MSE and Correlation Loss, for various pairs of $K,d$ with $d^K=16777216$, using $\phi= \mathrm{GeLU}$.}
    \label{fig:k2_vs_k3}
\end{figure}

\begin{figure}
    \centering
    \includegraphics[width=0.75\linewidth]{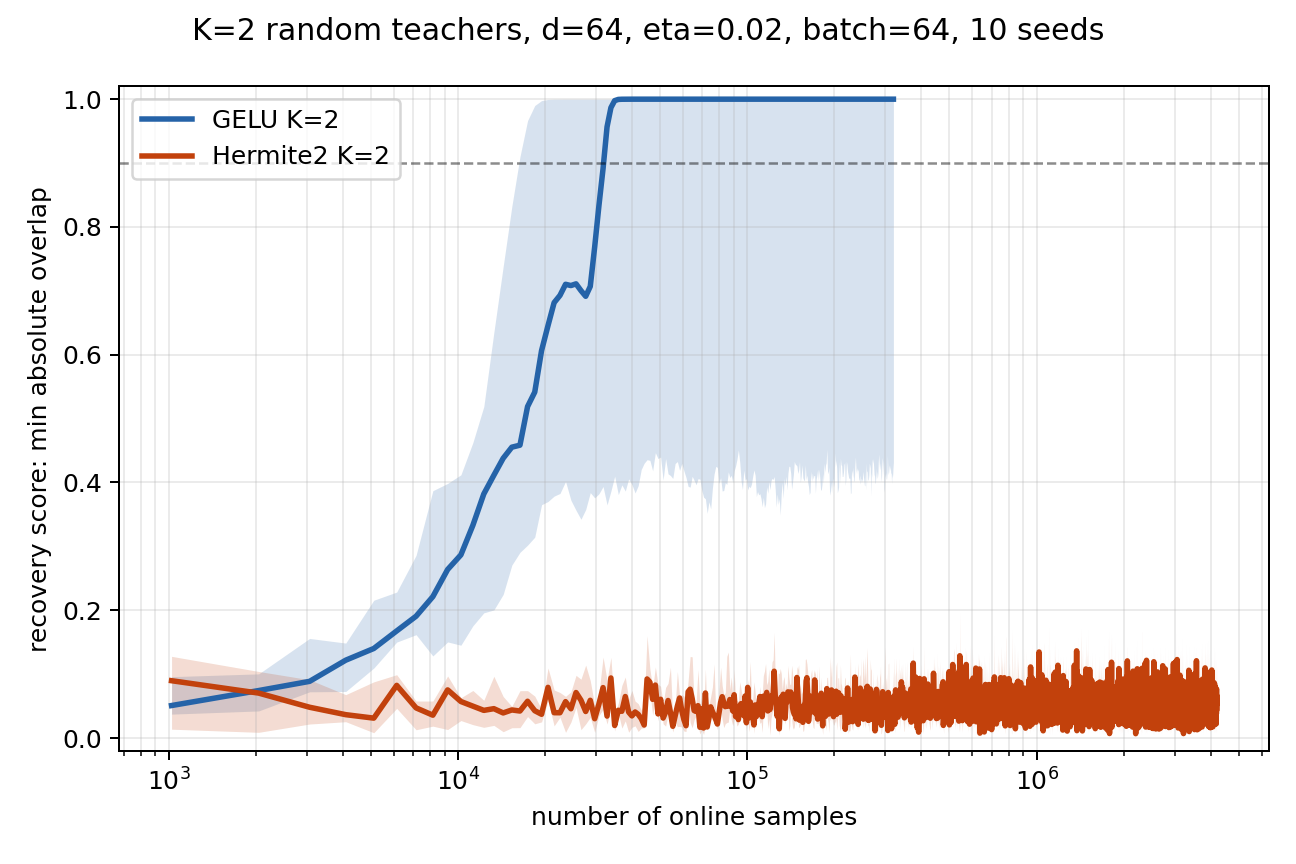}
    \caption{Minimum absolute overlap for $K=2$ with normalized GELU and with $\mathrm{He}_2$. We use $d=64$, $\eta=0.02$, batch size $64$, and $10$ seeds.}
    \label{fig:degeneracy_assumption}
\end{figure}

\paragraph{Testing non-degeneracy assumption} Finally, we investigate the effect of removing the non-degeneracy assumption, by using instead $\phi = \mathrm{He}_2$, which has information exponent $2$. Figure \ref{fig:degeneracy_assumption} shows that, in this setting, online SGD does not make significant progress, even at timescales $n \gg d^2$. 



\input{neurips/assumptions}

\section{Proof of Continuous-Time Population Dynamics}\label{sec:cts-time-population-dynamics-proof}

\subsection{Initialization Event}

\begin{lemma}[Delocalization of random spherical vectors]
\label{lem:random-spherical-delocalization}
Let \(u\sim \operatorname{Unif}(\mathbb S^{d-1})\). Then, for every fixed
\(K>0\), there exists a constant \(C_K<\infty\), independent of \(d\), such
that
\begin{align}\label{eq:good-initialization-probability}
\mathbb P\left(
\|u\|_\infty\le C_K\sqrt{\frac{\log d}{d}}
\quad\text{and}\quad
\|u\|_3^3\le C_Kd^{-1/2}
\right)
\ge 1-d^{-K}.
\end{align}
Consequently, if \(\theta_0,\theta^*\) are independent uniform random vectors
on \(\mathbb S^{d-1}\), then, after increasing \(C_K\), with probability at
least \(1-d^{-K}\),
\begin{align}\label{eq:theta-infty-bound}
\|\theta_0\|_\infty+\|\theta^*\|_\infty
\le
C_K\sqrt{\frac{\log d}{d}},
\end{align}
and
\begin{align}\label{eq:theta-l3-bound}
\|\theta_0\|_3^3+\|\theta^*\|_3^3
\le
C_Kd^{-1/2}.
\end{align}
\end{lemma}

\begin{proof}
Write
\[
u=\frac{g}{\|g\|_2},
\qquad
g=(g_1,\ldots,g_d),
\qquad
g_i\stackrel{\rm iid}{\sim}N(0,1).
\]
Standard Gaussian concentration gives, for every fixed \(K>0\), after choosing
\(C_K\) large enough,
\[
\max_{1\le i\le d}|g_i|\le C_K\sqrt{\log d},
\qquad
\|g\|_2^2\ge \frac d2
\]
with probability at least \(1-d^{-K}\). Hence
\[
\|u\|_\infty
=
\frac{\max_i|g_i|}{\|g\|_2}
\le
C_K\sqrt{\frac{\log d}{d}}.
\]

For the third moment,
\[
\|u\|_3^3
=
\sum_{i=1}^d |u_i|^3
=
\frac{\sum_{i=1}^d |g_i|^3}{\|g\|_2^3}.
\]
By a standard concentration bound for sums of iid variables with finite
moments, for every fixed \(K>0\)
there exists \(C_K<\infty\) such that
\[
\sum_{i=1}^d |g_i|^3\le C_Kd
\]
with probability at least \(1-d^{-K}\). Combining this with
\(\|g\|_2^2\ge d/2\), we get
\[
\|u\|_3^3
\le
\frac{C_Kd}{(d/2)^{3/2}}
\le
C_Kd^{-1/2}.
\]
This proves \eqref{eq:good-initialization-probability}. By a union
bound applied to \(\theta_0\) and \(\theta^*\), we get \eqref{eq:theta-infty-bound}
and \eqref{eq:theta-l3-bound}, respectively.
\end{proof}

\subsection{General Results on Dynamics}
\begin{lemma}[Uniform first-order gradient comparison]
\label{lem:uniform-first-order-gradient-comparison}
Assume \ref{ass:normalization} and \ref{ass:regularity-and-moments}, then there exist constants \(C_\beta^{(1)},C_\theta^{(1)}>0\), independent of \(d\), such that uniformly for all \((\theta,\beta)\) with \(\rho,\chi\ge0\),
\begin{align}
\left|
\left\langle
\nabla_{\mathbb S,\beta}(L-L_G)(\theta,\beta),
\beta^*
\right\rangle
\right|
&\le C_\beta^{(1)}\eta.
\label{eq:first-order-beta-gradient-comparison}\\
\left|
\left\langle
\nabla_{\mathbb S,\theta}(L-L_G)(\theta,\beta),
\theta^*
\right\rangle
\right|
&\le C_\theta^{(1)}\eta.
\label{eq:first-order-theta-gradient-comparison}
\end{align}
\end{lemma}

\begin{proof}
Throughout the proof, constants may change from line to line but depend only
on the model and moment constants, not on \(d,\theta,\theta^*,\beta,\beta^*\).

Set
\[
U_i:=x^{(i)}\cdot\beta,
\qquad
U_i^*:=x^{(i)}\cdot\beta^*,
\]
so that \((U_i,U_i^*)_{i=1}^d\) are iid centered Gaussian pairs with
correlation \(\rho\). Define
\[
X_i:=\phi(U_i),
\qquad
X_i^*:=\phi(U_i^*),
\]
and
\[
S:=\sum_{i=1}^d \theta_iX_i,
\qquad
S^*:=\sum_{i=1}^d \theta_i^*X_i^*.
\]
Then
\[
\E[X_i]=\E[X_i^*]=0,
\qquad
\E[X_i^2]=\E[(X_i^*)^2]=1,
\qquad
\E[X_iX_i^*]=\kappa(\rho).
\]
Let \((\Gamma_i,\Gamma_i^*)_{i=1}^d\) be iid Gaussian pairs with the same
covariance matrix
\[
\E[\Gamma_i^2]=\E[(\Gamma_i^*)^2]=1,
\qquad
\E[\Gamma_i\Gamma_i^*]=\kappa(\rho),
\]
and write
\[
S_G:=\sum_{i=1}^d\theta_i\Gamma_i,
\qquad
S_G^*:=\sum_{i=1}^d\theta_i^*\Gamma_i^*.
\]

\medskip
\noindent\emph{Step 1: the \(\theta\)-gradient comparison.}
Let
\[
w:=P_{\theta^\perp}\theta^*
=
\theta^*-\chi\theta .
\]
Then
\[
\langle \nabla_{\mathbb S,\theta}L(\theta,\beta),\theta^*\rangle
=
2\E\left[
(\psi(S)-\psi(S^*))\psi'(S)
\sum_{i=1}^d w_iX_i
\right].
\]
The same formula holds for \(L_G\), with \((X,X^*)\) replaced by
\((\Gamma,\Gamma^*)\). Hence it is enough to compare the expectations of
\[
h(x,x^*)
:=
2\bigl(\psi(x\cdot\theta)-\psi(x^*\cdot\theta^*)\bigr)
\psi'(x\cdot\theta)(x\cdot w).
\]

We apply Corollary~\ref{cor:stein-block}. Fix a block \(j\), condition on all
other blocks, and write
\[
s:=x_j,\qquad t:=x_j^*,
\]
\[
\ell_1=\theta_js+S_1,\qquad
\ell_2=w_js+S_2,\qquad
\ell_3=\theta_j^*t+S_3,
\]
where \(S_1,S_2,S_3\) are the leave-out sums. Then
\[
h(s,t)=2(\psi(\ell_1)-\psi(\ell_3))\psi'(\ell_1)\ell_2.
\]
Since \(\psi\in C_b^4\), every third partial derivative
\(\partial_s^a\partial_t^b h\), \(a+b=3\), is a finite sum of terms of the two
following types:
\[
C\,|\theta_j|^a|\theta_j^*|^b |\ell_2|,
\qquad a+b=3,
\]
and
\[
C\,|w_j|\,|\theta_j|^{a'}|\theta_j^*|^{b'},
\qquad a'+b'=2.
\]
At every hybrid vector in the Lindeberg interpolation,
\[
\E|\ell_2|\le C\|w\|_2\le C.
\]
Therefore
\[
\E\left[
\left|
\partial_s^a\partial_t^b h(W)
\right|
\right]
\le
C\left(
|\theta_j|^a|\theta_j^*|^b
+
|w_j|\sum_{a'+b'=2}
|\theta_j|^{a'}|\theta_j^*|^{b'}
\right).
\]
Summing over \(j\) and over \(a+b=3\), we get
\begin{align}\label{eq:3-d-lindeberg-sum}
\left|
\E h(X,X^*)-\E h(\Gamma,\Gamma^*)
\right|
\le
C\left[
\sum_j\sum_{a+b=3}|\theta_j|^a|\theta_j^*|^b
+
\sum_j |w_j|
\sum_{a'+b'=2}
|\theta_j|^{a'}|\theta_j^*|^{b'}
\right].
\end{align}

The first sum is bounded by
\[
C(\|\theta\|_\infty+\|\theta^*\|_\infty).
\]
Indeed,
\[
\sum_j|\theta_j|^3\le \|\theta\|_\infty,
\qquad
\sum_j|\theta_j|^2|\theta_j^*|\le \|\theta^*\|_\infty,
\]
and similarly for the remaining degree-three monomials. For the second sum,
use
\[
\|w\|_\infty
=
\|\theta^*-\chi\theta\|_\infty
\le
\|\theta^*\|_\infty+\|\theta\|_\infty
=
\eta,
\]
and \(\|w\|_2,\|\theta\|_2,\|\theta^*\|_2\le C\), to obtain
\[
\sum_j |w_j|
\sum_{a'+b'=2}
|\theta_j|^{a'}|\theta_j^*|^{b'}
\le
C\eta.
\]
Thus
\[
\left|
\left\langle
\nabla_{\mathbb S,\theta}(L-L_G)(\theta,\beta),
\theta^*
\right\rangle
\right|
\le
C\eta.
\]
This proves the \(\theta\)-gradient comparison.

\medskip
\noindent\emph{Step 2: exact formula for the \(\beta\)-gradient.}
Let
\[
b:=P_{\beta^\perp}\beta^*=\beta^*-\rho\beta.
\]
Then
\[
\langle\nabla_{\mathbb S,\beta}L(\theta,\beta),\beta^*\rangle
=
\langle\nabla_{\beta}L(\theta,\beta),b\rangle .
\]
Differentiating \(S\) in the direction \(b\) gives
\[
D_bS
=
\sum_{i=1}^d
\theta_i\phi'(U_i)H_i,
\qquad
H_i:=x^{(i)}\cdot b.
\]
Hence
\[
\langle\nabla_{\mathbb S,\beta}L(\theta,\beta),\beta^*\rangle
=
2\E\left[
(\psi(S)-\psi(S^*))\psi'(S)
\sum_{i=1}^d\theta_i\phi'(U_i)H_i
\right].
\]
For each \(i\), \(H_i=U_i^*-\rho U_i\), so \(H_i\) is centered Gaussian with
variance \(1-\rho^2\), independent of \(U_i\). Moreover \(U_i^*=\rho U_i+H_i\).
Applying Gaussian integration by parts in the \(H_i\)-variable gives
\[
\E\left[
(\psi(S)-\psi(S^*))\psi'(S)\theta_i\phi'(U_i)H_i
\right]
=
-(1-\rho^2)\theta_i\theta_i^*
\E\left[
\psi'(S)\psi'(S^*)\phi'(U_i)\phi'(U_i^*)
\right].
\]
Therefore
\begin{equation}\label{eq:true-beta-gradient-exact}
-\left\langle
\nabla_{\mathbb S,\beta}L(\theta,\beta),
\beta^*
\right\rangle
=
2(1-\rho^2)
\sum_{i=1}^d
\theta_i\theta_i^*
\E\left[
\psi'(S)\psi'(S^*)\phi'(U_i)\phi'(U_i^*)
\right].
\end{equation}

\medskip
\noindent\emph{Step 3: leave-one-out reduction for the true \(\beta\)-gradient.}
Set
\[
F(r,s):=\psi'(r)\psi'(s),
\qquad
q_i:=\phi'(U_i)\phi'(U_i^*),
\qquad
c_i:=\theta_i\theta_i^*.
\]
Also define
\[
S_{-i}:=S-\theta_iX_i,
\qquad
S_{-i}^*:=S^*-\theta_i^*X_i^*.
\]
Since \(S_{-i},S_{-i}^*\) are independent of \((U_i,U_i^*)\),
\[
\E[F(S_{-i},S_{-i}^*)q_i]
=
\E[F(S_{-i},S_{-i}^*)]\E[q_i].
\]
By the Hermite identity,
\[
\E[q_i]
=
\E[\phi'(U_i)\phi'(U_i^*)]
=
\kappa'(\rho).
\]
Moreover, since \(F\) is Lipschitz and the stated moment assumptions imply
\[
\sup_{\rho\in[0,1]}
\E\left[
(|X_i|+|X_i^*|)|q_i|
\right]<\infty,
\]
we have
\[
\left|
\E[F(S,S^*)q_i]
-
\E[F(S_{-i},S_{-i}^*)q_i]
\right|
\le
C(|\theta_i|+|\theta_i^*|).
\]
Multiplying by \(|c_i|\) and summing,
\[
\sum_{i=1}^d
|c_i|(|\theta_i|+|\theta_i^*|)
\le
\sum_i|\theta_i|^2|\theta_i^*|
+
\sum_i|\theta_i||\theta_i^*|^2
\le
\|\theta^*\|_\infty+\|\theta\|_\infty
=
\eta.
\]
Thus
\[
\sum_{i=1}^d c_i\E[F(S,S^*)q_i]
=
\kappa'(\rho)
\sum_{i=1}^d c_i\E[F(S_{-i},S_{-i}^*)]
+
O(\eta).
\]
Similarly,
\[
\left|
\E[F(S,S^*)]-\E[F(S_{-i},S_{-i}^*)]
\right|
\le
C(|\theta_i|+|\theta_i^*|),
\]
and the same summation estimate gives
\[
\sum_{i=1}^d c_i\E[F(S_{-i},S_{-i}^*)]
=
\left(\sum_{i=1}^d c_i\right)\E[F(S,S^*)]
+
O(\eta).
\]
Since \(\sum_i c_i=\theta\cdot\theta^*=\chi\), we conclude
\begin{equation}\label{eq:beta-true-loo}
\sum_{i=1}^d c_i\E[F(S,S^*)q_i]
=
\kappa'(\rho)\chi\,\E[F(S,S^*)]
+
O(\eta).
\end{equation}

\medskip
\noindent\emph{Step 4: Stein comparison for the two-dimensional statistic.}
Apply Corollary~\ref{cor:stein-block} to
\[
h(x,x^*)
:=
F(x\cdot\theta,x^*\cdot\theta^*).
\]
For a fixed block \(j\), the third partials have the form
\[
\partial_s^a\partial_t^b h
=
\theta_j^a(\theta_j^*)^b
\partial_1^a\partial_2^bF(\ell_1,\ell_2),
\qquad
a+b=3.
\]
Since \(F\in C_b^3(\mathbb R^2)\),
\[
\E\left[
|\partial_s^a\partial_t^b h(W)|
\right]
\le
C|\theta_j|^a|\theta_j^*|^b.
\]
Therefore
\[
\left|
\E[F(S,S^*)]
-
\E[F(S_G,S_G^*)]
\right|
\le
C\sum_j\sum_{a+b=3}|\theta_j|^a|\theta_j^*|^b
\le
C\eta.
\]
Combining this with \eqref{eq:beta-true-loo}, and using that
\(\kappa'(\rho)\) and \(\chi\) are bounded on the entry corridor, gives
\begin{equation}\label{eq:beta-sum-compared}
\sum_{i=1}^d c_i\E[F(S,S^*)q_i]
=
\kappa'(\rho)\chi\,\E[F(S_G,S_G^*)]
+
O(\eta).
\end{equation}

\medskip
\noindent\emph{Step 5: identify the Gaussian \(\beta\)-gradient.}
In the Gaussian equivalent model,
\[
L_G(\theta,\beta)
=
\Phi(\chi,\rho),
\qquad
\Phi(\chi,\rho)
=
2\sum_{k\ge1}\alpha_k^2
\left[
1-(\kappa(\rho)\chi)^k
\right].
\]
Equivalently, by Price's identity applied to the Gaussian pair
\((S_G,S_G^*)\),
\[
-\Phi_\rho(\chi,\rho)
=
2\kappa'(\rho)\chi
\E[\psi'(S_G)\psi'(S_G^*)].
\]
Since
\[
\nabla_{\mathbb S,\beta}L_G
=
\Phi_\rho(\chi,\rho)P_{\beta^\perp}\beta^*,
\]
we have
\begin{equation}\label{eq:gaussian-beta-gradient-exact}
-\left\langle
\nabla_{\mathbb S,\beta}L_G(\theta,\beta),
\beta^*
\right\rangle
=
2(1-\rho^2)\kappa'(\rho)\chi
\E[\psi'(S_G)\psi'(S_G^*)].
\end{equation}
Comparing \eqref{eq:true-beta-gradient-exact},
\eqref{eq:beta-sum-compared}, and
\eqref{eq:gaussian-beta-gradient-exact}, we obtain
\[
\left|
\left\langle
\nabla_{\mathbb S,\beta}(L-L_G)(\theta,\beta),
\beta^*
\right\rangle
\right|
\le
C(1-\rho^2)\eta
\le
C\eta.
\]
This proves the \(\beta\)-gradient comparison.

Combining the two block estimates gives the lemma.
\end{proof}

\begin{lemma}[Uniform positive-overlap drift]
\label{lem:uniform-positive-overlap-drift}
Assume Lemma~\ref{lem:uniform-first-order-gradient-comparison}. On the positive overlap region \(\rho,\chi\ge0\), the population gradient flow satisfies
\begin{align}
\dot\rho_t
\ge
a_\rho\chi_t(1-\rho_t^2)-b_\rho\eta_t,
\qquad
\dot\chi_t
\ge
a_\chi\rho_t(1-\chi_t^2)-b_\chi\eta_t,
\end{align}
where \(a_\rho,a_\chi,b_\rho,b_\chi>0\) are independent of \(d\).
\end{lemma}

\begin{proof}
With the population flow defined in \eqref{eq:popgradientflow}, we have
\[
\dot\rho_t
=
-\left\langle
\nabla_{\mathbb S,\beta}L(\theta_t,\beta_t),
\beta^*
\right\rangle,
\qquad
\dot\chi_t
=
-\left\langle
\nabla_{\mathbb S,\theta}L(\theta_t,\beta_t),
\theta^*
\right\rangle.
\]

We first compute the Gaussian drifts exactly. Recall that
\[
L_G(\theta,\beta)=\Phi(\chi,\rho),
\qquad
\Phi(\chi,\rho)
=
2\sum_{k\ge1}\alpha_k^2
\left[
1-(\kappa(\rho)\chi)^k
\right],
\]
where $\kappa(\rho)=\sum_{\ell\ge1}a_\ell^2\rho^\ell.$

For the first layer,
\[
\nabla_{\mathbb S,\beta}L_G
=
\Phi_\rho(\chi,\rho)P_{\beta^\perp}\beta^*
=
\Phi_\rho(\chi,\rho)(\beta^*-\rho\beta).
\]
Hence
\[
-\left\langle
\nabla_{\mathbb S,\beta}L_G,
\beta^*
\right\rangle
=
-\Phi_\rho(\chi,\rho)(1-\rho^2).
\]
Moreover,
\[
-\Phi_\rho(\chi,\rho)
=
2\chi\kappa'(\rho)
\sum_{k\ge1}
k\alpha_k^2(\kappa(\rho)\chi)^{k-1}.
\]
On the positive-overlap region \(\rho,\chi\ge0\), all terms are nonnegative.
Keeping only the \(k=1\) term gives
\[
-\Phi_\rho(\chi,\rho)
\ge
2\alpha_1^2\chi\kappa'(\rho).
\]
Since
\[
\kappa'(\rho)
=
\sum_{\ell\ge1}\ell a_\ell^2\rho^{\ell-1}
\ge
a_1^2
\qquad
(\rho\ge0),
\]
we obtain
\[
-\left\langle
\nabla_{\mathbb S,\beta}L_G,
\beta^*
\right\rangle
\ge
2\alpha_1^2a_1^2\chi(1-\rho^2).
\]

Similarly,
\[
\nabla_{\mathbb S,\theta}L_G
=
\Phi_\chi(\chi,\rho)P_{\theta^\perp}\theta^*
=
\Phi_\chi(\chi,\rho)(\theta^*-\chi\theta),
\]
so
\[
-\left\langle
\nabla_{\mathbb S,\theta}L_G,
\theta^*
\right\rangle
=
-\Phi_\chi(\chi,\rho)(1-\chi^2).
\]
Now
\[
-\Phi_\chi(\chi,\rho)
=
2\kappa(\rho)
\sum_{k\ge1}
k\alpha_k^2(\kappa(\rho)\chi)^{k-1}.
\]
Again all terms are nonnegative on \(\rho,\chi\ge0\), so
\[
-\Phi_\chi(\chi,\rho)
\ge
2\alpha_1^2\kappa(\rho).
\]
Since
\[
\kappa(\rho)
=
\sum_{\ell\ge1}a_\ell^2\rho^\ell
\ge
a_1^2\rho
\qquad
(\rho\ge0),
\]
we get
\[
-\left\langle
\nabla_{\mathbb S,\theta}L_G,
\theta^*
\right\rangle
\ge
2\alpha_1^2a_1^2\rho(1-\chi^2).
\]

Now transfer the Gaussian drift to the true population loss. By
\eqref{eq:first-order-beta-gradient-comparison} from Lemma~\ref{lem:uniform-first-order-gradient-comparison},
\[
\begin{aligned}
\dot\rho_t
&=
-\left\langle
\nabla_{\mathbb S,\beta}L_G(\theta_t,\beta_t),
\beta^*
\right\rangle
-
\left\langle
\nabla_{\mathbb S,\beta}(L-L_G)(\theta_t,\beta_t),
\beta^*
\right\rangle \\
&\ge
2\alpha_1^2a_1^2\chi_t(1-\rho_t^2)
-
C_\beta^{(1)}\eta_t.
\end{aligned}
\]
Likewise, by \eqref{eq:first-order-theta-gradient-comparison} from Lemma~\ref{lem:uniform-first-order-gradient-comparison},
\[
\begin{aligned}
\dot\chi_t
&=
-\left\langle
\nabla_{\mathbb S,\theta}L_G(\theta_t,\beta_t),
\theta^*
\right\rangle
-
\left\langle
\nabla_{\mathbb S,\theta}(L-L_G)(\theta_t,\beta_t),
\theta^*
\right\rangle \\
&\ge
2\alpha_1^2a_1^2\rho_t(1-\chi_t^2)
-
C_\theta^{(1)}\eta_t.
\end{aligned}
\]
Thus the claimed inequalities hold with
\[
a_\rho=a_\chi:=2\alpha_1^2a_1^2,
\qquad
b_\rho:=C_\beta^{(1)},
\qquad
b_\chi:=C_\theta^{(1)}.
\]
\end{proof}

\begin{lemma}[$\ell_\infty$ control]  \label{lemma:secondlayercontrol}
    Under our assumptions, 
    \begin{align}
        \dot{\eta}_t & \leq C_1 \rho_t \| \theta^* \|_\infty + C_2 \chi_t \rho_t  \eta_t + O(\eta_t^2)~. 
    \end{align}
\end{lemma}
\begin{proof}
    By definition, we have 
    $$\mathcal{L} = \mathbb{E}_{(z, z^*) \sim \phi_\# \gamma_{\beta \cdot \beta^*} } [|\psi( z \cdot \theta) - \psi( z^* \cdot \theta^*)|^2]$$
    so 
    $$\nabla_\theta \mathcal{L} = \mathbb{E}_{(z, z^*) \sim \phi_\# \gamma_{\beta \cdot \beta^*} } [(\psi( z \cdot \theta) - \psi( z^* \cdot \theta^*)) \psi'(z \cdot \theta) z]~. $$
    Call $\nu$ the law of $\phi(w)$, where $w \sim N(0,1)$. The random vector $z$ is then drawn from $\nu^{\otimes d}$. 
    Let us now analyze the two terms in the RHS. 
    
    The first term reads $\E[ F( z \cdot \theta) z]$ for $F$ of the form $F = \psi \psi'$. Assume that $F'$ is Lipschitz. A second-order Taylor expansion of $F$ yields
    \begin{align}
        F( z \cdot \theta) &= F( z_1 \theta_1 + z_{-1} \cdot \theta_{-1}) \\
        &= F( z_{-1} \cdot \theta_{-1} ) + F'( z_{-1} \cdot \theta_{-1} ) \theta_1 z_1 + \frac12 \theta_1^2 z_1^2 F''(M)~,
    \end{align}
    and therefore, for all $i \in [d]$, 
    \begin{align}
         \E[F( z \cdot \theta ) z_i ] &= \theta_i \E[F'( z_{-i} \cdot \theta_{-i})] + O(\theta_i^2) ~.
    \end{align}
    Now, for each $\theta$ and $i\in [d]$, consider $\mu_i = \text{Law}( \sum_{j \neq i} \theta_j z_j) \in \mathcal{P}(\R)$, where $z_j \sim \nu$ iid. We claim that
    \begin{align}
        \sup_{i, i'} W_1( \mu_i, \mu_{i'}) &= O( \| \theta\|_\infty)~.
    \end{align}
    Indeed, fix $i, i'$ and consider the coupling 
    $(Y_i, Y_{i'}) \sim \Gamma_{\mu_i, \mu_{i'}}$ given by 
    $Y_i = W + \theta_i Z$, $Y_{i'} = W + \theta_{i'} Z$, where $Z \sim \nu$ and  $W = \sum_{j \neq i, i'} \theta_j z_j$, with $z_j \sim \nu$ independently of $Z$. We then have 
    $$W_1(\mu_i, \mu_{i'}) \leq \E| Y_i - Y_{i'}| = |\theta_i - \theta_{i'}| \E |Z| \leq 2\sqrt{\E[|\phi(G)|^2]}  \| \theta\|_\infty \leq 2 \| \theta\|_\infty~.$$
    Now, since $F'$ is Lipschitz, we have 
\begin{align}
    \sup_{i, j} \left| \E[F'( z_{-i} \cdot \theta_{-i})] - \E[F'( z_{-j} \cdot \theta_{-j})] \right| & \leq \textrm{Lip}(F') \sup_{i, i'} W_1( \mu_i, \mu_{i'}) = O( \| \theta\|_\infty)~,
\end{align}
    and therefore, denoting by $\upsilon(\theta) = \frac1d \sum_i \E[F'( z_{-i} \cdot \theta_{-i})]$, we have
\begin{align}
    \E[F( z \cdot \theta ) z ] &= \theta \upsilon(\theta) + \varepsilon~,~\text{with}~ |\varepsilon_i| \leq C |\theta_i| \|\theta\|_\infty~.
\end{align}
  It follows that   
    \begin{align}
         P_{\theta^\perp} \E[F( z \cdot \theta ) z ]  &= P_{\theta^\perp} \varepsilon = \varepsilon - (\varepsilon \cdot \theta) \theta~ 
    \end{align}
    satisfies 
    \begin{align}
        \| P_{\theta^\perp} \E[F( z \cdot \theta ) z ] \|_\infty & \leq \tilde{C} \|\theta\|_\infty^2~. 
    \end{align}
    Now we repeat the argument for a term of the form 
    $\E[ G(z \cdot \theta, \tilde{z} \cdot \theta^*) z]$ for $G$ of the form $G(t_1, t_2) = \psi'(t_1) \psi(t_2)$. Doing a second-order Taylor expansion leads to 
\begin{align}
    G( z \cdot \theta, \tilde{z} \cdot \theta^*) &= G( z_1 \theta_1 + z_{-1} \cdot \theta_{-1}, \tilde{z}_1 \theta_1^* + \tilde{z}_{-1} \cdot \theta_{-1}^*) \\
    &= \psi'( z_{-1} \cdot \theta_{-1}) \psi(\tilde{z}_{-1} \cdot \theta_{-1}^*) + \psi''( z_{-1} \cdot \theta_{-1}) \psi(\tilde{z}_{-1} \cdot \theta_{-1}^*) \theta_1 z_1 \notag\\
    &\quad + \psi'( z_{-1} \cdot \theta_{-1})\psi'( \tilde{z}_{-1} \cdot \theta_{-1}^*) \tilde{z}_1 \theta_1^* \\
    & + \frac12 [ z_1 \theta_1 ~~\tilde{z}_1 \theta_1^* ]^\top \nabla^2 G(M) [ z_1 \theta_1 ~~ \tilde{z}_1 \theta_1^* ]~,
\end{align}
and thus, for all $i \in [d]$, denoting $\alpha = \E[ z_i \tilde{z}_i]$, and assuming that $\nabla G$ is Lipschitz, we have 
\begin{align}
    \E[G(z \cdot \theta, \tilde{z} \cdot \theta^*)z_i ] &= \theta_i \E[ \psi''( z_{-i} \cdot \theta_{-i}) \psi(\tilde{z}_{-i} \cdot \theta_{-i}^*)] \notag\\
    &\quad + \alpha \theta_i^* \E[\psi'( z_{-i} \cdot \theta_{-i})\psi'( \tilde{z}_{-i} \cdot \theta_{-i}^*)] + O(\theta_i^2 + (\theta_i^*)^2) ~.
\end{align}
Following our previous argument, we now consider for each $\theta, \theta^*$ and each $i\in [d]$ the bivariate distributions $\bar{\mu}_i = \mathrm{Law}( \sum_{j \neq i} \theta_j z_j, \sum_{j \neq i} \theta_j^* \tilde{z}_j)$, where $(z_i, \tilde{z}_i)$ are drawn i.i.d. from $\bar{\nu}$. 
We now have 
\begin{align}
    \sup_{i , i'} W_1( \bar{\mu}_i, \bar{\mu}_{i'}) = O( \| \theta\|_\infty + \|\theta^*\|_\infty)~,
\end{align}
leading to 
\begin{align}
    \E[G(z \cdot \theta, \tilde{z} \cdot \theta^*) z ] &= \theta \upsilon_1(\theta, \theta^*) + \alpha \theta^* \upsilon_2(\theta, \theta^*) + \varepsilon ~,~ 
\end{align}
with $|\varepsilon_i | \leq (C_1 |\theta_i|  + C_2 |\alpha| |\theta_i^*| ) (\| \theta\|_\infty + \| \theta^* \|_\infty) ~$ and 
$\upsilon_2=O(1)$. 
We then arrive at 
\begin{align}
    P_{\theta^\perp} \E[G(z \cdot \theta, \tilde{z} \cdot \theta^*) z ] &= \alpha \theta^* \upsilon_2(\theta, \theta^*) - \chi \alpha \upsilon_2 \theta + \varepsilon - (\varepsilon \cdot \theta) \theta~.
\end{align}
Again, since $|\varepsilon \cdot \theta| = O( \| \theta\|_\infty + \|\theta^*\|_\infty) $, we have 
\begin{align}\label{eq:thetaellinfty}
    \| \nabla_\theta^S \mathcal{L} \|_\infty & \leq  C_1 \rho \| \theta^*\|_\infty +  C_2 \rho \chi \| \theta\|_\infty  + O( \| \theta\|_\infty^2 + \| \theta^* \|_\infty^2)~,
\end{align}
proving the lemma. 
\end{proof}

\subsection{Weak Recovery}

At the weak recovery, the total overlap \(Z_t\) is still below the incoherence scale. In this regime the crude \(O(\eta_t)\) comparison error from Lemma~\ref{lem:uniform-positive-overlap-drift} is too large under spherical initialization. We instead track the third moment quantity $\omega_t:=\sum_{i=1}^d\bigl(|\theta_{t,i}|+|\theta_i^*|\bigr)^3$ and show it is controlled, at population level, under the order $\frac{1+t}{\sqrt{d}}.$

\begin{lemma}[First-order comparison in the small-overlap regime]
\label{lem:refined-first-order-comparison-small}
Assume the hypotheses of Lemma~\ref{lem:uniform-first-order-gradient-comparison}
and the regularity needed for Corollary~\ref{cor:stein-block}. Define
\[
\omega(\theta,\theta^*)
:=
\sum_{i=1}^d\bigl(|\theta_i|+|\theta_i^*|\bigr)^3 .
\]
Let
\[
\dot\rho
:=
-\left\langle \nabla_{\mathbb S,\beta}L(\theta,\beta),\beta^*\right\rangle,
\qquad
\dot\chi
:=
-\left\langle \nabla_{\mathbb S,\theta}L(\theta,\beta),\theta^*\right\rangle,
\]
and define \(\dot\rho_G,\dot\chi_G\) analogously with \(L_G\). Then, uniformly
on the small-overlap region \(0\le \rho,\chi\) and \(Z:=\rho+\chi\le c_{\rm sm}\),
\begin{align}
|\dot\rho-\dot\rho_G|+|\dot\chi-\dot\chi_G|
\le
C\omega(\theta,\theta^*),
\end{align}
where \(C\) is independent of \(d\).
\end{lemma}

\begin{proof}
Throughout, constants may change from line to line and depend only on the
model and moment constants. Set
\[
U_i:=x^{(i)}\cdot\beta,
\qquad
U_i^*:=x^{(i)}\cdot\beta^*,
\]
so that \((U_i,U_i^*)_{i=1}^d\) are iid centered Gaussian pairs with correlation
\(\rho\). Let
\[
X_i:=\phi(U_i),
\qquad
X_i^*:=\phi(U_i^*),
\]
and
\[
S:=\sum_{i=1}^d\theta_iX_i,
\qquad
S^*:=\sum_{i=1}^d\theta_i^*X_i^* .
\]
Let \((\Gamma_i,\Gamma_i^*)_{i=1}^d\) be iid centered Gaussian pairs with the
same block covariance as \((X_i,X_i^*)\):
\[
\E[\Gamma_i^2]=\E[(\Gamma_i^*)^2]=1,
\qquad
\E[\Gamma_i\Gamma_i^*]=\kappa(\rho),
\]
and write
\[
S_G:=\sum_{i=1}^d\theta_i\Gamma_i,
\qquad
S_G^*:=\sum_{i=1}^d\theta_i^*\Gamma_i^* .
\]

We prove the two block comparisons separately.

\textbf{The \(\theta\)-block.}
Let
\[
w:=P_{\theta^\perp}\theta^*=\theta^*-\chi\theta .
\]
Since \(w\in T_\theta\mathbb S^{d-1}\),
\[
\left\langle
\nabla_{\mathbb S,\theta}(L-L_G)(\theta,\beta),
\theta^*
\right\rangle
=
\left\langle
\nabla_\theta(L-L_G)(\theta,\beta),
w
\right\rangle .
\]
Thus the \(\theta\)-comparison is the difference between the true and Gaussian
expectations of a function of the form
\[
h(x,x^*)
=
2\bigl(\psi(x\cdot\theta)-\psi(x^*\cdot\theta^*)\bigr)
\psi'(x\cdot\theta)\,(x\cdot w).
\]
Apply Corollary~\ref{cor:stein-block} to \(h\). For a fixed block \(j\), every
third partial derivative \(\partial_s^a\partial_t^b h\), \(a+b=3\), is a finite
sum of terms of the following two types:
\[
C|\theta_j|^a|\theta_j^*|^b\,|\ell_2|,
\qquad a+b=3,
\]
and
\[
C|w_j|\,|\theta_j|^{a'}|\theta_j^*|^{b'},
\qquad a'+b'=2,
\]
where \(\ell_2\) denotes the linear form \(w\cdot W\) evaluated at the
corresponding Lindeberg hybrid vector. Since \(\|w\|_2\le 1\), the moment
assumptions imply
\[
\E|\ell_2|\le C .
\]
Therefore
\[
\left|
\left\langle
\nabla_{\mathbb S,\theta}(L-L_G)(\theta,\beta),
\theta^*
\right\rangle
\right|
\le
C\sum_{j=1}^d\sum_{a+b=3}|\theta_j|^a|\theta_j^*|^b
+
C\sum_{j=1}^d |w_j|
\sum_{a'+b'=2}|\theta_j|^{a'}|\theta_j^*|^{b'} .
\]
Now
\[
|w_j|
=
|\theta_j^*-\chi\theta_j|
\le
|\theta_j^*|+|\theta_j|,
\]
because \(|\chi|\le1\). Hence, for every \(j\),
\[
\sum_{a+b=3}|\theta_j|^a|\theta_j^*|^b
\le
C\bigl(|\theta_j|+|\theta_j^*|\bigr)^3,
\]
and
\[
|w_j|\sum_{a'+b'=2}|\theta_j|^{a'}|\theta_j^*|^{b'}
\le
C\bigl(|\theta_j|+|\theta_j^*|\bigr)^3 .
\]
Summing over \(j\), we obtain
\[
\left|
\left\langle
\nabla_{\mathbb S,\theta}(L-L_G)(\theta,\beta),
\theta^*
\right\rangle
\right|
\le
C\omega(\theta,\theta^*).
\]
Equivalently,
\[
|\dot\chi-\dot\chi_G|
\le
C\omega(\theta,\theta^*).
\]

\textbf{The \(\beta\)-block.}
Let
\[
b:=P_{\beta^\perp}\beta^*=\beta^*-\rho\beta .
\]
Then
\[
\left\langle
\nabla_{\mathbb S,\beta}L(\theta,\beta),
\beta^*
\right\rangle
=
\left\langle
\nabla_{\beta}L(\theta,\beta),
b
\right\rangle .
\]
As in the proof of Lemma~\ref{lem:uniform-first-order-gradient-comparison},
Gaussian integration by parts in the \(b\)-direction gives the exact formula
\[
-\left\langle
\nabla_{\mathbb S,\beta}L(\theta,\beta),
\beta^*
\right\rangle
=
2(1-\rho^2)
\sum_{i=1}^d
\theta_i\theta_i^*
\E\!\left[
\psi'(S)\psi'(S^*)\phi'(U_i)\phi'(U_i^*)
\right].
\]
Set
\[
F(s,t):=\psi'(s)\psi'(t),
\qquad
q_i:=\phi'(U_i)\phi'(U_i^*),
\qquad
c_i:=\theta_i\theta_i^* .
\]
We compare
\[
\sum_{i=1}^d c_i\E[F(S,S^*)q_i]
\]
with its Gaussian analogue.

Let
\[
S_{-i}:=S-\theta_iX_i,
\qquad
S_{-i}^*:=S^*-\theta_i^*X_i^* .
\]
Since \((S_{-i},S_{-i}^*)\) is independent of \((U_i,U_i^*)\),
\[
\E[F(S_{-i},S_{-i}^*)q_i]
=
\E[F(S_{-i},S_{-i}^*)]\E[q_i].
\]
By the Hermite identity,
\[
\E[q_i]
=
\E[\phi'(U_i)\phi'(U_i^*)]
=
\kappa'(\rho).
\]
Moreover, by the Lipschitz bound on \(F\) and the moment assumptions,
\[
\left|
\E[F(S,S^*)q_i]
-
\E[F(S_{-i},S_{-i}^*)q_i]
\right|
\le
C(|\theta_i|+|\theta_i^*|).
\]
Multiplying by \(|c_i|=|\theta_i\theta_i^*|\) and summing yields
\[
\sum_i |c_i|(|\theta_i|+|\theta_i^*|)
=
\sum_i |\theta_i\theta_i^*|(|\theta_i|+|\theta_i^*|)
\le
C\omega(\theta,\theta^*).
\]
Hence
\[
\sum_{i=1}^d c_i\E[F(S,S^*)q_i]
=
\kappa'(\rho)
\sum_{i=1}^d c_i\E[F(S_{-i},S_{-i}^*)]
+
O(\omega).
\]
The same leave-one-out argument gives
\[
\sum_{i=1}^d c_i\E[F(S_{-i},S_{-i}^*)]
=
\left(\sum_{i=1}^d c_i\right)\E[F(S,S^*)]
+
O(\omega).
\]
Since
\[
\sum_{i=1}^d c_i
=
\theta\cdot\theta^*
=
\chi,
\]
we obtain
\[
\sum_{i=1}^d c_i\E[F(S,S^*)q_i]
=
\kappa'(\rho)\chi\,\E[F(S,S^*)]+O(\omega).
\]

It remains to replace \((S,S^*)\) by \((S_G,S_G^*)\). Applying
Corollary~\ref{cor:stein-block} to
\[
h(x,x^*)=F(x\cdot\theta,x^*\cdot\theta^*)
\]
gives
\[
\left|
\E[F(S,S^*)]-\E[F(S_G,S_G^*)]
\right|
\le
C\sum_{j=1}^d\sum_{a+b=3}|\theta_j|^a|\theta_j^*|^b
\le
C\omega(\theta,\theta^*).
\]
Since \(\kappa'(\rho)\) and \(\chi\) are bounded, we conclude
\[
\sum_{i=1}^d c_i\E[F(S,S^*)q_i]
=
\kappa'(\rho)\chi\,\E[F(S_G,S_G^*)]+O(\omega).
\]

In the Gaussian equivalent model, Price's identity gives
\[
-\left\langle
\nabla_{\mathbb S,\beta}L_G(\theta,\beta),
\beta^*
\right\rangle
=
2(1-\rho^2)\kappa'(\rho)\chi
\E[F(S_G,S_G^*)].
\]
Comparing the last two displays and using \(0\le 1-\rho^2\le1\), we get
\[
\left|
\left\langle
\nabla_{\mathbb S,\beta}(L-L_G)(\theta,\beta),
\beta^*
\right\rangle
\right|
\le
C\omega(\theta,\theta^*).
\]
Equivalently,
\[
|\dot\rho-\dot\rho_G|
\le
C\omega(\theta,\theta^*).
\]

Combining the \(\theta\)- and \(\beta\)-block estimates proves
\[
|\dot\rho-\dot\rho_G|+|\dot\chi-\dot\chi_G|
\le
C\omega(\theta,\theta^*).
\]
\end{proof}

\begin{lemma}[Small-overlap drift estimates]
\label{lem:refined-small-overlap-drift}
Under the assumptions of Lemma~\ref{lem:uniform-first-order-gradient-comparison}, with the same definition of $\omega(\theta,\theta^*)$ in~\ref{lem:refined-first-order-comparison-small}. There exist constants \(c_{\rm sm},a,b,C>0\), independent of \(d\), such that whenever $\rho,\chi\ge0, \quad Z:=\rho+\chi\le c_{\rm sm},$ the population overlap drifts satisfy
\begin{equation}
\label{eq:small-overlap-coop-drift}
\dot\rho
\ge
a\chi-b\omega(\theta,\theta^*),
\qquad
\dot\chi
\ge
a\rho-b\omega(\theta,\theta^*).
\end{equation}
Moreover, with $\widetilde C:=2\alpha_1^2a_1^2>0,$
one has the refined two-sided expansion
\begin{equation}
\label{eq:small-overlap-two-sided-drift}
\dot\rho
=
\widetilde C\,\chi(1-\rho^2)+R_\rho,
\qquad
\dot\chi
=
\widetilde C\,\rho(1-\chi^2)+R_\chi,
\end{equation}
where $|R_\rho|+|R_\chi|
\le
C\bigl(Z^2+\omega(\theta,\theta^*)\bigr).$
Consequently, for \(D:=\rho-\chi\),
\begin{equation}
\label{eq:small-overlap-D-drift}
\dot D
=
-\widetilde C(1+\rho\chi)D+R_D,
\qquad
|R_D|
\le
C\bigl(Z^2+\omega(\theta,\theta^*)\bigr).
\end{equation}
\end{lemma}
\begin{proof}
Let \(L_G\) denote the Gaussian-equivalent population loss. As in the proof of
Lemma~\ref{lem:uniform-positive-overlap-drift}, the Gaussian overlap drifts are
\[
\dot\rho_G
=
2\chi\,\kappa'(\rho)A(q)(1-\rho^2),
\qquad
\dot\chi_G
=
2\kappa(\rho)A(q)(1-\chi^2),
\]
where
\[
q:=\kappa(\rho)\chi,
\qquad
A(q):=\sum_{k\ge1}k\alpha_k^2q^{k-1},
\qquad
\kappa(\rho):=\sum_{\ell\ge1}a_\ell^2\rho^\ell .
\]
Since \(a_1\ne0\) and \(\alpha_1\ne0\),
\[
\kappa'(0)=a_1^2>0,
\qquad
A(0)=\alpha_1^2>0.
\]

First we prove the cooperative lower bounds. On the positive-overlap region
\(\rho,\chi\ge0\), all terms in the Gaussian drift formulas are nonnegative.
Keeping only the leading terms gives
\[
\dot\rho_G
\ge
2\alpha_1^2a_1^2\,\chi(1-\rho^2),
\qquad
\dot\chi_G
\ge
2\alpha_1^2a_1^2\,\rho(1-\chi^2).
\]
Choose \(c_{\rm sm}>0\) sufficiently small so that \(Z\le c_{\rm sm}\) implies
\(\rho,\chi\le c_{\rm sm}\) and hence
\[
1-\rho^2\ge \frac12,
\qquad
1-\chi^2\ge \frac12.
\]
Therefore
\[
\dot\rho_G\ge c_0\chi,
\qquad
\dot\chi_G\ge c_0\rho
\]
for \(c_0:=\alpha_1^2a_1^2>0\).

It remains to transfer these estimates from \(L_G\) to the true population loss.
By Lemma~\ref{lem:refined-first-order-comparison-small},
\[
|\dot\rho-\dot\rho_G|+|\dot\chi-\dot\chi_G|
\le C\omega(\theta,\theta^*).
\]
Hence, after setting \(a:=c_0/2\) and increasing \(b\),
\[
\dot\rho\ge a\chi-b\omega(\theta,\theta^*),
\qquad
\dot\chi\ge a\rho-b\omega(\theta,\theta^*),
\]
which proves \eqref{eq:small-overlap-coop-drift}.

We now prove the two-sided expansion. Since \(Z\le c_{\rm sm}\), we have
\[
|\rho|+|\chi|\le c_{\rm sm}.
\]
Taylor expanding \(\kappa\), \(\kappa'\), and \(A\) near the origin gives
\[
\kappa(\rho)=a_1^2\rho+O(\rho^2),
\qquad
\kappa'(\rho)=a_1^2+O(|\rho|),
\]
and since
\[
q=\kappa(\rho)\chi=O(Z^2),
\]
also
\[
A(q)=\alpha_1^2+O(Z^2).
\]
Substituting these expansions into the exact Gaussian drift formulas yields
\[
\dot\rho_G
=
2\alpha_1^2a_1^2\,\chi(1-\rho^2)+O(Z^2),
\]
and
\[
\dot\chi_G
=
2\alpha_1^2a_1^2\,\rho(1-\chi^2)+O(Z^2).
\]
Indeed, in the first formula the error from
\(\kappa'(\rho)-a_1^2\) is \(O(|\rho||\chi|)=O(Z^2)\), while the error from
\(A(q)-\alpha_1^2\) is \(O(|\chi|Z^2)=O(Z^3)\). In the second formula, the error
from \(\kappa(\rho)-a_1^2\rho\) is \(O(\rho^2)=O(Z^2)\), and the error from
\(A(q)-\alpha_1^2\) is \(O(|\rho|Z^2)=O(Z^3)\).

Thus, with $\widetilde C:=2\alpha_1^2a_1^2,$
we have
\[
\dot\rho_G
=
\widetilde C\,\chi(1-\rho^2)+O(Z^2),
\qquad
\dot\chi_G
=
\widetilde C\,\rho(1-\chi^2)+O(Z^2).
\]
Using again the refined comparison estimate
\[
|\dot\rho-\dot\rho_G|
+
|\dot\chi-\dot\chi_G|
\le
C\omega(\theta,\theta^*),
\]
we obtain
\[
\dot\rho
=
\widetilde C\,\chi(1-\rho^2)+R_\rho,
\qquad
\dot\chi
=
\widetilde C\,\rho(1-\chi^2)+R_\chi,
\]
with $|R_\rho|+|R_\chi|
\le
C\bigl(Z^2+\omega(\theta,\theta^*)\bigr).$ This proves \eqref{eq:small-overlap-two-sided-drift}.

Finally, subtract the two expansions. Since
\[
\chi(1-\rho^2)-\rho(1-\chi^2)
=
\chi-\rho-\rho^2\chi+\rho\chi^2
=
-(\rho-\chi)-\rho\chi(\rho-\chi)
=
-(1+\rho\chi)D,
\]
we get
\[
\dot D
=
-\widetilde C(1+\rho\chi)D+R_D,
\]
where
\[
|R_D|
\le
|R_\rho|+|R_\chi|
\le
C\bigl(Z^2+\omega(\theta,\theta^*)\bigr).
\]
This proves \eqref{eq:small-overlap-D-drift}.
\end{proof}

\begin{lemma}[Third-moment control at population level]
\label{lem:population-third-moment-control}
Let $\Omega_t:=\|\theta_t\|_3^3=\sum_{i=1}^d|\theta_{t,i}|^3, \quad \omega_t:=\sum_{i=1}^d\bigl(|\theta_{t,i}|+|\theta_i^*|\bigr)^3.$ Work on the initialization event of Lemma~\ref{lem:random-spherical-delocalization}. Then, on the positive-overlap region \(\rho_t,\chi_t\ge0\), with $Z_t:=\rho_t+\chi_t,$ the population flow satisfies
\begin{equation}
\label{eq:Omega-dot-bound-corrected}
\dot\Omega_t
\le
C\left(
Z_t\|\theta^*\|_\infty
+
Z_t^2\eta_t
+
\eta_t^2
\right).
\end{equation}

Consequently, suppose that on an interval \(0\le t\le T\), with
\(T\le C_T\log d\), one has
\[
Z_t^2\le A\eta_t,
\qquad
\eta_t\le C_\eta\frac{\sqrt{\log d}+1+t}{\sqrt d}.
\]
Then, for all sufficiently large \(d\),
\begin{equation}
\label{eq:Omega-phase1-bound}
\Omega_t\le C_\Omega\frac{1+t}{\sqrt d},\quad \omega_t\le C_\omega\frac{1+t}{\sqrt d},
\quad 0\le t\le T.
\end{equation}
The constants \(C_\Omega,C_\omega\) may depend on \(K,A,C_\eta,C_T\), but not
on \(d\).
\end{lemma}

\begin{proof}
The population \(\theta\)-flow is
\[
\dot\theta_t=-\nabla_{\mathbb S,\theta}L(\theta_t,\beta_t).
\]
Let
\[
v_t=(|\theta_{t,i}|\theta_{t,i})_{i=1}^d.
\]
Since \(x\mapsto |x|^3\) is \(C^1\), we have
\[
\frac{d}{dt}\Omega_t
=
3\sum_i|\theta_{t,i}|\theta_{t,i}\dot\theta_{t,i}
=
-3\langle v_t,\nabla_{\mathbb S,\theta}L(\theta_t,\beta_t)\rangle.
\]
Moreover,
\[
\|v_t\|_1=\sum_i|\theta_{t,i}|^2=1.
\]
Hence
\[
\dot\Omega_t
\le
3\|\nabla_{\mathbb S,\theta}L(\theta_t,\beta_t)\|_\infty.
\]

By Lemma~\ref{lemma:secondlayercontrol}, more precisely by
\eqref{eq:thetaellinfty}, we have
\[
\|\nabla_{\mathbb S,\theta}L(\theta_t,\beta_t)\|_\infty
\le
C_1|\rho_t|\|\theta^*\|_\infty
+
C_2|\rho_t\chi_t|\|\theta_t\|_\infty
+
C_3\eta_t^2.
\]
On the positive-overlap region, \(\rho_t,\chi_t\ge0\), and hence
\[
|\rho_t|\le Z_t,
\qquad
|\rho_t\chi_t|\le Z_t^2,
\qquad
\|\theta_t\|_\infty\le \eta_t.
\]
Therefore
\[
\dot\Omega_t
\le
C\left(
Z_t\|\theta^*\|_\infty
+
Z_t^2\eta_t
+
\eta_t^2
\right),
\]
which proves \eqref{eq:Omega-dot-bound-corrected}.

Now with the conditions
\[
Z_t^2\le A\eta_t,
\qquad
\eta_t\le C_\eta\frac{\sqrt{\log d}+1+t}{\sqrt d},
\qquad
0\le t\le T\le C_T\log d.
\]
Then
\[
Z_t\|\theta^*\|_\infty
\le
\sqrt{A\eta_t}\,
C_K\sqrt{\frac{\log d}{d}}.
\]
Using the envelope for \(\eta_t\),
\[
Z_t\|\theta^*\|_\infty
\le
C
\sqrt{
\frac{\sqrt{\log d}+1+t}{\sqrt d}
}
\sqrt{\frac{\log d}{d}}.
\]
Since \(t\le C_T\log d\), the right-hand side is bounded by
\[
C\frac{\log d}{d^{3/4}}
\le
\frac{C}{\sqrt d}
\]
for all sufficiently large \(d\). Also,
\[
Z_t^2\eta_t\le A\eta_t^2.
\]
Therefore, throughout the Phase-I interval,
\[
\dot\Omega_t
\le
\frac{C}{\sqrt d}+C\eta_t^2.
\]

Integrating from \(0\) to \(t\le T\), we get
\[
\Omega_t
\le
\Omega_0+\frac{Ct}{\sqrt d}
+
C\int_0^t\eta_s^2\,ds.
\]
By Lemma~\ref{lem:random-spherical-delocalization},
\[
\Omega_0=\|\theta_0\|_3^3\le C_Kd^{-1/2}.
\]
Moreover,
\[
\int_0^t\eta_s^2\,ds
\le
\frac{C}{d}
\int_0^t(\sqrt{\log d}+1+s)^2\,ds.
\]
For \(t\le C_T\log d\),
\[
\int_0^t(\sqrt{\log d}+1+s)^2\,ds
\le
C(\log d)^3.
\]
Thus
\[
\int_0^t\eta_s^2\,ds
\le
\frac{C(\log d)^3}{d}
=
o(d^{-1/2}).
\]
Hence
\[
\Omega_t
\le
C_\Omega\frac{1+t}{\sqrt d},
\qquad
0\le t\le T.
\]

Finally, by the elementary inequality
\[
(a+b)^3\le 4a^3+4b^3
\qquad
(a,b\ge0),
\]
we have
\[
\omega_t
=
\sum_i\bigl(|\theta_{t,i}|+|\theta_i^*|\bigr)^3
\le
4\|\theta_t\|_3^3+4\|\theta^*\|_3^3.
\]
Using the bound on \(\Omega_t\) and the initialization event
\[
\|\theta^*\|_3^3\le C_Kd^{-1/2},
\]
we obtain
\[
\omega_t
\le
C_\omega\frac{1+t}{\sqrt d}.
\]
This proves \eqref{eq:Omega-phase1-bound}.
\end{proof}

\begin{lemma}[Escape from mediocrity at population level]
\label{lem:popescape}
Assume Lemmas~\ref{lemma:secondlayercontrol},
\ref{lem:refined-small-overlap-drift}, and
\ref{lem:population-third-moment-control}. Work on the random-spherical
delocalization event of Lemma~\ref{lem:random-spherical-delocalization}, namely
\[
\|\theta_0\|_\infty+\|\theta^*\|_\infty
\le
C_K\sqrt{\frac{\log d}{d}},
\qquad
\|\theta_0\|_3^3+\|\theta^*\|_3^3
\le
C_Kd^{-1/2},
\] and the favorable initialization event defined in \ref{ass:favorable-initialization}. Then there exist constants \(c_{\rm esc},C_{\rm esc}>0\), independent of \(d\),
such that for all sufficiently large \(d\), there is a time
\begin{align}
t_{\rm esc}\le C_{\rm esc}\log d
\end{align}
with
\begin{align}
\min\{\rho_{t_{\rm esc}},\chi_{t_{\rm esc}}\}\ge c_{\rm esc}.
\end{align}
Moreover, throughout the escape window,
\begin{align}
\eta_t\le C\frac{\log d}{\sqrt d}=o_d(1).
\end{align}
\end{lemma}

\begin{proof}
Fix a large constant \(C_T>0\), to be chosen below, and set
\[
T:=C_T\log d.
\]
Let \(A>0\) be a fixed large constant and define the stopping times
\[
\tau_A:=\inf\{t:Z_t^2\ge A\eta_t\},
\]
\[
\tau_B:=\inf\{t:\rho_t\ge 1/2\text{ or }\chi_t\ge 1/2\},
\]
\[
\tau_+:=\inf\{t:\rho_t<0\text{ or }\chi_t<0\}.
\]
We first work on Phase I, defined as when $t\in [0,\sigma]$ where $\sigma:=\tau_A\wedge\tau_B\wedge\tau_+\wedge T.$ 
\paragraph{Step 1: Incoherence Control in Phase I}
We begin with the Phase-I control of \(\eta_t\). By Lemma~\ref{lemma:secondlayercontrol}, on the positive-overlap region,
\[
\dot\eta_t
\le
C Z_t\|\theta^*\|_\infty
+
C Z_t^2\eta_t
+
C\eta_t^2.
\]
For \(t<\tau_A\), we have \(Z_t^2<A\eta_t\). Therefore
\[
Z_t\|\theta^*\|_\infty
\le
\sqrt{A\eta_t}\,\|\theta^*\|_\infty.
\]
Define
\[
H_I(t):=
K_I\frac{\sqrt{\log d}+1+t}{\sqrt d},
\]
where \(K_I\) is a large constant. We claim that
\[
\eta_t\le H_I(t)
\qquad
(0\le t\le\sigma).
\]
Indeed, at initialization,
\[
\eta_0
=
\|\theta_0\|_\infty+\|\theta^*\|_\infty
\le
C_K\sqrt{\frac{\log d}{d}}
\le
\frac12H_I(0)
\]
after choosing \(K_I\) large. Suppose the estimate holds up to a first
violation time. Since \(t\le T=C_T\log d\), the bootstrap bound gives
\[
\eta_t\le H_I(t)\le C\frac{\log d}{\sqrt d}.
\]
Thus
\[
Z_t\|\theta^*\|_\infty
\le
\sqrt{A H_I(t)}\,C_K\sqrt{\frac{\log d}{d}}
\le
C\frac{\log d}{d^{3/4}}
\le
\frac{C}{\sqrt d}
\]
for all sufficiently large \(d\). Also,
\[
Z_t^2\eta_t\le A\eta_t^2.
\]
Consequently, up to the first violation time,
\[
\dot\eta_t
\le
\frac{C}{\sqrt d}+C\eta_t^2.
\]
Integrating,
\[
\eta_t
\le
\eta_0+\frac{Ct}{\sqrt d}
+
C\int_0^t\eta_s^2\,ds.
\]
Under the bootstrap \(\eta_s\le H_I(s)\),
\[
\int_0^t\eta_s^2\,ds
\le
\frac{C}{d}
\int_0^{C_T\log d}
(\sqrt{\log d}+1+s)^2\,ds
\le
\frac{C(\log d)^3}{d}
=
o\!\left(\frac{\sqrt{\log d}+1+t}{\sqrt d}\right).
\]
Choosing \(K_I\) large gives
\[
\eta_t\le \frac12H_I(t)
\]
at the putative first violation, a contradiction. Hence for $0\le t\le\sigma,$
\[
\eta_t\le H_I(t).
\]
So by Lemma~\ref{lem:population-third-moment-control}, we have for \(0\le t\le\sigma,\)
\[
\Omega_t:=\|\theta_t\|_3^3
\le
C_\Omega\frac{1+t}{\sqrt d},
\]
and
\[
\omega_t
\le
C_\omega\frac{1+t}{\sqrt d}.
\]

\paragraph{Step 2: Phase I Positivity, Exponential Pull-Up, and Balancing.}

We next prove positivity and exponential pull-up in Phase I. On
\(t<\sigma\), we have
\[
0\le \rho_t,\chi_t\le 1/2.
\]
By Lemma~\ref{lem:refined-small-overlap-drift}, since
\(0\le\rho_t,\chi_t\le 1/2\) on the stopped Phase-I interval and \(d\) is large,
after decreasing \(a>0\) and increasing \(b>0\),
\[
\dot\rho_t\ge a\chi_t-b\omega_t,
\qquad
\dot\chi_t\ge a\rho_t-b\omega_t.
\]
Let
\[
u_t:=(\rho_t,\chi_t)^\top,
\qquad
J:=
\begin{pmatrix}
0&1\\
1&0
\end{pmatrix},
\qquad
\mathbf 1:=(1,1)^\top.
\]
Then, componentwise,
\[
\dot u_t\ge aJu_t-b\omega_t\mathbf 1.
\]
Since \(e^{aJr}\) has nonnegative entries and
\[
e^{aJr}\mathbf 1=e^{ar}\mathbf 1,
\]
variation of constants gives, for \(0\le t\le\sigma\),
\[
u_t
\ge
e^{aJt}u_0
-
b\int_0^t e^{aJ(t-s)}\omega_s\mathbf 1\,ds.
\]
Using $\rho_0,\chi_0\ge z_{\rm init}d^{-1/2},$ from \ref{ass:favorable-initialization}, we get componentwise
\[
e^{aJt}u_0
\ge
e^{at}\frac{z_{\rm init}}{\sqrt d}\mathbf 1.
\]
Also,
\[
b\int_0^t e^{aJ(t-s)}\omega_s\mathbf 1\,ds
=
b e^{at}
\left(\int_0^t e^{-as}\omega_s\,ds\right)\mathbf 1.
\]
Since
\[
\omega_s\le C_\omega\frac{1+s}{\sqrt d},
\]
we have
\[
\int_0^\infty e^{-as}\omega_s\,ds
\le
\frac{C}{\sqrt d}.
\]
Let
\[
M_\omega:=
bC_\omega\int_0^\infty e^{-as}(1+s)\,ds.
\]
By choosing \(z_{\rm init}>2M_\omega\), we obtain
\[
\rho_t,\chi_t
\ge
e^{at}\frac{z_{\rm init}-M_\omega}{\sqrt d}
\ge
\frac{c_-}{\sqrt d}e^{at}
\qquad
(0\le t\le\sigma),
\]
for some \(c_->0\). In particular,
\[
Z_t\ge \frac{2c_-}{\sqrt d}e^{at}
\qquad
(0\le t\le\sigma),
\]
and \(\tau_+\) cannot be the first stopping time.

Choose
\[
t_0:=C_0\log d
\]
with \(C_0>0\) sufficiently large, and then choose \(C_T>C_0\). Here we show that
\begin{align}\label{eq:sigmat0bound}
\sigma\le t_0.
\end{align}
Suppose instead that \(\sigma>t_0\). Then \(t_0<\tau_A\), so
\[
Z_{t_0}^2<A\eta_{t_0}.
\]
But the lower bound above gives
\[
Z_{t_0}^2
\ge
\frac{c}{d}e^{2at_0}
=
c\,d^{2aC_0-1},
\]
whereas the Phase-I \(\eta\)-bound gives
\[
A\eta_{t_0}
\le
C\frac{\sqrt{\log d}+1+t_0}{\sqrt d}
\le
C\frac{\log d}{\sqrt d}.
\]
For \(C_0\) sufficiently large, this is impossible for large \(d\). Hence \eqref{eq:sigmat0bound} holds.

Since \(t_0<T\) and \(\tau_+\) cannot occur first, the first stopping time is
either \(\tau_A\) or \(\tau_B\). We claim it must be \(\tau_A\). If
\(\sigma=\tau_B\le \tau_A\), then by continuity and the definition of
\(\tau_B\),
\[
Z_{\tau_B}\ge \frac12.
\]
On the other hand, since \(\tau_B\le\tau_A\),
\[
Z_{\tau_B}^2\le A\eta_{\tau_B}
\le
C\frac{\log d}{\sqrt d},
\]
and hence
\[
Z_{\tau_B}\le C\sqrt{\log d}\,d^{-1/4}=o(1),
\]
a contradiction. Therefore
\[
\sigma=\tau_A,
\qquad
\tau_A\le C_0\log d,
\qquad
\tau_A<\tau_B.
\]
Since \(Z_0\le C_{\rm init}d^{-1/2}\) by assumption and given the lower bound $\eta_0\ge \|\theta^*\|_\infty\ge d^{-1/2},$ we have \(Z_0^2<A\eta_0\) for all large \(d\). Thus, by continuity,
\[
Z_{\tau_A}^2=A\eta_{\tau_A}.
\]
Moreover,
\[
\eta_{\tau_A}\ge \|\theta^*\|_\infty\ge d^{-1/2},
\]
and from the Phase-I bound,
\[
\eta_{\tau_A}\le C\frac{\log d}{\sqrt d}.
\]
Therefore
\[
Z_{\tau_A}\ge c_A d^{-1/4},
\qquad
Z_{\tau_A}\le C_A\sqrt{\log d}\,d^{-1/4}=o(1),
\]
and
\[
\eta_{\tau_A}\le C\frac{\log d}{\sqrt d}.
\]

We also need the two overlaps to be balanced at \(\tau_A\). On
\([0,\tau_A]\), by the \(D\)-drift consequence \eqref{eq:small-overlap-D-drift} of Lemma~\ref{lem:refined-small-overlap-drift}, we have
\begin{align}\label{eq:D-balance-popescape}
\dot D_t
=
-\widetilde C(1+\rho_t\chi_t)D_t
+
O(Z_t^2+\omega_t).
\end{align}
Integrating \eqref{eq:D-balance-popescape} on \([0,\tau_A]\), and using
\(Z_t^2\le A\eta_t\), gives
\[
|D_{\tau_A}|
\le
|D_0|
+
C\int_0^{\tau_A}(Z_s^2+\omega_s)\,ds.
\]
Using
\[
|D_0|\le C_{\rm init}d^{-1/2},
\qquad
\eta_s\le C\frac{\sqrt{\log d}+1+s}{\sqrt d},
\qquad
\omega_s\le C\frac{1+s}{\sqrt d},
\]
and \(\tau_A\le C_0\log d\), we obtain
\[
|D_{\tau_A}|
\le
\frac{C}{\sqrt d}
+
C\int_0^{C_0\log d}
\left(
\frac{\sqrt{\log d}+1+s}{\sqrt d}
+
\frac{1+s}{\sqrt d}
\right)ds
\le
C\frac{(\log d)^2}{\sqrt d}.
\]
Since $Z_{\tau_A}\ge c_A d^{-1/4},$
we have
\[
|D_{\tau_A}|=o(Z_{\tau_A}).
\]
In particular, for all sufficiently large \(d\),
\[
|D_{\tau_A}|\le \frac18 Z_{\tau_A}.
\]

\paragraph{Step 3: Phase II Incoherence Control and Drift Pull-Up}

Exiting Phase I, we next show that \(Z_t\) will be pulled from the \(d^{-1/4}\)-scale to a fixed constant. We define Phase II as when $t\in [\tau_A, t^*],$ where $t^*:=\inf\{t\ge \tau_A:Z_t\ge c_Z\}.$ Choose a small constant \(c_Z>0\), to be fixed below, and define
\[
H:=K_2\frac{\log d}{\sqrt d},
\qquad
T_2:=\tau_A+C_2\log d,
\]
where \(K_2,C_2\) are large constants.  Define the bootstrap stopping time
\[
\sigma_2:=
\inf\left\{
t\ge\tau_A:
\eta_t>H
\text{ or }
Z_t>2c_Z
\text{ or }
|D_t|>\frac12Z_t
\text{ or }
t=T_2
\right\}.
\]

At \(t=\tau_A\), none of the first three clauses holds for large \(d\). Hence
\(\sigma_2>\tau_A\). For \(t<\sigma_2\), we have
\[
|D_t|\le \frac12 Z_t,
\]
and therefore
\[
\frac14 Z_t\le \rho_t,\chi_t\le \frac34 Z_t.
\]
Also \(Z_t\le 2c_Z\).

We first obtain a crude bound on \(\omega_t\) on the Phase-II bootstrap
interval. From the third-moment differential inequality,
\[
\dot\Omega_t
\le
C\left(
Z_t\|\theta^*\|_\infty
+
Z_t^2\eta_t
+
\eta_t^2
\right).
\]
For \(t<\sigma_2\), we have \(Z_t\le 2c_Z\), \(\eta_t\le H\), and
\(t-\tau_A\le C_2\log d\). Thus
\[
\Omega_t
\le
\Omega_{\tau_A}
+
C\|\theta^*\|_\infty\int_{\tau_A}^t Z_s\,ds
+
C\int_{\tau_A}^t Z_s^2\eta_s\,ds
+
C\int_{\tau_A}^t\eta_s^2\,ds.
\]
Using
\[
\Omega_{\tau_A}\le C\frac{1+\tau_A}{\sqrt d}
\le C\frac{\log d}{\sqrt d},
\]
\[
\|\theta^*\|_\infty\le C_K\sqrt{\frac{\log d}{d}},
\]
and the crude bounds \(Z_s\le 2c_Z\), \(\eta_s\le H\), we get
\[
\Omega_t
\le
C\frac{\log d}{\sqrt d}
+
C c_Z\frac{(\log d)^{3/2}}{\sqrt d}
+
C c_Z^2\frac{(\log d)^2}{\sqrt d}
+
C\frac{(\log d)^3}{d}.
\]
After increasing the constant, this yields
\[
\Omega_t\le C\frac{(\log d)^2}{\sqrt d}
\qquad
(t<\sigma_2).
\]
Since $\omega_t \le 4\Omega_t+4\|\theta^*\|_3^3$ and $\|\theta^*\|_3^3\le C_Kd^{-1/2}$ by Lemma~\ref{lem:random-spherical-delocalization}, we obtain
\[
\omega_t\le W_d:=C\frac{(\log d)^2}{\sqrt d}
\qquad
(t<\sigma_2).
\]

We next prove that \(Z_t\) grows exponentially on the Phase-II bootstrap interval.
Let
\[
\tau_-:=\inf\{t\ge \tau_A:Z_t<Z_{\tau_A}/2\}.
\]
On \([\tau_A,\sigma_2\wedge\tau_-)\), we have \(Z_t\ge Z_{\tau_A}/2\), the
balance bounds imply \(\rho_t,\chi_t\ge Z_t/4\), and, by choosing \(c_Z\) small,
\(\rho_t,\chi_t\le 1/2\). Hence the refined drift gives
\[
\dot Z_t
=
\dot\rho_t+\dot\chi_t
\ge
a_ZZ_t-b_Z\omega_t
\ge
a_ZZ_t-b_ZW_d.
\]
Since
\[
\frac{W_d}{Z_{\tau_A}}
\le
C\frac{(\log d)^2/\sqrt d}{d^{-1/4}}
=
C\frac{(\log d)^2}{d^{1/4}}
=
o(1),
\]
we have, for all sufficiently large \(d\),
\[
b_ZW_d\le \frac12a_ZZ_t
\qquad
\text{on }[\tau_A,\sigma_2\wedge\tau_-).
\]
Thus
\[
\dot Z_t\ge \mu Z_t
\qquad
(\tau_A\le t<\sigma_2\wedge\tau_-)
\]
for some \(\mu>0\). This implies \(\tau_->\sigma_2\), and hence
\[
Z_t\ge Z_{\tau_A}e^{\mu(t-\tau_A)}
\qquad
(\tau_A\le t<\sigma_2).
\]
In particular, for \(s\le t<\sigma_2\),
\[
Z_s\le e^{-\mu(t-s)}Z_t.
\]

We now close the \(\eta\)-bootstrap before \(t^*\). For
\(\tau_A\le t<\sigma_2\wedge t^*\), Lemma~\ref{lemma:secondlayercontrol} gives
\[
\eta_t
\le
\eta_{\tau_A}
+
C\|\theta^*\|_\infty\int_{\tau_A}^t Z_s\,ds
+
C\int_{\tau_A}^t Z_s^2\eta_s\,ds
+
C\int_{\tau_A}^t\eta_s^2\,ds.
\]
Since \(t<t^*\), we have \(Z_t<c_Z\). Using the exponential comparison
\(Z_s\le e^{-\mu(t-s)}Z_t\), we obtain
\[
\int_{\tau_A}^t Z_s\,ds\le Cc_Z,
\qquad
\int_{\tau_A}^t Z_s^2\,ds\le Cc_Z^2.
\]
Therefore
\[
\eta_t
\le
C\frac{\log d}{\sqrt d}
+
C c_Z\sqrt{\frac{\log d}{d}}
+
C c_Z^2H
+
C H^2\log d.
\]
Choose \(c_Z>0\) small enough that
\[
C c_Z^2H\le \frac18H.
\]
Then choose \(K_2\) large enough so that
\[
C\frac{\log d}{\sqrt d}\le \frac14H.
\]
Moreover,
\[
C c_Z\sqrt{\frac{\log d}{d}}=o(H),
\qquad
H^2\log d
=
K_2^2\frac{(\log d)^3}{d}
=
o(H).
\]
Thus, for all sufficiently large \(d\),
\[
\eta_t\le \frac12H
\qquad
(\tau_A\le t<\sigma_2\wedge t^*).
\]
By a first-violation argument, the clause \(\eta_t>H\) cannot trigger before
\(t^*\).

Next we close the \(D/Z\)-balance bootstrap. On the balanced small-overlap
band, the same subtraction estimate as above gives
\[
\dot D_t
=
-\widetilde C(1+\rho_t\chi_t)D_t
+
O(Z_t^2+\omega_t).
\]
Since \(c_Z\) is small, \(1+\rho_t\chi_t\ge 1/2\). Hence, for
\(\tau_A\le t<\sigma_2\wedge t^*\),
\[
|D_t|
\le
e^{-c(t-\tau_A)}|D_{\tau_A}|
+
C\int_{\tau_A}^t e^{-c(t-s)}(Z_s^2+\omega_s)\,ds.
\]
The entry term is bounded by
\[
e^{-c(t-\tau_A)}|D_{\tau_A}|
\le
|D_{\tau_A}|
\le
\frac18 Z_{\tau_A}
\le
\frac18 Z_t.
\]
The \(\omega\)-term is bounded by
\[
C W_d
=
o(Z_{\tau_A})
\le
o(Z_t).
\]
For the \(Z_s^2\)-term, using
\[
Z_s\le e^{-\mu(t-s)}Z_t,
\]
we get
\[
\int_{\tau_A}^t e^{-c(t-s)}Z_s^2\,ds
\le
C Z_t^2
\le
C c_Z Z_t.
\]
Choosing \(c_Z\) smaller if necessary, and then \(d\) large, gives
\[
|D_t|\le \frac38 Z_t<\frac12 Z_t
\qquad
(\tau_A\le t<\sigma_2\wedge t^*).
\]
Thus the clause \(|D_t|>Z_t/2\) cannot trigger before \(t^*\).

We now show that \(t^*\le T_2\). Suppose, for contradiction, that
\(t^*>T_2\). Then \(Z_t<c_Z\) for all \(t\le T_2\). The clauses
\(\eta_t>H\) and \(|D_t|>Z_t/2\) cannot trigger before \(t^*\), as just shown,
and the clause \(Z_t>2c_Z\) cannot trigger while \(Z_t<c_Z\). Hence
\[
T_2<\sigma_2.
\]
The exponential lower bound then gives
\[
Z_{T_2}
\ge
Z_{\tau_A}e^{\mu C_2\log d}
\ge
c_A d^{-1/4}d^{\mu C_2}.
\]
Choosing \(C_2>1/(4\mu)\) large enough gives
\[
Z_{T_2}\ge c_Z
\]
for all sufficiently large \(d\), contradicting \(t^*>T_2\). Therefore
\[
t^*\le T_2.
\]

By continuity,
\[
Z_{t^*}=c_Z.
\]
The bootstrap balance gives
\[
|D_{t^*}|\le \frac12 Z_{t^*}.
\]
Therefore
\[
\min\{\rho_{t^*},\chi_{t^*}\}
=
\frac{Z_{t^*}-|D_{t^*}|}{2}
\ge
\frac{c_Z}{4}.
\]
Set
\[
t_{\rm esc}:=t^*,
\qquad
c_{\rm esc}:=\frac{c_Z}{4}.
\]
Since
\[
\tau_A\le C_0\log d
\]
and
\[
t^*\le \tau_A+C_2\log d,
\]
we have
\[
t_{\rm esc}\le (C_0+C_2)\log d.
\]
Thus the theorem holds with
\[
C_{\rm esc}:=C_0+C_2.
\]

Finally, the escape-window \(\eta_t\)-bound follows by combining the Phase-I
estimate
\[
\eta_t\le K_I\frac{\sqrt{\log d}+1+t}{\sqrt d}
\le
C\frac{\log d}{\sqrt d}
\qquad
(0\le t\le \tau_A)
\]
with the Phase-II bootstrap bound
\[
\eta_t\le H=K_2\frac{\log d}{\sqrt d}
\qquad
(\tau_A\le t\le t^*).
\]
Hence, throughout weak recovery window \(0\le t\le t_{\rm esc}\),
\[
\eta_t\le C\frac{\log d}{\sqrt d}=o_d(1).
\]
This completes the proof.
\end{proof}

\subsection{Strong Recovery}

Building on Lemma~\ref{lem:popescape}, we show that after weak recovery the
population dynamics enters the fixed local strong-convexity cap around the
teacher in an additional \(O(1)\) time. Once inside this cap, local strong
convexity gives exponential contraction to the teacher, achiving strong recovery.

\begin{lemma}[Entry into the local cap after weak recovery]
\label{lem:entry-local-cap-after-corrected-weak}
Let \(r_{\rm loc}>0\) be the radius from
Lemma~\ref{lem:true-local-product-strong-convexity}. After possibly shrinking
\(r_{\rm loc}\), assume
\[
r_{\rm loc}<\frac{\pi}{2}.
\]
Set
\[
s_{\rm loc}:=\cos(r_{\rm loc}/4).
\]

Assume ~\ref{lem:popescape} with
\(t_{\rm esc}\), \(c_{\rm esc}\), and \(C_{\rm esc}\). Under the same assumptions as
Lemmas~\ref{lem:uniform-positive-overlap-drift} and
\ref{lemma:secondlayercontrol}, for all sufficiently large \(d\), there exists
a time
\[
t_{\rm loc}\le t_{\rm esc}+C_{\rm in}
\]
where \(C_{\rm in}<\infty\) is independent of \(d\), such that
\[
\rho_{t_{\rm loc}}\ge s_{\rm loc},
\qquad
\chi_{t_{\rm loc}}\ge s_{\rm loc}.
\]
Consequently,
\[
\operatorname{dist}_{\mathcal M}(y_{t_{\rm loc}},y^*)^2
\le
\frac{r_{\rm loc}^2}{8}.
\]
In particular, since \(t_{\rm esc}\le C_{\rm esc}\log d\),
\[
t_{\rm loc}\le C_{\rm loc}\log d
\]
for some constant \(C_{\rm loc}<\infty\) independent of \(d\).
\end{lemma}

\begin{proof}
The proof is deterministic on the favorable event supplied by
Lemma~\ref{lem:popescape}. By that lemma, there are constants
\(c_{\rm esc}\in(0,1)\) and \(C_{\rm esc}<\infty\), independent of \(d\), and a
time \(t_{\rm esc}\) such that
\[
t_{\rm esc}\le C_{\rm esc}\log d,
\qquad
\rho_{t_{\rm esc}}\ge c_{\rm esc},
\qquad
\chi_{t_{\rm esc}}\ge c_{\rm esc},
\qquad
\eta_{t_{\rm esc}}=o_d(1).
\]

Set
\[
c_-:=\frac{c_{\rm esc}}2.
\]
Since \(r_{\rm loc}>0\), we have
\[
s_{\rm loc}=\cos(r_{\rm loc}/4)<1.
\]
Hence
\[
m_{\rm loc}:=1-s_{\rm loc}^2>0.
\]

By Lemma~\ref{lem:uniform-positive-overlap-drift}, on the positive-overlap
region \(\rho_t,\chi_t\ge0\), the flow satisfies
\[
\dot\rho_t
\ge
a_\rho\chi_t(1-\rho_t^2)-b_\rho\eta_t,
\]
\[
\dot\chi_t
\ge
a_\chi\rho_t(1-\chi_t^2)-b_\chi\eta_t,
\]
where \(a_\rho,a_\chi,b_\rho,b_\chi>0\) are independent of \(d\).

Choose \(\varepsilon_*>0\), independent of \(d\), so small that
\[
b_\rho\varepsilon_*
\le
\frac12a_\rho c_-m_{\rm loc},
\qquad
b_\chi\varepsilon_*
\le
\frac12a_\chi c_-m_{\rm loc},
\]
and also
\[
b_\rho\varepsilon_*
\le
\frac12a_\rho c_-(1-c_-^2),
\qquad
b_\chi\varepsilon_*
\le
\frac12a_\chi c_-(1-c_-^2).
\]
Define
\[
v_\rho:=\frac12a_\rho c_-m_{\rm loc},
\qquad
v_\chi:=\frac12a_\chi c_-m_{\rm loc}.
\]
Let
\[
T_\rho:=\frac{(s_{\rm loc}-c_{\rm esc})_+}{v_\rho},
\qquad
T_\chi:=\frac{(s_{\rm loc}-c_{\rm esc})_+}{v_\chi},
\]
and set
\[
C_{\rm in}:=1+\max\{T_\rho,T_\chi\}.
\]
This constant depends only on the fixed population constants and on
\(r_{\rm loc}\), and therefore is independent of \(d\).

We first show that \(\eta_t\) remains smaller than \(\varepsilon_*\) on the
fixed window
\[
[t_{\rm esc},t_{\rm esc}+C_{\rm in}].
\]
By Lemma~\ref{lemma:secondlayercontrol},
\[
\dot\eta_t
\le
C_1|\rho_t|\|\theta^*\|_\infty
+
C_2|\chi_t\rho_t|\eta_t
+
C_3\eta_t^2.
\]
Since \(|\rho_t|,|\chi_t|\le1\), on the present fixed window this gives
\[
\dot\eta_t
\le
A_d+B\eta_t+C\eta_t^2,
\]
where
\[
A_d:=C_1\|\theta^*\|_\infty
\le
C\sqrt{\frac{\log d}{d}}
=
o_d(1),
\]
and \(B,C<\infty\) are independent of \(d\).

Define
\[
\tau_\eta
:=
\inf\{t\ge t_{\rm esc}:\eta_t\ge\varepsilon_*\}
\wedge
(t_{\rm esc}+C_{\rm in}).
\]
Since \(\eta_{t_{\rm esc}}=o_d(1)\), for all sufficiently large \(d\),
\[
\eta_{t_{\rm esc}}<\varepsilon_*.
\]
Thus \(\tau_\eta>t_{\rm esc}\). On
\([t_{\rm esc},\tau_\eta]\), we have \(\eta_t\le\varepsilon_*\), and therefore
for almost every \(t\in[t_{\rm esc},\tau_\eta]\),
\[
\dot\eta_t
\le
A_d+(B+C\varepsilon_*)\eta_t.
\]
Set
\[
B_*:=B+C\varepsilon_*.
\]
By Gronwall's inequality, for every
\(t\in[t_{\rm esc},\tau_\eta]\),
\[
\eta_t
\le
e^{B_*(t-t_{\rm esc})}\eta_{t_{\rm esc}}
+
A_d\int_{t_{\rm esc}}^t e^{B_*(t-u)}\,du.
\]
Hence
\[
\eta_t
\le
e^{B_*C_{\rm in}}\eta_{t_{\rm esc}}
+
A_dC_{\rm in}e^{B_*C_{\rm in}}.
\]
The right-hand side is \(o_d(1)\). Therefore, for all sufficiently large \(d\),
\[
\eta_t<\varepsilon_*
\qquad
\text{for all }
t\in[t_{\rm esc},\tau_\eta].
\]
By the definition of \(\tau_\eta\), this forces
\[
\tau_\eta=t_{\rm esc}+C_{\rm in}.
\]
Consequently,
\[
\eta_t\le\varepsilon_*
\qquad
\text{for all }
t\in[t_{\rm esc},t_{\rm esc}+C_{\rm in}].
\]

Next we prove that the rectangle
\[
\mathcal R_-:=\{(\rho,\chi):\rho\ge c_-,\ \chi\ge c_-\}
\]
is forward invariant on the same window. At time \(t_{\rm esc}\),
\[
\rho_{t_{\rm esc}}\ge c_{\rm esc}>c_-,
\qquad
\chi_{t_{\rm esc}}\ge c_{\rm esc}>c_-.
\]
Suppose, for contradiction, that the trajectory exits \(\mathcal R_-\) before
\(t_{\rm esc}+C_{\rm in}\). Let
\[
\tau_-:=
\inf\left\{
t\in[t_{\rm esc},t_{\rm esc}+C_{\rm in}]:
\min\{\rho_t,\chi_t\}\le c_-
\right\}.
\]
Then \(\tau_-<\infty\), and by continuity,
\[
\rho_{\tau_-}\ge c_-,
\qquad
\chi_{\tau_-}\ge c_-,
\]
with at least one equality.

If \(\rho_{\tau_-}=c_-\), then
\(\rho_{\tau_-},\chi_{\tau_-}\ge0\), so
Lemma~\ref{lem:uniform-positive-overlap-drift} applies. Using also
\(\eta_{\tau_-}\le\varepsilon_*\), we get
\[
\dot\rho_{\tau_-}
\ge
a_\rho c_-(1-c_-^2)-b_\rho\varepsilon_*
\ge
\frac12a_\rho c_-(1-c_-^2)
>
0.
\]
Thus \(\rho_t\) cannot cross downward through \(c_-\) at \(\tau_-\). Similarly,
if \(\chi_{\tau_-}=c_-\), then
\[
\dot\chi_{\tau_-}
\ge
a_\chi c_-(1-c_-^2)-b_\chi\varepsilon_*
\ge
\frac12a_\chi c_-(1-c_-^2)
>
0,
\]
so \(\chi_t\) cannot cross downward through \(c_-\). This contradicts the
definition of \(\tau_-\). Therefore,
\[
\rho_t,\chi_t\ge c_-
\qquad
\text{for all }
t\in[t_{\rm esc},t_{\rm esc}+C_{\rm in}].
\]

We now show that \(\rho_t\) reaches \(s_{\rm loc}\) within time \(T_\rho\) and
then cannot fall below \(s_{\rm loc}\) during the window. Define
\[
\tau_\rho:=\inf\{t\ge t_{\rm esc}:\rho_t\ge s_{\rm loc}\}.
\]
If \(\rho_{t_{\rm esc}}\ge s_{\rm loc}\), then
\(\tau_\rho=t_{\rm esc}\). Otherwise, before \(\rho_t\) reaches
\(s_{\rm loc}\), we have \(\rho_t<s_{\rm loc}\). On this interval, the
rectangle invariance gives \(\chi_t\ge c_-\), and the previous step gives
\(\eta_t\le\varepsilon_*\). Hence Lemma~\ref{lem:uniform-positive-overlap-drift}
implies
\[
\dot\rho_t
\ge
a_\rho c_-(1-\rho_t^2)-b_\rho\varepsilon_*.
\]
Since \(\rho_t<s_{\rm loc}\),
\[
1-\rho_t^2\ge 1-s_{\rm loc}^2=m_{\rm loc}.
\]
Therefore
\[
\dot\rho_t
\ge
a_\rho c_-m_{\rm loc}-b_\rho\varepsilon_*
\ge
\frac12a_\rho c_-m_{\rm loc}
=
v_\rho.
\]
Thus \(\rho_t\) reaches \(s_{\rm loc}\) no later than
\[
t_{\rm esc}+T_\rho.
\]

Moreover, after \(\rho_t\) reaches \(s_{\rm loc}\), it cannot cross downward
through \(s_{\rm loc}\) inside the same window. Indeed, at a hypothetical first
downward crossing through \(s_{\rm loc}\), we would have
\[
\rho_t=s_{\rm loc},
\qquad
\chi_t\ge c_-,
\qquad
\eta_t\le\varepsilon_*.
\]
Therefore
\[
\dot\rho_t
\ge
a_\rho c_-(1-s_{\rm loc}^2)-b_\rho\varepsilon_*
\ge
v_\rho
>
0,
\]
which is incompatible with a first downward crossing.

The same argument applies to \(\chi_t\). Define
\[
\tau_\chi:=\inf\{t\ge t_{\rm esc}:\chi_t\ge s_{\rm loc}\}.
\]
If \(\chi_{t_{\rm esc}}\ge s_{\rm loc}\), then
\(\tau_\chi=t_{\rm esc}\). Otherwise, before \(\chi_t\) reaches
\(s_{\rm loc}\), we have \(\chi_t<s_{\rm loc}\), while
\(\rho_t\ge c_-\) and \(\eta_t\le\varepsilon_*\). Hence
\[
\dot\chi_t
\ge
a_\chi c_-(1-\chi_t^2)-b_\chi\varepsilon_*
\ge
a_\chi c_-m_{\rm loc}-b_\chi\varepsilon_*
\ge
v_\chi.
\]
Thus \(\chi_t\) reaches \(s_{\rm loc}\) no later than
\[
t_{\rm esc}+T_\chi.
\]
Once it reaches \(s_{\rm loc}\), it cannot cross downward through
\(s_{\rm loc}\), because at such a crossing the same estimate gives
\[
\dot\chi_t\ge v_\chi>0.
\]

Now set
\[
t_{\rm loc}:=\max\{\tau_\rho,\tau_\chi\}.
\]
By the previous estimates,
\[
t_{\rm loc}
\le
t_{\rm esc}+\max\{T_\rho,T_\chi\}
\le
t_{\rm esc}+C_{\rm in},
\]
and by the no-downcrossing arguments,
\[
\rho_{t_{\rm loc}}\ge s_{\rm loc},
\qquad
\chi_{t_{\rm loc}}\ge s_{\rm loc}.
\]

It remains to convert the overlap bounds into a product-manifold distance
bound. Since
\[
\rho_{t_{\rm loc}}
=
\langle \beta_{t_{\rm loc}},\beta^*\rangle
\ge
s_{\rm loc}
=
\cos(r_{\rm loc}/4),
\]
and since \(\arccos\) is decreasing on \([-1,1]\),
\[
\operatorname{dist}_{\mathbb S^{d-1}}(\beta_{t_{\rm loc}},\beta^*)
=
\arccos(\rho_{t_{\rm loc}})
\le
\frac{r_{\rm loc}}4.
\]
Similarly,
\[
\operatorname{dist}_{\mathbb S^{d-1}}(\theta_{t_{\rm loc}},\theta^*)
=
\arccos(\chi_{t_{\rm loc}})
\le
\frac{r_{\rm loc}}4.
\]
Using the product metric on
\(\mathcal M=\mathbb S^{d-1}\times\mathbb S^{d-1}\), we get
\[
\operatorname{dist}_{\mathcal M}(y_{t_{\rm loc}},y^*)^2
=
\operatorname{dist}_{\mathbb S^{d-1}}(\theta_{t_{\rm loc}},\theta^*)^2
+
\operatorname{dist}_{\mathbb S^{d-1}}(\beta_{t_{\rm loc}},\beta^*)^2.
\]
Therefore,
\[
\operatorname{dist}_{\mathcal M}(y_{t_{\rm loc}},y^*)^2
\le
\left(\frac{r_{\rm loc}}4\right)^2
+
\left(\frac{r_{\rm loc}}4\right)^2
=
\frac{r_{\rm loc}^2}{8}.
\]

Finally, since Lemma~\ref{lem:popescape} gives $t_{\rm esc}\le C_{\rm esc}\log d,$
we have
\[
t_{\rm loc}
\le
C_{\rm esc}\log d+C_{\rm in}.
\]
For \(d\ge e\),
\[
C_{\rm in}\le C_{\rm in}\log d,
\]
and hence
\[
t_{\rm loc}
\le
(C_{\rm esc}+C_{\rm in})\log d.
\]
Thus the last claim holds with
\[
C_{\rm loc}:=C_{\rm esc}+C_{\rm in}.
\]
The proof is complete.
\end{proof}

\paragraph{Curvature near the teacher.}
Here, we prove the local curvature estimate used for the contraction phase.
The comparison with the Gaussian model is needed only at the teacher; the
extension to a small cap uses Hessian continuity of the true loss.

\begin{lemma}[Gaussian local strong convexity]\label{lem:gaussian-local-strong-convexity}
Let $\mathcal M:=\mathbb S^{d-1}\times\mathbb S^{d-1}, y=(\theta,\beta), y^*=(\theta^*,\beta^*).$ Let \(L_G\) be the Gaussian population loss
\[
L_G(\theta,\beta)
=
\mathbb E\left[
\left(
\psi(g\cdot\theta)-\psi(g^*\cdot\theta^*)
\right)^2
\right],
\]
where \((g,g^*)\in\mathbb R^d\times\mathbb R^d\) is centered Gaussian with
$ \mathbb E[g_ig_j]=\delta_{ij}, \mathbb E[g_i^*g_j^*]=\delta_{ij}, \mathbb E[g_i g_j^*]=\kappa(\rho)\delta_{ij},$
with
\[
\rho:=\beta\cdot\beta^*,
\qquad
\kappa(\rho):=\sum_{\ell\ge1}a_\ell^2\rho^\ell.
\]
Assume \(\|\phi\|_{L^2(\gamma)}=1\), so \(\kappa(1)=1\), and assume the information-exponent-one conditions $\alpha_1\neq0, a_1\neq0,$
where
\[
\psi=\sum_{k\ge0}\alpha_k h_k,
\qquad
\phi=\sum_{\ell\ge1}a_\ell h_\ell.
\]
Assume also enough regularity so that the function below is \(C^2\) in a neighborhood of \((1,1)\).

Then there exist constants \(r_G>0\) and \(\mu_G>0\), independent of \(d\), such that for every
\[
(\theta,\beta)\in\mathcal U_{r_G}
:=
\left\{
(\theta,\beta):
\operatorname{dist}_{\mathbb S}(\theta,\theta^*)^2+
\operatorname{dist}_{\mathbb S}(\beta,\beta^*)^2
\le r_G^2
\right\},
\]
and every product tangent vector $\xi=(u,v)\in T_\theta\mathbb S^{d-1}\times T_\beta\mathbb S^{d-1},$ one has
\begin{align}
    \operatorname{Hess}_{\mathcal M}L_G(\theta,\beta)[\xi,\xi] \ge \mu_G\left(\|u\|^2+\|v\|^2\right).
\end{align}
\end{lemma}

\begin{proof} In the Gaussian model, the scalar pair $(g\cdot\theta,\ g^*\cdot\theta^*)$ is centered Gaussian with unit variances and correlation $m=\kappa(\rho)\chi.$
Therefore, using the Hermite expansion of \(\psi\),
\[
\mathbb E[\psi(g\cdot\theta)\psi(g^*\cdot\theta^*)]
=
\sum_{k\ge0}\alpha_k^2(\kappa(\rho)\chi)^k.
\]
Since both marginals have variance one,
\[
\mathbb E[\psi(g\cdot\theta)^2]
=
\mathbb E[\psi(g^*\cdot\theta^*)^2]
=
\sum_{k\ge0}\alpha_k^2.
\]
Hence
\begin{align}\label{eq:L_G}
L_G(\theta,\beta)
=
2\sum_{k\ge0}\alpha_k^2
\left[
1-(\kappa(\rho)\chi)^k
\right],
\end{align}
with \(k=0\) term is zero. We define 
\[
\Phi(\chi,\rho)
:=
2\sum_{k\ge1}\alpha_k^2
\left[
1-(\kappa(\rho)\chi)^k
\right]=L_G(\theta,\beta).
\]
Now we first compute the Hessian at the planted point $(\theta,\beta)=(\theta^*,\beta^*):$ we have $\chi=1, \rho=1, \kappa(1)=1.$ Also $\Phi_\chi(1,1)=-2\sum_{k\ge1}k\alpha_k^2,$ and $\Phi_\rho(1,1)=-2\kappa'(1)\sum_{k\ge1}k\alpha_k^2.$
Define
\[
\lambda_\theta
:=
-\Phi_\chi(1,1)
=
2\sum_{k\ge1}k\alpha_k^2,
\]
and
\[
\lambda_\beta
:=
-\Phi_\rho(1,1)
=
2\kappa'(1)\sum_{k\ge1}k\alpha_k^2.
\]
The information-exponent-one condition \(\alpha_1\neq0\) implies $\lambda_\theta>0.$
Moreover,
\[
\kappa'(1)=\sum_{\ell\ge1}\ell a_\ell^2>0,
\]
because \(a_1\neq0\). Hence $\lambda_\beta>0.$

Now let $u\in T_{\theta^*}\mathbb S^{d-1}, v\in T_{\beta^*}\mathbb S^{d-1}$ and let \((\theta_t,\beta_t)\) be the product geodesic with
\[
(\theta_0,\beta_0)=(\theta^*,\beta^*),
\qquad
(\dot\theta_0,\dot\beta_0)=(u,v).
\]
Then
\[
\chi(t):=\theta_t\cdot\theta^*
=
\cos(t\|u\|),
\]
and
\[
\rho(t):=\beta_t\cdot\beta^*
=
\cos(t\|v\|).
\]
Thus
\[
\chi'(0)=0,
\qquad
\chi''(0)=-\|u\|^2,
\]
and
\[
\rho'(0)=0,
\qquad
\rho''(0)=-\|v\|^2.
\]
Since the curve is a product geodesic, the Riemannian Hessian is computed by
the second derivative along this curve:
\[
\operatorname{Hess}_{\mathcal M}L_G(y^*)[(u,v),(u,v)]
=
\left.\frac{d^2}{dt^2}\right|_{t=0}
\Phi(\chi(t),\rho(t)).
\]
By the chain rule and the fact that \(\chi'(0)=\rho'(0)=0\),
\[
\left.\frac{d^2}{dt^2}\right|_{t=0}
\Phi(\chi(t),\rho(t))
=
\Phi_\chi(1,1)\chi''(0)
+
\Phi_\rho(1,1)\rho''(0).
\]
Therefore
\[
\operatorname{Hess}_{\mathcal M}L_G(y^*)[(u,v),(u,v)]
=
\lambda_\theta\|u\|^2+\lambda_\beta\|v\|^2.
\]
Consequently, with
\[
\lambda_0:=\min\{\lambda_\theta,\lambda_\beta\}>0,
\]
we have
\[
\operatorname{Hess}_{\mathcal M}L_G(y^*)[(u,v),(u,v)]
\ge
\lambda_0(\|u\|^2+\|v\|^2).
\]

It remains to extend this lower bound to a small product cap around \(y^*\).
We do this explicitly.

Let
\[
(\theta,\beta)\in\mathcal M,
\qquad
u\in T_\theta\mathbb S^{d-1},
\qquad
v\in T_\beta\mathbb S^{d-1}.
\]
Let \((\theta_t,\beta_t)\) be the product geodesic starting from
\((\theta,\beta)\) with initial velocity \((u,v)\).
Then
\[
\chi'(0)=u\cdot\theta^*,
\qquad
\chi''(0)=-\chi\|u\|^2,
\]
and
\[
\rho'(0)=v\cdot\beta^*,
\qquad
\rho''(0)=-\rho\|v\|^2.
\]
Therefore
\begin{align}
    & \operatorname{Hess}_{\mathcal M}L_G(\theta,\beta)[(u,v),(u,v)] \\ 
&=
\Phi_{\chi\chi}(\chi,\rho)(u\cdot\theta^*)^2
+
2\Phi_{\chi\rho}(\chi,\rho)(u\cdot\theta^*)(v\cdot\beta^*)
+
\Phi_{\rho\rho}(\chi,\rho)(v\cdot\beta^*)^2
\notag\\
&\quad -\chi\Phi_\chi(\chi,\rho)\|u\|^2
-\rho\Phi_\rho(\chi,\rho)\|v\|^2.
\end{align}

We now estimate the terms in this expression. Choose a small neighborhood
\[
\mathcal N:=\{(\chi,\rho):|\chi-1|\le r_0,\ |\rho-1|\le r_0\}
\]
on which all second derivatives of \(\Phi\) are bounded:
\[
|\Phi_{\chi\chi}|+|\Phi_{\chi\rho}|+|\Phi_{\rho\rho}|
\le M.
\]
By continuity of \(\Phi_\chi,\Phi_\rho\), after shrinking \(r_0\) if necessary, we also have
\[
-\chi\Phi_\chi(\chi,\rho)\ge \frac{\lambda_\theta}{2},
\qquad
-\rho\Phi_\rho(\chi,\rho)\ge \frac{\lambda_\beta}{2}
\]
throughout \(\mathcal N\).

Now suppose
\[
\operatorname{dist}_{\mathbb S}(\theta,\theta^*)\le r,
\qquad
\operatorname{dist}_{\mathbb S}(\beta,\beta^*)\le r,
\]
with \(r>0\) small enough so that \((\chi,\rho)\in\mathcal N\). Since
\(u\perp\theta\),
\[
|u\cdot\theta^*|
=
|u\cdot P_{\theta^\perp}\theta^*|
\le
\|u\|\|P_{\theta^\perp}\theta^*\|
=
\|u\|\sqrt{1-\chi^2}
\le
r\|u\|.
\]
Similarly,
\[
|v\cdot\beta^*|\le r\|v\|.
\]
Hence the second-derivative terms are bounded in absolute value by
\[
M r^2\|u\|^2
+
2M r^2\|u\|\|v\|
+
M r^2\|v\|^2
\le
2M r^2(\|u\|^2+\|v\|^2).
\]
Therefore,
\[
\operatorname{Hess}_{\mathcal M}L_G(\theta,\beta)[(u,v),(u,v)]
\ge
\frac{\lambda_\theta}{2}\|u\|^2
+
\frac{\lambda_\beta}{2}\|v\|^2
-
2M r^2(\|u\|^2+\|v\|^2).
\]
Choose \(r_G>0\) small enough that $2M r_G^2\le \frac14\min\{\lambda_\theta,\lambda_\beta\}.$ Then, for every \(r\le r_G\),
\[
\operatorname{Hess}_{\mathcal M}L_G(\theta,\beta)[(u,v),(u,v)]
\ge
\frac14\min\{\lambda_\theta,\lambda_\beta\}
(\|u\|^2+\|v\|^2).
\]
Thus the lemma holds with
\[
\mu_G:=\frac14\min\{\lambda_\theta,\lambda_\beta\}>0.
\]
\end{proof}

\begin{lemma}[Teacher Hessian block formula]
\label{lem:teacher-hessian-formula}
Let \(y^*:=(\theta^*,\beta^*)\). Define
\[
G_i:=x^{(i)}\cdot \beta^*,
\qquad
S:=\sum_{i=1}^d \theta_i^*\phi(G_i).
\]
For \(u\in T_{\theta^*}\mathbb S^{d-1}\), set
\[
A_u:=\sum_{i=1}^d u_i\phi(G_i).
\]
Then, for
\(u_1,u_2\in T_{\theta^*}\mathbb S^{d-1}\) and
\(v_1,v_2\in T_{\beta^*}\mathbb S^{d-1}\), the Hessian blocks of \(L\) at
the teacher satisfy
\begin{align}
H_{L,\theta\theta}(y^*)[u_1,u_2]
&=
2\mathbb E\!\left[
\psi'(S)^2 A_{u_1}A_{u_2}
\right],\label{eq:thetatheta}\\
H_{L,\theta\beta}(y^*)[u,v]&=0,\label{eq:thetabeta}\\
H_{L,\beta\beta}(y^*)[v_1,v_2]
&=2\langle v_1,v_2\rangle
\sum_{i=1}^d
(\theta_i^*)^2
\mathbb E\!\left[
\psi'(S)^2\phi'(G_i)^2
\right].\label{eq:betabeta}
\end{align}

Consequently, for every
\((u,v)\in T_{\theta^*}\mathbb S^{d-1}\times T_{\beta^*}\mathbb S^{d-1}\),
\begin{align}
\operatorname{Hess}_{\mathcal M}L(y^*)[(u,v),(u,v)]
=
2\mathbb E\!\left[
\psi'(S)^2A_u^2
\right]
+
2\|v\|^2
\sum_{i=1}^d
(\theta_i^*)^2
\mathbb E\!\left[
\psi'(S)^2\phi'(G_i)^2
\right].
\end{align}
\end{lemma}

\begin{proof}
We first compute the full quadratic form at the teacher and then polarize.

Let \(y_t=(\theta_t,\beta_t)\) be any product geodesic on
\(\mathcal M=\mathbb S^{d-1}\times\mathbb S^{d-1}\) with
\[
y_0=y^*,
\qquad
\dot y_0=(u,v).
\]
Write
\[
f_t(x):=f_{\beta_t,\theta_t}(x),
\qquad
f_*(x):=f_{\beta^*,\theta^*}(x).
\]
Since \(f_0=f_*\) pointwise,
\[
L(y_t)=\mathbb E[(f_t-f_*)^2],
\]
and therefore
\[
\frac{d^2}{dt^2}\Big|_{t=0}L(y_t)
=
2\mathbb E[\dot f_0^2]
+
2\mathbb E[(f_0-f_*)\ddot f_0]
=
2\mathbb E[\dot f_0^2].
\]
Because \(y_t\) is a geodesic, this second derivative is exactly the
Riemannian Hessian:
\[
\operatorname{Hess}_{\mathcal M}L(y^*)[(u,v),(u,v)]
=
2\mathbb E[\dot f_0^2].
\]

Now compute \(\dot f_0\). At \(y^*\),
\[
f_t(x)
=
\psi\!\left(
\sum_{i=1}^d \theta_{t,i}\phi(x^{(i)}\cdot\beta_t)
\right).
\]
Thus
\[
\dot f_0
=
\psi'(S)
\left[
\sum_{i=1}^d u_i\phi(G_i)
+
\sum_{i=1}^d
\theta_i^*\phi'(G_i)(x^{(i)}\cdot v)
\right].
\]
Define, for the proof,
\[
B_v:=\sum_{i=1}^d
\theta_i^*\phi'(G_i)H_i(v),
\qquad
H_i(v):=x^{(i)}\cdot v.
\]
Then
\[
\dot f_0=\psi'(S)(A_u+B_v),
\]
and hence
\begin{align}\label{eq:Hess-1}
\operatorname{Hess}_{\mathcal M}L(y^*)[(u,v),(u,v)]
=
2\mathbb E\!\left[
\psi'(S)^2(A_u+B_v)^2
\right].
\end{align}

Since the right-hand side of \eqref{eq:Hess-1} is a quadratic form in \((u,v)\), its
associated symmetric bilinear form is obtained by polarization:
\begin{align}\label{eq:Hess-2}
\operatorname{Hess}_{\mathcal M}L(y^*)[(u_1,v_1),(u_2,v_2)]
=
2\mathbb E\!\left[
\psi'(S)^2
(A_{u_1}+B_{v_1})(A_{u_2}+B_{v_2})
\right].
\end{align}

Now, we derive the three Hessian blocks: First, take \(v_1=v_2=0\) in \eqref{eq:Hess-2}. This gives
\[
H_{L,\theta\theta}(y^*)[u_1,u_2]
=
2\mathbb E\!\left[
\psi'(S)^2A_{u_1}A_{u_2}
\right],
\]
which proves \eqref{eq:thetatheta}, the \(\theta\theta\)-block.

Second, take \((u_1,v_1)=(u,0)\) and \((u_2,v_2)=(0,v)\). Then
\[
H_{L,\theta\beta}(y^*)[u,v]
=
2\mathbb E\!\left[
\psi'(S)^2A_uB_v
\right].
\]
We show this expectation is zero. Since \(v\in T_{\beta^*}\mathbb S^{d-1}\),
we have \(v\perp \beta^*\). For each \(i\), the pair
\[
G_i=x^{(i)}\cdot\beta^*,
\qquad
H_i(v)=x^{(i)}\cdot v
\]
is jointly Gaussian with covariance
\[
\mathbb E[G_iH_i(v)]
=
\langle \beta^*,v\rangle
=
0.
\]
Hence \(G_i\) and \(H_i(v)\) are independent. Moreover, conditional on
\((G_1,\ldots,G_d)\), the variables \(H_i(v)\) are centered Gaussian and
independent across \(i\). Therefore
\[
\mathbb E[B_v\mid G_1,\ldots,G_d]
=
\sum_{i=1}^d
\theta_i^*\phi'(G_i)
\mathbb E[H_i(v)\mid G_1,\ldots,G_d]
=
0.
\]
Since \(\psi'(S)^2A_u\) is measurable with respect to
\(\sigma(G_1,\ldots,G_d)\), we obtain
\[
\mathbb E[\psi'(S)^2A_uB_v]
=
\mathbb E\!\left[
\psi'(S)^2A_u
\mathbb E[B_v\mid G_1,\ldots,G_d]
\right]
=
0.
\]
Thus we have \eqref{eq:thetabeta}, the \(\theta\beta\)-block:
\[
H_{L,\theta\beta}(y^*)[u,v]=0.
\]

Finally, take \(u_1=u_2=0\) in \eqref{eq:Hess-2}. Then
\[
H_{L,\beta\beta}(y^*)[v_1,v_2]
=
2\mathbb E\!\left[
\psi'(S)^2B_{v_1}B_{v_2}
\right].
\]
Conditioning again on \(G:=(G_1,\ldots,G_d)\),
\[
B_{v_1}B_{v_2}
=
\sum_{i,j=1}^d
\theta_i^*\theta_j^*
\phi'(G_i)\phi'(G_j)
H_i(v_1)H_j(v_2).
\]
For \(i\neq j\), conditional independence and centering give
\[
\mathbb E[H_i(v_1)H_j(v_2)\mid G]=0.
\]
For \(i=j\), since \(v_1,v_2\perp\beta^*\),
\[
\mathbb E[H_i(v_1)H_i(v_2)\mid G]
=
\langle v_1,v_2\rangle.
\]
Hence
\[
\mathbb E[B_{v_1}B_{v_2}\mid G]
=
\langle v_1,v_2\rangle
\sum_{i=1}^d
(\theta_i^*)^2\phi'(G_i)^2.
\]
Because \(\psi'(S)^2\) is \(G\)-measurable,
\[
\mathbb E[\psi'(S)^2B_{v_1}B_{v_2}]
=
\langle v_1,v_2\rangle
\sum_{i=1}^d
(\theta_i^*)^2
\mathbb E\!\left[
\psi'(S)^2\phi'(G_i)^2
\right].
\]
Therefore we have the final \(\beta\beta\)-block in \eqref{eq:betabeta}:
\[
H_{L,\beta\beta}(y^*)[v_1,v_2]
=
2\langle v_1,v_2\rangle
\sum_{i=1}^d
(\theta_i^*)^2
\mathbb E\!\left[
\psi'(S)^2\phi'(G_i)^2
\right].
\]

Combining the \(\theta\theta\)-block, the vanishing mixed block, and the
\(\beta\beta\)-block with \(u_1=u_2=u\), \(v_1=v_2=v\), gives the final
quadratic formula. This completes the proof.
\end{proof}

\paragraph{Stein inputs for the teacher Hessian.}
We next state the two blockwise teacher-level comparisons used below. Their
proofs are deferred to Subsection~\ref{subsec:stein-bounds}.

\begin{lemma}[Teacher Hessian comparison]\label{lem:teacher-full-hessian-comparison}
Let
\[
A_\psi:=\mathbb E[\psi'(G)^2],
\qquad
B_\phi:=\mathbb E[\phi'(G)^2]=\kappa'(1),
\qquad
\lambda_\theta:=2A_\psi,
\qquad
\lambda_\beta:=2A_\psi B_\phi.
\]
Under the regularity assumptions of
Lemma~\ref{lem:teacher-theta-hessian-comparison} and
Lemma~\ref{lem:teacher-beta-hessian-comparison}, there exists
\(C_*>0\), independent of \(d\), such that
\begin{align}
\sup_{\|\xi\|=1}
\left|
\operatorname{Hess}_{\mathcal M}L(y^*)[\xi,\xi]
-
\operatorname{Hess}_{\mathcal M}L_G(y^*)[\xi,\xi]
\right|
\le
C_*\|\theta^*\|_\infty.
\end{align}
Moreover,
\begin{align}
\operatorname{Hess}_{\mathcal M}L_G(y^*)[(u,v),(u,v)]
=
\lambda_\theta\|u\|^2+\lambda_\beta\|v\|^2.
\end{align}
\end{lemma}

\begin{proof}
Let
\[
\xi=(u,v)\in T_{y^*}\mathcal M,
\qquad
\|\xi\|^2=\|u\|^2+\|v\|^2.
\]

First, the Gaussian teacher Hessian formula follows from the explicit
Gaussian loss:
\[
L_G(\theta,\beta)
=
2\sum_{k\ge1}\alpha_k^2
\left(1-(\kappa(\rho)\chi)^k\right).
\]
At \(y^*\), tangent geodesics satisfy
\[
\chi''(0)=-\|u\|^2,
\qquad
\rho''(0)=-\|v\|^2,
\qquad
\chi'(0)=\rho'(0)=0.
\]
Therefore
\[
\operatorname{Hess}_{\mathcal M}L_G(y^*)[(u,v),(u,v)]
=
\lambda_\theta\|u\|^2+\lambda_\beta\|v\|^2,
\]
with
\[
\lambda_\theta=2\sum_{k\ge1}k\alpha_k^2=2A_\psi,
\qquad
\lambda_\beta
=
2\kappa'(1)\sum_{k\ge1}k\alpha_k^2
=
2A_\psi B_\phi.
\]

For the true loss, decompose the Hessian into blocks. By
Lemma~\ref{lem:teacher-hessian-formula},
\[
H_{L,\theta\beta}(y^*)=0.
\]
The same geodesic computation above gives
\[
H_{L_G,\theta\beta}(y^*)=0.
\]
For the \(\theta\theta\)-block, Lemma~\ref{lem:teacher-theta-hessian-comparison}
at \((\theta,\beta)=(\theta^*,\beta^*)\) gives
\[
\left|
H_{L,\theta\theta}(y^*)[u,u]
-
H_{L_G,\theta\theta}(y^*)[u,u]
\right|
\le
C_\theta\|\theta^*\|_\infty\|u\|^2.
\]
For the \(\beta\beta\)-block,
Lemma~\ref{lem:teacher-beta-hessian-comparison} gives
\[
\left|
H_{L,\beta\beta}(y^*)[v,v]
-
H_{L_G,\beta\beta}(y^*)[v,v]
\right|
\le
C_\beta\|\theta^*\|_\infty\|v\|^2.
\]
Combining the three block estimates,
\[
\left|
H_L(y^*)[\xi,\xi]
-
H_{L_G}(y^*)[\xi,\xi]
\right|
\le
(C_\theta+C_\beta)\|\theta^*\|_\infty
(\|u\|^2+\|v\|^2).
\]
Taking \(\|\xi\|=1\) and setting \(C_*:=C_\theta+C_\beta\) proves the claim.
\end{proof}

\begin{lemma}[Uniform Hessian continuity near the teacher]
\label{lem:uniform-hessian-continuity}
Assume \(\psi\in C_b^3(\mathbb R)\), and assume that \(\phi\in C^3(\mathbb R)\)
satisfies
\[
\max_{0\le j\le 3}
\mathbb E\left[
(1+|G|)^{12}|\phi^{(j)}(G)|^{12}
\right]
<\infty,
\qquad G\sim N(0,1).
\]
Assume also
\[
\mathbb E[\phi(G)]=0,
\qquad
\mathbb E[\phi(G)^2]=1.
\]
Then there exist constants \(r_0>0\) and \(C_3>0\), independent of \(d\), such
that for every \(y\in\mathcal U_{r_0}\),
\begin{align}
\left\|
P_{y\to y^*}
\operatorname{Hess}_{\mathcal M}L(y)
P_{y^*\to y}
-
\operatorname{Hess}_{\mathcal M}L(y^*)
\right\|_{\rm op}
\le
C_3\operatorname{dist}_{\mathcal M}(y,y^*).
\end{align}
Here \(P_{y\to y^*}\) denotes parallel transport along the minimizing product
geodesic from \(y\) to \(y^*\).
\end{lemma}

\begin{proof}
It is enough to prove that the covariant derivative of the Hessian is
uniformly bounded on a fixed product cap. Indeed, let
\(\gamma:[0,\ell]\to\mathcal M\) be the minimizing product geodesic from
\(y^*\) to \(y\), parametrized by arclength, so that
\[
\ell=\operatorname{dist}_{\mathcal M}(y,y^*).
\]
Let \(\xi_0\in T_{y^*}\mathcal M\) be a unit vector, and let \(\xi_t\) be its
parallel transport along \(\gamma\). Then
\[
\frac{d}{dt}
\operatorname{Hess}_{\mathcal M}L(\gamma(t))[\xi_t,\xi_t]
=
(\nabla\operatorname{Hess}_{\mathcal M}L)(\gamma(t))
[\dot\gamma(t),\xi_t,\xi_t].
\]
Thus, if
\[
\sup_{\gamma(t)\in\mathcal U_{r_0}}
\sup_{\|\zeta\|=\|\xi\|=1}
\left|
(\nabla\operatorname{Hess}_{\mathcal M}L)(\gamma(t))
[\zeta,\xi,\xi]
\right|
\le C_3,
\]
then integration along \(\gamma\) gives the desired transported-Hessian
Lipschitz bound.

We prove this uniform bound. Write
\[
s_y(x):=
\sum_{i=1}^d\theta_i\phi(x^{(i)}\cdot\beta),
\qquad
s_*(x):=
\sum_{i=1}^d\theta_i^*\phi(x^{(i)}\cdot\beta^*),
\]
so that
\[
L(y)=\mathbb E\left[
\bigl(\psi(s_y(x))-\psi(s_*(x))\bigr)^2
\right].
\]
Let \(y_t=(\theta_t,\beta_t)\) be any unit-speed product geodesic contained in
\(\mathcal U_{r_0}\), and set
\[
s_t:=s_{y_t}(x).
\]
We first record the basic moment estimate
\begin{equation}\label{eq:uniform-st-moments}
\sup_{t}
\left(
\|\dot s_t\|_{L^6}
+
\|\ddot s_t\|_{L^3}
+
\|s_t^{(3)}\|_{L^1}
\right)
\le C,
\end{equation}
where \(C\) is independent of \(d\) and of the chosen unit-speed product
geodesic.

To prove \eqref{eq:uniform-st-moments}, observe that
\[
s_t=\sum_{i=1}^d \theta_{t,i}\phi(U_{t,i}),
\qquad
U_{t,i}:=x^{(i)}\cdot\beta_t.
\]
Along a unit-speed spherical geodesic,
\[
\|\dot\theta_t\|+\|\ddot\theta_t\|+\|\theta_t^{(3)}\|\le C,
\qquad
\|\dot\beta_t\|+\|\ddot\beta_t\|+\|\beta_t^{(3)}\|\le C.
\]
Moreover,
\[
\dot U_{t,i}=x^{(i)}\cdot\dot\beta_t,\qquad
\ddot U_{t,i}=x^{(i)}\cdot\ddot\beta_t,\qquad
U_{t,i}^{(3)}=x^{(i)}\cdot\beta_t^{(3)}.
\]
The variables coming from distinct blocks \(i\) are independent, and each
block has moments bounded uniformly in \(t\). Differentiating
\[
\theta_{t,i}\phi(U_{t,i})
\]
up to order three gives a finite sum of terms whose coefficient vectors have
uniformly bounded \(\ell_2\)-norm and whose block variables have uniformly
bounded moments by the assumptions on \(\phi\).

The only point that needs care is centering. For each \(i\),
\[
\mathbb E[\phi(U_{t,i})]=\mathbb E[\phi(G)]=0
\]
for every \(t\), because \(U_{t,i}\sim N(0,1)\). Differentiating this identity
in \(t\) shows that the differentiated block variables appearing in
\(\dot s_t,\ddot s_t,s_t^{(3)}\) are centered. Therefore Rosenthal's inequality
applies to each differentiated sum and gives the bounds in
\eqref{eq:uniform-st-moments}, uniformly in \(d\).

Now define
\[
R_t:=\psi(s_t)-\psi(s_*).
\]
Since \(\psi\) and its derivatives up to order three are bounded,
\[
\dot R_t=\psi'(s_t)\dot s_t,
\]
\[
\ddot R_t=\psi''(s_t)\dot s_t^2+\psi'(s_t)\ddot s_t,
\]
and
\[
R_t^{(3)}
=
\psi^{(3)}(s_t)\dot s_t^3
+
3\psi''(s_t)\dot s_t\ddot s_t
+
\psi'(s_t)s_t^{(3)}.
\]
Since \(L(y_t)=\mathbb E[R_t^2]\),
\[
\frac{d^3}{dt^3}L(y_t)
=
2\mathbb E\left[
3\dot R_t\ddot R_t+R_tR_t^{(3)}
\right].
\]
Using the boundedness of \(\psi,\psi',\psi'',\psi^{(3)}\), together with
\eqref{eq:uniform-st-moments} and Hölder's inequality, we obtain
\[
\sup_t
\left|
\frac{d^3}{dt^3}L(y_t)
\right|
\le C.
\]
The same argument, applied to geodesic variations and then polarized, gives
\[
\sup_{y\in\mathcal U_{r_0}}
\sup_{\|\zeta\|=\|\xi\|=1}
\left|
(\nabla\operatorname{Hess}_{\mathcal M}L)(y)[\zeta,\xi,\xi]
\right|
\le C.
\]
Equivalently, the transported Hessian is \(C\)-Lipschitz along every
minimizing product geodesic in \(\mathcal U_{r_0}\). Integrating this bound
along the geodesic from \(y^*\) to \(y\) gives
\[
\left\|
P_{y\to y^*}
\operatorname{Hess}_{\mathcal M}L(y)
P_{y^*\to y}
-
\operatorname{Hess}_{\mathcal M}L(y^*)
\right\|_{\rm op}
\le
C\operatorname{dist}_{\mathcal M}(y,y^*).
\]
This proves the lemma, with \(C_3=C\).
\end{proof}

\begin{lemma}[True local product strong convexity]
\label{lem:true-local-product-strong-convexity}
Assume the teacher Hessian comparison of
Lemma~\ref{lem:teacher-full-hessian-comparison} and the uniform Hessian
continuity of Lemma~\ref{lem:uniform-hessian-continuity}. On the random-spherical delocalization event of Lemma~\ref{lem:random-spherical-delocalization}, so that
\[
\|\theta^*\|_\infty
\le
C_K\sqrt{\frac{\log d}{d}},
\]
there exist constants \(r_{\rm loc}>0\) and \(\mu_{\rm loc}>0\), independent of
\(d\), such that for all sufficiently large \(d\) and every
\(y\in\mathcal U_{r_{\rm loc}}\),
\begin{align}
    \operatorname{Hess}_{\mathcal M}L(y)\succeq \mu_{\rm loc}I.
\end{align}
\end{lemma}

\begin{proof}
Let
\[
\lambda_0:=\min\{\lambda_\theta,\lambda_\beta\}
=
2A_\psi\min\{1,B_\phi\}.
\]
By the information-exponent-one assumptions, \(A_\psi>0\) and \(B_\phi>0\),
so \(\lambda_0>0\).

By Lemma~\ref{lem:teacher-full-hessian-comparison}, for every
\(\xi\in T_{y^*}\mathcal M\),
\[
\operatorname{Hess}_{\mathcal M}L(y^*)[\xi,\xi]
\ge
\operatorname{Hess}_{\mathcal M}L_G(y^*)[\xi,\xi]
-
C_*\|\theta^*\|_\infty\|\xi\|^2.
\]
Since by Lemma~\ref{lem:gaussian-local-strong-convexity}
\[
\operatorname{Hess}_{\mathcal M}L_G(y^*)[\xi,\xi]
\ge
\lambda_0\|\xi\|^2,
\]
we obtain
\[
\operatorname{Hess}_{\mathcal M}L(y^*)[\xi,\xi]
\ge
\left(\lambda_0-C_*\|\theta^*\|_\infty\right)\|\xi\|^2.
\]
By Lemma~\ref{lem:random-spherical-delocalization},
\[
\|\theta^*\|_\infty
\le
C_K\sqrt{\frac{\log d}{d}}.
\]
Hence, for all sufficiently large \(d\),
\[
C_*\|\theta^*\|_\infty
\le
C_*C_K\sqrt{\frac{\log d}{d}}
\le
\frac{\lambda_0}{4}.
\]
Hence
\[
\operatorname{Hess}_{\mathcal M}L(y^*)\succeq \frac{3\lambda_0}{4}I.
\]

Now choose
\[
r_{\rm loc}
\le
r_0
\qquad\text{and}\qquad
C_3r_{\rm loc}\le \frac{\lambda_0}{4}.
\]
Let \(y\in\mathcal U_{r_{\rm loc}}\), and let
\(\xi\in T_y\mathcal M\) be a unit tangent vector. Parallel transport \(\xi\)
to \(T_{y^*}\mathcal M\), writing
\[
\widetilde \xi:=P_{y\to y^*}\xi.
\]
Then \(\|\widetilde\xi\|=1\). By Lemma~\ref{lem:uniform-hessian-continuity},
\[
\operatorname{Hess}_{\mathcal M}L(y)[\xi,\xi]
\ge
\operatorname{Hess}_{\mathcal M}L(y^*)[\widetilde\xi,\widetilde\xi]
-
C_3\operatorname{dist}_{\mathcal M}(y,y^*).
\]
Since \(y\in\mathcal U_{r_{\rm loc}}\),
\[
\operatorname{dist}_{\mathcal M}(y,y^*)\le r_{\rm loc}.
\]
Therefore
\[
\operatorname{Hess}_{\mathcal M}L(y)[\xi,\xi]
\ge
\frac{3\lambda_0}{4}
-
C_3r_{\rm loc}
\ge
\frac{\lambda_0}{2}.
\]
Thus the lemma holds with
\[
\mu_{\rm loc}:=\frac{\lambda_0}{2}.
\]
\end{proof}

\paragraph{Strong Recovery Dynamics.}

\begin{lemma}[Local contraction under product strong convexity]
\label{lem:product-contraction}
Let \(\mathcal M=\mathbb S^{d-1}\times\mathbb S^{d-1}\) equipped with the product
Riemannian metric, and let \(y^*=(\theta^*,\beta^*)\in\mathcal M\). Fix
\(r\in(0,\pi/2)\) and suppose \(L:\mathcal M\to\mathbb R\) is \(C^2\) and
satisfies
\[
\operatorname{Hess}_{\mathcal M}L(y)\succeq \mu I
\qquad\text{for all }
y\in\mathcal U_r:=\{y\in\mathcal M:\operatorname{dist}_{\mathcal M}(y,y^*)\le r\},
\]
for some \(\mu>0\) (Lemma~\ref{lem:true-local-product-strong-convexity},  with
\(r=r_{\rm loc}\) and \(\mu=\mu_{\rm loc}\)), and that \(y^*\) is a critical point of \(L\), i.e.,
\(\nabla_{\mathcal M}L(y^*)=0\).

Let \(y_t\) be the population gradient flow
\[
\dot y_t=-\nabla_{\mathcal M}L(y_t),
\qquad
y_{t_0}\in\mathcal M,
\qquad
\operatorname{dist}_{\mathcal M}(y_{t_0},y^*)\le r/2.
\]
Then \(y_t\in\mathcal U_r\) for all \(t\ge t_0\), and
\begin{align}
\operatorname{dist}_{\mathcal M}(y_t,y^*)^2
\le
e^{-2\mu(t-t_0)}
\operatorname{dist}_{\mathcal M}(y_{t_0},y^*)^2.
\end{align}
\end{lemma}

\begin{proof}
Write
\[
D_t:=\operatorname{dist}_{\mathcal M}(y_t,y^*),
\qquad
E_t:=\frac12D_t^2.
\]

Since \(r<\pi/2\), every point \(y=(\theta,\beta)\in\mathcal U_r\) satisfies
\[
\operatorname{dist}_{\mathbb S}(\theta,\theta^*)<\pi/2,
\qquad
\operatorname{dist}_{\mathbb S}(\beta,\beta^*)<\pi/2.
\]
Thus neither component is antipodal to its target. Hence the minimizing
geodesic from \(y\) to \(y^*\) is unique and is obtained by taking the
componentwise minimizing geodesics on the two spheres. In particular,
\(\operatorname{Log}_y(y^*)\) is well-defined, and
\[
\|\operatorname{Log}_y(y^*)\|
=
\operatorname{dist}_{\mathcal M}(y,y^*).
\]
Moreover, if \(\gamma:[0,1]\to\mathcal M\) is this minimizing product geodesic
from \(y\) to \(y^*\), then
\[
\operatorname{dist}_{\mathcal M}(\gamma(s),y^*)
=
(1-s)\operatorname{dist}_{\mathcal M}(y,y^*),
\]
so \(\gamma(s)\in\mathcal U_r\) whenever \(y\in\mathcal U_r\).

Define the exit time
\[
\tau:=\inf\{t\ge t_0:\ D_t>r\}.
\]
For \(t\in[t_0,\tau)\), we have \(y_t\in\mathcal U_r\). We will prove
\[
\frac{d}{dt}E_t\le -2\mu E_t
\qquad
\text{for all }t\in[t_0,\tau).
\]
Once this is shown, Gronwall gives
\[
E_t\le e^{-2\mu(t-t_0)}E_{t_0}
\qquad
\text{for all }t\in[t_0,\tau).
\]
Since \(D_{t_0}\le r/2\), this implies \(D_t\le r/2\) on
\([t_0,\tau)\). If \(\tau<\infty\), then by continuity \(D_\tau\le r/2<r\),
contradicting the definition of \(\tau\). Hence \(\tau=\infty\), and the
trajectory stays in \(\mathcal U_r\) for all \(t\ge t_0\).

It remains to prove the differential inequality. Fix \(t<\tau\), and let
\(\gamma:[0,1]\to\mathcal M\) be the minimizing geodesic from \(y_t\) to
\(y^*\), parametrized with constant speed:
\[
\gamma(0)=y_t,\qquad
\gamma(1)=y^*,\qquad
\|\gamma'(s)\|\equiv D_t.
\]
Then
\[
\gamma'(0)=\operatorname{Log}_{y_t}(y^*).
\]
Set
\[
\varphi(s):=L(\gamma(s)).
\]
Since \(\gamma\) is a geodesic,
\[
\varphi''(s)
=
\operatorname{Hess}_{\mathcal M}L(\gamma(s))[\gamma'(s),\gamma'(s)].
\]
The geodesic remains in \(\mathcal U_r\), so the Hessian lower bound gives
\[
\varphi''(s)\ge \mu\|\gamma'(s)\|^2=\mu D_t^2.
\]
Integrating from \(0\) to \(1\),
\[
\varphi'(1)-\varphi'(0)\ge \mu D_t^2.
\]
Because \(y^*\) is critical,
\[
\varphi'(1)
=
\langle\nabla_{\mathcal M}L(y^*),\gamma'(1)\rangle
=
0.
\]
Therefore
\[
\varphi'(0)
\le
-\mu D_t^2.
\]
But
\[
\varphi'(0)
=
\left\langle
\nabla_{\mathcal M}L(y_t),
\operatorname{Log}_{y_t}(y^*)
\right\rangle.
\]
Thus
\[
\left\langle
\nabla_{\mathcal M}L(y_t),
\operatorname{Log}_{y_t}(y^*)
\right\rangle
\le
-\mu D_t^2.
\]

Finally, the first-variation formula for the squared distance gives
\[
\frac{d}{dt}E_t
=
-\left\langle
\dot y_t,
\operatorname{Log}_{y_t}(y^*)
\right\rangle.
\]
Since \(\dot y_t=-\nabla_{\mathcal M}L(y_t)\),
\[
\frac{d}{dt}E_t
=
\left\langle
\nabla_{\mathcal M}L(y_t),
\operatorname{Log}_{y_t}(y^*)
\right\rangle
\le
-\mu D_t^2
=
-2\mu E_t.
\]
This proves the desired differential inequality, and hence the contraction
estimate.
\end{proof}

\begin{theorem}[Continuous-time strong recovery]
\label{thm:continuous-weak-to-strong-recovery}
Assume the weak-recovery conclusion of
Lemma~\ref{lem:popescape}, together with the assumptions of
Lemmas~\ref{lem:uniform-positive-overlap-drift} and
\ref{lemma:secondlayercontrol}. Assume also the regularity and incoherence
assumptions under which Lemmas~\ref{lem:teacher-full-hessian-comparison},
\ref{lem:uniform-hessian-continuity},
\ref{lem:true-local-product-strong-convexity}, and
\ref{lem:product-contraction} hold.

Let $ y_t:=(\theta_t,\beta_t)\in
\mathcal M:=\mathbb S^{d-1}\times\mathbb S^{d-1}, \quad
y^*:=(\theta^*,\beta^*),$ and write $R_t:=\operatorname{dist}_{\mathcal M}(y_t,y^*).$
Then there exist constants $r_{\rm loc}>0,\quad \mu_{\rm loc}>0,\quad C_{\rm loc}<\infty,$
independent of \(d\), such that for all sufficiently large \(d\), on the
favorable initialization event of \ref{ass:favorable-initialization}, there exists a
time $t_{\rm loc}\le C_{\rm loc}\log d$
for which
\begin{align}\label{eq:R-t-loc}
R_{t_{\rm loc}}^2\le
\frac{r_{\rm loc}^2}{8}.
\end{align}
Moreover, for every \(t\ge t_{\rm loc}\),
\[
R_t^2
\le
e^{-2\mu_{\rm loc}(t-t_{\rm loc})}
R_{t_{\rm loc}}^2.
\]
Consequently,
\[
(1-\chi_t)+(1-\rho_t)
\le
\frac{r_{\rm loc}^2}{16}
e^{-2\mu_{\rm loc}(t-t_{\rm loc})}.
\]
In particular, there exists \(\varepsilon_0>0\), independent of \(d\), such
that for every \(\varepsilon\in(0,\varepsilon_0)\),
\begin{align}\label{eq:strong-recovery-achieved}
    \min\{\chi_t,\rho_t\}\ge 1-\varepsilon
\end{align}
whenever
\[
t\ge C\log d+C\log(1/\varepsilon),
\]
where \(C<\infty\) is independent of \(d\) and \(\varepsilon\).
\end{theorem}

\begin{proof}
By Lemma~\ref{lem:true-local-product-strong-convexity}, there exist constants
\(r_{\rm loc}>0\) and \(\mu_{\rm loc}>0\), independent of \(d\), such that for
all sufficiently large \(d\),
\[
\operatorname{Hess}_{\mathcal M}L(y)\succeq \mu_{\rm loc}I
\qquad
\text{for every }
y\in\mathcal U_{r_{\rm loc}}
:=
\{y\in\mathcal M:
\operatorname{dist}_{\mathcal M}(y,y^*)\le r_{\rm loc}\}.
\]
After shrinking \(r_{\rm loc}\), if necessary, we may assume
\[
r_{\rm loc}<\frac{\pi}{2}.
\]
The same Hessian lower bound remains valid on the smaller neighborhood.

Apply Lemma~\ref{lem:entry-local-cap-after-corrected-weak} with this value of
\(r_{\rm loc}\). On the favorable event of Lemma~\ref{lem:popescape}, it gives
a time \(t_{\rm loc}\) satisfying
\[
t_{\rm loc}\le t_{\rm esc}+C_{\rm in},
\]
where \(C_{\rm in}<\infty\) is independent of \(d\), and
\[
\rho_{t_{\rm loc}}\ge \cos(r_{\rm loc}/4),
\qquad
\chi_{t_{\rm loc}}\ge \cos(r_{\rm loc}/4).
\]
Equivalently
\[R_{t_{\rm loc}}^2
=
\operatorname{dist}_{\mathcal M}(y_{t_{\rm loc}},y^*)^2
\le
\frac{r_{\rm loc}^2}{8},
\]
which proves \eqref{eq:R-t-loc}.
Since Lemma~\ref{lem:popescape} gives
\[
t_{\rm esc}\le C_{\rm esc}\log d,
\]
we have, for \(d\ge e\),
\[
t_{\rm loc}
\le
C_{\rm esc}\log d+C_{\rm in}
\le
(C_{\rm esc}+C_{\rm in})\log d.
\]
Thus the claimed bound holds with
\[
C_{\rm loc}:=C_{\rm esc}+C_{\rm in}.
\]

Next, observe that \(y^*\) is a critical point of \(L\). Indeed, at the planted
pair the residual vanishes pointwise, so
\[
L(y^*)=0.
\]
Since \(L\ge 0\), \(y^*\) is a global minimizer of \(L\), and therefore
\[
\nabla_{\mathcal M}L(y^*)=0.
\]

The entry estimate gives
\[
R_{t_{\rm loc}}
\le
\frac{r_{\rm loc}}{2\sqrt2}
<
\frac{r_{\rm loc}}2.
\]
Hence we may apply Lemma~\ref{lem:product-contraction} with
\[
r=r_{\rm loc},
\qquad
\mu=\mu_{\rm loc},
\qquad
t_0=t_{\rm loc}.
\]
It follows that the trajectory remains in
\(\mathcal U_{r_{\rm loc}}\) for all \(t\ge t_{\rm loc}\), and that
\[
R_t^2
\le
e^{-2\mu_{\rm loc}(t-t_{\rm loc})}
R_{t_{\rm loc}}^2
\qquad
\text{for every }t\ge t_{\rm loc}.
\]

We now convert the product-manifold distance decay into overlap decay. Since $\chi_t= \langle \theta_t,\theta^*\rangle =\cos\left( \operatorname{dist}_{\mathbb S^{d-1}}(\theta_t,\theta^*)\right),$ and $\rho_t=\langle \beta_t,\beta^*\rangle=\cos\left(\operatorname{dist}_{\mathbb S^{d-1}}(\beta_t,\beta^*)\right),$ the elementary bound $1-\cos x\le \frac{x^2}{2}
\quad (x\ge0)$ gives
\[
1-\chi_t
\le
\frac12
\operatorname{dist}_{\mathbb S^{d-1}}(\theta_t,\theta^*)^2,
\]
and
\[
1-\rho_t
\le
\frac12
\operatorname{dist}_{\mathbb S^{d-1}}(\beta_t,\beta^*)^2.
\]
Summing these two estimates and using the product metric, we have
\[
(1-\chi_t)+(1-\rho_t)
\le
\frac12
\operatorname{dist}_{\mathcal M}(y_t,y^*)^2
=
\frac12 R_t^2.
\]
Therefore,
\[
(1-\chi_t)+(1-\rho_t)
\le
\frac12
e^{-2\mu_{\rm loc}(t-t_{\rm loc})}
R_{t_{\rm loc}}^2.
\]
Using \ref{eq:R-t-loc} that we proved, we obtain
\[
(1-\chi_t)+(1-\rho_t)
\le
\frac{r_{\rm loc}^2}{16}
e^{-2\mu_{\rm loc}(t-t_{\rm loc})}.
\]

Finally, set
\[
\varepsilon_0:=\frac{r_{\rm loc}^2}{16}.
\]
Let \(\varepsilon\in(0,\varepsilon_0)\). If
\[
\frac{r_{\rm loc}^2}{16}
e^{-2\mu_{\rm loc}(t-t_{\rm loc})}
\le
\varepsilon,
\]
then
\[
(1-\chi_t)+(1-\rho_t)\le \varepsilon.
\]
Since both summands are nonnegative, this implies
\[
1-\chi_t\le \varepsilon,
\qquad
1-\rho_t\le \varepsilon,
\]
and hence we arrive at \eqref{eq:strong-recovery-achieved}:
\[
\min\{\chi_t,\rho_t\}\ge 1-\varepsilon.
\]

The displayed exponential condition is equivalent to
\[
t-t_{\rm loc}
\ge
\frac{1}{2\mu_{\rm loc}}
\log\frac{r_{\rm loc}^2}{16\varepsilon}.
\]
Using $t_{\rm loc}\le C_{\rm loc}\log d,$ we conclude that
\[
\min\{\chi_t,\rho_t\}\ge 1-\varepsilon
\]
whenever
\[
t
\ge
C_{\rm loc}\log d
+
\frac{1}{2\mu_{\rm loc}}
\log\frac{r_{\rm loc}^2}{16\varepsilon}.
\]
Since \(r_{\rm loc}\) and \(\mu_{\rm loc}\) are fixed constants independent of
\(d\) and \(\varepsilon\), this is of the form
\[
t\ge C\log d+C\log(1/\varepsilon),
\]
for a constant \(C<\infty\) independent of \(d\) and \(\varepsilon\). This
proves the theorem.
\end{proof}

\section{Proof of Discrete-Time Online SGD Dynamics}\label{sec:dis-time-sgd-dynamics-proof}
We now move to
analyze the learning dynamics under SGD discretization.

\subsection{Weak Recovery Setups}

\ignore{\paragraph{Necessary bounds} Note that Lemmas~\ref{lem:sgd-tech-bounds}, \ref{lem:sgd-martingale-Z}, and~\ref{lem:sgd-discretization} are mainly inherited
from~\cite{bruna2023singleindexmodelsgaussian}.}

\begin{assumption}[SGD regime and initialization for weak recovery]\label{ass:sgd-regime}
Fix $0<\varepsilon\le \varepsilon_*$. Consider online \emph{blockwise} spherical SGD on
$\mathbb S^{d-1}\times\mathbb S^{d-1}$ follows \eqref{eq:eachlayerupdate}. The horizon and step-size follows
\[
T\delta \le C_T\log d,
\qquad
\delta \le \frac{c_0\varepsilon}{d\log d},
\qquad
0<\varepsilon\le \varepsilon_* .
\]
for an absolute constant $c_0>0$, so we have the weak recovery horizon $T:=\lceil C_T\delta^{-1}\log d \rceil \in\mathbb N.$ Initialization condition \ref{ass:favorable-initialization} also follows.
\end{assumption}

We start the discretizations analysis by deriving the one-step evolution of the overlaps under SGD:

\begin{lemma}[One-step evolution of the overlaps under SGD]\label{lem:sgd-one-step}
Under Assumption~\ref{ass:sgd-regime}, on the event
\begin{align}\label{event:taylor}
E_t^{(T)}:=\{\delta\|g_t^{(\beta)}\|\le 1/2,\ \delta\|g_t^{(\theta)}\|\le 1/2\},
\end{align}
there exist \(\mathcal F_t\)-measurable predictable drifts \(F_t^{(\rho)},F_t^{(\chi)}\), martingale increments \(M_t^{(\rho)},M_t^{(\chi)}\), and remainders \(R_t^{(\rho)},R_t^{(\chi)}\) such that for all $t$, \begin{align}
    \rho_{t+1}=\rho_t + \delta F_t^{(\rho)}+\delta M_t^{(\rho)} +R_t^{(\rho)}\label{eq:rho-one-step}\\
    \chi_{t+1}=\chi_t + \delta F_t^{(\chi)}+\delta M_t^{(\chi)} +R_t^{(\chi)}\label{eq:chi-one-step}
\end{align} 
where $\E[M_t^{(\cdot)}\mid\mathcal F_t]=0$ for each component, $R_t^{(\rho)}=O(\delta^3\|g_t^{(\beta)}\|^3),$ and $R_t^{(\chi)}=O(\delta^3\|g_t^{(\theta)}\|^3).$
\end{lemma}

\begin{proof}
Let $g_t^{(\beta)}:=\nabla_{\beta}^Sl(\beta_t,\theta_t;x_t)$ and $g_t^{(\theta)}:=\nabla_{\theta}^Sl(\beta_t,\theta_t;x_t)$ be the spherical gradients taken with respect $\beta$ and $\theta$ respectively, which implies $g_t^{(\beta)}\perp \beta_t$ and $g_t^{(\theta)}\perp \theta_t$.From \eqref{eq:eachlayerupdate} and under $\mathcal E_t,$ we have the normalized update as
\begin{align}
    \beta_{t+1} 
    &=\frac{\beta_t-\delta g_t^{(\beta)}}{\|\beta_t-\delta g_t^{(\beta)}\|} =\beta_t-\delta g_t^{(\beta)}-\frac{\delta^2}{2}\|g_t^{(\beta)}\|^2\beta_t+O(\delta^3\|g_t^{(\beta)}\|^3) \\
\theta_{t+1} 
    &=\frac{\theta_t-\delta g_t^{(\theta)}}{\|\theta_t-\delta g_t^{(\theta)}\|} =\theta_t-\delta g_t^{(\theta)}-\frac{\delta^2}{2}\|g_t^{(\theta)}\|^2\theta_t+O(\delta^3\|g_t^{(\theta)}\|^3)
\end{align}
Hence taking overlap with the teacher direction $\beta^*$ and $\theta^*,$ respectively, we have 
\begin{align} \label{eq:firstlayerprojection}
        \langle \beta_{t+1},\beta^* \rangle &=\langle \beta_t,\beta^* \rangle - \delta \langle g_t^{(\beta)}, \beta^* \rangle - \frac{\delta^2}{2}\|g_t^{(\beta)}\|^2\langle \beta_t,\beta^*\rangle+O(\delta^3\|g_t^{(\beta)}\|^3)\\
    \langle \theta_{t+1},\theta^* \rangle 
    &=\langle \theta_t,\theta^* \rangle - \delta \langle g_t^{(\theta)}, \theta^* \rangle - \frac{\delta^2}{2}\|g_t^{(\theta)}\|^2\langle \theta_t,\theta^*\rangle+O(\delta^3\|g_t^{(\theta)}\|^3), \label{eq:secondlayerprojection}
\end{align}
where $\rho_t:=\langle \beta_t,\beta^* \rangle$ and $\chi_t:=\langle \theta_t,\theta^* \rangle$ Now, we decompose the middle two terms involving spherical gradient as the population drift plus the martingale increment, namely,
\textcolor{black}{
\begin{align}
\langle g_t^{(\beta)},\beta^*\rangle + \frac{\delta}{2}\|g_t^{(\beta)}\|^2\rho_t &=\underbrace{\mathbb{E}\left[\langle g_t^{(\beta)},\beta^*\rangle + \frac{\delta}{2}\|g_t^{(\beta)}\|^2\rho_t|\mathcal{F}_t\right]}_{-F_t^{(\rho)}}\notag\\
&\quad +\underbrace{\left(\begin{aligned} &\langle g_t^{(\beta)},\beta^*\rangle + \frac{\delta}{2}\|g_t^{(\beta)}\|^2\rho_t\\[-2pt] &\quad -\mathbb{E}[\langle g_t^{(\beta)},\beta^*\rangle + \frac{\delta}{2}\|g_t^{(\beta)}\|^2\rho_t|\mathcal{F}_t]\end{aligned}\right)}_{-M_t^{(\rho)}}\label{eq:generalizedgradientprojection}\\
\langle g_t^{(\theta)},\theta^*\rangle + \frac{\delta}{2}\|g_t^{(\theta)}\|^2\chi_t &=\underbrace{\mathbb{E}\left[\langle g_t^{(\theta)},\theta^*\rangle + \frac{\delta}{2}\|g_t^{(\theta)}\|^2\chi_t|\mathcal{F}_t\right]}_{-F_t^{(\chi)}}\notag\\
&\quad +\underbrace{\left(\begin{aligned} &\langle g_t^{(\theta)},\theta^*\rangle + \frac{\delta}{2}\|g_t^{(\theta)}\|^2\chi_t\\[-2pt] &\quad -\mathbb{E}[\langle g_t^{(\theta)},\theta^*\rangle + \frac{\delta}{2}\|g_t^{(\theta)}\|^2\chi_t|\mathcal{F}_t]\end{aligned}\right)}_{-M_t^{(\chi)}}.
\end{align}}
Therefore, we can write \eqref{eq:firstlayerprojection} and \eqref{eq:secondlayerprojection} respectively as
\begin{align} 
    \rho_{t+1}=\rho_t + \delta F_t^{(\rho)}+\delta M_t^{(\rho)} +O(\delta^3\|g_t^{(\beta)}\|^3)\\
    \chi_{t+1}=\chi_t + \delta F_t^{(\chi)}+\delta M_t^{(\chi)} +O(\delta^3\|g_t^{(\theta)}\|^3)
\end{align} 
Defining $R_t^{(\rho)}:=O(\delta^3\|g_t^{(\beta)}\|^3)$ and $R_t^{(\chi)}:=O(\delta^3\|g_t^{(\theta)}\|^3)$ gives the claimed expression.
\end{proof}

Next we provide a class of concentration inequalities to better control the iteration of $(\rho_t,\chi_t)$ that has been used in studying the aggregate SGD dynamics, on top of \eqref{eq:rho-one-step} and \eqref{eq:chi-one-step}. In particular, Lemma~\ref{lem:sgd-martingale-Z} controls the aggregate martingale term, and Lemma~\ref{lem:sgd-discretization} controls the aggregate discretization remainder term. Besides, As the one-step expansion in Lemma~\ref{lem:sgd-one-step} relies on the event $\mathcal E_t,$ we provide Lemma~\ref{lem:uniform-taylor-weak} to show that $\mathcal E_t$ is true with high probability. All of these inequalities are obtained via using the technical bounds specified in Lemma~\ref{lem:sgd-tech-bounds} in \ref{subsec:gradient-bounds}.

\subsubsection{Taylor Expansion Error Control}

\begin{lemma}[Uniform Taylor event on the Weak-Recovery Horizon]\label{lem:uniform-taylor-weak}
    Assume the moment bounds of Lemma~\ref{lem:sgd-tech-bounds} and following Assumption~\ref{ass:sgd-regime} suppose
\[
T\delta \le C_T\log d,
\qquad
\delta \le \frac{c_0\varepsilon}{d\log d},
\qquad
0<\varepsilon\le \varepsilon_* .
\]
Then there is a constant $K_{\mathrm{Tay}}$, depending only on model constants, \(C_T,c_0,\varepsilon_*\), such that for all sufficiently large $d$, the Taylor Event $E_T^{(T)}$ satisfies: 
\begin{align}\label{eq:taylor-bound}
\mathbb P((E_T^{(T)})^c)
\le
32K T\delta^4 d^2
\le
K_{\mathrm{Tay}}\frac{\varepsilon^3}{d\log^2 d}
\le
K_{\mathrm{Tay}}\varepsilon.
\end{align}
Moreover, on $E_T^{(T)}$, the one-step expansions \eqref{eq:rho-one-step} and \eqref{eq:chi-one-step} in Lemma~\ref{lem:sgd-one-step} are valid simultaneously for every $t<T$, with remainders
\[
|R_t^{(\rho)}|
\le
C\delta^3\|g_t^{(\beta)}\|^3,
\qquad
|R_t^{(\chi)}|
\le
C\delta^3\|g_t^{(\theta)}\|^3.
\]
\end{lemma}
\begin{proof}
By Lemma~\ref{lem:sgd-tech-bounds}, conditionally on $\mathcal F_t,$ we have
\[
\mathbb E\left[\|g_t^{(\beta)}\|^4\mid \mathcal F_t\right]\le Kd^2,
\qquad
\mathbb E\left[\|g_t^{(\theta)}\|^4\mid \mathcal F_t\right]\le Kd^2.
\]
Hence, by Markov’s inequality,
$
\mathbb P\left(\delta\|g_t^{(\beta)}\|>\frac12\mid \mathcal F_t\right)
\le
16\delta^4
\mathbb E\left[\|g_t^{(\beta)}\|^4\mid \mathcal F_t\right]
\le
16K\delta^4 d^2.
$
The same bound holds for the \(\theta\)-block. Therefore
\[
\mathbb P({E_t^{(T)}}^c)
\le
32K\delta^4d^2.
\]
Union bounding over $t=0,\dots,T-1,$
\[
\mathbb P({E_T^{(T)}}^c)
\le
32KT\delta^4d^2.
\]
Since $T\delta\le C_T\log d,$
\[
T\delta^4d^2
=(T\delta)\delta^3d^2
\le
C_T\delta^3d^2\log d.
\]
Using $\delta \le \frac{c_0\varepsilon}{d\log d},$
we get
\[
\delta^3d^2\log d
\le
\frac{c_0^3\varepsilon^3}{d\log^2 d}.
\]
Thus
\[
\mathbb P({E_T^{(T)}}^c)
\le
K_{\mathrm{Tay}}\frac{\varepsilon^3}{d\log^2 d}
\le
K_{\mathrm{Tay}}\varepsilon
\]
for all sufficiently large $d$. This proves the lemma.
\end{proof}

\subsubsection{Martingale and Discretization Error Control for \texorpdfstring{$Z_t$}{Zt}}

The Martingale Error for $Z_t$ depends on this intermediate Corollary, derived from Lemma~\ref{lem:sgd-tech-bounds}.

\begin{corollary}[Negligible remainder for drift] \label{cor:mtgandhotremainders}
With the martingale increments $M_t^{(\rho)}$ and $M_t^{(\chi)}$ we defined in Lemma~\ref{lem:sgd-one-step} and adopting the technical gradient bounds in Lemma~\ref{lem:sgd-tech-bounds}, we have 
\begin{align}
\E\!\left[(M_t^{(\rho)})^2\mid \mathcal F_t\right]\le K_M^{\rho},
\
\E\!\left[(M_t^{(\chi)})^2\mid \mathcal F_t\right]\le K_M^{\chi},
\end{align}
where $K_M^{\rho},K_M^{\chi}$ absolute for large $d$ under step size $\delta\leq c_0\varepsilon/{d\log d}$.

\end{corollary}
\begin{proof}
    Let $A_t^{(\rho)}=\langle g_t^{(\beta)}, \beta^*\rangle +\frac{\delta}{2}\|g_t^{(\beta)}\|^2\rho_t,$ then the martingale term writes as $M_t^{(\rho)}=-(A_t^{(\rho)}-\mathbb{E}[A_t^{(\rho)}\mid\mathcal{F}_t]),$ so the conditional second moment of $M_t^{(\rho)}$ can be written as
\begin{align} \label{eq:secondmomentmartingale}
    \mathbb{E}[(M_t^{(\rho)})^2\mid\mathcal{F}_t] 
    &= \mathbb{E}\big[\big(A_t^{(\rho)}-\mathbb{E}[A_t^{(\rho)}\mid \mathcal{F}_t]\big)^2 \mid\mathcal{F}_t\big]
\end{align}
Since we have: 
\begin{align*}
    A_t^{(\rho)}-\mathbb{E}[A_t^{(\rho)}\mid\mathcal{F}_t] 
    &=\big(\langle g_t^{(\beta)}, \beta^*\rangle +\frac{\delta}{2}\|g_t^{(\beta)}\|^2\rho_t\big) \\
    &\quad -\mathbb{E}\big[\langle g_t^{(\beta)}, \beta^*\rangle +\frac{\delta}{2}\|g_t^{(\beta)}\|^2\rho_t\mid \mathcal{F}_t \big]\\
    &= \big(\langle g_t^{(\beta)}, \beta^*\rangle- \mathbb{E}[\langle g_t^{(\beta)}, \beta^*\rangle\mid \mathcal{F}_t]\big) \\
    &\quad +\frac{\delta}{2}\rho_t \big(\|g_t^{(\beta)}\|^2-\mathbb{E}[\|g_t^{(\beta)}\|^2 \mid \mathcal{F}_t] \big) \text{ by regrouping} \\
    &= \langle \Delta g_t^{(\beta)},\beta^*\rangle+\frac{\delta}{2}\rho_t\Delta q_t^{(\beta)},
\end{align*}
where we defined $\Delta g_t^{(\beta)}:=g_t^{(\beta)}-\mathbb{E}[g_t^{(\beta)}\mid\mathcal{F}_t],$ and $\Delta q_t^{(\beta)}:=\|g_t^{(\beta)}\|^2-\mathbb{E}[\|g_t^{(\beta)}\|^2 \mid \mathcal{F}_t],$
\eqref{eq:secondmomentmartingale} can be written as: 
\begin{align}
    \mathbb{E}[(M_t^{(\rho)})^2\mid\mathcal{F}_t]  
    &= \mathbb{E}[\big(\langle \Delta g_t^{(\beta)},\beta^*\rangle+\frac{\delta}{2}\rho_t\Delta q_t^{(\beta)}\big)^2 \mid\mathcal{F}_t]\\
    &= 2 \mathbb{E}[\langle \Delta g_t^{(\beta)},\beta^*\rangle^2\mid\mathcal{F}_t] +\frac{\delta^2}{2}\rho_t^2 \mathbb{E}[(\Delta q_t^{(\beta)})^2] \text{, given }(a+b)^2\leq 2a^2+2b^2 \label{eq:twotermssecondmoment}
\end{align}
By Lemma~\ref{lem:sgd-tech-bounds}, we can bound the first term of \eqref{eq:twotermssecondmoment} by $K$. For the second term of \eqref{eq:twotermssecondmoment}, note that $(\Delta q)^2=\big(\|g_t^{(\beta)}\|^2-\mathbb{E}[\|g_t^{(\beta)}\|^2 \mid \mathcal{F}_t]\big)^2\leq \|g_t^{(\beta)}\|^4,$ we can use the forth moment bound in Lemma~\ref{lem:sgd-tech-bounds} to obtain the $Kd^2$ bound, so 
\begin{align}
    \mathbb{E}[(M_t^{(\rho)})^2\mid\mathcal{F}_t] 
    &\leq 2K+\frac{(\delta\rho_t)^2}{2}Kd^2\\
    &\leq 2K+\frac{\delta^2}{2}Kd^2.
\end{align}
Given our step size $\delta\leq \frac{c_0\varepsilon}{d \log d}$ by Assumption~\ref{ass:sgd-regime}, $\frac{\delta^2}{2}Kd^2\leq \frac{c_0^2 \varepsilon^2}{2\log d}K$, enabling us to write the bound 
\begin{align}
    \mathbb{E}[(M_t^{(\rho)})^2\mid\mathcal{F}_t] \leq K_M^{\rho},
\end{align}
for an absolute $K_M^{\rho}$ that depends only on model constants and $c_0,\varepsilon_*.$ The exact argument for $\chi$ gives 
\begin{align}
    \mathbb{E}[(M_t^{(\chi)})^2\mid\mathcal{F}_t] \leq K_M^{\chi},
\end{align}
for an absolute $K_M^{\chi}$.
\end{proof}

\begin{lemma}[Martingale inequality for $Z_t$]\label{lem:sgd-martingale-Z}
Let
\[
\mathcal M_t^{(Z)} := \sum_{k=0}^{t-1}
\big(M_k^{(\rho)} + M_k^{(\chi)}\big),
\qquad t\ge 0.
\]
Under Assumption~\ref{ass:sgd-regime} and the bounds of Corollary~\ref{cor:mtgandhotremainders},
for all $T\in\mathbb N$ and all $\lambda>0$,
\begin{align}
\mathbb P\Big(
\sup_{0\le t\le T}
\big|\delta \mathcal M_t^{(Z)}\big|
\;\ge\; \lambda
\Big)
\;\le\;
\frac{K T \delta^2}{\lambda^2},
\end{align}
for some constant $K>0$ depending only on the model. 
\end{lemma}

\begin{proof}
Let $D_k:=M_k^{(\rho)}+M_k^{(\chi)}$, so that
\(
\mathcal M_t^{(Z)}=\sum_{k=0}^{t-1}D_k
\)
is a martingale with respect to $(\mathcal F_t)$.
By Doob's $L^2$ maximal inequality,
\begin{equation}\label{eq:doob}
\mathbb E\Big[\sup_{0\le t\le T}\big|\mathcal M_t^{(Z)}\big|^2\Big]
\le 4\,\mathbb E\big[\big|\mathcal M_T^{(Z)}\big|^2\big].
\end{equation}
Moreover, using $(a+b)^2\le 2a^2+2b^2$ and Corollary~\ref{cor:mtgandhotremainders},
\[
\mathbb E[D_k^2\mid \mathcal F_k]
\le 2\mathbb E[(M_k^{(\rho)})^2\mid\mathcal F_k]
   +2\mathbb E[(M_k^{(\chi)})^2\mid\mathcal F_k]
\le 4K.
\]
Since martingale increments are orthogonal in $L^2$,
\[
\mathbb E\big[\big|\mathcal M_T^{(Z)}\big|^2\big]
= \sum_{k=0}^{T-1}\mathbb E[D_k^2]
\le \sum_{k=0}^{T-1} 4K
=4KT.
\]
Combining with~\eqref{eq:doob} and applying Markov's inequality yields
\begin{align*}
\mathbb P\Big(\sup_{0\le t\le T}\big|\delta \mathcal M_t^{(Z)}\big|\ge \lambda\Big)
&\le \frac{\delta^2}{\lambda^2}\,
\mathbb E\Big[\sup_{0\le t\le T}\big|\mathcal M_t^{(Z)}\big|^2\Big]\\
&\le \frac{\delta^2}{\lambda^2}\cdot 4\,\mathbb E\big[\big|\mathcal M_T^{(Z)}\big|^2\big]
\le \frac{16KT\delta^2}{\lambda^2}.
\end{align*}
Absorbing the numerical factor into the constant completes the proof.
\end{proof}

\begin{remark}With $\delta\leq \frac{c_0\varepsilon}{d\log d}$ and $T:=\lceil C_T\delta^{-1}\log d \rceil$ and picking $\lambda = C_0/\sqrt{d}$, we have \[
\mathbb P\Big(
\sup_{0\le t\le T}
\big|\delta \mathcal M_t^{(Z)}\big|
\;\ge\; \frac{C_0}{\sqrt{d}}
\Big)
\;\le\;\frac{KT\delta^2}{\lambda^2}
=\frac{KT\delta^2d}{C_0^2}
\le
\frac{KC_T\delta d\log d}{C_0^2}
\le
\frac{KC_Tc_0}{C_0^2}\varepsilon.
\]
\end{remark}
\begin{lemma}[Discretization terms for $Z_t$]\label{lem:sgd-discretization}
Under Assumption~\ref{ass:sgd-regime} and Lemma~\ref{lem:sgd-tech-bounds}, there exists $K>0$ such that for all $T\in\mathbb N$ and all $\lambda>0$,
\[
    \mathbb P\left(
    E_T^{(T)}
    \cap
    \left\{
    \sup_{t\le T}|\mathcal R_t^{(Z)}|\ge \lambda
    \right\}
    \right)
    \le
    \frac{KT\delta^3 d^{3/2}}{\lambda}.
\]
Consequently, under the assumptions of Lemma~\ref{lem:uniform-taylor-weak},
\begin{align}
        \mathbb P\left(
    \sup_{t\le T}|\mathcal R_t^{(Z)}|\ge \lambda
    \right)
    \le
    \mathbb P\left((E_T^{(T)})^c\right)
    +
    \frac{KT\delta^3 d^{3/2}}{\lambda}.
\end{align}
\end{lemma}

\begin{proof}
Let
\[
r_k^{(Z)}:=R_k^{(\rho)}+R_k^{(\chi)},
\qquad
\mathcal R_t^{(Z)}:=\sum_{k=0}^{t-1}r_k^{(Z)}.
\]
On the Taylor event $E_T^{(T)}$, Lemma~\ref{lem:uniform-taylor-weak} gives, simultaneously for all $k<T$,
\[
|R_k^{(\rho)}|
\le
C\delta^3\|g_k^{(\beta)}\|^3,
\qquad
|R_k^{(\chi)}|
\le
C\delta^3\|g_k^{(\theta)}\|^3.
\]
Therefore, on $E_T^{(T)}$,
\[
|r_k^{(Z)}|
\le
C\delta^3
\left(
\|g_k^{(\beta)}\|^3+\|g_k^{(\theta)}\|^3
\right).
\]
Hence
\[
E_T^{(T)}
\cap
\left\{
\sup_{t\le T}|\mathcal R_t^{(Z)}|\ge \lambda
\right\}
\subseteq
\left\{
C\delta^3
\sum_{k<T}
\left(
\|g_k^{(\beta)}\|^3+\|g_k^{(\theta)}\|^3
\right)
\ge \lambda
\right\}.
\]
By Markov's inequality,
\[
\mathbb P\left(
E_T^{(T)}
\cap
\left\{
\sup_{t\le T}|\mathcal R_t^{(Z)}|\ge \lambda
\right\}
\right)
\le
\frac{
C\delta^3
\sum_{k<T}
\mathbb E\left[
\|g_k^{(\beta)}\|^3+\|g_k^{(\theta)}\|^3
\right]
}{\lambda}.
\]
By Hölder's inequality and Lemma~\ref{lem:sgd-tech-bounds},
\[
\mathbb E\left[
\|g_k^{(\beta)}\|^3
\mid \mathcal F_k
\right]
\le
\left(
\mathbb E\left[
\|g_k^{(\beta)}\|^4
\mid \mathcal F_k
\right]
\right)^{3/4}
\le
(Kd^2)^{3/4}
\le
K d^{3/2}.
\]
The same bound holds for $g_k^{(\theta)}$. Therefore
\[
\sum_{k<T}
\mathbb E\left[
\|g_k^{(\beta)}\|^3+\|g_k^{(\theta)}\|^3
\right]
\le
KTd^{3/2}.
\]
Thus
\[
\mathbb P\left(
E_T^{(T)}
\cap
\left\{
\sup_{t\le T}|\mathcal R_t^{(Z)}|\ge \lambda
\right\}
\right)
\le
\frac{KT\delta^3d^{3/2}}{\lambda}.
\]

Finally,
\[
\left\{
\sup_{t\le T}|\mathcal R_t^{(Z)}|\ge \lambda
\right\}
\subseteq
(E_T^{(T)})^c
\cup
\left(
E_T^{(T)}
\cap
\left\{
\sup_{t\le T}|\mathcal R_t^{(Z)}|\ge \lambda
\right\}
\right),
\]
so
\[
\mathbb P\left(
\sup_{t\le T}|\mathcal R_t^{(Z)}|\ge \lambda
\right)
\le
\mathbb P\left((E_T^{(T)})^c\right)
+
\frac{KT\delta^3d^{3/2}}{\lambda}.
\]
The second displayed bound follows from Lemma~\ref{lem:uniform-taylor-weak}.
\end{proof}
\begin{remark} Taking $T\delta\le c_2\log d$, $\delta\le c_0\varepsilon/(d\log d)$, and
$\lambda=C_0/\sqrt d$, Lemma~\ref{lem:sgd-discretization} gives
\[
\frac{KT\delta^3d^{3/2}}{\lambda}
=
\frac{KT\delta^3d^2}{C_0}
\le
\frac{Kc_2}{C_0}\delta^2d^2\log d
\le
\frac{Kc_2c_0^2}{C_0}
\frac{\varepsilon^2}{\log d}.
\]
Together with Lemma~\ref{lem:uniform-taylor-weak},
\[
\mathbb P\left((E_T^{(T)})^c\right)
\le
K_{\mathrm{Tay}}
\frac{\varepsilon^3}{d\log^2 d}.
\]
Hence the discretization contribution is absorbed into the existing $O(\varepsilon)$ good-event failure probability.
\end{remark}

\paragraph{Good Events for $Z_t$} Now, we formalize the definition and the high-likelihood of good events for $Z_t$ iterate: From the two-layer aggregate martingale and discretization remainder for each time step defined as $M_t^{(Z)}:=M_t^{(\rho)}+M_t^{(\chi)}, r_t^{(Z)}:=R_t^{(\rho)}+R_t^{(\chi)},$ we write their time accumulation as 
\[
\mathcal M_t^{(Z)}:=\sum_{k<t}M_k^{(Z)},
\qquad
\mathcal R_t^{(Z)}:=\sum_{k<t}r_k^{(Z)},
\qquad
N_t^{(Z)}:=\delta\mathcal M_t^{(Z)}+\mathcal R_t^{(Z)}.
\]
For a threshold \(\lambda_Z=C_0/\sqrt d\), define
\[
E_T^{(Z,M)}
:=
\left\{
\sup_{t\le T}|\delta\mathcal M_t^{(Z)}|\le \lambda_Z
\right\},
\qquad
E_T^{(Z,R)}
:=
\left\{
\sup_{t\le T}|\mathcal R_t^{(Z)}|\le \lambda_Z
\right\},
\]
and
\begin{align}\label{event:Z}
    E_T^{(Z)}:=E_T^{(Z,M)}\cap E_T^{(Z,R)}.
\end{align}
Then on \(E_T^{(Z)}\),
\begin{align}\label{eq:sup-N-lambda}
    \sup_{t\le T}|N_t^{(Z)}|\le 2\lambda_Z.
\end{align}
By Lemma~\ref{lem:sgd-martingale-Z} and Lemma~\ref{lem:sgd-discretization},
together with Lemma~\ref{lem:uniform-taylor-weak}, we have
\begin{align}\label{eq:Z-union-bound}
    \mathbb P\big((E_T^{(Z)})^c\big) \leq \mathbb P\big((E_T^{(Z,M)})^c\big)+ \mathbb P \big((E_T^{(Z,R)})^c\big) \le K\varepsilon,
\end{align}
for some constant $K$.
\paragraph{Good Events for $D_t$}
Besides, we keep track of the good events for $D_t:=\rho_t-\chi_t:$ for $a\in\{\rho,\chi\},$ we have the individual-layer time accumulation as:
\[
\mathcal M_t^{(a)}:=\sum_{k<t}M_k^{(a)},
\qquad
\mathcal R_t^{(a)}:=\sum_{k<t}R_k^{(a)},
\qquad
N_t^{(a)}:=\delta\mathcal M_t^{(a)}+\mathcal R_t^{(a)}.
\]
We then track their difference $N_t^{(D)}:=N_t^{(\rho)}-N_t^{(\chi)}:$
With parameter $\lambda_\rho=\lambda_\chi:=\frac{C_{\rm ind}}{\sqrt d}, \qquad \lambda_D:=\lambda_\rho+\lambda_\chi,$ we define the good events here as
\begin{align}\label{event:rho-chi}
E_T^{(\rho,\chi)}
:=
\left\{
\sup_{t\le T}|N_t^{(\rho)}|\le 2\lambda_\rho
\right\}
\cap
\left\{
\sup_{t\le T}|N_t^{(\chi)}|\le 2\lambda_\chi
\right\}, 
\end{align}
So on $E_T^{(\rho,\chi)},$
\[
\sup_{t\le T}|N_t^{(D)}|
\le
\sup_{t\le T}|N_t^{(\rho)}|
+
\sup_{t\le T}|N_t^{(\chi)}|
\le
2(\lambda_\rho+\lambda_\chi)
=
2\lambda_D.
\]
By the individual versions of Lemmas~\ref{lem:sgd-martingale-Z} and~\ref{lem:sgd-discretization}, choosing \(C_{\rm ind}\) large enough gives
\begin{align}\label{eq:rho-chi-bound}
\mathbb P\big((E_T^{(\rho,\chi)})^c\big)
\le K\varepsilon.
\end{align}
\ignore{\begin{lemma}[Individual overlap-error control]
\label{lem:individual-overlap-error-control}
Under Assumption~\ref{ass:sgd-regime} and Lemma~\ref{lem:sgd-tech-bounds},
there is a constant \(C_{\rm ind}\) such that, with
\[
\lambda_\rho=\lambda_\chi=\frac{C_{\rm ind}}{\sqrt d},
\]
the event \(E_T^{(\rho,\chi)}\) satisfies
\[
\mathbb P\big((E_T^{(\rho,\chi)})^c\big)\le K\varepsilon.
\]
\end{lemma}}

\subsubsection{Martingale and Discretization Error Control for \texorpdfstring{$\eta_t$}{eta	extsubscript{t}}} \label{subsec:eta}

Deriving the discrete-time evolution $\eta_t$ requires a different stragegy compares to that of $Z_t.$ To start with, we derive the one-step evolution $\eta_t$:

\begin{lemma}[Evolution of $\eta_t$]\label{lem:sgd-evolution-eta}
Work under Assumption~\ref{ass:sgd-regime}. Let
$g_t^{(\theta)}:=\nabla_\theta^{\mathcal S}\ell(\beta_t,\theta_t;x_t)$
so that $g_t^{(\theta)}\perp \theta_t$, and define the mean-field spherical gradient $\bar g_t^{(\theta)}:=\E[g_t^{(\theta)}\mid\mathcal F_t]$ and
$\Delta g_t^{(\theta)}:=g_t^{(\theta)}-\bar g_t^{(\theta)}$.
Then for all $t$, we have
\begin{align}
    \|\theta_t\|_{\infty} &\leq \|\theta_0\|_{\infty}+\delta \sum_{k<t}\| \bar{g}_{k}\|_{\infty}+\max_i\big|\delta\sum_{k<t}\xi_{k,i}\big|+\frac{\delta^2}{2}\sum_{k<t}\|g_k\|_2^2\|\theta_k\|_{\infty}+\frac{\delta^3}{2}\sum_{k<t}\|g_k\|_2^3.
\label{eq:eta-onestep}
\end{align}
\end{lemma}

\begin{proof}
Since $g_t^{(\theta)}\perp \theta_t$ and $\|\theta_t\|_2=1$, we have
$a_t:=1/\|\theta_t-\delta g_t^{(\theta)}\|_2=1/\sqrt{1+\delta^2\|g_t^{(\theta)}\|_2^2}$ and
\[
\theta_{t+1}=a_t(\theta_t-\delta g_t^{(\theta)}).
\]
Define the "telescoping correlation" 
\begin{align}
    r_t=\theta_{t+1}-(\theta_t-\delta g_t^{(\theta)})=(a_t-1)(\theta_t-\delta g_t^{(\theta)}),\label{eq:telescopingcorrelation}
\end{align} 
so as $\theta_{t+1}=\theta_t-\delta g_t^{(\theta)}+r_t,$ under time-cumulation we have 
\[\theta_{t}=\theta_0-\delta \sum_{k<t}g_k^{(\theta)}+\sum_{k<t}r_k.\]
Now we decompose the gradient $g_{k}^{(\theta)}$ as the mean-field and martingale noise $g_{k}^{(\theta)}=\bar{g}_{k}^{(\theta)}+\xi_{k}^{(\theta)},$ where $\bar{g}_{k}:=\mathbb{E}[g_{k}^{(\theta)}\mid\mathcal{F}_k]$ and $\xi_{k}:=g_{k}-\bar{g}_{k}.$ Thus, taking the ${l}_{\infty}$-norm
\begin{align}
    \|\theta_t\|_{\infty}\leq \|\theta_0\|_{\infty}+\delta \sum_{k<t}\| \bar{g}_{k}^{(\theta)}\|_{\infty}+\max_i\big|\delta\sum_{k<t}\xi_{k,i}^{(\theta)}\big|+\sum_{k<t}\|r_k^{(\theta)}\|_{\infty}
\end{align}
To bound $\|r_k^{(\theta)}\|_{\infty},$ taking $l_{\infty}$-norm of \eqref{eq:telescopingcorrelation}:
\begin{align}
    \|r_t^{(\theta)}\|_{\infty} &\leq |a_t-1|\cdot\|\theta_t-\delta g_t^{(\theta)}\|_{\infty} \\&\leq |1-a_t|\cdot (\|\theta_t\|_{\infty}+\delta\|g_t^{(\theta)}\|_{\infty}) \\
    &\leq \frac{\delta^2}{2}\|g_t^{(\theta)}\|_2^2\cdot (\|\theta_t\|_{\infty}+\delta\|g_t^{(\theta)}\|_{\infty}),\text{ since for } x\geq0,  \sqrt{1+x}\geq1+x/2.\\
    &=\frac{\delta^2}{2}\|g_t^{(\theta)}\|_2^2\|\theta_t\|_{\infty}+\frac{\delta^3}{2}\|g_t^{(\theta)}\|_2^3,\text{ given }\|g_t^{(\theta)}\|_{\infty}\leq \|g_t^{(\theta)}\|.
\end{align}
So we have the bound
\begin{align}
    \|\theta_t\|_{\infty} 
    &\leq \|\theta_0\|_{\infty}+\delta \sum_{k<t}\| \bar{g}_{k}^{(\theta)}\|_{\infty}+\max_i\big|\delta\sum_{k<t}\xi_{k,i}^{(\theta)}\big|+\frac{\delta^2}{2}\sum_{k<t}\|g_k^{(\theta)}\|_2^2\|\theta_k\|_{\infty}+\frac{\delta^3}{2}\sum_{k<t}\|g_k^{(\theta)}\|_2^3.
\end{align}
\end{proof}

Now, we analyze the aggregation through time: let $\Theta_t:=\max_{s\leq t}\|\theta_s\|_{\infty}$ to be the maximum over the trajectory, we then have for any $t\leq T,$
\begin{align}\label{eq:4termsofetaevolution}
\Theta_t \leq \|\theta_0\|_{\infty}
    +\underbrace{\delta \sum_{k<t}\| \bar{g}_{k}^{(\theta)}\|_{\infty}}_{1}
    +\underbrace{\max_i \sup_{s\leq t}\big|\delta\sum_{k<s}\xi_{k,i}^{(\theta)}\big|}_{2}
    +\underbrace{\Big(\frac{\delta^2}{2}\sum_{k<t}\|g_k^{(\theta)}\|_2^2\Big)\Theta_t}_{3}
    +\underbrace{\frac{\delta^3}{2}\sum_{k<t}\|g_k^{(\theta)}\|_2^3}_{4}.\end{align}

To bound term $3$ in \eqref{eq:4termsofetaevolution}, we have by Markov Inequality, $$\mathbb{P}\Big(\frac{\delta^2}{2}\sum_{k<T}\|g_k^{(\theta)}\|_2^2 \geq \lambda \Big)\leq \frac{\mathbb{E}\big[\frac{\delta^2}{2}\sum_{k<T}\|g_k^{(\theta)}\|_2^2\big]}{\lambda}\leq \frac{\delta^2TKd}{\lambda}$$
Taking $\lambda = 1/4,$ we have
\begin{align}\label{eq:eta-2-bound}
\mathbb{P}\Big(\frac{\delta^2}{2}\sum_{k<T}\|g_k^{(\theta)}\|_2^2\geq \frac{1}{4} \Big)=O(\varepsilon),
\end{align}
giving that w.p. $1-O(\varepsilon),$ 
$$\Theta_t \leq \|\theta_0\|_{\infty}
    +\underbrace{\delta \sum_{k<t}\| \bar{g}_{k}^{(\theta)}\|_{\infty}}_{1}
    +\underbrace{\max_i \sup_{s\leq t}\big|\delta\sum_{k<s}\xi_{k,i}^{(\theta)}\big|}_{2}
    +\frac{1}{4}\Theta_t
    +\underbrace{\frac{\delta^3}{2}\sum_{k<t}\|g_k^{(\theta)}\|_2^3}_{4}.$$
$$(1-\frac{1}{4})\Theta_t \leq \|\theta_0\|_{\infty}
    +\underbrace{\delta \sum_{k<t}\| \bar{g}_{k}^{(\theta)}\|_{\infty}}_{1}
    +\underbrace{\max_i \sup_{s\leq t}\big|\delta\sum_{k<s}\xi_{k,i}^{(\theta)}\big|}_{2}
    +\underbrace{\frac{\delta^3}{2}\sum_{k<t}\|g_k^{(\theta)}\|_2^3}_{4}.$$ 
\begin{align}\label{eq:bigthetaupperbound}
    \Theta_t \leq \frac{4}{3}\Big(\|\theta_0\|_{\infty}
    +\underbrace{\delta \sum_{k<t}\| \bar{g}_{k}^{(\theta)}\|_{\infty}}_{1}
    +\underbrace{\max_i \sup_{s\leq t}\big|\delta\sum_{k<s}\xi_{k,i}^{(\theta)}\big|}_{2}
    +\underbrace{\frac{\delta^3}{2}\sum_{k<t}\|g_k^{(\theta)}\|_2^3}_{4}\Big).
\end{align}

We start by bounding term 2 using a Freedman Inequality. Yet to show it, we need this corollary below:

\begin{corollary}[Coordinate variance and $8$th-moment bounds for $\xi^{(\theta)}$]
\label{cor:xi-moments}
Work under Assumption~\ref{ass:sgd-regime} and Lemma~\ref{lem:sgd-tech-bounds}.
Let $\xi_t^{(\theta)}:=g_t^{(\theta)}-\E[g_t^{(\theta)}\mid\mathcal F_t]$
and write $\xi_{t,i}^{(\theta)}$ for its $i$-th coordinate.
Then there exist constants $\sigma^2>0$ and $K_8>0$ (depending only on model constants)
such that for all $t$ and all $i\le d$,
\begin{align}
\E\!\left[(\xi_{t,i}^{(\theta)})^2\mid\mathcal F_t\right]\le \sigma^2,
\qquad
\E\!\left[|\xi_{t,i}^{(\theta)}|^8\mid\mathcal F_t\right]\le K_8.
\end{align}
\end{corollary}

\begin{proof}
\textbf{Conditional variance.}
Lemma~\ref{lem:sgd-tech-bounds} gives the directional variance bound
\[
\sup_{\|u\|=1}\E\big[((\Delta g_t^{(\theta)})\cdot u)^2\mid\mathcal F_t\big]\le K.
\]
Taking $u=e_i$ yields $\E[(\xi_{t,i}^{(\theta)})^2\mid\mathcal F_t]\le K$.
Set $\sigma^2:=K$.

\textbf{8-th moment.} Recall that in Lemma~\ref{lem:sgd-tech-bounds}, from the per-sample gradient $\nabla_\theta \ell(\beta_t,\theta_t;x_t)=A_t z_t$
with $A_t:=2(f_t-f_t^*)\psi'(z_t\cdot\theta_t)$ defined in \eqref{eq:persamplegradient}, using the boundedness of $\psi,\psi'$ we showed $|A_t|\le C_A$ in \eqref{eq:Abound}. From there, the spherical gradient is
\[
g_t^{(\theta)}=(I-\theta_t\theta_t^\top)\nabla_\theta \ell
=A_t\big(z_t-(z_t\cdot\theta_t)\theta_t\big),
\]
so for each coordinate
\begin{align}\label{eq:gthetadimensionwisebound}
|g_{t,i}^{(\theta)}|
\le |A_t|\big(|z_{t,i}|+|z_t\cdot\theta_t|\big)
\le C_A\big(|z_{t,i}|+|z_t\cdot\theta_t|\big).
\end{align}

Condition on $\mathcal F_t$. Then $\theta_t$ is deterministic with $\|\theta_t\|_2=1$ and
$z_t=(z_{t,1},\dots,z_{t,d})$ has i.i.d.\ coordinates $z_{t,j}=\phi(\gamma_{t,j})$ with
$\gamma_{t,j}\sim\mathcal N(0,1)$. In particular,
\[
\mu_2:=\E[z_{t,1}^2]=\E[\phi(\gamma)^2]<\infty,
\qquad
\mu_8:=\E[|z_{t,1}|^8]=\E[|\phi(\gamma)|^8]<\infty.
\]
Let $X_j:=\theta_{t,j}z_{t,j}$. Then $(X_j)_{j\le d}$ are independent and mean-zero, and
\[
\sum_{j=1}^d \E[X_j^2\mid\mathcal F_t]
=
\sum_{j=1}^d \theta_{t,j}^2\,\E[z_{t,1}^2]
=
\mu_2\sum_{j=1}^d\theta_{t,j}^2
=
\mu_2.
\]
By Rosenthal's inequality for $p=8$, there exists a constant $C_8>0$ such that
\[
\E\Big[\Big|\sum_{j=1}^d X_j\Big|^8\ \Big|\ \mathcal F_t\Big]
\le
C_8\Big(\Big(\sum_{j=1}^d \E[X_j^2\mid\mathcal F_t]\Big)^4
+\sum_{j=1}^d \E[|X_j|^8\mid\mathcal F_t]\Big).
\]
Moreover,
\[
\sum_{j=1}^d \E[|X_j|^8\mid\mathcal F_t]
=
\sum_{j=1}^d |\theta_{t,j}|^8\,\E[|z_{t,1}|^8]
=
\mu_8\sum_{j=1}^d |\theta_{t,j}|^8
\le
\mu_8\sum_{j=1}^d \theta_{t,j}^2
=
\mu_8,
\]
since $|\theta_{t,j}|\le 1$ implies $|\theta_{t,j}|^8\le |\theta_{t,j}|^2$.
Therefore,
\[
\E\big[|z_t\cdot\theta_t|^8\mid\mathcal F_t\big]
=
\E\Big[\Big|\sum_{j=1}^d X_j\Big|^8\ \Big|\ \mathcal F_t\Big]
\le
C_8(\mu_2^4+\mu_8)
=:C,
\]
which is a constant independent of $t$ and $d$. Hence, plugging into \eqref{eq:gthetadimensionwisebound},  we have $\E[|g_{t,i}^{(\theta)}|^8\mid\mathcal F_t]\le K_8'$ uniformly.

Finally, by Jensen and $(a-b)^8\le 2^7(a^8+b^8)$,
\[
\E[|\xi_{t,i}^{(\theta)}|^8\mid\mathcal F_t]
\le 2^8\E[|g_{t,i}^{(\theta)}|^8\mid\mathcal F_t]
\le 2^8K_8'.
\]
Set $K_8:=2^8K_8'$ gives the claimed result.
\end{proof}

Now we move to prove the Freedman bound:

\begin{lemma}[Coordinate martingale control for the \(\eta\)-bootstrap] 
\label{lem:freedman-bound}
\label{lem:eta-coordinate-martingale}
Assume Corollary~\ref{cor:xi-moments}. Let \(T\delta\le C_T\log d\) and \(\delta\le c_0\varepsilon/(d\log d)\). Then there is a constant \(C_M\) such that 
\begin{align}\label{eq:eta-coordinate-martingale}
    \mathbb P\left( \max_{i\le d}\sup_{t\le T} \left| \delta\sum_{k<t}\xi_{k,i}^{(\theta)} \right| > C_M\sqrt{\frac{\log(40d/\varepsilon)}{d}} \right) \le \varepsilon/5 .
\end{align}
\end{lemma}

\begin{proof}
Let
\[
L:=\log\frac{40d}{\varepsilon}.
\]
The case \(T=0\) is trivial, so assume \(T\ge 1\). Let \(\sigma^2,K_8\) be the constants from Corollary~\ref{cor:xi-moments}. Define the truncation level
\[
B:=\left(\frac{40dTK_8}{\varepsilon}\right)^{1/8}.
\]
For \(k<T\) and \(i\le d\), write
\[
\xi_{k,i}:=\xi_{k,i}^{(\theta)}.
\]
Define the centered truncated martingale difference
\[
Y_{k,i}
:=
\xi_{k,i}\mathbf 1_{\{|\xi_{k,i}|\le B\}}
-
\E\!\left[
\xi_{k,i}\mathbf 1_{\{|\xi_{k,i}|\le B\}}
\mid \mathcal F_k
\right].
\]
Then
\[
\E[Y_{k,i}\mid \mathcal F_k]=0,
\qquad
|Y_{k,i}|\le 2B,
\]
and
\[
\E[Y_{k,i}^2\mid \mathcal F_k]
\le
\E[\xi_{k,i}^2\mid \mathcal F_k]
\le
\sigma^2.
\]
Hence, for each fixed \(i\), the process
\[
M_t^{(i)}:=\sum_{k<t}Y_{k,i}
\]
is a martingale with increments bounded by \(2B\) and predictable quadratic variation at most \(\sigma^2t\le \sigma^2T\).

By the maximal form of Freedman's inequality, for every \(x>0\),
\[
\mathbb P\left(
\sup_{t\le T}M_t^{(i)}\ge x
\right)
\le
\exp\left(
-\frac{x^2}{2(\sigma^2T+2Bx/3)}
\right),
\]
and the same estimate holds for \(-M_t^{(i)}\). Set
\[
x_T:=\sqrt{2\sigma^2TL}+\frac{4}{3}BL.
\]
Then
\[
\frac{x_T^2}{2(\sigma^2T+2Bx_T/3)}\ge L.
\]
Indeed, writing \(a=\sigma^2T\), \(u=\sqrt{2aL}\), and
\(v=\frac43BL\), we have \(x_T=u+v\), while
\[
2L\left(a+\frac{2Bx_T}{3}\right)
=
2aL+v(u+v).
\]
Since
\[
x_T^2=(u+v)^2=2aL+2uv+v^2,
\]
we get
\[
x_T^2-2L\left(a+\frac{2Bx_T}{3}\right)=uv\ge0.
\]
Therefore
\[
\mathbb P\left(
\sup_{t\le T}|M_t^{(i)}|\ge x_T
\right)
\le 2e^{-L}.
\]
Taking a union bound over \(i\le d\), and using \(L=\log(40d/\varepsilon)\), gives
\[
\mathbb P\left(
\max_{i\le d}\sup_{t\le T}|M_t^{(i)}|\ge x_T
\right)
\le 2de^{-L}
=
\frac{\varepsilon}{20}.
\]

It remains to compare the original martingale to the truncated one. Let
\[
\mathcal J
:=
\left\{
\max_{k<T}\max_{i\le d}|\xi_{k,i}|\le B
\right\}.
\]
By Corollary~\ref{cor:xi-moments} and Markov's inequality,
\[
\mathbb P(\mathcal J^c)
\le
\sum_{k<T}\sum_{i\le d}
\mathbb P(|\xi_{k,i}|>B)
\le
dT\frac{K_8}{B^8}
=
\frac{\varepsilon}{40}.
\]
Now define
\[
b_{k,i}
:=
\E\!\left[
\xi_{k,i}\mathbf 1_{\{|\xi_{k,i}|>B\}}
\mid \mathcal F_k
\right].
\]
Since \(\E[\xi_{k,i}\mid \mathcal F_k]=0\),
\[
\E\!\left[
\xi_{k,i}\mathbf 1_{\{|\xi_{k,i}|\le B\}}
\mid \mathcal F_k
\right]
=
-b_{k,i}.
\]
Hence
\[
Y_{k,i}
=
\xi_{k,i}\mathbf 1_{\{|\xi_{k,i}|\le B\}}
+
b_{k,i}.
\]
On \(\mathcal J\), the indicator \(\mathbf 1_{\{|\xi_{k,i}|\le B\}}\) is identically one for all \(k<T\) and \(i\le d\). Therefore, on \(\mathcal J\),
\[
\xi_{k,i}=Y_{k,i}-b_{k,i}.
\]
Moreover, by the conditional eighth-moment bound,
\[
|b_{k,i}|
\le
\E\!\left[
|\xi_{k,i}|\mathbf 1_{\{|\xi_{k,i}|>B\}}
\mid \mathcal F_k
\right]
\le
\frac{\E[|\xi_{k,i}|^8\mid \mathcal F_k]}{B^7}
\le
\frac{K_8}{B^7}.
\]
Thus, on \(\mathcal J\),
\[
\sup_{t\le T}
\left|
\sum_{k<t}\xi_{k,i}
\right|
\le
\sup_{t\le T}|M_t^{(i)}|
+
\frac{TK_8}{B^7}.
\]
Combining the previous estimates, with probability at least
\(1-\varepsilon/20-\varepsilon/40\), we have
\[
\max_{i\le d}\sup_{t\le T}
\left|
\sum_{k<t}\xi_{k,i}
\right|
\le
x_T+\frac{TK_8}{B^7}.
\]
Multiplying by \(\delta\), it remains to show that
\[
\delta x_T+\delta\frac{TK_8}{B^7}
\le
C_M\sqrt{\frac{L}{d}}.
\]

First,
\[
\delta\sqrt{2\sigma^2TL}
=
\sqrt{2\sigma^2L\,\delta(\delta T)}
\le
\sqrt{2\sigma^2C_Tc_0\varepsilon}\sqrt{\frac{L}{d}}
\le
C\sqrt{\frac{L}{d}},
\]
because \(T\delta\le C_T\log d\), \(\delta\le c_0\varepsilon/(d\log d)\), and \(\varepsilon\le1\).

Next, using \(T\le C_T(\log d)/\delta\),
\[
\delta BL
=
\delta L\left(\frac{40dTK_8}{\varepsilon}\right)^{1/8}
\le
C L\,\delta^{7/8}
\left(\frac{d\log d}{\varepsilon}\right)^{1/8}.
\]
Using again \(\delta\le c_0\varepsilon/(d\log d)\), this gives
\[
\delta BL
\le
C c_0^{7/8}
\varepsilon^{3/4}
L\,d^{-3/4}(\log d)^{-3/4}.
\]
Since \(L=\log(40d/\varepsilon)\) and \(0<\varepsilon\le1\),
\[
\varepsilon^{3/4}L^{1/2}\le C(\log d)^{1/2}
\]
for \(d\ge3\). Hence
\[
\delta BL
\le
C c_0^{7/8}
L^{1/2}d^{-3/4}(\log d)^{-1/4}
\le
C\sqrt{\frac{L}{d}}.
\]

Finally,
\[
\delta\frac{TK_8}{B^7}
=
\delta TK_8
\left(\frac{\varepsilon}{40dTK_8}\right)^{7/8}
\le
C\delta T^{1/8}\varepsilon^{7/8}d^{-7/8}.
\]
Using \(T\le C_T(\log d)/\delta\) and
\(\delta\le c_0\varepsilon/(d\log d)\), we get
\[
\delta\frac{TK_8}{B^7}
\le
C c_0^{7/8}
\varepsilon^{7/4}
d^{-7/4}(\log d)^{-3/4}
\le
C\sqrt{\frac{L}{d}}.
\]
Therefore, after increasing \(C_M\) to absorb constants depending only on
\(\sigma^2,K_8,C_T,c_0\), we have
\[
\delta x_T+\delta\frac{TK_8}{B^7}
\le
C_M\sqrt{\frac{L}{d}}
=
C_M\sqrt{\frac{\log(40d/\varepsilon)}{d}}.
\]
Since
\[
\frac{\varepsilon}{20}+\frac{\varepsilon}{40}
<
\frac{\varepsilon}{5},
\]
the desired bound follows.
\end{proof}

We plug this proved Freedman bound into \eqref{eq:bigthetaupperbound} to obtain 
\begin{align}
    \Theta_t \leq \frac{4}{3}\Big(\|\theta_0\|_{\infty}
    +\underbrace{\delta \sum_{k<t}\| \bar{g}_{k}^{(\theta)}\|_{\infty}}_{1}
    + \Lambda_\eta
    +\underbrace{\frac{\delta^3}{2}\sum_{k<t}\|g_k^{(\theta)}\|_2^3}_{4}\Big).
\end{align}
Since $\eta_t=\|\theta_t\|_\infty+\|\theta^*\|_\infty\leq \Theta_t+\|\theta^*\|_\infty$, we re-write it as:
\begin{align}
    \eta_t \leq \frac{4}{3}\Big(\eta_0
    +\underbrace{\delta \sum_{k<t}\| \bar{g}_{k}^{(\theta)}\|_{\infty}}_{1}
    + \Lambda_\eta
    +\underbrace{\frac{\delta^3}{2}\sum_{k<t}\|g_k^{(\theta)}\|_2^3}_{4}\Big).
\end{align}

Next we bound $4$: notice that since $\frac{\delta^3}{2}\sum_{k<t}\|g_k^{(\theta)}\|_2^3$ is non decreasing in $t,$ we have $$\sup_{t\leq T}\frac{\delta^3}{2}\sum_{k<t}\|g_k^{(\theta)}\|_2^3 = \frac{\delta^3}{2}\sum_{k<T}\|g_k^{(\theta)}\|_2^3.$$
On the other hand,
\begin{align}
    \mathbb{E}\big[\|g_k^{(\theta)}\|_2^3\big] &\leq \mathbb{E}\big[\|g_k^{(\theta)}\|_2^4\big]^{3/4} \text{ by Holder inequality }\\ 
    &\leq (Kd^2)^{3/4} \text{ by the technical bounds in Lemma \ref{lem:sgd-tech-bounds}} \\
    &= K^{3/4}d^{3/2} \label{eq:gbetacubebound}.
\end{align}
Therefore, for a given $\lambda_{\text{cub}},$
\begin{align}
    \mathbb{P}\Big(\sup_{t\leq T }\frac{\delta^3}{2}\sum_{k<t}\|g_k^{(\theta)}\|_2^3 \geq \lambda_{\text{cub}} \Big) &= \mathbb{P}\Big(\frac{\delta^3}{2}\sum_{k<T}\|g_k^{(\theta)}\|_2^3 \geq \lambda_{\text{cub}} \Big) \text{ by monotonicity} \\ &\leq\frac{\mathbb{E}\big[\frac{\delta^3}{2}\sum_{k<t}\|g_k^{(\theta)}\|_2^3\big]}{\lambda_{\text{cub}}},\text{ by Markov Inequality} \\
    & \leq \frac{K^{3/4} d^{3/2}T\frac{\delta^3}{2}}{\lambda_{\text{cub}}}, \text{ following \eqref{eq:gbetacubebound}}
\end{align}
Taking $\lambda_{\text{cub}} = C_{\text{cub}}/\sqrt{d},$ and plugging in $T\leq c_2(\log d)/\delta$ and $\delta\leq c_0\varepsilon/(d\log d),$ we have 
\begin{align} 
    \mathbb{P}\left(\sup_{t\leq T }\frac{\delta^3}{2}\sum_{k<t}\|g_k^{(\theta)}\|_2^3 \geq \frac{C_{\text{cub}}}{\sqrt{d}} \right)
    &\leq \frac{c_2,c_0^2,K^{3/4}}{2C_{\text{cub}}}\cdot \frac{\varepsilon^2}{\log d} .
\end{align}
Choose $C_{\text{cub}}$ large enough (absorbing constants) and take $d$ sufficiently large so that $\frac{\varepsilon^2}{\log d}\le \varepsilon/10,$ we obtain 
\begin{align}\label{eq:term4finalbound}
\mathbb{P}\left(\sup_{t\leq T }\frac{\delta^3}{2}\sum_{k<t}\|g_k^{(\theta)}\|_2^3 \geq \frac{C_{\text{cub}}}{\sqrt{d}} \right) \leq \frac{\varepsilon}{10}
\end{align}
Plugging it in the $\eta$ bound in \eqref{eq:bigthetaupperbound} it becomes
\begin{align} \label{eq:etaboundbeforedynamics}
    \eta_t \leq C_{\eta}\Big(\eta_0
    +\underbrace{\delta \sum_{k<t}\| \bar{g}_{k}^{(\theta)}\|_{\infty}}_{1}
    + \Lambda_\eta
    + \frac{C_{\text{cub}}}{\sqrt{d}}\Big),
\end{align}
with high probability. The bound of $1$ is contingent to dynamics's phase that we will derive it later. We later bound them in Step~1 of the Phase~I proof (Lemma~\ref{lem:sgd-phase1}) and Step~1(a) of the Phase~II proof (Lemma~\ref{lem:sgd-phase2}).

\paragraph{Good event for $\eta$ iteration} 

Define
\begin{align}\label{event:eta}
E_T^{(\eta)}
:=
E_T^{(\eta,M)}
\cap E_T^{(\eta,2)}
\cap E_T^{(\eta,3)},
\end{align}
where
\begin{align}
E_T^{(\eta,M)}
&:=
\left\{
\max_{i\le d}\sup_{t\le T}
\left|
\delta\sum_{k<t}\xi_{k,i}^{(\theta)}
\right|
\le \Lambda_\eta
\right\}\label{event:sgd-tech-bounds},\\
E_T^{(\eta,2)}
&:=
\left\{
\frac{\delta^2}{2}\sum_{k<T}\|g_k^{(\theta)}\|_2^2
\le \frac14
\right\},\label{event:eta-2-bound}\\
E_T^{(\eta,3)}
&:=
\left\{
\sup_{t\le T}
\frac{\delta^3}{2}\sum_{k<t}\|g_k^{(\theta)}\|_2^3
\le \frac{C_{\rm cub}}{\sqrt d}
\right\}.\label{event:term4finalbound}
\end{align}
Therefore, with Bound~\eqref{eq:eta-coordinate-martingale} from Lemma~\ref{lem:sgd-tech-bounds} to ensure \eqref{event:sgd-tech-bounds}, Bound~\eqref{eq:eta-2-bound} to ensure \eqref{event:eta-2-bound}, and Bound~\eqref{eq:term4finalbound} to ensure \eqref{event:term4finalbound}, we obtain a union bound,
\begin{align}\label{eq:eta-union-bound}
\mathbb P\big((E_T^{(\eta)})^c\big)
\le
\mathbb P\big((E_T^{(\eta,M)})^c\big)
+
\mathbb P\big((E_T^{(\eta,2)})^c\big)
+
\mathbb P\big((E_T^{(\eta,3)})^c\big)
\le K\varepsilon ,
\end{align}
after increasing \(K\).

\subsubsection{Martingale and Discretization Error Control for \texorpdfstring{$\Omega_t$}{Omega	extsubscript{t}} and \texorpdfstring{$\omega_t$}{omega extsubscript{t}}} \label{subsec:Omega-omega}

As a discrete analogy of Lemma~\ref{lem:population-third-moment-control}, we next provide the control of third moment quantities $\Omega_t:=\|\theta_t\|_3^3=\sum_{i=1}^d|\theta_{t,i}|^3,$ and $\omega_t:=\sum_{i=1}^d\bigl(|\theta_{t,i}|+|\theta_i^*|\bigr)^3,$ upon their martingale and discretizatio noises.

\begin{lemma}[Third-moment control for SGD through Phase I]
\label{lem:sgd-third-moment-control}
Let $\Omega_t$ and $\omega_t$ follow the definition made in  Lemma~\ref{lem:population-third-moment-control}. Fix \(T\in\mathbb N\) with \(T\delta\le C_T\log d\). Under
Assumption~\ref{ass:sgd-regime}, Lemma~\ref{lem:sgd-tech-bounds}, and
Corollary~\ref{cor:xi-moments}, there is an event \(E_T^{(\Omega)}\) satisfying
\[
\mathbb P\big((E_T^{(\Omega)})^c\big)\le K_\Omega\varepsilon
\]
such that the following holds. Work on $\mathcal E_0\cap E_T^{(\Omega)}.$
Let
\[
\begin{gathered}
\tau_A:=\inf\{t:Z_t^2\ge A\eta_t\},\\
\tau_B:=\inf\{t:\rho_t\ge1/2\text{ or }\chi_t\ge1/2\},\\
\tau_+:=\inf\{t:\rho_t<0\text{ or }\chi_t<0\},
\end{gathered}
\]
and set
\[
\sigma:=\tau_A\wedge\tau_B\wedge\tau_+\wedge T.
\]
Suppose that, for some constant \(C_\eta>0\),
\begin{align}\label{eq:third-moment-eta-input}
\eta_t &\le C_\eta\frac{\sqrt{\log d}+1+t\delta}{\sqrt d} \qquad (0\le t\le\sigma).
\end{align}
Then, for all sufficiently large \(d\),
\begin{align}
\Omega_t\le C_\Omega\frac{1+t\delta}{\sqrt d},
\qquad
\omega_t\le C_\omega\frac{1+t\delta}{\sqrt d},
\qquad
0\le t\le\sigma.
\end{align}
The constants \(K_\Omega,C_\Omega,C_\omega\) depend only on model constants,
\(A,C_\eta,C_T\), and the step-size regime.
\end{lemma}

\begin{proof}
Write \(g_t:=g_t^{(\theta)}\), \(\bar g_t:=\E[g_t\mid\mathcal F_t]\), and
\(\xi_t:=g_t-\bar g_t\). Since \(g_t\perp\theta_t\), the normalized spherical
update is
\[
\theta_{t+1}=a_t(\theta_t-\delta g_t),
\qquad
a_t=(1+\delta^2\|g_t\|_2^2)^{-1/2}\le1.
\]
Hence
\[
\Omega_{t+1}
=
\|\theta_{t+1}\|_3^3
\le
\|\theta_t-\delta g_t\|_3^3.
\]
We use the elementary inequality
\[
|x-y|^3
\le
|x|^3-3|x|x\,y+C\bigl(|x|y^2+|y|^3\bigr),
\qquad x,y\in\mathbb R.
\]
Applying it coordinatewise with \(x=\theta_{t,i}\) and \(y=\delta g_{t,i}\),
and letting $v_t:=\bigl(|\theta_{t,i}|\theta_{t,i}\bigr)_{i=1}^d,$ gives
\begin{equation}
\label{eq:Omega-one-step-sgd}
\Omega_{t+1}
\le
\Omega_t
-3\delta\langle v_t,\bar g_t\rangle
-3\delta\langle v_t,\xi_t\rangle
+C\delta^2\sum_{i=1}^d|\theta_{t,i}|g_{t,i}^2
+C\delta^3\|g_t\|_3^3.
\end{equation}
Since \(\|v_t\|_1=\sum_i|\theta_{t,i}|^2=1\),
\[
-3\delta\langle v_t,\bar g_t\rangle
\le
3\delta\|\bar g_t\|_\infty.
\]
Summing \eqref{eq:Omega-one-step-sgd} gives
\begin{equation}
\label{eq:Omega-path-sgd}
\Omega_t
\le
\Omega_0
+
3\delta\sum_{k<t}\|\bar g_k\|_\infty
+
\mathcal M_t^{(\Omega)}
+
C\mathcal Q_t^{(\Omega)}
+
C\mathcal C_t^{(\Omega)},
\end{equation}
where
\[
\mathcal M_t^{(\Omega)}
:=
-3\delta\sum_{k<t}\langle v_k,\xi_k\rangle,
\]
\[
\mathcal Q_t^{(\Omega)}
:=
\delta^2\sum_{k<t}\sum_{i=1}^d|\theta_{k,i}|g_{k,i}^2,
\qquad
\mathcal C_t^{(\Omega)}
:=
\delta^3\sum_{k<t}\|g_k\|_3^3.
\]

We first define the good event \(E_T^{(\Omega)}\). Let
\[
\lambda_{\Omega,M}:=\frac{C_{\Omega,M}}{\sqrt d},
\qquad
\lambda_{\Omega,Q}:=\frac{C_{\Omega,Q}}{\sqrt d},
\qquad
\lambda_{\Omega,C}:=\frac{C_{\Omega,C}}{\sqrt d},
\]
where the constants will be chosen large. Set
\begin{equation}
\label{eq:third-moment-good-event-sgd}
E_T^{(\Omega)}
:=
\left\{
\sup_{t\le T}|\mathcal M_t^{(\Omega)}|\le \lambda_{\Omega,M}
\right\}
\cap
\left\{
\sup_{t\le T}\mathcal Q_t^{(\Omega)}\le \lambda_{\Omega,Q}
\right\}
\cap
\left\{
\sup_{t\le T}\mathcal C_t^{(\Omega)}\le \lambda_{\Omega,C}
\right\}.
\end{equation}

We verify that this event has high probability. The process
\(\mathcal M_t^{(\Omega)}\) is a martingale. Moreover,
\[
\|v_k\|_2^2
=
\sum_i|\theta_{k,i}|^4
\le
\sum_i|\theta_{k,i}|^2
=
1.
\]
By Lemma~\ref{lem:sgd-tech-bounds},
\[
\E\bigl[\langle v_k,\xi_k\rangle^2\mid\mathcal F_k\bigr]\le K.
\]
Thus Doob's \(L^2\) inequality gives
\[
\mathbb P\left(
\sup_{t\le T}|\mathcal M_t^{(\Omega)}|\ge \lambda_{\Omega,M}
\right)
\le
\frac{K T\delta^2}{\lambda_{\Omega,M}^2}
=
\frac{K T\delta^2 d}{C_{\Omega,M}^2}
\le
\frac{K C_Tc_0}{C_{\Omega,M}^2}\varepsilon.
\]
Choosing \(C_{\Omega,M}\) large makes this contribution \(O(\varepsilon)\).

Next, by the coordinate moment bound proved in Corollary~\ref{cor:xi-moments}
and \eqref{eq:gthetadimensionwisebound}, we have
\[
\E[g_{k,i}^2\mid\mathcal F_k]\le K
\qquad
\text{uniformly in }i,k.
\]
Therefore
\[
\E\left[
\sum_i|\theta_{k,i}|g_{k,i}^2
\,\middle|\,
\mathcal F_k
\right]
\le
K\sum_i|\theta_{k,i}|
\le
K\sqrt d.
\]
Hence
\[
\E[\mathcal Q_T^{(\Omega)}]
\le
K T\delta^2\sqrt d.
\]
By Markov's inequality,
\[
\mathbb P\left(
\sup_{t\le T}\mathcal Q_t^{(\Omega)}\ge \lambda_{\Omega,Q}
\right)
\le
\frac{K T\delta^2\sqrt d}{C_{\Omega,Q}/\sqrt d}
=
\frac{K T\delta^2 d}{C_{\Omega,Q}}
\le
\frac{K C_Tc_0}{C_{\Omega,Q}}\varepsilon.
\]
Choosing \(C_{\Omega,Q}\) large makes this contribution \(O(\varepsilon)\).

Similarly, the same coordinate moment bounds imply
\[
\E[|g_{k,i}|^3\mid\mathcal F_k]\le K,
\qquad
\E[\|g_k\|_3^3\mid\mathcal F_k]\le Kd.
\]
Thus
\[
\E[\mathcal C_T^{(\Omega)}]
\le
K T\delta^3 d.
\]
Again by Markov's inequality,
\[
\mathbb P\left(
\sup_{t\le T}\mathcal C_t^{(\Omega)}\ge \lambda_{\Omega,C}
\right)
\le
\frac{K T\delta^3d}{C_{\Omega,C}/\sqrt d}
=
\frac{K T\delta^3d^{3/2}}{C_{\Omega,C}}.
\]
Since \(T\delta\le C_T\log d\) and
\(\delta\le c_0\varepsilon/(d\log d)\), the last display is
\(o(\varepsilon)\) for \(d\) large. Therefore, after increasing the constants
if necessary,
\begin{align}\label{eq:omega-bound}
    \mathbb P\big((E_T^{(\Omega)})^c\big)\le K_\Omega\varepsilon.
\end{align}
It remains to prove the pathwise bound on \(\Omega_t\). Work on
\(\mathcal E_0\cap E_T^{(\Omega)}\). From \eqref{eq:Omega-path-sgd} and
\eqref{eq:third-moment-good-event-sgd},
\begin{equation}
\label{eq:Omega-path-good}
\Omega_t
\le
\Omega_0
+
3\delta\sum_{k<t}\|\bar g_k\|_\infty
+
\frac{C}{\sqrt d},
\qquad
t\le T.
\end{equation}
We now bound the mean-field term for \(t\le\sigma\). Since \(k<\sigma\) implies
\(k<\tau_+\), we have \(\rho_k,\chi_k\ge0\). Since \(k<\sigma\) also implies
\(k<\tau_A\), we have \(Z_k^2<A\eta_k\). By \eqref{eq:thetaellinfty} in Lemma~\ref{lemma:secondlayercontrol},
\[
\|\bar g_k\|_\infty
\le
C Z_k\|\theta^*\|_\infty
+
C Z_k^2\eta_k
+
C\eta_k^2.
\]
On \(\mathcal E_0\), \(\|\theta^*\|_\infty\le C\sqrt{\log d/d}\). Using
\(Z_k^2<A\eta_k\) and the assumed envelope for \(\eta_k\), we obtain
\[
Z_k\|\theta^*\|_\infty
\le
\sqrt{A\eta_k}\,\|\theta^*\|_\infty
\le
C\sqrt{\frac{\sqrt{\log d}+1+k\delta}{\sqrt d}}
\sqrt{\frac{\log d}{d}}.
\]
Since \(k\delta\le T\delta\le C_T\log d\), the right-hand side is bounded by
\(C(\log d)d^{-3/4}\le C d^{-1/2}\) for all sufficiently large \(d\). Also,
\(Z_k^2\eta_k\le A\eta_k^2\). Hence, for \(k<\sigma\),
\begin{equation}
\label{eq:Omega-mean-field-bound}
\|\bar g_k\|_\infty
\le
\frac{C}{\sqrt d}+C\eta_k^2.
\end{equation}
Therefore, for \(t\le\sigma\),
\[
\delta\sum_{k<t}\|\bar g_k\|_\infty
\le
C\frac{t\delta}{\sqrt d}
+
C\delta\sum_{k<t}\eta_k^2.
\]
Using the assumed \(\eta\)-envelope,
\[
\delta\sum_{k<t}\eta_k^2
\le
\frac{C}{d}
\delta\sum_{k<t}
\bigl(\sqrt{\log d}+1+k\delta\bigr)^2
\le
\frac{C(\log d)^3}{d}
=
o(d^{-1/2}).
\]
Plugging this into \eqref{eq:Omega-path-good}, and using
\(\Omega_0\le C/\sqrt d\) on \(\mathcal E_0\), gives
\[
\Omega_t
\le
C\frac{1+t\delta}{\sqrt d},
\qquad
0\le t\le\sigma.
\]
This proves the desired \(\Omega_t\)-bound.

Finally,
\[
\omega_t
=
\sum_i\bigl(|\theta_{t,i}|+|\theta_i^*|\bigr)^3
\le
4\sum_i|\theta_{t,i}|^3+4\sum_i|\theta_i^*|^3
=
4\Omega_t+4\|\theta^*\|_3^3.
\]
On \(\mathcal E_0\), \(\|\theta^*\|_3^3\le C/\sqrt d\). Therefore
\[
\omega_t
\le
C_\omega\frac{1+t\delta}{\sqrt d},
\qquad
0\le t\le\sigma.
\]
This completes the proof.
\end{proof}

\subsection{Weak Recovery Dynamics}
\paragraph{Good Events for Weak Recovery}

Now, we define the good event for weak recovery by taking the intersection of $\mathcal{E}_{\rm deloc}$ from \ref{eeee} $E_T^{(T)}$ from \eqref{event:taylor}, $E_T^{(Z)}$ from \eqref{event:Z}, $E_T^{(\rho,\chi)}$ from \eqref{event:rho-chi}, $E_T^{(\eta)}$ from \eqref{event:eta}, and $E_T^{(\Omega)}$ from \eqref{eq:third-moment-good-event-sgd}:
\begin{align}\label{eq:good-event-weak}
\mathcal G_T^{\rm weak}
:=
\mathcal E_0
\cap E_T^{(T)}
\cap E_T^{(Z)}
\cap E_T^{(\rho,\chi)}
\cap E_T^{(\eta)}
\cap E_T^{(\Omega)}.
\end{align}
With their respective probabilistic controls made in \eqref{eq:good-initialization-probability}, \eqref{eq:taylor-bound}, \eqref{eq:Z-union-bound}, \eqref{eq:rho-chi-bound}, \eqref{eq:eta-union-bound}, \eqref{eq:omega-bound}, they are true with probability $(1-K_{\rm alg}\varepsilon)(1-d^{-K_{\rm ini}}.)$

\subsubsection{Phase I Dynamics, Weak Recovery}

\begin{lemma}[Phase I: pull-up to the signal-dominant scale] \label{lem:sgd-phase1} Work on \(\mathcal G_T^{\rm weak}\). Then, up to \(\tau_A\wedge\tau_B\), 
\begin{align}
    \eta_t\le C_\eta\frac{\sqrt{\log d}+1+t\delta}{\sqrt d}, \qquad \omega_t\le C_\omega\frac{1+t\delta}{\sqrt d}. 
\end{align}
Moreover,
\begin{align}
\tau_A<\tau_B,\qquad
\tau_+>\tau_A,\qquad
\tau_A\delta\le C\log d,\qquad
Z_{\tau_A}\ge c_A d^{-1/4}.
\end{align}
Equivalently,
\begin{align}
\rho_t,\chi_t>0
\qquad
\text{for all }0\le t\le\tau_A.
\end{align}
Finally,
\begin{align}
    Z_{\tau_A}\le C_A\sqrt{\log d}\,d^{-1/4}+o(1).
\end{align}
\end{lemma}

\begin{proof}
Work throughout on \(\mathcal G_T^{\rm weak}\). We use the favorable
initialization event in the form
\[
\rho_0,\chi_0\ge \frac{z_-}{\sqrt d},
\qquad
\eta_0\le C\sqrt{\frac{\log d}{d}},
\]
where \(z_->0\) is fixed large enough below. Recall that
\[
\tau_+:=\inf\{t:\rho_t<0\text{ or }\chi_t<0\}.
\]
Set
\[
\sigma:=\tau_A\wedge\tau_B\wedge\tau_+\wedge T .
\]It is enough to prove the desired estimates up to \(\sigma\), and then show
that \(\sigma=\tau_A\).

\paragraph{Step 1: Phase-I \(\eta\)-envelope.}

Let
\[
\Theta_t:=\max_{0\le s\le t}\eta_s,
\qquad
H_I(t):=K_I\frac{\sqrt{\log d}+1+t\delta}{\sqrt d},
\]
where \(K_I\) will be chosen large. By the discrete \(\eta\)-estimate
\eqref{eq:etaboundbeforedynamics}, on \(E_T^{(\eta)}\),
\[
\eta_t
\le
C\left(
\eta_0
+
\delta\sum_{k<t}\|\bar g_k^{(\theta)}\|_\infty
+
\Lambda_\eta
+
\frac{C_{\rm cub}}{\sqrt d}
\right),
\qquad t\le T .
\]
The definition of \(E_T^{(\eta)}\) gives
\[
\Lambda_\eta
\le
C\sqrt{\frac{\log(40d/\varepsilon)}{d}}
\le
C\frac{\sqrt{\log d}+1+t\delta}{\sqrt d},
\qquad t\le T ,
\]

For \(k<\sigma\), we have \(k<\tau_+\), hence
\(\rho_k,\chi_k\ge0\), and also \(k<\tau_A\), hence
\(Z_k^2<A\eta_k\). The population \(\ell_\infty\)-gradient bound
\eqref{eq:thetaellinfty} gives
\[
\|\bar g_k^{(\theta)}\|_\infty
\le
C Z_k\|\theta^*\|_\infty
+
C Z_k^2\eta_k
+
C\eta_k^2 .
\]
Using \(Z_k^2<A\eta_k\), this becomes
\[
\|\bar g_k^{(\theta)}\|_\infty
\le
C\sqrt{\eta_k}\,\|\theta^*\|_\infty
+
C\eta_k^2 .
\]
Since \(\|\theta^*\|_\infty\le C\sqrt{\log d/d}\), and since throughout the
bootstrap \(k\delta\le T\delta\le C_T\log d\), the implication
\(\eta_k\le H_I(k)\) gives
\[
\sqrt{\eta_k}\,\|\theta^*\|_\infty
\le
C\frac{\log d}{d^{3/4}}
\le
\frac{C}{\sqrt d}
\]
for all sufficiently large \(d\). Thus, under the bootstrap,
\[
\|\bar g_k^{(\theta)}\|_\infty
\le
\frac{C}{\sqrt d}+C\eta_k^2 .
\]

Let
\[
\tau_H:=\inf\{t\le\sigma:\Theta_t>H_I(t)\}.
\]
Suppose for contradiction that \(\tau_H\le\sigma\). Then for every
\(k<\tau_H\), \(\eta_k\le \Theta_k\le H_I(k)\), and therefore
\[
\delta\sum_{k<\tau_H}\|\bar g_k^{(\theta)}\|_\infty
\le
C\frac{\tau_H\delta}{\sqrt d}
+
C\delta\sum_{k<\tau_H}H_I(k)^2 .
\]
Moreover,
\[
\delta\sum_{k<\tau_H}H_I(k)^2
\le
\frac{C K_I^2}{d}
\delta\sum_{k<\tau_H}
\bigl(\sqrt{\log d}+1+k\delta\bigr)^2
\le
\frac{C(\log d)^3}{d}
=
o\!\left(
\frac{\sqrt{\log d}+1+\tau_H\delta}{\sqrt d}
\right).
\]
Combining the previous estimates with the bound on \(\eta_t\), we get
\[
\Theta_{\tau_H}
\le
C\frac{\sqrt{\log d}+1+\tau_H\delta}{\sqrt d}
+
o\!\left(
\frac{\sqrt{\log d}+1+\tau_H\delta}{\sqrt d}
\right).
\]
Choosing \(K_I\) large enough, and then \(d\) large enough, gives
\[
\Theta_{\tau_H}\le \frac12 H_I(\tau_H),
\]
contradicting the definition of \(\tau_H\). Hence \(\tau_H>\sigma\), and
therefore
\begin{equation}
\label{eq:phase1-eta-bound-clean}
\eta_t
\le
C_\eta\frac{\sqrt{\log d}+1+t\delta}{\sqrt d},
\qquad
0\le t\le\sigma,
\end{equation}
which is exactly the assumption \eqref{eq:third-moment-eta-input} made in Lemma~\ref{lem:sgd-third-moment-control}. Alongside the condition \(Z_t^2<A\eta_t\)
for \(t<\tau_A\), we can apply Lemma~\ref{lem:sgd-third-moment-control} to obtain
\begin{equation}
\label{eq:phase1-omega-bound-clean}
\omega_t
\le
C_\omega\frac{1+t\delta}{\sqrt d},
\qquad
0\le t\le\sigma .
\end{equation}

\paragraph{Step 2: Positivity and scalar \(Z\)-pull-up.}
For \(t<\sigma\), we have \(t<\tau_A\), hence $Z_t^2<A\eta_t.$
By Step 1,
\[
\eta_t\le C_\eta\frac{\sqrt{\log d}+1+t\delta}{\sqrt d}
\le C\frac{\log d}{\sqrt d}.
\]
Thus
\[
Z_t\le C\sqrt{\log d}\,d^{-1/4}=o(1).
\]
For \(d\) large, \(Z_t\le c_{\rm sm}\), so we can apply Lemma~\ref{lem:refined-small-overlap-drift}
\[
\dot\rho_t\ge a\chi_t-b\omega_t,
\qquad
\dot\chi_t\ge a\rho_t-b\omega_t .
\]
Combining this with Lemma~\ref{lem:sgd-one-step}, the Taylor event
\(E_T^{(T)}\), and Lemma~\ref{lem:sgd-tech-bounds}, we obtain
\[
u_{t+1}
\ge
\left[(1-C\delta^2d)I+a\delta J\right]u_t
-
C\delta\omega_t{\bf 1}
+
\varepsilon_t^{(u)},
\qquad
u_t:=(\rho_t,\chi_t)^\top,
\]
where
\[
J=
\begin{pmatrix}
0&1\\
1&0
\end{pmatrix},
\qquad
{\bf 1}=(1,1)^\top,
\qquad
\varepsilon_t^{(u)}
=
\bigl(N_{t+1}^{(\rho)}-N_t^{(\rho)},
N_{t+1}^{(\chi)}-N_t^{(\chi)}\bigr)^\top .
\]
Let
\[
B:=(1-C\delta^2d)I+a\delta J .
\]
For \(d\) large, \(B\) has nonnegative entries. Its eigenvalues in the
\({\bf 1}\)- and \((1,-1)^\top\)-directions are
\[
\gamma_+:=1+a\delta-C\delta^2d,
\qquad
\gamma_-:=1-a\delta-C\delta^2d .
\]
After decreasing \(a\), we may assume
\[
\gamma_+\ge 1+\beta\delta,
\qquad
0<\gamma_-<1,
\qquad
\beta>0 .
\]

Unrolling the previous cooperative recursion and applying the summation-by-parts
bounds to the \(Z\)-noise and \(D\)-noise gives, componentwise,
\begin{equation}
\label{eq:phase1-coordinate-positivity-clean}
\rho_t,\chi_t
\ge
\gamma_+^t
\left[
\frac{z_-}{\sqrt d}
-
C\sum_{k<t}\gamma_+^{-(k+1)}\delta\omega_k
-
C(\lambda_Z+\lambda_\rho+\lambda_\chi)
\right],
\qquad
0\le t\le\sigma .
\end{equation}
Indeed, the \(Z\)-noise is controlled by Corollary~\ref{lem:sbp}, while the
contracting \(D\)-noise is controlled by Corollary~\ref{lem:sbp-increasing}.

By \eqref{eq:phase1-omega-bound-clean},
\[
\sum_{k<t}\gamma_+^{-(k+1)}\delta\omega_k
\le
\frac{C}{\sqrt d}
\left(
\delta\sum_{k\ge0}\gamma_+^{-(k+1)}
+
\delta^2\sum_{k\ge0}k\gamma_+^{-(k+1)}
\right)
\le
\frac{C}{\sqrt d}.
\]
Since
\[
\lambda_Z+\lambda_\rho+\lambda_\chi=O(d^{-1/2}),
\]
choosing \(z_-\) larger than the constants in the bracket yields
\begin{equation}
\label{eq:phase1-coordinate-positive-clean}
\rho_t,\chi_t>0,
\qquad
0\le t\le\sigma .
\end{equation}
In particular, \(\sigma\neq\tau_+\).

Now sum the two one-step inequalities. For
\[
\varepsilon_t^{(Z)}:=N_{t+1}^{(Z)}-N_t^{(Z)},
\]
we get
\[
Z_{t+1}
\ge
(1+\beta\delta)Z_t
-
C\delta\omega_t
+
\varepsilon_t^{(Z)},
\qquad
t<\sigma .
\]
Set \(\gamma:=1+\beta\delta\). Multiplying by \(\gamma^{-(t+1)}\) and summing,
\[
\gamma^{-t}Z_t
\ge
Z_0
-
C\sum_{k<t}\gamma^{-(k+1)}\delta\omega_k
+
\sum_{k<t}\gamma^{-(k+1)}\varepsilon_k^{(Z)} .
\]
The weighted \(\omega\)-sum is \(O(d^{-1/2})\) by
\eqref{eq:phase1-omega-bound-clean}. Also, on \(E_T^{(Z)}\),
\[
\sup_{u\le T}|N_u^{(Z)}|\le 2\lambda_Z,
\]
so Corollary~\ref{lem:sbp} gives
\[
\left|
\sum_{k<t}\gamma^{-(k+1)}\varepsilon_k^{(Z)}
\right|
\le
4\lambda_Z .
\]
Since \(Z_0\ge 2z_-d^{-1/2}\), increasing \(z_-\) if necessary gives a
constant \(c_->0\) such that
\begin{equation}
\label{eq:phase1-Z-growth-clean}
Z_t
\ge
\frac{c_-}{\sqrt d}\gamma^t,
\qquad
0\le t\le\sigma .
\end{equation}

\paragraph{Step 3: Some stopping event occurs by \(O(\delta^{-1}\log d)\).}

Let
\[
t_0:=\left\lceil c_6\delta^{-1}\log d\right\rceil ,
\]
where \(c_6\) will be chosen large. Choose the weak-recovery horizon constant
\(C_T\) so that \(t_0<T\). Suppose, for contradiction, that \(\sigma>t_0\).
Then \(t_0<\tau_A\), hence
\[
Z_{t_0}^2<A\eta_{t_0}.
\]
Using \eqref{eq:phase1-Z-growth-clean} and
\eqref{eq:phase1-eta-bound-clean},
\[
\frac{c_-^2}{d}\gamma^{2t_0}
\le
Z_{t_0}^2
<
A\eta_{t_0}
\le
C\frac{\sqrt{\log d}+1+t_0\delta}{\sqrt d}
\le
C\frac{\log d}{\sqrt d}.
\]
Since \(\log\gamma\ge c\delta\) for some \(c>0\), we have
\[
\gamma^{2t_0}\ge d^{2cc_6+o(1)} .
\]
Choosing \(c_6\) so large that \(2cc_6>1/2\) gives a contradiction for all
large \(d\). Therefore
\begin{equation}
\label{eq:phase1-sigma-time-clean}
\sigma\le t_0,
\qquad
\sigma\delta\le c_6\log d+O(\delta).
\end{equation}

\paragraph{Step 4: The first stopping event is \(\tau_A\).}

Since \(t_0<T\), \eqref{eq:phase1-sigma-time-clean} implies
\(\sigma\neq T\). By \eqref{eq:phase1-coordinate-positive-clean},
\(\sigma\neq\tau_+\). Thus \(\sigma\) is either \(\tau_A\) or \(\tau_B\).

Suppose for contradiction that \(\sigma=\tau_B\le\tau_A\). Since
\(\tau_B\le t_0\), at time \(\tau_B-1\) we have
\[
Z_{\tau_B-1}^2<A\eta_{\tau_B-1}.
\]
Thus, by \eqref{eq:phase1-eta-bound-clean},
\[
Z_{\tau_B-1}
\le
C\frac{\sqrt{\log d}}{d^{1/4}}
=o(1).
\]
The one-step expansion gives
\[
Z_{\tau_B}
=
Z_{\tau_B-1}
+
\delta F_{\tau_B-1}^{(Z)}
+
\varepsilon_{\tau_B-1}^{(Z)} .
\]
By Lemma~\ref{lem:sgd-tech-bounds},
\[
|F_t^{(Z)}|\le C\sqrt d+C\delta d,
\]
and hence
\[
\delta |F_t^{(Z)}|\le C\delta\sqrt d+C\delta^2d=o(1).
\]
Also,
\[
|\varepsilon_{\tau_B-1}^{(Z)}|
\le
|N_{\tau_B}^{(Z)}|+|N_{\tau_B-1}^{(Z)}|
\le
4\lambda_Z
=o(1).
\]
Therefore \(Z_{\tau_B}=o(1)\). But
\eqref{eq:phase1-coordinate-positive-clean} gives
\(\rho_{\tau_B},\chi_{\tau_B}>0\), and by definition of \(\tau_B\), one of
\(\rho_{\tau_B},\chi_{\tau_B}\) is at least \(1/2\). Hence
\(Z_{\tau_B}\ge1/2\), a contradiction. Therefore
\[
\sigma=\tau_A,
\qquad
\tau_A<\tau_B,
\qquad
\tau_A\delta\le C\log d .
\]
The bounds
\[
\eta_t\le C_\eta\frac{\sqrt{\log d}+1+t\delta}{\sqrt d},
\qquad
\omega_t\le C_\omega\frac{1+t\delta}{\sqrt d}
\]
now hold for all \(t\le\tau_A\), and since \(\tau_A<\tau_B\), also up to
\(\tau_A\wedge\tau_B\).

\paragraph{Step 5: Size of \(Z_{\tau_A}\).}

By definition of \(\tau_A\),
\[
Z_{\tau_A}^2\ge A\eta_{\tau_A}.
\]
Since \(\theta_{\tau_A}\in\mathbb S^{d-1}\),
\[
\|\theta_{\tau_A}\|_\infty\ge d^{-1/2},
\]
and therefore \(\eta_{\tau_A}\ge d^{-1/2}\). Hence
\[
Z_{\tau_A}
\ge
\sqrt A\,d^{-1/4}.
\]
This proves the lower bound with \(c_A:=\sqrt A\).

For the upper bound, first note that \(\tau_A\ge1\) for large \(d\), since
\(Z_0^2=O(d^{-1})\) while \(\eta_0\ge d^{-1/2}\). Hence
\(\tau_A-1<\tau_A\), and so
\[
Z_{\tau_A-1}^2<A\eta_{\tau_A-1}.
\]
Using the \(\eta\)-bound and \(\tau_A\delta\le C\log d\),
\[
Z_{\tau_A-1}
\le
C\frac{\sqrt{\log d}}{d^{1/4}} .
\]
The one-step expansion gives
\[
Z_{\tau_A}
\le
Z_{\tau_A-1}
+
\delta |F_{\tau_A-1}^{(Z)}|
+
|\varepsilon_{\tau_A-1}^{(Z)}| .
\]
As above,
\[
\delta |F_{\tau_A-1}^{(Z)}|
\le
C\delta\sqrt d+C\delta^2d=o(1),
\qquad
|\varepsilon_{\tau_A-1}^{(Z)}|\le4\lambda_Z=o(1).
\]
Therefore
\[
Z_{\tau_A}
\le
C_A\sqrt{\log d}\,d^{-1/4}
+
o(1).
\]
This completes the proof.
\end{proof}

\begin{lemma}[Entry balance at \(\tau_A\)]
\label{lem:sgd-entry-balance}
Work on \(\mathcal G_T^{\rm weak}\), and assume the conclusions of
Lemma~\ref{lem:sgd-phase1}. Let
\[
D_t:=\rho_t-\chi_t,
\qquad
\lambda_D:=\lambda_\rho+\lambda_\chi .
\]
Then
\[
|D_{\tau_A}|
\le
C\frac{(\log d)^2}{\sqrt d}
+
C\lambda_D
+
C\delta d\,\frac{\sqrt{\log d}}{d^{1/4}} .
\]
In particular,
\[
|D_{\tau_A}|=o(Z_{\tau_A}).
\]
Hence, for all sufficiently large \(d\),
\begin{align}
    |D_{\tau_A}|\le \frac18 Z_{\tau_A}.
\end{align}
\end{lemma}

\begin{proof}
By Lemma~\ref{lem:sgd-phase1}, on \(0\le t\le \tau_A\) we have
\[
\tau_A<\tau_B,
\qquad
\tau_+>\tau_A,
\qquad
\tau_A\delta\le C\log d,
\]
and for \(t\le \tau_A\),
\[
\eta_t\le C_\eta\frac{\sqrt{\log d}+1+t\delta}{\sqrt d}.
\]
Moreover, for \(t<\tau_A\),
\[
Z_t^2<A\eta_t.
\]
Therefore, for \(t<\tau_A\),
\[
Z_t
\le
C\frac{\sqrt{\log d}}{d^{1/4}}
=o(1).
\]
Thus, throughout the interval \(0\le t<\tau_A\), the overlaps are nonnegative
and lie in the small-overlap regime. Hence, by \eqref{eq:small-overlap-D-drift} from
Lemma~\ref{lem:refined-small-overlap-drift},
\[
\dot D_t
=
-\widetilde C(1+\rho_t\chi_t)D_t+E_t,
\qquad
|E_t|\le C(Z_t^2+\omega_t).
\]
Now use the one-step SGD expansion. Let
\[
a_t:=\mathbb E[\|g_t^{(\beta)}\|^2\mid\mathcal F_t],
\qquad
b_t:=\mathbb E[\|g_t^{(\theta)}\|^2\mid\mathcal F_t].
\]
Subtracting the one-step overlap expansions gives
\[
D_{t+1}
=
D_t
+
\delta(\dot\rho_t-\dot\chi_t)
-
\frac{\delta^2}{2}(a_t\rho_t-b_t\chi_t)
+
\varepsilon_t^{(D)},
\]
where
\[
\varepsilon_t^{(D)}
:=
N_{t+1}^{(D)}-N_t^{(D)},
\qquad
N_t^{(D)}:=N_t^{(\rho)}-N_t^{(\chi)}.
\]
Decompose
\[
a_t\rho_t-b_t\chi_t
=
\frac{a_t+b_t}{2}D_t+\frac{a_t-b_t}{2}Z_t .
\]
Therefore
\[
D_{t+1}
=
(1-\alpha_t\delta)D_t
+
\delta e_t
+
\varepsilon_t^{(D)},
\tag{1}
\label{eq:entry-D-recursion}
\]
where
\[
\alpha_t
:=
\widetilde C(1+\rho_t\chi_t)
+
\frac{\delta}{4}(a_t+b_t),
\]
and
\[
e_t
:=
E_t-\frac{\delta}{4}(a_t-b_t)Z_t .
\]
By Lemma~\ref{lem:sgd-tech-bounds}, \(a_t,b_t\le Kd\). Hence
\[
|e_t|
\le
C(Z_t^2+\eta_t)+C\delta d\,Z_t .
\tag{2}
\label{eq:entry-e-bound}
\]
Also, since \(\rho_t,\chi_t\ge0\),
\[
\alpha_t\ge \widetilde C.
\]
After reducing the step-size constant, we may assume
\[
0\le 1-\alpha_t\delta\le 1-\alpha_0\delta<1
\]
for a fixed \(\alpha_0>0\).

Unroll \eqref{eq:entry-D-recursion} up to \(t=\tau_A\). Define
\[
P_{m,t}:=\prod_{\ell=m}^{t-1}(1-\alpha_\ell\delta),
\qquad
P_{t,t}:=1.
\]
Then
\[
D_{\tau_A}
=
P_{0,\tau_A}D_0
+
\sum_{k=0}^{\tau_A-1}P_{k+1,\tau_A}\delta e_k
+
\sum_{k=0}^{\tau_A-1}P_{k+1,\tau_A}\varepsilon_k^{(D)} .
\]
Since \(0\le P_{k+1,\tau_A}\le1\),
\[
|P_{0,\tau_A}D_0|\le |D_0|\le C d^{-1/2},
\]
by the favorable initialization event.

For the first forcing term, using \(Z_k^2\le A\eta_k\) for \(k<\tau_A\),
\[
\delta\sum_{k<\tau_A}(Z_k^2+\eta_k)
\le
C\delta\sum_{k<\tau_A}\eta_k .
\]
By the Phase-I \(\eta\)-bound and \(\tau_A\delta\le C\log d\),
\[
\delta\sum_{k<\tau_A}\eta_k
\le
\frac{C}{\sqrt d}
\delta\sum_{k<\tau_A}
\bigl(\sqrt{\log d}+1+k\delta\bigr)
\le
C\frac{(\log d)^2}{\sqrt d}.
\tag{3}
\label{eq:entry-eta-sum}
\]

For the discretization forcing term, since
\[
P_{k+1,\tau_A}
\le
(1-\alpha_0\delta)^{\tau_A-1-k},
\]
we have
\[
\delta\sum_{k<\tau_A}P_{k+1,\tau_A}Z_k
\le
\left(\sup_{k<\tau_A}Z_k\right)
\delta\sum_{j\ge0}(1-\alpha_0\delta)^j
\le
C\sup_{k<\tau_A}Z_k .
\]
As observed above,
\[
\sup_{k<\tau_A}Z_k
\le
C\frac{\sqrt{\log d}}{d^{1/4}}.
\]
Therefore
\[
\delta\sum_{k<\tau_A}P_{k+1,\tau_A}\,\delta d\,Z_k
\le
C\delta d\,\frac{\sqrt{\log d}}{d^{1/4}}.
\tag{4}
\label{eq:entry-discrete-sum}
\]

It remains to control the noise term. For fixed \(\tau_A\), the weights
\[
P_{k+1,\tau_A},
\qquad
0\le k\le \tau_A-1,
\]
lie in \([0,1]\) and are nondecreasing in \(k\). On \(E_T^{(\rho,\chi)}\),
\[
\sup_{u\le T}|N_u^{(D)}|
\le
2(\lambda_\rho+\lambda_\chi)
=
2\lambda_D.
\]
Thus Lemma~\ref{lem:sbp-monotone} gives
\[
\left|
\sum_{k=0}^{\tau_A-1}P_{k+1,\tau_A}\varepsilon_k^{(D)}
\right|
\le
4\lambda_D.
\tag{5}
\label{eq:entry-noise-sum}
\]

Combining \eqref{eq:entry-e-bound}--\eqref{eq:entry-noise-sum}, and absorbing
\(|D_0|\le Cd^{-1/2}\) into the first term, gives
\[
|D_{\tau_A}|
\le
C\frac{(\log d)^2}{\sqrt d}
+
C\lambda_D
+
C\delta d\,\frac{\sqrt{\log d}}{d^{1/4}} .
\]

Finally, Lemma~\ref{lem:sgd-phase1} gives
\[
Z_{\tau_A}\ge c_A d^{-1/4}.
\]
Hence
\[
\frac{|D_{\tau_A}|}{Z_{\tau_A}}
\le
C\frac{(\log d)^2}{d^{1/4}}
+
C\frac{\lambda_D}{d^{-1/4}}
+
C\delta d\sqrt{\log d}.
\]
Since \(\lambda_D=O(d^{-1/2})\) and
\[
\delta d\le \frac{c_0\varepsilon}{\log d},
\]
the right-hand side tends to zero. Therefore
\[
|D_{\tau_A}|=o(Z_{\tau_A}).
\]
In particular, for all sufficiently large \(d\),
\[
|D_{\tau_A}|\le \frac18 Z_{\tau_A}.
\]
\end{proof}

Before Phase II, we provide another lemma below that controls the difference of drift

\begin{lemma}[Balanced \(D\)-drift on the Phase-II band]
\label{lem:D-drift}
Assume Lemma~\ref{lem:uniform-first-order-gradient-comparison}. Let
\(D:=\rho-\chi\), \(Z:=\rho+\chi\), and define
\[
\widetilde C:=2\alpha_1^2a_1^2>0.
\]
There exist constants \(c_{\rm bd}\in(0,1/8)\) and \(C_E>0\), depending only
on the model, such that the following holds. Suppose
\begin{equation}
\label{eq:D-drift-balance-hyp}
0<Z\le 2c_{\rm bd},
\qquad
|D|\le \frac12 Z .
\end{equation}
Then \(\rho,\chi\ge0\), \(|\rho|,|\chi|\le 3c_{\rm bd}/2\), and the population
drifts satisfy
\begin{equation}
\label{eq:D-drift-statement}
\dot\rho-\dot\chi
=
-\widetilde C(1+\rho\chi)D+E,
\end{equation}
where the error obeys the sharper bound
\begin{equation}
\label{eq:D-drift-sharp-remainder}
|E|\le C_E(Z^2+\eta).
\end{equation}
In particular, since \(Z\le 2c_{\rm bd}\), one also has the weaker bound
\begin{equation}
\label{eq:D-drift-band-remainder}
|E|\le C_E(c_{\rm bd}^2+\eta).
\end{equation}
\end{lemma}

\begin{proof}
From \(Z=\rho+\chi\) and \(D=\rho-\chi\),
\[
\rho=\frac{Z+D}{2},
\qquad
\chi=\frac{Z-D}{2}.
\]
The balance condition \(|D|\le Z/2\) gives
\[
\frac14Z\le \rho,\chi\le \frac34Z.
\]
Thus \(\rho,\chi\ge0\), and since \(Z\le2c_{\rm bd}\), we also have
\begin{equation}
\label{eq:D-drift-rhochi-size}
|\rho|,|\chi|\le \frac34Z\le \frac32c_{\rm bd}<\frac12.
\end{equation}

Let \(L_G\) denote the Gaussian-equivalent population loss. As in the proof of
Lemma~\ref{lem:uniform-positive-overlap-drift},
\[
L_G(\theta,\beta)=\Phi(\chi,\rho),
\qquad
\Phi(\chi,\rho)
=
2\sum_{k\ge1}\alpha_k^2\left[1-(\kappa(\rho)\chi)^k\right],
\]
where \(\kappa(\rho)=\sum_{\ell\ge1}a_\ell^2\rho^\ell\). Set
\[
q:=\kappa(\rho)\chi,
\qquad
A(q):=\sum_{k\ge1}k\alpha_k^2q^{k-1}.
\]
Then \(A(0)=\alpha_1^2\). By the Gaussian \(\beta\)-gradient identity
\eqref{eq:gaussian-beta-gradient-exact}, equivalently by differentiating
\(\Phi\) in \(\rho\), and by the corresponding \(\chi\)-derivative displayed in
the proof of Lemma~\ref{lem:uniform-positive-overlap-drift}, the Gaussian-model
overlap drifts are
\begin{equation}
\label{eq:D-drift-gaussian-exact}
\dot\rho_G
=
2\chi\,\kappa'(\rho)A(q)(1-\rho^2),
\qquad
\dot\chi_G
=
2\kappa(\rho)A(q)(1-\chi^2).
\end{equation}

We now expand these identities on the small band
\eqref{eq:D-drift-balance-hyp}. Since \(|\rho|\le 3c_{\rm bd}/2<1/2\), the
Hermite covariance function satisfies
\begin{equation}
\label{eq:D-drift-kappa-expansions}
\kappa(\rho)=a_1^2\rho+O(\rho^2),
\qquad
\kappa'(\rho)=a_1^2+O(|\rho|).
\end{equation}
Moreover, by \eqref{eq:D-drift-rhochi-size}, \(q=\kappa(\rho)\chi=O(Z^2)\), so
\begin{equation}
\label{eq:D-drift-A-expansion}
A(q)=\alpha_1^2+O(Z^2).
\end{equation}
Using \eqref{eq:D-drift-gaussian-exact}--\eqref{eq:D-drift-A-expansion},
\[
\dot\rho_G
=
2\alpha_1^2a_1^2\,\chi(1-\rho^2)+O(Z^2),
\]
because the error from \(\kappa'(\rho)-a_1^2\) is
\(O(|\chi||\rho|)=O(Z^2)\), and the error from \(A(q)-\alpha_1^2\) is
\(O(|\chi|Z^2)=O(Z^3)\). Similarly,
\[
\dot\chi_G
=
2\alpha_1^2a_1^2\,\rho(1-\chi^2)+O(Z^2),
\]
because \(\kappa(\rho)-a_1^2\rho=O(\rho^2)=O(Z^2)\).

Thus, with \(\widetilde C:=2\alpha_1^2a_1^2\),
\begin{equation}
\label{eq:D-drift-gaussian-expanded}
\dot\rho_G
=
\widetilde C\,\chi(1-\rho^2)+O(Z^2),
\qquad
\dot\chi_G
=
\widetilde C\,\rho(1-\chi^2)+O(Z^2).
\end{equation}

It remains to transfer the Gaussian drift estimates to the true population loss.
By \eqref{eq:first-order-beta-gradient-comparison} and
\eqref{eq:first-order-theta-gradient-comparison} from
Lemma~\ref{lem:uniform-first-order-gradient-comparison},
\[
|\dot\rho-\dot\rho_G|\le C\eta,
\qquad
|\dot\chi-\dot\chi_G|\le C\eta.
\]
Combining this with \eqref{eq:D-drift-gaussian-expanded}, we get
\[
\dot\rho-\dot\chi
=
\widetilde C\bigl[\chi(1-\rho^2)-\rho(1-\chi^2)\bigr]
+
O(Z^2+\eta).
\]
Finally,
\begin{equation}
\label{eq:D-drift-cross-identity}
\chi(1-\rho^2)-\rho(1-\chi^2)
=
(\chi-\rho)(1+\rho\chi)
=
-D(1+\rho\chi).
\end{equation}
Therefore
\[
\dot\rho-\dot\chi
=
-\widetilde C(1+\rho\chi)D+E,
\qquad
|E|\le C_E(Z^2+\eta),
\]
which proves \eqref{eq:D-drift-statement} and
\eqref{eq:D-drift-sharp-remainder}. Since \(Z\le 2c_{\rm bd}\),
\eqref{eq:D-drift-band-remainder} follows immediately.
\end{proof}

\subsubsection{Phase II Dynamics, Weak Recovery}

\begin{lemma}[Phase II: signal-dominated amplification and balance]
\label{lem:sgd-phase2}
Assume Lemmas~\ref{lem:sgd-phase1}, \ref{lem:sgd-entry-balance}, and
\ref{lem:D-drift}. Work on \(\mathcal G_T^{\rm weak}\). There exist constants
\[
c_Z\in(0,c_{\rm bd}\wedge 1/8),
\qquad
C_2,C_{\eta,{\rm II}},C_{\rm II}>0,
\]
depending only on the model constants, such that the following holds for all
sufficiently large \(d\). If
\[
C_{\rm II}\log d\le T\delta\le C_T\log d,
\]
and
\[
t^*:=\inf\{t\ge \tau_A: Z_t\ge c_Z\},
\]
then
\[
t^*\le \tau_A+\left\lceil C_2\delta^{-1}\log d\right\rceil,
\qquad
Z_{t^*}\in[c_Z,2c_Z].
\]
Moreover, for every \(\tau_A\le t\le t^*\),
\[
\eta_t\le C_{\eta,{\rm II}}\frac{\log d}{\sqrt d},
\qquad
|D_t|\le \frac12 Z_t.
\]
Consequently,
\begin{align}
\rho_t,\chi_t\ge \frac14 Z_t
\qquad
(\tau_A\le t\le t^*),
\end{align}
with
\begin{align}
t^*<\tau_B,
\qquad
\min\{\rho_{t^*},\chi_{t^*}\}\ge \frac{c_Z}{4}.
\end{align}
\end{lemma}

\begin{proof}
Work on \(\mathcal G_T^{\rm weak}\). By Lemma~\ref{lem:sgd-phase1} and
Lemma~\ref{lem:sgd-entry-balance},
\begin{equation}
\label{eq:phase2-inputs-clean}
\tau_A<\tau_B,\qquad
\tau_A\delta\le C\log d,\qquad
Z_{\tau_A}\ge c_A d^{-1/4},
\end{equation}
and
\begin{equation}
\label{eq:phase2-entry-clean}
\eta_{\tau_A}\le C\frac{\log d}{\sqrt d},
\qquad
Z_{\tau_A}=o(1),
\qquad
|D_{\tau_A}|\le \frac18 Z_{\tau_A}.
\end{equation}
Also, on \(\mathcal G_T^{\rm weak}\),
\[
\lambda_Z,\lambda_D=O(d^{-1/2})=o(Z_{\tau_A}).
\]

Choose
\[
H:=C_{\eta,{\rm II}}\frac{\log d}{\sqrt d},
\qquad
T_2:=\tau_A+\left\lceil C_2\delta^{-1}\log d\right\rceil .
\]
The constants \(C_{\eta,{\rm II}}\), \(c_Z\), and \(C_2\) will be fixed below,
and \(C_{\rm II}\) is chosen large enough so that \(T_2\le T\), using
\(\tau_A\delta\le C\log d\).

Define the bootstrap stopping time
\[
\sigma:=
\inf\left\{
t\ge\tau_A:
\eta_t>H
\ \text{or}\
Z_t>2c_Z
\ \text{or}\
|D_t|>\frac12 Z_t
\ \text{or}\
t=T_2
\right\}.
\]
At \(t=\tau_A\), none of the first three clauses holds for large \(d\). For
\(t<\sigma\),
\[
0<Z_t\le 2c_Z,
\qquad
|D_t|\le \frac12Z_t.
\]
Hence
\begin{equation}
\label{eq:phase2-balanced-band-clean}
\rho_t=\frac{Z_t+D_t}{2},
\qquad
\chi_t=\frac{Z_t-D_t}{2},
\qquad
\frac14Z_t\le \rho_t,\chi_t\le \frac34Z_t.
\end{equation}
Taking \(c_Z<1/3\), this implies \(\rho_t,\chi_t<1/2\), so all one-step
expansions below are valid before \(\sigma\).

\paragraph{Step 1: \(Z\)-growth and no-drop.}

Let
\[
\varepsilon_t^{(Z)}:=N_{t+1}^{(Z)}-N_t^{(Z)}.
\]
For \(t<\sigma\), the one-step overlap expansion and the aggregate drift bound
give
\[
Z_{t+1}
\ge
Z_t+\delta(C_9Z_t-C_{11}\eta_t)
-C\delta^2d\,Z_t
+\varepsilon_t^{(Z)}.
\]
Since \(\delta d=o(1)\), for large \(d\),
\[
C_9-C\delta d\ge \frac34C_9.
\]

Set
\[
\tau_-:=\inf\left\{t\ge\tau_A: Z_t<\frac12Z_{\tau_A}\right\}.
\]
For \(t<\sigma\wedge\tau_-\), we have \(Z_t\ge Z_{\tau_A}/2\) and
\(\eta_t\le H\). Therefore
\[
\frac{\eta_t}{Z_t}
\le
\frac{2H}{Z_{\tau_A}}
\le
C\frac{\log d}{d^{1/4}}
=o(1).
\]
Thus, for large \(d\),
\[
C_{11}\eta_t\le \frac14C_9Z_t,
\]
and hence
\begin{equation}
\label{eq:phase2-Z-step-clean}
Z_{t+1}\ge (1+\beta\delta)Z_t+\varepsilon_t^{(Z)},
\qquad
t<\sigma\wedge\tau_-,
\end{equation}
where \(\beta:=C_9/2\). Let \(\gamma:=1+\beta\delta\).

Unrolling \eqref{eq:phase2-Z-step-clean}, for every \(s\ge0\) with
\(\tau_A+s<\sigma\wedge\tau_-\),
\[
Z_{\tau_A+s}
\ge
\gamma^s
\left(
Z_{\tau_A}
+
\sum_{k=0}^{s-1}
\gamma^{-(k+1)}\varepsilon_{\tau_A+k}^{(Z)}
\right).
\]
On \(E_T^{(Z)}\), \eqref{eq:sup-N-lambda} gives 
\[
\sup_{u\le T}|N_u^{(Z)}|\le 2\lambda_Z.
\]
By Corollary~\ref{lem:sbp} applied on the shifted interval starting at
\(\tau_A\),
\[
\left|
\sum_{k=0}^{s-1}
\gamma^{-(k+1)}\varepsilon_{\tau_A+k}^{(Z)}
\right|
\le 4\lambda_Z.
\]
Therefore
\begin{equation}
\label{eq:phase2-Z-exp-clean}
Z_{\tau_A+s}
\ge
\bigl(Z_{\tau_A}-4\lambda_Z\bigr)\gamma^s.
\end{equation}
Since \(\lambda_Z=o(Z_{\tau_A})\), the right side is at least
\((3/4)Z_{\tau_A}\gamma^s\). This rules out \(\tau_-\le\sigma\). Hence
\[
Z_t\ge \frac12Z_{\tau_A}
\qquad
(\tau_A\le t<\sigma),
\]
and \eqref{eq:phase2-Z-exp-clean} holds for every \(\tau_A+s<\sigma\).

We shall use the following summability estimates. Summing
\eqref{eq:phase2-Z-step-clean} from \(\tau_A\) to \(t-1\), for \(t<\sigma\),
gives
\[
\beta\delta\sum_{k=\tau_A}^{t-1}Z_k
\le
Z_t-Z_{\tau_A}
-
\bigl(N_t^{(Z)}-N_{\tau_A}^{(Z)}\bigr).
\]
Since \(Z_t\le 2c_Z\) and
\(|N_t^{(Z)}-N_{\tau_A}^{(Z)}|\le 4\lambda_Z\),
\begin{equation}
\label{eq:phase2-Z-sum-clean}
\delta\sum_{k=\tau_A}^{t-1}Z_k\le Cc_Z.
\end{equation}
Also, since \(Z_k\le2c_Z\),
\begin{equation}
\label{eq:phase2-Z2-sum-clean}
\delta\sum_{k=\tau_A}^{t-1}Z_k^2\le Cc_Z^2.
\end{equation}

Finally, the backward form of \eqref{eq:phase2-Z-step-clean} gives, for
\(\tau_A\le k<t<\sigma\),
\begin{equation}
\label{eq:phase2-backward-Z-clean}
Z_k\le \gamma^{-(t-k)}Z_t+C\lambda_Z.
\end{equation}
Indeed, unroll \eqref{eq:phase2-Z-step-clean} from \(k\) to \(t\), divide by
\(\gamma^{t-k}\), and apply Corollary~\ref{lem:sbp} to the shifted weighted
noise sum.

\paragraph{Step 2: Closing the \(\eta\)-bootstrap.}

Since \(E_T^{(\eta)}\) controls the cumulative coordinate martingales from
time \(0\), for every \(t\ge\tau_A\) and every \(i\le d\),
\[
\left|
\delta\sum_{k=\tau_A}^{t-1}\xi_{k,i}^{(\theta)}
\right|
\le
\left|
\delta\sum_{k<t}\xi_{k,i}^{(\theta)}
\right|
+
\left|
\delta\sum_{k<\tau_A}\xi_{k,i}^{(\theta)}
\right|
\le 2\Lambda_\eta .
\]
Given that the quadratic and cubic remainder sums over \([\tau_A,t]\) are bounded by their
global good-event bounds, therefore, applying Lemma~\ref{lem:sgd-evolution-eta} on the interval
\([\tau_A,t]\), for every \(\tau_A\le t\le\sigma\),
\[
\eta_t
\le
C\left(
\eta_{\tau_A}
+
\delta\sum_{k=\tau_A}^{t-1}\|\bar g_k^{(\theta)}\|_\infty
+
2\Lambda_\eta
+
\lambda_{\rm cub}
\right).
\]
After increasing \(C\), we write this as
\[
\eta_t
\le
C\left(
\eta_{\tau_A}
+
\delta\sum_{k=\tau_A}^{t-1}\|\bar g_k^{(\theta)}\|_\infty
+
\Lambda_\eta
+
\lambda_{\rm cub}
\right).
\]
On the Phase-II band, the mean-field \(\ell_\infty\)-gradient bound gives
\[
\|\bar g_k^{(\theta)}\|_\infty
\le
C Z_k\|\theta^*\|_\infty
+
C Z_k^2\eta_k
+
C\eta_k^2.
\]
Using \(\|\theta^*\|_\infty\le C\sqrt{\log d/d}\) and
\eqref{eq:phase2-Z-sum-clean},
\[
\delta\sum_{k=\tau_A}^{t-1}Z_k\|\theta^*\|_\infty
\le
Cc_Z\sqrt{\frac{\log d}{d}}
=o(H).
\]
By \eqref{eq:phase2-Z2-sum-clean} and the bootstrap bound \(\eta_k\le H\),
\[
\delta\sum_{k=\tau_A}^{t-1}Z_k^2\eta_k
\le
H\delta\sum_{k=\tau_A}^{t-1}Z_k^2
\le
Cc_Z^2H.
\]
Also,
\[
\delta\sum_{k=\tau_A}^{t-1}\eta_k^2
\le
H^2(t-\tau_A)\delta
\le
C\frac{(\log d)^3}{d}
=o(H).
\]
Together with
\[
\eta_{\tau_A}+\lambda_\eta(t)+\lambda_{\rm cub}
\le
C\frac{\log d}{\sqrt d},
\]
we obtain
\[
\eta_t
\le
C\frac{\log d}{\sqrt d}
+
Cc_Z^2H
+
o(H).
\]
Choose \(c_Z\) small enough that \(Cc_Z^2\le1/8\), then choose
\(C_{\eta,{\rm II}}\) large enough that \(C(\log d)/\sqrt d\le H/4\). For
large \(d\),
\[
\eta_t\le \frac12H
\qquad
(\tau_A\le t\le\sigma).
\]
Thus the stopping clause \(\eta_t>H\) cannot be the first active clause before
\(t^*\).

\paragraph{Step 3: Closing the \(D/Z\)-bootstrap.}

For \(t<\sigma\), choose \(c_Z\le c_{\rm bd}\). Then
\eqref{eq:phase2-balanced-band-clean} allows us to apply Lemma~\ref{lem:D-drift}:
\[
\dot D_t
=
-\widetilde C(1+\rho_t\chi_t)D_t+E_t,
\qquad
|E_t|\le C(Z_t^2+\eta_t).
\]
The discrete one-step expansion for \(D_t\) has the form
\begin{equation}
\label{eq:phase2-D-rec-clean}
D_{t+1}
=
(1-\alpha_t\delta)D_t+\delta e_t+\varepsilon_t^{(D)},
\end{equation}
where
\[
\varepsilon_t^{(D)}:=N_{t+1}^{(D)}-N_t^{(D)},
\]
\[
\alpha_t
=
\widetilde C(1+\rho_t\chi_t)
+
\frac{\delta}{4}
\left(
\E[\|g_t^{(\beta)}\|^2\mid\mathcal F_t]
+
\E[\|g_t^{(\theta)}\|^2\mid\mathcal F_t]
\right),
\]
and
\begin{equation}
\label{eq:phase2-e-bound-clean}
|e_t|\le C(Z_t^2+\eta_t)+C\delta d\,Z_t.
\end{equation}
Here the last term comes from the difference of the two normalization
corrections. Since \(c_Z\) is small, \(\alpha_t\ge\alpha_0>0\), and after
decreasing the step-size constant,
\[
0\le 1-\alpha_t\delta\le 1-\alpha_0\delta=:\gamma_D<1.
\]

Unroll \eqref{eq:phase2-D-rec-clean}. For \(\tau_A\le t\le\sigma\), define
\[
P_{k+1,t}:=\prod_{\ell=k+1}^{t-1}(1-\alpha_\ell\delta),
\qquad
\tau_A\le k\le t-1,
\]
with \(P_{t,t}=1\). Then
\[
D_t
=
P_{\tau_A,t}D_{\tau_A}
+
\sum_{k=\tau_A}^{t-1}P_{k+1,t}\delta e_k
+
\sum_{k=\tau_A}^{t-1}P_{k+1,t}\varepsilon_k^{(D)}.
\]

The entry term is controlled by Lemma~\ref{lem:sgd-entry-balance} and the
no-drop estimate:
\[
|P_{\tau_A,t}D_{\tau_A}|
\le
|D_{\tau_A}|
\le
\frac18Z_{\tau_A}
\le
\frac14Z_t.
\]

For the noise term, the weights \(P_{k+1,t}\) lie in \([0,1]\) and are
nondecreasing in \(k\). On \(E_T^{(\rho,\chi)}\),
\[
\sup_{u\le T}|N_u^{(D)}|\le 2\lambda_D.
\]
Therefore Lemma~\ref{lem:sbp-monotone} gives
\[
\left|
\sum_{k=\tau_A}^{t-1}P_{k+1,t}\varepsilon_k^{(D)}
\right|
\le
4\lambda_D
=
o(Z_{\tau_A})
=
o(Z_t).
\]

It remains to bound the forcing. Using \eqref{eq:phase2-backward-Z-clean},
\[
Z_k\le \gamma^{-(t-k)}Z_t+C\lambda_Z.
\]
Since \(P_{k+1,t}\le \gamma_D^{t-1-k}\), we have
\[
\delta\sum_{k=\tau_A}^{t-1}P_{k+1,t}Z_k
\le
C\delta\sum_{j\ge0}\gamma_D^j\gamma^{-j}Z_t
+
C\lambda_Z\delta\sum_{j\ge0}\gamma_D^j
\le
CZ_t+C\lambda_Z.
\]
Since \(\lambda_Z=o(Z_t)\), this is \(CZ_t+o(Z_t)\). Similarly,
\[
\delta\sum_{k=\tau_A}^{t-1}P_{k+1,t}Z_k^2
\le
CZ_t^2+C\lambda_Z^2
=
CZ_t^2+o(Z_t).
\]
Also, by the \(\eta\)-bootstrap,
\[
\delta\sum_{k=\tau_A}^{t-1}P_{k+1,t}\eta_k
\le
CH
=
o(Z_{\tau_A})
=
o(Z_t).
\]
Using \eqref{eq:phase2-e-bound-clean}, we conclude
\[
\sum_{k=\tau_A}^{t-1}P_{k+1,t}\delta |e_k|
\le
Cc_ZZ_t+o(Z_t)+C\delta d\,Z_t.
\]
Since \(\delta d=o(1)\),
\[
\sum_{k=\tau_A}^{t-1}P_{k+1,t}\delta |e_k|
\le
Cc_ZZ_t+o(Z_t).
\]
Thus
\[
|D_t|
\le
\frac14Z_t+Cc_ZZ_t+o(Z_t).
\]
Choose \(c_Z\) smaller, if necessary, so that \(Cc_Z\le1/16\), and then take
\(d\) large enough so that the \(o(Z_t)\)-term is at most \(Z_t/16\). Hence
\[
|D_t|\le \frac38Z_t<\frac12Z_t
\qquad
(\tau_A\le t\le\sigma).
\]
Therefore the stopping clause \(|D_t|>Z_t/2\) cannot be the first active clause
before \(t^*\).

\paragraph{Step 4: Hitting \(c_Z\) before \(T_2\).}

From \eqref{eq:phase2-Z-exp-clean},
\[
Z_{\tau_A+s}
\ge
\bigl(Z_{\tau_A}-4\lambda_Z\bigr)\gamma^s
\]
as long as \(\tau_A+s<\sigma\). Define
\[
s_*:=
\left\lceil
\frac{\log\!\left(c_Z/(Z_{\tau_A}-4\lambda_Z)\right)}{\log\gamma}
\right\rceil.
\]
Since \(Z_{\tau_A}-4\lambda_Z\ge c d^{-1/4}\) and
\(\log\gamma\ge c\delta\), we have
\[
s_*\le C\delta^{-1}\log d.
\]
Choose \(C_2\) larger than this constant.

Suppose, for contradiction, that \(t^*>\tau_A+s_*\). Then \(Z_t<c_Z\) up to
\(\tau_A+s_*\), so the \(Z_t>2c_Z\) clause cannot trigger before that time.
By Steps 2 and 3, the \(\eta\)- and \(D/Z\)-clauses cannot trigger before
\(t^*\). By the choice of \(C_2\), the time clause \(t=T_2\) also cannot trigger
before \(\tau_A+s_*\). Hence \(\tau_A+s_*<\sigma\), and
\[
Z_{\tau_A+s_*}
\ge
\bigl(Z_{\tau_A}-4\lambda_Z\bigr)\gamma^{s_*}
\ge c_Z,
\]
contradicting the definition of \(t^*\). Therefore
\[
t^*\le \tau_A+s_*
\le
\tau_A+\left\lceil C_2\delta^{-1}\log d\right\rceil .
\]

\paragraph{Step 5: Overshoot and conclusion.}

Since \(Z_{\tau_A}=o(1)\), for large \(d\), \(Z_{\tau_A}<c_Z\), so
\(t^*>\tau_A\) and \(Z_{t^*-1}<c_Z\). The one-step expansion gives
\[
Z_{t^*}
=
Z_{t^*-1}
+
\delta F_{t^*-1}^{(Z)}
+
\varepsilon_{t^*-1}^{(Z)}.
\]
By Lemma~\ref{lem:sgd-tech-bounds},
\[
|F_t^{(Z)}|
\le C\sqrt d+C\delta d.
\]
Hence
\[
\delta |F_t^{(Z)}|
\le C\delta\sqrt d+C\delta^2d=o(1).
\]
Also, on \(E_T^{(Z)}\),
\[
|\varepsilon_t^{(Z)}|
\le
|N_{t+1}^{(Z)}|+|N_t^{(Z)}|
\le 4\lambda_Z=o(1).
\]
Therefore
\[
Z_{t^*}\le c_Z+o(1)\le 2c_Z
\]
for all large \(d\). By definition, \(Z_{t^*}\ge c_Z\), so
\[
Z_{t^*}\in[c_Z,2c_Z].
\]

The bootstrap conclusions give, for \(\tau_A\le t\le t^*\),
\[
\eta_t\le H=C_{\eta,{\rm II}}\frac{\log d}{\sqrt d},
\qquad
|D_t|\le\frac12Z_t.
\]
Thus
\[
\rho_t=\frac{Z_t+D_t}{2}\ge \frac14Z_t,
\qquad
\chi_t=\frac{Z_t-D_t}{2}\ge \frac14Z_t.
\]
At \(t=t^*\),
\[
\min\{\rho_{t^*},\chi_{t^*}\}
\ge
\frac14Z_{t^*}
\ge
\frac{c_Z}{4}.
\]
Moreover,
\[
\max\{\rho_{t^*},\chi_{t^*}\}
\le
\frac{Z_{t^*}+|D_{t^*}|}{2}
\le
\frac34Z_{t^*}
\le
\frac32c_Z
<
\frac12,
\]
after choosing \(c_Z<1/3\). Hence \(t^*<\tau_B\). This completes the proof.
\end{proof}

\subsubsection{Weak Recovery Theorem}

Now, we're prepared to derive the final theorem for weak recovery, which states:
\begin{theorem}[Escape from mediocrity under discrete-time SGD]
\label{thm:sgd-escape-mediocrity-pullup}
Assume Assumption~\ref{ass:sgd-regime} and the auxiliary estimates used in
Lemmas~\ref{lem:sgd-phase1}, \ref{lem:sgd-entry-balance}, and
\ref{lem:sgd-phase2}. Let \(\mathcal E_0\) be the favorable initialization
event. Then there exist constants
\[
c_{\rm esc},C_{\rm time},K>0
\]
depending only on the model constants such that, for all sufficiently large
\(d\), the following holds. Let
\[
T:=\left\lceil C_{\rm time}\delta^{-1}\log d\right\rceil .
\]
Conditional on \(\mathcal E_0\), with probability at least \(1-K\varepsilon\),
there exists a time
\begin{align}
    t_{\rm esc}\le C_{\rm time}\delta^{-1}\log d
\end{align}
such that
\begin{align}
    \min\{\rho_{t_{\rm esc}},\chi_{t_{\rm esc}}\}\ge c_{\rm esc}.
\end{align}
Moreover, throughout the escape window,
\begin{equation}
\label{eq:sgd-weak-eta-final}
\eta_t\le C\frac{\log d}{\sqrt d},
\qquad
0\le t\le t_{\rm esc}.
\end{equation}
Equivalently, if \(\mathbb P(\mathcal E_0)\ge p_0>0\), then
\begin{align}
\mathbb P\left(
\exists\,t_{\rm esc}\le C_{\rm time}\delta^{-1}\log d:
\min\{\rho_{t_{\rm esc}},\chi_{t_{\rm esc}}\}\ge c_{\rm esc}
\right)
\ge
p_0(1-K\varepsilon).
\end{align}
\end{theorem}

\begin{proof}
Choose \(C_{\rm time}\) large enough so that
\[
C_{\rm time}>C_{\rm I}+C_{\rm II}+2,
\]
where \(C_{\rm I}\) is the Phase-I time constant from
Lemma~\ref{lem:sgd-phase1}, and \(C_{\rm II}\) is the Phase-II time constant
from Lemma~\ref{lem:sgd-phase2}. Increasing \(C_{\rm time}\) if necessary, the
horizon \(T\) satisfies all horizon assumptions required in the Phase-I,
entry-balance, and Phase-II lemmas.

Define the weak-recovery good event, excluding initialization, by
\[
\mathcal H_T^{\rm weak}
:=
E_T^{(T)}
\cap E_T^{(Z)}
\cap E_T^{(\eta)}
\cap E_T^{(\rho,\chi)}
\cap E_T^{(\Omega)} .
\]
By bounds of the Taylor \eqref{eq:taylor-bound}, accumulative \eqref{eq:Z-union-bound} and difference \eqref{eq:rho-chi-bound} martingale and discretization control, \(\eta\)-control \eqref{eq:eta-union-bound}, \eqref{eq:omega-bound} and third-moment estimates,
\begin{equation}
\label{eq:weak-good-probability}
\mathbb P\left((\mathcal H_T^{\rm weak})^c\mid\mathcal E_0\right)
\le K\varepsilon .
\end{equation}
Set
\[
\mathcal G_T^{\rm weak}:=\mathcal E_0\cap\mathcal H_T^{\rm weak}.
\]
We now work on \(\mathcal G_T^{\rm weak}\).

By Lemma~\ref{lem:sgd-phase1},
\begin{equation}
\label{eq:weak-thm-phase1-inputs}
\tau_A<\tau_B,
\qquad
\tau_A\delta\le C_{\rm I}\log d,
\qquad
Z_{\tau_A}\ge c_A d^{-1/4}.
\end{equation}
The same lemma also gives, for \(0\le t\le\tau_A\),
\begin{equation}
\label{eq:weak-thm-phase1-eta}
\eta_t
\le
C_\eta\frac{\sqrt{\log d}+1+t\delta}{\sqrt d}.
\end{equation}

Next, Lemma~\ref{lem:sgd-entry-balance} gives
\[
|D_{\tau_A}|=o(Z_{\tau_A}).
\]
Hence, for all sufficiently large \(d\),
\begin{equation}
\label{eq:weak-thm-entry-balance}
|D_{\tau_A}|\le \frac18 Z_{\tau_A}.
\end{equation}

Now apply Lemma~\ref{lem:sgd-phase2}. Define
\[
t^*
:=
\inf\{t\ge \tau_A:Z_t\ge c_Z\},
\]
where \(c_Z>0\) is the constant from Lemma~\ref{lem:sgd-phase2}. The Phase-II
lemma gives
\[
t^*
\le
\tau_A+\left\lceil C_{\rm II}\delta^{-1}\log d\right\rceil,
\]
\[
Z_{t^*}\in[c_Z,2c_Z],
\]
and, for every \(\tau_A\le t\le t^*\),
\begin{equation}
\label{eq:weak-thm-phase2-bounds}
\eta_t\le C_{\eta,{\rm II}}\frac{\log d}{\sqrt d},
\qquad
|D_t|\le \frac12Z_t.
\end{equation}
Therefore,
\[
t^*
\le
C_{\rm time}\delta^{-1}\log d
\]
for all sufficiently large \(d\).

Set
\[
t_{\rm esc}:=t^*,
\qquad
c_{\rm esc}:=\frac{c_Z}{4}.
\]
Since
\[
\rho_t=\frac{Z_t+D_t}{2},
\qquad
\chi_t=\frac{Z_t-D_t}{2},
\]
the balance estimate at \(t=t^*\) gives
\[
\min\{\rho_{t_{\rm esc}},\chi_{t_{\rm esc}}\}
=
\frac{Z_{t_{\rm esc}}-|D_{t_{\rm esc}}|}{2}
\ge
\frac14 Z_{t_{\rm esc}}
\ge
\frac{c_Z}{4}
=
c_{\rm esc}.
\]
This proves weak recovery on \(\mathcal G_T^{\rm weak}\).

It remains to verify the claimed \(\eta_t\)-bound. For \(0\le t\le\tau_A\),
\eqref{eq:weak-thm-phase1-eta} and \(\tau_A\delta\le C_{\rm I}\log d\) imply
\[
\eta_t\le C\frac{\log d}{\sqrt d}.
\]
For \(\tau_A\le t\le t_{\rm esc}\), the same bound follows from
\eqref{eq:weak-thm-phase2-bounds}. Hence
\[
\eta_t\le C\frac{\log d}{\sqrt d},
\qquad
0\le t\le t_{\rm esc}.
\]

Finally, the conditional probability statement follows from
\eqref{eq:weak-good-probability}. If \(\mathbb P(\mathcal E_0)\ge p_0\), then
\[
\mathbb P(\mathcal E_0\cap\mathcal H_T^{\rm weak})
=
\mathbb P(\mathcal H_T^{\rm weak}\mid\mathcal E_0)\mathbb P(\mathcal E_0)
\ge
p_0(1-K\varepsilon),
\]
which gives the unconditional statement.
\end{proof}

\subsection{Strong Recovery Setups}

\begin{remark}[Local strong-recovery step-size regime]\label{rmk:optimal-delta-strong}
    Entering strong recovery phase, we changed the step size to \[\delta_{\rm loc}:=\frac{c_{\rm loc}}d,\] where \(c_{\rm loc}>0\) is chosen sufficiently small so that $C_{\rm ret}C_{\rm sq}\,\delta_{\rm loc}d\le \mu_{\rm loc}$ for the optimum convergence rate: our one-step recursion is $Q_{t+1} \leq Q_t +\delta \langle \bar{g}_t, \text{Log}_t \rangle+\delta \Xi_t+C_{\text{ret}}\delta^2 G_t$ With $\langle \bar{g}_t, \text{Log}_t \rangle \leq -2\mu_{\text{loc}}Q_t$ and $\mathbb{E}[G_t\mid \mathcal{F}_t]\leq C_{\text{sq}}dQ_t,$ we have in expectation $\mathbb{E}[Q_{t+1}\mid \mathcal{F}_t]\leq (1-2\mu_{\text{loc}}Q_t+Cd\delta^2)Q_t,$ which means by taking $\delta\asymp \frac{1}{d},$ we will have the maximized contraction rate of $O(d \log \frac{1}{\varepsilon})$ samples. 
\end{remark}

\paragraph{Population inputs reused from the continuous-time analysis.}
Throughout this subsection, let \(r_{\rm loc}>0\) and \(\mu_{\rm loc}>0\) be the
constants from Lemma~\ref{lem:true-local-product-strong-convexity}. In
particular, for every
\[
y\in \mathcal U_{r_{\rm loc}},
\qquad
\mathcal U_{r_{\rm loc}}
:=
\{y\in\mathcal M:\operatorname{dist}_{\mathcal M}(y,y^*)\le r_{\rm loc}\},
\]
we have
\[
\operatorname{Hess}_{\mathcal M}L(y)\succeq \mu_{\rm loc} I.
\]

\paragraph{Discrete-time basin-entry.}

Mirroring  Lemma~\ref{lem:entry-local-cap-after-corrected-weak}, the following lemmas establish that after an additional \(O(d)\) iterations beyond weak recovery, the trajectory enters the local strongly-convex cap in discrete-time:

\begin{lemma}[Entry-window delocalization and overlap-noise control]
\label{lem:entry-window-control}
Work on the weak-recovery event at \(t_{\rm esc}\). Assume the incoherence bound \ref{ass:favorable-initialization}. After \(t_{\rm esc}\), run SGD with the local step size $\delta_{\rm loc}=\frac{c_{\rm loc}}d,$
where \(c_{\rm loc}>0\) is chosen sufficiently small depending only on the
model constants, \(C_{\rm in}\), and the desired entry-window failure level.
Fix \(C_{\rm in}<\infty\), and set $N_{\rm in}:=\left\lceil C_{\rm in}\delta_{\rm loc}^{-1}\right\rceil .$
Then, conditionally on \(\mathcal F_{t_{\rm esc}}\), there exists an event
\(E_{\rm in}\), measurable with respect to the samples in
\([t_{\rm esc},t_{\rm esc}+N_{\rm in}]\), such that
\begin{align}
\mathbb P(E_{\rm in}\mid \mathcal F_{t_{\rm esc}})
\ge
1-\nu_{\rm in}.
\end{align}
On \(E_{\rm in}\),
\begin{align}
\sup_{t_{\rm esc}\le t\le t_{\rm esc}+N_{\rm in}}\eta_t
\le
\bar\eta_d,
\end{align}
where, with \(L_{\nu,d}:=1+\log(d/\nu_{\rm in})\),
\begin{align}
\bar\eta_d
\le
C\left(
\frac{\log d}{\sqrt d}
+
\sqrt{\frac{\log d}{d}}
+
\sqrt{\delta_{\rm loc}L_{\nu,d}}
+
L_{\nu,d}\delta_{\rm loc}^2d^{3/2}
\right).
\end{align}
In particular, for fixed or polynomially small \(\nu_{\rm in}\),
\begin{align}
    \bar\eta_d
=
O\left(\frac{\log d}{\sqrt d}\right)
=
o_d(1).
\end{align}
Moreover, for \(j\in\{\rho,\chi\}\),
\begin{align}
\sup_{t_{\rm esc}\le s\le u\le t_{\rm esc}+N_{\rm in}}
\left|
\sum_{k=s}^{u-1}\Delta N_{k+1}^{(j)}
\right|
\le
\Lambda_d,
\end{align}
where
\begin{align}
\Lambda_d
\le
C\left(
\sqrt{\delta_{\rm loc}\log(1/\nu_{\rm in})}
+
L_{\nu,d}\delta_{\rm loc}^2d^{3/2}
\right)
=
o_d(1)
\end{align}
for fixed or polynomially small \(\nu_{\rm in}\).
\end{lemma}

\begin{proof}
Write
\[
t_0:=t_{\rm esc},
\qquad
\delta:=\delta_{\rm loc},
\qquad
N:=N_{\rm in}
=
\left\lceil C_{\rm in}\delta^{-1}\right\rceil .
\]
Throughout the proof we condition on \(\mathcal F_{t_0}\) and on the
weak-recovery event stated in the lemma. Since the samples after \(t_0\) are
fresh and independent of \(\mathcal F_{t_0}\), the martingale and moment
estimates from the weak-recovery section apply conditionally on
\(\mathcal F_{t_0}\), with the same constants.

We shall use the normalized spherical one-step expansion
\[
\rho_{t+1}
=
\rho_t+\delta F_t^{(\rho)}+\Delta N_{t+1}^{(\rho)},
\qquad
\chi_{t+1}
=
\chi_t+\delta F_t^{(\chi)}+\Delta N_{t+1}^{(\chi)},
\]
where
\[
\Delta N_{t+1}^{(j)}
=
\delta M_t^{(j)}+R_t^{(j)},
\qquad
j\in\{\rho,\chi\}.
\]
Here \(M_t^{(j)}\) is a scalar martingale increment and \(R_t^{(j)}\) is the
normalization/retraction remainder.

\medskip
\noindent
\textbf{Step 1: overlap-noise control.}

By the scalar martingale moment bounds used in the weak-recovery proof, there
is \(K_M<\infty\), independent of \(d\), such that
\[
\mathbb E\!\left[(M_t^{(j)})^2\mid\mathcal F_t\right]
\le
K_M(1+\delta^2d^2),
\qquad
j\in\{\rho,\chi\}.
\]
Since \(\delta d=c_{\rm loc}\), this is bounded by another constant \(K_M'\).

For \(u\in[t_0,t_0+N]\), define
\[
S_u^{(j)}
:=
\delta\sum_{k=t_0}^{u-1}M_k^{(j)}.
\]
Then
\[
\langle S^{(j)}\rangle_{t_0+N}
\le
K_M'\delta^2N
\le
C\delta,
\]
because \(N\le C\delta^{-1}\). By the same conditional Freedman/truncation
argument used for the weak-recovery overlap martingales, with conditional
probability at least \(1-\nu_{\rm in}/8\),
\[
\sup_{t_0\le u\le t_0+N}
\left|
\delta\sum_{k=t_0}^{u-1}M_k^{(j)}
\right|
\le
C\sqrt{\delta\log(1/\nu_{\rm in})}.
\]
Consequently,
\[
\sup_{t_0\le s\le u\le t_0+N}
\left|
\delta\sum_{k=s}^{u-1}M_k^{(j)}
\right|
\le
C\sqrt{\delta\log(1/\nu_{\rm in})},
\]
after increasing \(C\).

Next consider the remainders. On the local Taylor event for the normalized
spherical step,
\[
|R_t^{(\rho)}|
\le
C\delta^3\|g_t^{(\beta)}\|^3,
\qquad
|R_t^{(\chi)}|
\le
C\delta^3\|g_t^{(\theta)}\|^3.
\]
Moreover, the gradient moment bounds give
\[
\mathbb E[\|g_t^{(\beta)}\|^3\mid\mathcal F_t]
+
\mathbb E[\|g_t^{(\theta)}\|^3\mid\mathcal F_t]
\le
Cd^{3/2}.
\]
Hence, for \(j\in\{\rho,\chi\}\),
\[
\mathbb E\!\left[
\sum_{k=t_0}^{t_0+N-1}|R_k^{(j)}|
\,\middle|\,
\mathcal F_{t_0}
\right]
\le
CN\delta^3d^{3/2}
\le
C\delta^2d^{3/2}.
\]
Using the same high-moment/truncation estimate as in the weak-recovery
discretization control, there is an event of conditional probability at least
\(1-\nu_{\rm in}/8\) on which
\[
\sup_{t_0\le s\le u\le t_0+N}
\left|
\sum_{k=s}^{u-1}R_k^{(j)}
\right|
\le
CL_{\nu,d}\delta^2d^{3/2},
\qquad
L_{\nu,d}:=1+\log(d/\nu_{\rm in}).
\]
Therefore, on an event \(E_{\rm ov}\) of conditional probability at least
\(1-\nu_{\rm in}/2\), for both \(j\in\{\rho,\chi\}\),
\[
\sup_{t_0\le s\le u\le t_0+N}
\left|
\sum_{k=s}^{u-1}\Delta N_{k+1}^{(j)}
\right|
\le
\Lambda_d,
\]
where
\[
\Lambda_d
:=
C\left(
\sqrt{\delta\log(1/\nu_{\rm in})}
+
L_{\nu,d}\delta^2d^{3/2}
\right).
\]
Since \(\delta=c_{\rm loc}/d\),
\[
\Lambda_d
\le
C\sqrt{\frac{\log(1/\nu_{\rm in})}{d}}
+
C L_{\nu,d}\frac1{\sqrt d}
=
o_d(1)
\]
for fixed or polynomially small \(\nu_{\rm in}\).

\medskip
\noindent
\textbf{Step 2: \(\eta_t\)-control on the entry window.}

To control $\eta_t=\|\theta_t\|_\infty+\|\theta^*\|_\infty,$ we use the same \(\ell_\infty\) inequality as in the weak-recovery analysis:
\[
\|\theta_t\|_\infty
\le
\|\theta_{t_0}\|_\infty
+
\delta\sum_{k=t_0}^{t-1}\|\bar g_k^{(\theta)}\|_\infty
+
\max_{i\le d}
\left|
\delta\sum_{k=t_0}^{t-1}\xi_{k,i}^{(\theta)}
\right|
+
\frac{\delta^2}{2}
\sum_{k=t_0}^{t-1}\|g_k^{(\theta)}\|^2\|\theta_k\|_\infty
+
\frac{\delta^3}{2}
\sum_{k=t_0}^{t-1}\|g_k^{(\theta)}\|^3,
\]
where
\[
\bar g_k^{(\theta)}
:=
\mathbb E[g_k^{(\theta)}\mid\mathcal F_k],
\qquad
\xi_k^{(\theta)}
:=
g_k^{(\theta)}-\bar g_k^{(\theta)}.
\]
Set
\[
\Theta_u
:=
\sup_{t_0\le t\le u}\eta_t,
\qquad
t_0\le u\le t_0+N.
\]

By the conditional coordinate Freedman bound, with conditional probability at
least \(1-\nu_{\rm in}/8\),
\[
\max_{i\le d}
\sup_{t_0\le u\le t_0+N}
\left|
\delta\sum_{k=t_0}^{u-1}\xi_{k,i}^{(\theta)}
\right|
\le
C\sqrt{\delta L_{\nu,d}}.
\]
The gradient moment estimates imply
\[
\mathbb E\!\left[
\delta^2\sum_{k=t_0}^{t_0+N-1}\|g_k^{(\theta)}\|^2
\,\middle|\,
\mathcal F_{t_0}
\right]
\le
C\delta^2Nd
\le
C\delta d
=
Cc_{\rm loc}.
\]
Choosing \(c_{\rm loc}>0\) sufficiently small, the same concentration/Markov
argument used in the weak phase gives, with conditional probability at least
\(1-\nu_{\rm in}/8\),
\[
\frac{\delta^2}{2}
\sum_{k=t_0}^{t_0+N-1}\|g_k^{(\theta)}\|^2
\le
\frac14 .
\]
Similarly,
\[
\mathbb E\!\left[
\delta^3\sum_{k=t_0}^{t_0+N-1}\|g_k^{(\theta)}\|^3
\,\middle|\,
\mathcal F_{t_0}
\right]
\le
C\delta^3Nd^{3/2}
\le
C\delta^2d^{3/2},
\]
and hence, on an event of conditional probability at least \(1-\nu_{\rm in}/8\),
\[
\delta^3
\sum_{k=t_0}^{t_0+N-1}\|g_k^{(\theta)}\|^3
\le
CL_{\nu,d}\delta^2d^{3/2}.
\]

Let \(E_{\eta,{\rm noise}}\) be the intersection of these three
\(\eta\)-control events. On \(E_{\eta,{\rm noise}}\), for every
\(u\in[t_0,t_0+N]\),
\[
\Theta_u
\le
C\eta_{t_0}
+
C\delta\sum_{k=t_0}^{u-1}\|\bar g_k^{(\theta)}\|_\infty
+
C\sqrt{\delta L_{\nu,d}}
+
CL_{\nu,d}\delta^2d^{3/2}
+
\frac14\Theta_u.
\]
Absorbing the last term,
\begin{equation}
\label{eq:entry-eta-prefield}
\Theta_u
\le
C\eta_{t_0}
+
C\delta\sum_{k=t_0}^{u-1}\|\bar g_k^{(\theta)}\|_\infty
+
C\sqrt{\delta L_{\nu,d}}
+
CL_{\nu,d}\delta^2d^{3/2}.
\end{equation}

It remains to bound the mean-field term. The population \(\ell_\infty\)
gradient estimate gives, uniformly along the entry window,
\[
\|\bar g_k^{(\theta)}\|_\infty
\le
C\|\theta^*\|_\infty
+
C\eta_k
+
C\eta_k^2.
\]
Therefore, since \(N\delta\le C_{\rm in}+1\), \eqref{eq:entry-eta-prefield}
implies
\[
\Theta_u
\le
A_d
+
C\delta\sum_{k=t_0}^{u-1}\Theta_k
+
C\delta\sum_{k=t_0}^{u-1}\Theta_k^2,
\]
where
\[
A_d
:=
C\left(
\eta_{t_0}
+
\|\theta^*\|_\infty
+
\sqrt{\delta L_{\nu,d}}
+
L_{\nu,d}\delta^2d^{3/2}
\right).
\]

Now insert the scales. On the weak-recovery event,
\[
\eta_{t_0}
\le
C_{\eta,{\rm esc}}\frac{\log d}{\sqrt d},
\]
and by the teacher incoherence assumption,
\[
\|\theta^*\|_\infty
\le
C_*\sqrt{\frac{\log d}{d}}.
\]
Also \(\delta=c_{\rm loc}/d\). Hence, for fixed or polynomially small
\(\nu_{\rm in}\),
\[
A_d
\le
C\left(
\frac{\log d}{\sqrt d}
+
\sqrt{\frac{\log d}{d}}
+
\sqrt{\frac{L_{\nu,d}}d}
+
\frac{L_{\nu,d}}{\sqrt d}
\right)
=
O\left(\frac{\log d}{\sqrt d}\right)
=
o_d(1).
\]

We close the bootstrap. Define
\[
\tau_\eta
:=
\inf\{u\in[t_0,t_0+N]:\Theta_u\ge 1\}
\wedge(t_0+N).
\]
For \(u\le\tau_\eta\), one has \(\Theta_k^2\le\Theta_k\), and therefore
\[
\Theta_u
\le
A_d
+
C\delta\sum_{k=t_0}^{u-1}\Theta_k.
\]
The discrete Gronwall inequality gives
\[
\Theta_u
\le
A_d\exp(C(u-t_0)\delta)
\le
A_d\exp(C(C_{\rm in}+1))
\le
C_{\rm in}'A_d,
\qquad
u\le\tau_\eta.
\]
Since \(A_d=o_d(1)\), for all sufficiently large \(d\) the right-hand side is
strictly smaller than \(1\). Hence \(\tau_\eta=t_0+N\), and
\[
\sup_{t_0\le t\le t_0+N}\eta_t
=
\Theta_{t_0+N}
\le
\bar\eta_d,
\]
where
\[
\bar\eta_d
:=
C_{\rm in}'A_d
\le
C\left(
\frac{\log d}{\sqrt d}
+
\sqrt{\frac{\log d}{d}}
+
\sqrt{\delta L_{\nu,d}}
+
L_{\nu,d}\delta^2d^{3/2}
\right).
\]
This proves the claimed delocalization bound.

\medskip
\noindent
\textbf{conclusion.}

Define
\[
E_{\rm in}:=
E_{\rm ov}\cap E_{\eta,{\rm noise}}.
\]
By the conditional probability bounds above and a union bound, after adjusting
constants,
\[
\mathbb P(E_{\rm in}\mid\mathcal F_{t_0})
\ge
1-\nu_{\rm in}.
\]
On \(E_{\rm in}\), both conclusions hold:
\[
\sup_{t_0\le t\le t_0+N}\eta_t\le\bar\eta_d
\]
and, for \(j\in\{\rho,\chi\}\),
\[
\sup_{t_0\le s\le u\le t_0+N}
\left|
\sum_{k=s}^{u-1}\Delta N_{k+1}^{(j)}
\right|
\le
\Lambda_d.
\]
Substituting back \(t_0=t_{\rm esc}\), \(N=N_{\rm in}\), and
\(\delta=\delta_{\rm loc}\) completes the proof.
\end{proof}

\begin{lemma}[One-step overlap recursion on the entry corridor]
\label{lem:entry-overlap-recursion}
Assume Lemma~\ref{lem:uniform-positive-overlap-drift}. After \(t_{\rm esc}\),
run SGD with step size \(\delta=\delta_{\rm loc}\). On any time \(t\) in the
entry window for which the one-step Taylor expansion is valid and
\[
\rho_t\ge0,\qquad \chi_t\ge0,
\]
the overlap recursions can be written as
\[
\rho_{t+1}
=
\rho_t+\delta F_t^{(\rho)}+\Delta N_{t+1}^{(\rho)},
\qquad
\chi_{t+1}
=
\chi_t+\delta F_t^{(\chi)}+\Delta N_{t+1}^{(\chi)},
\]
where
\[
\Delta N_{t+1}^{(\rho)}
=
\delta M_t^{(\rho)}+R_t^{(\rho)},
\qquad
\Delta N_{t+1}^{(\chi)}
=
\delta M_t^{(\chi)}+R_t^{(\chi)}.
\]
Here \(M_t^{(\rho)},M_t^{(\chi)}\) are martingale increments and the remainders
satisfy
\[
|R_t^{(\rho)}|
\le
C\delta^3\|g_t^{(\beta)}\|^3,
\qquad
|R_t^{(\chi)}|
\le
C\delta^3\|g_t^{(\theta)}\|^3.
\]
Moreover, the predictable drifts satisfy
\[
F_t^{(\rho)}
\ge
a_\rho\chi_t(1-\rho_t^2)-b_\rho\eta_t-C\delta d,
\]
and
\[
F_t^{(\chi)}
\ge
a_\chi\rho_t(1-\chi_t^2)-b_\chi\eta_t-C\delta d.
\]
Consequently, on the event \(E_{\rm in}\) from
Lemma~\ref{lem:entry-window-control},
\[
F_t^{(\rho)}
\ge
a_\rho\chi_t(1-\rho_t^2)-B_d,
\qquad
F_t^{(\chi)}
\ge
a_\chi\rho_t(1-\chi_t^2)-B_d,
\]
where
\[
B_d:=C(\bar\eta_d+\delta d).
\]
In particular, for \(\delta=c_{\rm loc}/d\),
\[
B_d
\le
C\left(\frac{\log d}{\sqrt d}+c_{\rm loc}\right)
\]
for fixed or polynomially small \(\nu_{\rm in}\).
\end{lemma}

\begin{proof}
We prove the \(\rho\)-recursion; the proof for \(\chi\) is identical.

By the normalized spherical update,
\[
\beta_{t+1}
=
\frac{\beta_t-\delta g_t^{(\beta)}}
{\|\beta_t-\delta g_t^{(\beta)}\|}.
\]
Since \(g_t^{(\beta)}\perp\beta_t\), on the one-step Taylor event
\(\delta\|g_t^{(\beta)}\|\le 1/2\), we have
\[
\beta_{t+1}
=
\beta_t
-\delta g_t^{(\beta)}
-\frac{\delta^2}{2}\|g_t^{(\beta)}\|^2\beta_t
+
O(\delta^3\|g_t^{(\beta)}\|^3).
\]
Taking inner product with \(\beta^*\), and using
\(\rho_t=\langle\beta_t,\beta^*\rangle\), gives
\[
\rho_{t+1}
=
\rho_t
-\delta\langle g_t^{(\beta)},\beta^*\rangle
-\frac{\delta^2}{2}\rho_t\|g_t^{(\beta)}\|^2
+
R_t^{(\rho)},
\]
where
\[
|R_t^{(\rho)}|
\le
C\delta^3\|g_t^{(\beta)}\|^3.
\]
Equivalently,
\[
\rho_{t+1}
=
\rho_t
-
\delta
\left(
\langle g_t^{(\beta)},\beta^*\rangle
+
\frac{\delta}{2}\rho_t\|g_t^{(\beta)}\|^2
\right)
+
R_t^{(\rho)}.
\]

Define
\[
F_t^{(\rho)}
:=
-
\mathbb E\!\left[
\langle g_t^{(\beta)},\beta^*\rangle
+
\frac{\delta}{2}\rho_t\|g_t^{(\beta)}\|^2
\,\middle|\,
\mathcal F_t
\right],
\]
and
\[
M_t^{(\rho)}
:=
-
\left[
\langle g_t^{(\beta)},\beta^*\rangle
+
\frac{\delta}{2}\rho_t\|g_t^{(\beta)}\|^2
-
\mathbb E\!\left[
\langle g_t^{(\beta)},\beta^*\rangle
+
\frac{\delta}{2}\rho_t\|g_t^{(\beta)}\|^2
\,\middle|\,
\mathcal F_t
\right]
\right].
\]
Then
\[
\mathbb E[M_t^{(\rho)}\mid\mathcal F_t]=0,
\]
and
\[
\rho_{t+1}
=
\rho_t+\delta F_t^{(\rho)}+\delta M_t^{(\rho)}+R_t^{(\rho)}.
\]
Thus
\[
\Delta N_{t+1}^{(\rho)}
:=
\delta M_t^{(\rho)}+R_t^{(\rho)}.
\]

It remains to lower-bound \(F_t^{(\rho)}\). By definition,
\[
F_t^{(\rho)}
=
-
\mathbb E[\langle g_t^{(\beta)},\beta^*\rangle\mid\mathcal F_t]
-
\frac{\delta}{2}\rho_t
\mathbb E[\|g_t^{(\beta)}\|^2\mid\mathcal F_t].
\]
The first term is the population spherical-gradient drift:
\[
-
\mathbb E[\langle g_t^{(\beta)},\beta^*\rangle\mid\mathcal F_t]
=
-
\left\langle
\nabla_{\mathbb S,\beta}L(\theta_t,\beta_t),
\beta^*
\right\rangle.
\]
Since \(\rho_t,\chi_t\ge0\), Lemma~\ref{lem:uniform-positive-overlap-drift}
implies
\[
-
\left\langle
\nabla_{\mathbb S,\beta}L(\theta_t,\beta_t),
\beta^*
\right\rangle
\ge
a_\rho\chi_t(1-\rho_t^2)-b_\rho\eta_t.
\]
For the normalization correction, use \(|\rho_t|\le1\) and the conditional
second-moment gradient bound:
\[
-\frac{\delta}{2}\rho_t
\mathbb E[\|g_t^{(\beta)}\|^2\mid\mathcal F_t]
\ge
-\frac{\delta}{2}
\mathbb E[\|g_t^{(\beta)}\|^2\mid\mathcal F_t]
\ge
-C\delta d.
\]
Combining the two estimates gives
\[
F_t^{(\rho)}
\ge
a_\rho\chi_t(1-\rho_t^2)-b_\rho\eta_t-C\delta d.
\]

The \(\chi\)-recursion follows from the same argument applied to
\[
\theta_{t+1}
=
\frac{\theta_t-\delta g_t^{(\theta)}}
{\|\theta_t-\delta g_t^{(\theta)}\|}.
\]
Thus
\[
\chi_{t+1}
=
\chi_t+\delta F_t^{(\chi)}+\Delta N_{t+1}^{(\chi)},
\]
with
\[
F_t^{(\chi)}
\ge
a_\chi\rho_t(1-\chi_t^2)-b_\chi\eta_t-C\delta d.
\]

Finally, on \(E_{\rm in}\),
\[
\eta_t\le\bar\eta_d
\]
throughout the entry window. Increasing \(C\) to absorb \(b_\rho,b_\chi\), the
two drift bounds become
\[
F_t^{(\rho)}
\ge
a_\rho\chi_t(1-\rho_t^2)-C(\bar\eta_d+\delta d),
\]
and
\[
F_t^{(\chi)}
\ge
a_\chi\rho_t(1-\chi_t^2)-C(\bar\eta_d+\delta d).
\]
This proves the stated bounds with
\[
B_d:=C(\bar\eta_d+\delta d).
\]
Since Lemma~\ref{lem:entry-window-control} gives
\(\bar\eta_d=O(\log d/\sqrt d)\) for fixed or polynomially small
\(\nu_{\rm in}\), and since \(\delta=c_{\rm loc}/d\), we obtain
\[
B_d
\le
C\left(\frac{\log d}{\sqrt d}+c_{\rm loc}\right).
\]
The proof is complete.
\end{proof}

\begin{proposition}[Discrete entry into the local product basin]
\label{prop:discrete-entry-local-basin}
Let \(r_{\rm loc}\) be the radius from
Lemma~\ref{lem:true-local-product-strong-convexity}. After possibly shrinking
\(r_{\rm loc}\), assume \(0<r_{\rm loc}<\pi\). Define
\[
s_b:=\cos(r_{\rm loc}/8),
\qquad
s_+:=\frac{1+s_b}{2}.
\]
Apply Lemma~\ref{lem:entry-window-control} with a constant
\(C_{\rm in}\) at least as large as the fixed constant \(C_{\rm ent}\)
constructed in the proof below.

Then there exist constants \(c_{\rm loc,0}>0\), \(d_0<\infty\), and
\(C_{\rm ent}<\infty\), depending only on \(r_{\rm loc}\), \(c_{\rm esc}\),
and the model constants, such that the following holds. If
\[
\delta_{\rm loc}=\frac{c_{\rm loc}}d,
\qquad
0<c_{\rm loc}\le c_{\rm loc,0},
\qquad
d\ge d_0,
\]
then on the event \(E_{\rm in}\) from Lemma~\ref{lem:entry-window-control},
there exists a time $t_1\le t_{\rm esc}+C_{\rm ent}\delta_{\rm loc}^{-1}$
such that
\[
\rho_{t_1}\ge s_b,
\qquad
\chi_{t_1}\ge s_b.
\]
Consequently,
\[
Q_{t_1}
=
\frac12\operatorname{dist}_{\mathbb S}(\theta_{t_1},\theta^*)^2
+
\frac12\operatorname{dist}_{\mathbb S}(\beta_{t_1},\beta^*)^2
\le
\frac{r_{\rm loc}^2}{64}
\le
\frac{r_{\rm loc}^2}{32}.
\]
\end{proposition}

\begin{proof}
Write
\[
t_0:=t_{\rm esc},
\qquad
\delta:=\delta_{\rm loc},
\qquad
c_-:=\frac{c_{\rm esc}}2.
\]
We work deterministically on the event \(E_{\rm in}\), together with the weak
recovery event at \(t_{\rm esc}\). Thus
\[
\rho_{t_0}\ge c_{\rm esc},
\qquad
\chi_{t_0}\ge c_{\rm esc},
\]
and, throughout the entry window,
\[
\eta_t\le\bar\eta_d,
\qquad
\sup_{t_0\le s\le u\le t_0+C_{\rm in}\delta^{-1}}
\left|
\sum_{k=s}^{u-1}\Delta N_{k+1}^{(j)}
\right|
\le \Lambda_d,
\qquad j\in\{\rho,\chi\}.
\]

By Lemma~\ref{lem:entry-overlap-recursion}, on \(E_{\rm in}\),
\[
F_t^{(\rho)}
\ge
a_\rho\chi_t(1-\rho_t^2)-B_d,
\qquad
F_t^{(\chi)}
\ge
a_\chi\rho_t(1-\chi_t^2)-B_d,
\]
where
\[
B_d:=C(\bar\eta_d+\delta d).
\]
Since
\[
\bar\eta_d
=
O\left(\frac{\log d}{\sqrt d}\right),
\qquad
\delta d=c_{\rm loc},
\]
we may choose \(c_{\rm loc}>0\) sufficiently small and then \(d\) sufficiently
large so that all smallness conditions below hold.

Define
\[
m_+:=1-s_+^2>0,
\]
and set
\[
\gamma_\rho:=\frac12 a_\rho c_-m_+,
\qquad
\gamma_\chi:=\frac12 a_\chi c_-m_+.
\]
These are fixed positive constants independent of \(d\). Also define
\[
\alpha_*:=
\frac14
\min\left\{
c_{\rm esc},
s_+-s_b,
\gamma_\rho,
\gamma_\chi
\right\}
>0.
\]
Let
\[
T_\rho
:=
\frac{(s_+-c_{\rm esc})_+ +\alpha_*}{\gamma_\rho},
\qquad
T_\chi
:=
\frac{(s_+-c_{\rm esc})_+ +\alpha_*}{\gamma_\chi},
\]
and choose
\[
C_{\rm ent}:=
2+\max\{T_\rho,T_\chi\}.
\]
We choose \(c_{\rm loc}\) small enough and then \(d\) large enough so that
\begin{align}
B_d&\le \min\{\gamma_\rho,\gamma_\chi\},
\label{eq:entry-local-small-1}\\
\Lambda_d&\le \alpha_*,
\label{eq:entry-local-small-2}\\
C_{\rm ent}B_d+\Lambda_d&\le \alpha_*.
\label{eq:entry-local-small-3}
\end{align}
These conditions are consistent because \(B_d\le C(\log d/\sqrt d+c_{\rm loc})\)
and \(\Lambda_d=o_d(1)\).

Let
\[
N_{\rm ent}:=\left\lceil \max\{T_\rho,T_\chi\}\delta^{-1}\right\rceil .
\]
Since \(\delta\le1\), we have
\[
N_{\rm ent}\delta\le C_{\rm ent}.
\]
By the assumption \(C_{\rm in}\ge C_{\rm ent}\), all estimates from
\(E_{\rm in}\) hold on \([t_0,t_0+N_{\rm ent}]\).

\medskip
\noindent
\textbf{Step 1: the overlaps stay above the lower barrier.}

We claim that
\[
\rho_t\ge c_-,
\qquad
\chi_t\ge c_-,
\qquad
t_0\le t\le t_0+N_{\rm ent}.
\]
Suppose not, and define the first lower-barrier exit time
\[
\tau_-:=
\inf\left\{
t\in[t_0,t_0+N_{\rm ent}]:
\min\{\rho_t,\chi_t\}<c_-
\right\}.
\]
Since \(\rho_{t_0},\chi_{t_0}\ge c_{\rm esc}>c_-\), one has
\(\tau_->t_0\) if \(\tau_-<\infty\).

For every \(k<\tau_-\), both overlaps are at least \(c_-\), and hence
nonnegative. Therefore the crude drift bounds
\[
F_k^{(\rho)}\ge -B_d,
\qquad
F_k^{(\chi)}\ge -B_d
\]
hold. Summing the \(\rho\)-recursion from \(t_0\) to \(\tau_-\), and using the
cumulative noise bound from \(E_{\rm in}\), gives
\[
\rho_{\tau_-}
\ge
\rho_{t_0}
-
(\tau_--t_0)\delta B_d
-
\Lambda_d.
\]
Since \((\tau_--t_0)\delta\le C_{\rm ent}\), \eqref{eq:entry-local-small-3}
implies
\[
\rho_{\tau_-}
\ge
c_{\rm esc}-C_{\rm ent}B_d-\Lambda_d
\ge
c_{\rm esc}-\alpha_*
\ge
\frac34c_{\rm esc}
>
c_-.
\]
The same argument gives \(\chi_{\tau_-}>c_-\). This contradicts the definition
of \(\tau_-\). Hence no such exit occurs.

\medskip
\noindent
\textbf{Step 2: each overlap reaches the buffer level \(s_+\).}

First consider \(\rho_t\). If \(\rho_{t_0}\ge s_+\), set
\(\tau_\rho:=t_0\). Otherwise, suppose for contradiction that
\[
\rho_t<s_+
\qquad
\text{for all }
t_0\le t\le t_0+\left\lceil T_\rho\delta^{-1}\right\rceil .
\]
By Step 1, \(\chi_t\ge c_-\) on this interval. Hence, for every such \(k\),
\[
F_k^{(\rho)}
\ge
a_\rho c_-(1-\rho_k^2)-B_d.
\]
Since \(\rho_k<s_+\), we have \(1-\rho_k^2\ge 1-s_+^2=m_+\), and therefore
\[
F_k^{(\rho)}
\ge
a_\rho c_-m_+-B_d
=
2\gamma_\rho-B_d
\ge
\gamma_\rho
\]
by \eqref{eq:entry-local-small-1}. Summing the recursion gives
\[
\rho_{t_0+\lceil T_\rho\delta^{-1}\rceil}
\ge
\rho_{t_0}
+
\gamma_\rho T_\rho
-
\Lambda_d.
\]
Using \(\rho_{t_0}\ge c_{\rm esc}\), the definition of \(T_\rho\), and
\eqref{eq:entry-local-small-2},
\[
\rho_{t_0+\lceil T_\rho\delta^{-1}\rceil}
\ge
c_{\rm esc}
+
(s_+-c_{\rm esc})_+
+
\alpha_*
-
\Lambda_d
\ge
s_+.
\]
This contradicts the assumption that \(\rho_t<s_+\) throughout the interval.
Thus there exists
\[
\tau_\rho
\le
t_0+\left\lceil T_\rho\delta^{-1}\right\rceil
\]
such that
\[
\rho_{\tau_\rho}\ge s_+.
\]

The proof for \(\chi_t\) is identical. If \(\chi_{t_0}\ge s_+\), set
\(\tau_\chi:=t_0\). Otherwise, using Step 1 and the lower bound for
\(F_t^{(\chi)}\), one obtains a time
\[
\tau_\chi
\le
t_0+\left\lceil T_\chi\delta^{-1}\right\rceil
\]
such that
\[
\chi_{\tau_\chi}\ge s_+.
\]

\medskip
\noindent
\textbf{Step 3: retain both overlaps above \(s_b\).}

Define
\[
t_1:=\max\{\tau_\rho,\tau_\chi\}.
\]
Then
\[
t_1
\le
t_0+N_{\rm ent}
\le
t_{\rm esc}+C_{\rm ent}\delta^{-1}.
\]

We show that both overlaps are at least \(s_b\) at time \(t_1\). Since
\(\rho_{\tau_\rho}\ge s_+\), and since Step 1 gives the lower barrier on the
whole window, the crude drift bound \(F_k^{(\rho)}\ge -B_d\) holds on
\([\tau_\rho,t_1]\). Therefore
\[
\rho_{t_1}
\ge
\rho_{\tau_\rho}
-
(t_1-\tau_\rho)\delta B_d
-
\Lambda_d.
\]
As \((t_1-\tau_\rho)\delta\le C_{\rm ent}\), \eqref{eq:entry-local-small-3}
gives
\[
\rho_{t_1}
\ge
s_+-C_{\rm ent}B_d-\Lambda_d
\ge
s_+-\alpha_*.
\]
By the definition of \(\alpha_*\),
\[
\alpha_*\le \frac14(s_+-s_b),
\]
so
\[
\rho_{t_1}\ge s_b.
\]
The same argument gives
\[
\chi_{t_1}\ge s_b.
\]

\medskip
\noindent
\textbf{Step 4: convert overlap entry into product-distance entry.}

Since
\[
\rho_{t_1}=\langle \beta_{t_1},\beta^*\rangle\ge s_b=\cos(r_{\rm loc}/8),
\]
and \(0<r_{\rm loc}<\pi\),
\[
\operatorname{dist}_{\mathbb S}(\beta_{t_1},\beta^*)
=
\arccos(\rho_{t_1})
\le
\frac{r_{\rm loc}}8.
\]
Similarly,
\[
\operatorname{dist}_{\mathbb S}(\theta_{t_1},\theta^*)
\le
\frac{r_{\rm loc}}8.
\]
Therefore
\[
Q_{t_1}
=
\frac12\operatorname{dist}_{\mathbb S}(\theta_{t_1},\theta^*)^2
+
\frac12\operatorname{dist}_{\mathbb S}(\beta_{t_1},\beta^*)^2
\le
\frac12\left(\frac{r_{\rm loc}}8\right)^2
+
\frac12\left(\frac{r_{\rm loc}}8\right)^2
=
\frac{r_{\rm loc}^2}{64}.
\]
This proves the proposition.
\end{proof}

\paragraph{Deriving \(Q_t\)-iteration and its controlling bounds.}

\begin{lemma}[One-step local distance recursion]
\label{lem:one-step-distance-recursion}
Let
\[
y_t=(\theta_t,\beta_t),
\qquad
y^*=(\theta^*,\beta^*),
\qquad
g_t=(g_t^{(\theta)},g_t^{(\beta)}),
\]
and define
\[
Q_t
:=
\frac12\operatorname{dist}_{\mathcal M}(y_t,y^*)^2
=
\frac12\operatorname{dist}_{\mathbb S}(\theta_t,\theta^*)^2
+
\frac12\operatorname{dist}_{\mathbb S}(\beta_t,\beta^*)^2.
\]
Assume \(r_{\rm loc}<\pi/4\). There exists a constant \(C_{\rm ret}<\infty\),
depending only on \(r_{\rm loc}\), such that whenever
\(y_t\in\mathcal U_{r_{\rm loc}}\), the normalized SGD step satisfies
\[
Q_{t+1}
\le
Q_t
+
\delta\langle g_t,\mathcal L_t\rangle
+
C_{\rm ret}\delta^2\|g_t\|^2,
\]
where
\[
\mathcal L_t
:=
\left(
\operatorname{Log}_{\theta_t}(\theta^*),
\operatorname{Log}_{\beta_t}(\beta^*)
\right).
\]
Consequently, if
\[
\bar g_t:=\mathbb E[g_t\mid\mathcal F_t],
\qquad
\Xi_t:=\langle g_t-\bar g_t,\mathcal L_t\rangle,
\qquad
G_t:=\|g_t\|^2,
\]
then
\begin{equation}
\label{eq:q-t-one-step-recursion}
Q_{t+1}
\le
Q_t
+
\delta\langle \bar g_t,\mathcal L_t\rangle
+
\delta\Xi_t
+
C_{\rm ret}\delta^2G_t.
\end{equation}
\end{lemma}

\begin{proof}
We use the product metric on
\[
\mathcal M=\mathbb S^{d-1}\times\mathbb S^{d-1},
\]
namely
\[
\langle (a_\theta,a_\beta),(b_\theta,b_\beta)\rangle
:=
\langle a_\theta,b_\theta\rangle
+
\langle a_\beta,b_\beta\rangle,
\]
and
\[
\|(a_\theta,a_\beta)\|^2
:=
\|a_\theta\|^2+\|a_\beta\|^2.
\]

Fix \(t\) such that \(y_t\in\mathcal U_{r_{\rm loc}}\). We prove the one-step
estimate on one sphere and then add the two block estimates.

Let \(u\in\mathbb S^{d-1}\), let \(v\in\mathbb S^{d-1}\), and let
\(h\in T_u\mathbb S^{d-1}\). Consider the normalized spherical step
\[
\widehat u
:=
\frac{u-\delta h}{\|u-\delta h\|}.
\]
Since \(h\perp u\),
\[
\|u-\delta h\|^2=1+\delta^2\|h\|^2.
\]
Define
\[
\tau(h)
:=
\begin{cases}
\dfrac{\arctan(\delta\|h\|)}{\|h\|}, & h\neq 0,\\[1.2ex]
0, & h=0.
\end{cases}
\]
Then
\[
\widehat u
=
\operatorname{Exp}_u(-\tau(h)h).
\]
Indeed, when \(h\neq0\), the normalized point lies on the great circle through
\(u\) in direction \(-h/\|h\|\), with angular displacement
\(\arctan(\delta\|h\|)\). The case \(h=0\) is immediate.

Now set
\[
q_v(u):=\frac12\operatorname{dist}_{\mathbb S}(u,v)^2.
\]
Since \(u\) lies in the local cap around \(v\), and since the normalized step
has spherical length at most \(\pi/2\), after shrinking \(r_{\rm loc}\) if
necessary the geodesic segment used above remains in a compact subset away
from the cut locus of \(v\). Therefore \(q_v\) has uniformly bounded Hessian
on this region:
\[
\|\operatorname{Hess}_{\mathbb S}q_v\|\le C_{\rm sph}.
\]
Taylor expansion along
\[
s\mapsto \operatorname{Exp}_u(-s\tau(h)h)
\]
gives
\[
q_v(\widehat u)
\le
q_v(u)
+
\left\langle \nabla_{\mathbb S}q_v(u),-\tau(h)h\right\rangle
+
\frac{C_{\rm sph}}2\tau(h)^2\|h\|^2.
\]
Since
\[
\nabla_{\mathbb S}q_v(u)
=
-\operatorname{Log}_u(v),
\]
we get
\[
q_v(\widehat u)
\le
q_v(u)
+
\tau(h)\langle h,\operatorname{Log}_u(v)\rangle
+
\frac{C_{\rm sph}}2\tau(h)^2\|h\|^2.
\]

We compare \(\tau(h)\) with \(\delta\). Since
\[
0\le \tau(h)\le \delta,
\]
and, writing \(x=\delta\|h\|\),
\[
0\le \delta\|h\|-\arctan(\delta\|h\|)
=
x-\arctan x
\le Cx^2,
\]
we have
\[
|\delta-\tau(h)|\,\|h\|
\le
C\delta^2\|h\|^2.
\]
Also \(u\) is within distance \(r_{\rm loc}\) of \(v\), so
\[
\|\operatorname{Log}_u(v)\|
=
\operatorname{dist}_{\mathbb S}(u,v)
\le r_{\rm loc}.
\]
Therefore
\[
\tau(h)\langle h,\operatorname{Log}_u(v)\rangle
\le
\delta\langle h,\operatorname{Log}_u(v)\rangle
+
Cr_{\rm loc}\delta^2\|h\|^2.
\]
Moreover,
\[
\tau(h)^2\|h\|^2
\le
\delta^2\|h\|^2.
\]
Thus
\[
q_v(\widehat u)
\le
q_v(u)
+
\delta\langle h,\operatorname{Log}_u(v)\rangle
+
C_{\rm ret}^{(1)}\delta^2\|h\|^2,
\]
where \(C_{\rm ret}^{(1)}\) depends only on \(r_{\rm loc}\) and \(C_{\rm sph}\).

Apply this estimate first with
\[
u=\theta_t,
\qquad
v=\theta^*,
\qquad
h=g_t^{(\theta)},
\]
and then with
\[
u=\beta_t,
\qquad
v=\beta^*,
\qquad
h=g_t^{(\beta)}.
\]
The normalized SGD updates are
\[
\theta_{t+1}
=
\frac{\theta_t-\delta g_t^{(\theta)}}
{\|\theta_t-\delta g_t^{(\theta)}\|},
\qquad
\beta_{t+1}
=
\frac{\beta_t-\delta g_t^{(\beta)}}
{\|\beta_t-\delta g_t^{(\beta)}\|}.
\]
Hence
\[
\frac12\operatorname{dist}_{\mathbb S}(\theta_{t+1},\theta^*)^2
\le
\frac12\operatorname{dist}_{\mathbb S}(\theta_t,\theta^*)^2
+
\delta
\left\langle
g_t^{(\theta)},
\operatorname{Log}_{\theta_t}(\theta^*)
\right\rangle
+
C_{\rm ret}^{(1)}\delta^2\|g_t^{(\theta)}\|^2,
\]
and
\[
\frac12\operatorname{dist}_{\mathbb S}(\beta_{t+1},\beta^*)^2
\le
\frac12\operatorname{dist}_{\mathbb S}(\beta_t,\beta^*)^2
+
\delta
\left\langle
g_t^{(\beta)},
\operatorname{Log}_{\beta_t}(\beta^*)
\right\rangle
+
C_{\rm ret}^{(1)}\delta^2\|g_t^{(\beta)}\|^2.
\]
Adding the two estimates gives
\[
Q_{t+1}
\le
Q_t
+
\delta\langle g_t,\mathcal L_t\rangle
+
C_{\rm ret}\delta^2\|g_t\|^2,
\]
after increasing the constant \(C_{\rm ret}\).

Finally,
\[
g_t=\bar g_t+(g_t-\bar g_t),
\]
so
\[
\langle g_t,\mathcal L_t\rangle
=
\langle \bar g_t,\mathcal L_t\rangle
+
\langle g_t-\bar g_t,\mathcal L_t\rangle
=
\langle \bar g_t,\mathcal L_t\rangle+\Xi_t.
\]
With \(G_t=\|g_t\|^2\), this yields
\[
Q_{t+1}
\le
Q_t
+
\delta\langle \bar g_t,\mathcal L_t\rangle
+
\delta\Xi_t
+
C_{\rm ret}\delta^2G_t,
\]
which proves \eqref{eq:q-t-one-step-recursion}.
\end{proof}

\begin{lemma}[Sharp local self-quenching]
\label{lem:sharp-local-self-quenching-sgd}
Inside \(\mathcal U_{r_{\rm loc}}\), there exists \(C_{\rm sq}>0\), independent
of \(d\), such that
\[
\mathbb E[G_t\mid\mathcal F_t]\le C_{\rm sq}d\,Q_t,
\]
\[
\mathbb E[\Xi_t^2\mid\mathcal F_t]\le C_{\rm sq}Q_t^2,
\]
and
\[
\mathbb E\!\left[
\left(G_t-\mathbb E[G_t\mid\mathcal F_t]\right)^2
\mid\mathcal F_t
\right]
\le
C_{\rm sq}d^2Q_t^2.
\]
\end{lemma}

\begin{proof}
Condition on \(\mathcal F_t\), so that \((\theta_t,\beta_t)\) are deterministic.
To lighten notation, write
\[
\theta:=\theta_t,\qquad \beta:=\beta_t,
\]
and set
\[
d_\theta:=\operatorname{dist}_{\mathbb S}(\theta,\theta^*),\qquad
d_\beta:=\operatorname{dist}_{\mathbb S}(\beta,\beta^*).
\]
Since
\[
Q_t=\frac12\bigl(d_\theta^2+d_\beta^2\bigr),
\]
we have
\[
d_\theta^2+d_\beta^2=2Q_t.
\]

Let
\[
a:=\operatorname{Log}_\theta(\theta^*)\in T_\theta\mathbb S^{d-1},
\qquad
b:=\operatorname{Log}_\beta(\beta^*)\in T_\beta\mathbb S^{d-1},
\]
so that
\[
\|a\|=d_\theta,\qquad \|b\|=d_\beta.
\]

Write the one-sample prediction as
\[
f_{\beta,\theta}(x)=\psi(s),
\qquad
s:=z_\beta(x)\cdot\theta
=
\sum_{i=1}^d \theta_i\,\phi(u_i),
\qquad
u_i:=x^{(i)}\cdot\beta,
\]
and the teacher prediction as
\[
f_{\beta^*,\theta^*}(x)=\psi(s^*),
\qquad
s^*:=z_{\beta^*}(x)\cdot\theta^*
=
\sum_{i=1}^d \theta_i^*\,\phi(u_i^*),
\qquad
u_i^*:=x^{(i)}\cdot\beta^*.
\]
Define the residual
\[
R:=f_{\beta,\theta}(x)-f_{\beta^*,\theta^*}(x)=\psi(s)-\psi(s^*).
\]

The spherical one-sample gradients are
\[
g_t^{(\theta)}
=
2R\,\psi'(s)\,P_\theta z_\beta(x),
\]
and
\[
g_t^{(\beta)}
=
2R\,\psi'(s)\,
P_\beta\Big(\sum_{i=1}^d \theta_i \phi'(u_i)x^{(i)}\Big).
\]
Set
\[
g^{(\theta)}:=g_t^{(\theta)},\qquad
g^{(\beta)}:=g_t^{(\beta)},
\qquad
g_t=(g^{(\theta)},g^{(\beta)}),
\]
and
\[
G_t:=\|g_t\|^2=\|g^{(\theta)}\|^2+\|g^{(\beta)}\|^2.
\]
Also recall
\[
\Xi_t
=
\langle g_t-\bar g_t,(a,b)\rangle,
\qquad
\bar g_t:=\mathbb E[g_t\mid\mathcal F_t].
\]

All moment estimates below follow from Assumption~\ref{ass:regularity-and-moments} and
Remark~\ref{rem:activation-moment-consequences}.

\medskip
\noindent
{\bf residual self-quenching.}

We claim that for \(m=2,4,8\),
\[
\mathbb E[|R|^m\mid\mathcal F_t]\le C_m Q_t^{m/2}.
\]

Since \(\psi'\) is bounded,
\[
|R|
\le
\|\psi'\|_\infty\,|s-s^*|.
\]
Decompose
\[
s-s^*=A+B,
\]
where
\[
A:=\sum_{i=1}^d (\theta_i-\theta_i^*)\phi(u_i),
\]
and
\[
B:=\sum_{i=1}^d \theta_i^*\bigl(\phi(u_i)-\phi(u_i^*)\bigr).
\]

For \(A\): the variables \(\phi(u_i)\) are iid, centered, and have finite
moments up to order \(8\). Hence by Rosenthal's inequality, for
\(m\in\{2,4,8\}\),
\[
\|A\|_{L^m(\mathbb{P}(\cdot\mid\mathcal F_t))}
\le
C_m \|\theta-\theta^*\|_2.
\]
Since Euclidean distance is bounded by geodesic distance on the sphere,
\[
\|\theta-\theta^*\|_2\le d_\theta,
\]
we get
\[
\|A\|_{L^m(\mathbb{P}(\cdot\mid\mathcal F_t))}\le C_m d_\theta.
\]

For \(B\): the blocks \(x^{(i)}\) are independent, so the summands
\[
Y_i:=\theta_i^*\bigl(\phi(u_i)-\phi(u_i^*)\bigr)
\]
are independent across \(i\). They are centered because
\(u_i\sim N(0,1)\) and \(u_i^*\sim N(0,1)\), hence
\[
\mathbb E[\phi(u_i)-\phi(u_i^*)]=0.
\]
Using the Lipschitzness of \(\phi\),
\[
|Y_i|
\le
|\theta_i^*|\,\operatorname{Lip}(\phi)\,|u_i-u_i^*|
=
|\theta_i^*|\,\operatorname{Lip}(\phi)\,|x^{(i)}\cdot(\beta-\beta^*)|.
\]
Thus for \(m\in\{2,4,8\}\),
\[
\sum_{i=1}^d \mathbb E[|Y_i|^m\mid\mathcal F_t]
\le
C_m \|\beta-\beta^*\|_2^m \sum_{i=1}^d |\theta_i^*|^m
\le
C_m d_\beta^m,
\]
since \(\sum_i |\theta_i^*|^m\le \|\theta^*\|_2^m=1\) for \(m\ge2\).
Applying Rosenthal again,
\[
\|B\|_{L^m(\mathbb{P}(\cdot\mid\mathcal F_t))}
\le
C_m d_\beta,
\qquad m=2,4,8.
\]

Combining the bounds for \(A\) and \(B\),
\[
\|s-s^*\|_{L^m(\mathbb{P}(\cdot\mid\mathcal F_t))}
\le
C_m(d_\theta+d_\beta)
\le
C_m\sqrt{d_\theta^2+d_\beta^2}
=
C_m\sqrt{2Q_t}.
\]
Hence
\[
\|R\|_{L^m(\mathbb{P}(\cdot\mid\mathcal F_t))}
\le
C_m\sqrt{Q_t},
\qquad m=2,4,8,
\]
which is equivalent to
\[
\mathbb E[|R|^m\mid\mathcal F_t]\le C_m Q_t^{m/2},
\qquad m=2,4,8.
\]

\medskip
\noindent
{\bf feature-side moment bounds.}

Define  $$z := z_\beta(x) = (\phi(u_1),\dots,\phi(u_d)), \qquad h := P_\beta\sum_{i=1}^d\theta_i\phi'(u_i)x^{(i)},$$  so that  $$g^{(\theta)} = A\cdot P_\theta z, \qquad g^{(\beta)} = A\cdot h, \qquad A := 2\psi'(s) R,$$  and $G_t \le C R^2 W$ where $W := \|z\|^2 + \|h\|^2$.   The same Gaussian moment computations used in the proof of Lemma \ref{lem:sgd-tech-bounds} give, for $k=1,2,4$,  $$\mathbb E[\|z\|^{2k}\mid\mathcal F_t]\le Cd^k, \qquad \mathbb E[\|h\|^{2k}\mid\mathcal F_t]\le Cd^k.$$  Hence $\mathbb E[W^k\mid\mathcal F_t]\le Cd^k$ for $k = 1, 2, 4$.

\medskip
\noindent
{\bf bound on \(\mathbb E[G_t\mid\mathcal F_t]\).}

Since \(G_t\le C R^2 W\), Cauchy--Schwarz gives
\[
\mathbb E[G_t\mid\mathcal F_t]
\le
C\,
\Bigl(\mathbb E[R^4\mid\mathcal F_t]\Bigr)^{1/2}
\Bigl(\mathbb E[W^2\mid\mathcal F_t]\Bigr)^{1/2}.
\]
Using Step 1 with \(m=4\) and Step 2,
\[
\mathbb E[G_t\mid\mathcal F_t]
\le
C\,(Q_t^2)^{1/2}(d^2)^{1/2}
=
C d\,Q_t.
\]

\medskip
\noindent
{\bf bound on \(\mathbb E[\Xi_t^2\mid\mathcal F_t]\).}

Since conditional variance is bounded by the second moment,
\[
\mathbb E[\Xi_t^2\mid\mathcal F_t]
\le
\mathbb E\!\left[
\bigl(
\langle g^{(\theta)},a\rangle+\langle g^{(\beta)},b\rangle
\bigr)^2
\middle|\mathcal F_t
\right].
\]
Hence
\[
\mathbb E[\Xi_t^2\mid\mathcal F_t]
\le
2\,\mathbb E[\langle g^{(\theta)},a\rangle^2\mid\mathcal F_t]
+
2\,\mathbb E[\langle g^{(\beta)},b\rangle^2\mid\mathcal F_t].
\]

For the \(\theta\)-block, \(a\in T_\theta\mathbb S^{d-1}\), so
\[
\langle P_\theta z,a\rangle=\langle z,a\rangle.
\]
Thus
\[
\langle g^{(\theta)},a\rangle
=
2R\,\psi'(s)\,\langle z,a\rangle.
\]
Therefore
\[
\mathbb E[\langle g^{(\theta)},a\rangle^2\mid\mathcal F_t]
\le
C\,
\Bigl(\mathbb E[R^4\mid\mathcal F_t]\Bigr)^{1/2}
\Bigl(\mathbb E[\langle z,a\rangle^4\mid\mathcal F_t]\Bigr)^{1/2}.
\]
Now \(\langle z,a\rangle=\sum_i a_i\phi(u_i)\) is a linear statistic of iid
centered variables with finite fourth moment, so Rosenthal yields
\[
\mathbb E[\langle z,a\rangle^4\mid\mathcal F_t]\le C\|a\|^4.
\]
Hence
\[
\mathbb E[\langle g^{(\theta)},a\rangle^2\mid\mathcal F_t]
\le
C\,Q_t\,\|a\|^2
\le
C\,Q_t^2.
\]

For the \(\beta\)-block,
\[
\langle g^{(\beta)},b\rangle
=
2R\,\psi'(s)\,\langle h,b\rangle.
\]
Conditionally on \((u_1,\dots,u_d)\), the scalar \(\langle h,b\rangle\) is
Gaussian with variance
\[
\|b\|^2 \sigma^2,
\qquad
\sigma^2=\sum_{i=1}^d \theta_i^2 \phi'(u_i)^2.
\]
Therefore
\[
\mathbb E[\langle h,b\rangle^4\mid u,\mathcal F_t]
\le
C\|b\|^4\sigma^4.
\]
Averaging and using \(\mathbb E[\sigma^4\mid\mathcal F_t]\le C\),
\[
\mathbb E[\langle h,b\rangle^4\mid\mathcal F_t]\le C\|b\|^4.
\]
Thus
\[
\mathbb E[\langle g^{(\beta)},b\rangle^2\mid\mathcal F_t]
\le
C\,
\Bigl(\mathbb E[R^4\mid\mathcal F_t]\Bigr)^{1/2}
\Bigl(\mathbb E[\langle h,b\rangle^4\mid\mathcal F_t]\Bigr)^{1/2}
\le
C\,Q_t\,\|b\|^2
\le
C\,Q_t^2.
\]

Combining the two block estimates,
\[
\mathbb E[\Xi_t^2\mid\mathcal F_t]\le C Q_t^2.
\]

\medskip
\noindent
{\bf centered normalization fluctuation.}

Since
\[
\mathbb E\!\left[
\left(G_t-\mathbb E[G_t\mid\mathcal F_t]\right)^2
\middle|\mathcal F_t
\right]
\le
\mathbb E[G_t^2\mid\mathcal F_t],
\]
it suffices to bound \(\mathbb E[G_t^2\mid\mathcal F_t]\). Using
\(G_t\le C R^2W\),
\[
G_t^2\le C R^4 W^2.
\]
Hence, by Cauchy--Schwarz,
\[
\mathbb E[G_t^2\mid\mathcal F_t]
\le
C\,
\Bigl(\mathbb E[R^8\mid\mathcal F_t]\Bigr)^{1/2}
\Bigl(\mathbb E[W^4\mid\mathcal F_t]\Bigr)^{1/2}.
\]
Applying Step 1 with \(m=8\) and Step 2,
\[
\mathbb E[G_t^2\mid\mathcal F_t]
\le
C\,(Q_t^4)^{1/2}(d^4)^{1/2}
=
C d^2 Q_t^2.
\]
Therefore
\[
\mathbb E\!\left[
\left(G_t-\mathbb E[G_t\mid\mathcal F_t]\right)^2
\middle|\mathcal F_t
\right]
\le
C d^2 Q_t^2.
\]
This proves the lemma.
\end{proof}

\subsection{Strong Recovery Dynamics}

With all the controlling lemmas needed, we now move to prove the strong recovery dynamics.

\begin{proposition}[One local SGD epoch contracts]
\label{prop:one-local-epoch}
Assume Lemma~\ref{lem:product-contraction},
Lemma~\ref{lem:one-step-distance-recursion}, and
Lemma~\ref{lem:sharp-local-self-quenching-sgd}. Let the local step size satisfy
\[
\delta=\delta_{\rm loc}=\frac{c_{\rm loc}}d,
\qquad
C_{\rm ret}C_{\rm sq}\delta d\le \mu_{\rm loc}.
\]
Let \(s\) be a stopping time, and suppose that, conditionally on
\(\mathcal F_s\),
\[
Q_s\le B,
\qquad
0\le B\le \frac{r_{\rm loc}^2}{32}.
\]
Define $\tau_B:=\inf\{t\ge s:Q_t>2B\}.$ 
Then there exist constants \(C_{\rm ep}<\infty\), \(C<\infty\), and
\(d_0<\infty\), depending only on the model constants, such that for all
\(d\ge d_0\), with
\[
L_{\rm ep}:=
\left\lceil \frac{C_{\rm ep}}{\delta}\right\rceil ,
\]
one has
\[
\mathbb P\left(
Q_{s+L_{\rm ep}}\le \frac B2
\ \text{and}\
\tau_B>s+L_{\rm ep}
\;\middle|\;
\mathcal F_s
\right)
\ge
1-C\delta.
\]
Equivalently, with probability at least \(1-C\delta\),
\[
Q_{s+L_{\rm ep}}\le \frac B2
\qquad\text{and}\qquad
Q_t\le 2B
\quad\text{for every }s\le t\le s+L_{\rm ep}.
\]
\end{proposition}

\begin{proof}
If \(B=0\), then \(Q_s=0\), so \(y_s=y^*\). By
Lemma~\ref{lem:sharp-local-self-quenching-sgd},
\[
\mathbb E[G_s\mid\mathcal F_s]\le C_{\rm sq}dQ_s=0,
\]
hence \(G_s=0\) almost surely conditionally on \(\mathcal F_s\). Thus the SGD
step is trivial at \(y^*\), and the same argument iterates. The conclusion is
therefore immediate. Hence assume \(B>0\).

Throughout the proof we condition on \(\mathcal F_s\). Since the samples after
time \(s\) are fresh, all martingale and conditional moment estimates hold with
the same constants.

\medskip
\noindent
\textbf{Step 1: stopped multiplicative recursion.}

For every \(t<\tau_B\), one has \(Q_t\le 2B\). Hence
\[
\operatorname{dist}_{\mathcal M}(y_t,y^*)^2
=
2Q_t
\le
4B
\le
\frac{r_{\rm loc}^2}{8}
<
r_{\rm loc}^2.
\]
Thus \(y_t\in\mathcal U_{r_{\rm loc}}\), and the local one-step recursion is
valid. By Lemma~\ref{lem:one-step-distance-recursion},
\[
Q_{t+1}
\le
Q_t
+
\delta\langle \bar g_t,\mathcal L_t\rangle
+
\delta\Xi_t
+
C_{\rm ret}\delta^2G_t.
\]
Since \(\bar g_t=\nabla_{\mathcal M}L(y_t)\), Lemma~\ref{lem:product-contraction}
gives
\[
\langle \bar g_t,\mathcal L_t\rangle
=
\langle \nabla_{\mathcal M}L(y_t),\mathcal L_t\rangle
\le
-2\mu_{\rm loc}Q_t.
\]
Define
\[
\zeta_t
:=
G_t-\mathbb E[G_t\mid\mathcal F_t].
\]
Then, for \(t<\tau_B\),
\[
Q_{t+1}
\le
Q_t
-
2\mu_{\rm loc}\delta Q_t
+
\delta\Xi_t
+
C_{\rm ret}\delta^2\mathbb E[G_t\mid\mathcal F_t]
+
C_{\rm ret}\delta^2\zeta_t.
\]
By Lemma~\ref{lem:sharp-local-self-quenching-sgd},
\[
\mathbb E[G_t\mid\mathcal F_t]
\le
C_{\rm sq}dQ_t.
\]
Therefore
\[
Q_{t+1}
\le
\left(1-\bigl(2\mu_{\rm loc}-C_{\rm ret}C_{\rm sq}\delta d\bigr)\delta\right)Q_t
+
\delta\Xi_t
+
C_{\rm ret}\delta^2\zeta_t.
\]
The local step-size condition implies
\[
2\mu_{\rm loc}-C_{\rm ret}C_{\rm sq}\delta d
\ge
\mu_{\rm loc}.
\]
Set
\[
\lambda:=\mu_{\rm loc},
\qquad
\alpha:=1-\lambda\delta.
\]
For \(d\) large enough, \(\alpha\in(0,1)\). Thus, for every \(t<\tau_B\),
\begin{equation}
\label{eq:local-epoch-stopped-recursion}
Q_{t+1}
\le
\alpha Q_t+\Delta N_{t+1},
\end{equation}
where
\[
\Delta N_{t+1}
:=
\delta\Xi_t+C_{\rm ret}\delta^2\zeta_t.
\]
Moreover,
\[
\mathbb E[\Delta N_{t+1}\mid\mathcal F_t]=0.
\]

\medskip
\noindent
\textbf{Step 2: martingale variance on the stopped epoch.}

By Lemma~\ref{lem:sharp-local-self-quenching-sgd},
\[
\mathbb E[\Xi_t^2\mid\mathcal F_t]
\le
C_{\rm sq}Q_t^2,
\]
and
\[
\mathbb E[\zeta_t^2\mid\mathcal F_t]
=
\mathbb E\!\left[
\left(G_t-\mathbb E[G_t\mid\mathcal F_t]\right)^2
\middle|\mathcal F_t
\right]
\le
C_{\rm sq}d^2Q_t^2.
\]
Hence, using \((a+b)^2\le 2a^2+2b^2\),
\[
\mathbb E[(\Delta N_{t+1})^2\mid\mathcal F_t]
\le
C\delta^2Q_t^2
+
C\delta^4d^2Q_t^2.
\]
Since \(\delta d\) is bounded under the local step-size regime,
\[
\mathbb E[(\Delta N_{t+1})^2\mid\mathcal F_t]
\le
C\delta^2Q_t^2.
\]
On \(\{t<\tau_B\}\), \(Q_t\le 2B\), and therefore
\[
\mathbb E[(\Delta N_{t+1})^2\mid\mathcal F_t]
\le
C\delta^2B^2.
\]

Define the stopped martingale increments
\[
\Delta \widetilde N_{t+1}
:=
\mathbf 1_{\{t<\tau_B\}}\Delta N_{t+1},
\]
and for \(u=s,\dots,s+L_{\rm ep}\),
\[
\widetilde N_u
:=
\sum_{k=s}^{u-1}\Delta \widetilde N_{k+1},
\qquad
\widetilde N_s:=0.
\]
Then \((\widetilde N_u)_{u=s}^{s+L_{\rm ep}}\) is a martingale, and
\[
\mathbb E\!\left[
|\widetilde N_{s+L_{\rm ep}}|^2
\middle|\mathcal F_s
\right]
=
\mathbb E\!\left[
\sum_{k=s}^{s+L_{\rm ep}-1}
\mathbb E[(\Delta \widetilde N_{k+1})^2\mid\mathcal F_k]
\middle|\mathcal F_s
\right]
\le
C L_{\rm ep}\delta^2B^2.
\]
Since
\[
L_{\rm ep}
=
\left\lceil \frac{C_{\rm ep}}{\delta}\right\rceil
\le
\frac{C}{\delta}
\]
for \(d\) large enough,
\[
\mathbb E\!\left[
|\widetilde N_{s+L_{\rm ep}}|^2
\middle|\mathcal F_s
\right]
\le
C\delta B^2.
\]
By Doob's \(L^2\) maximal inequality,
\[
\mathbb E\!\left[
\sup_{s\le u\le s+L_{\rm ep}}
|\widetilde N_u|^2
\middle|\mathcal F_s
\right]
\le
C\delta B^2.
\]
Therefore, by Markov's inequality,
\[
\mathbb P\left(
\sup_{s\le u\le s+L_{\rm ep}}
|\widetilde N_u|
>
\frac B8
\;\middle|\;
\mathcal F_s
\right)
\le
C\delta.
\]

\medskip
\noindent
\textbf{Step 3: weighted martingale control.}

For \(1\le \ell\le L_{\rm ep}\), define the stopped geometric convolution
\[
\widetilde S_\ell
:=
\sum_{j=0}^{\ell-1}
\alpha^{\ell-1-j}\Delta \widetilde N_{s+j+1}.
\]
We claim that
\[
\sup_{1\le \ell\le L_{\rm ep}}
|\widetilde S_\ell|
\le
2\sup_{s\le u\le s+L_{\rm ep}}|\widetilde N_u|.
\]
Indeed, for fixed \(\ell\), write
\[
A_j:=\widetilde N_{s+j},
\qquad
j=0,\dots,\ell.
\]
Then \(A_0=0\) and
\[
\Delta \widetilde N_{s+j+1}=A_{j+1}-A_j.
\]
By summation by parts,
\[
\widetilde S_\ell
=
A_\ell
-
(1-\alpha)
\sum_{j=1}^{\ell-1}
\alpha^{\ell-1-j}A_j.
\]
Thus
\[
|\widetilde S_\ell|
\le
\sup_{0\le j\le \ell}|A_j|
+
(1-\alpha)
\sum_{j=1}^{\ell-1}\alpha^{\ell-1-j}
\sup_{0\le j\le \ell}|A_j|
\le
2\sup_{0\le j\le \ell}|A_j|.
\]
Taking the supremum over \(\ell\) proves the claim.

Define the epoch good event
\[
\mathcal E_{\rm ep}
:=
\left\{
\sup_{1\le \ell\le L_{\rm ep}}
|\widetilde S_\ell|
\le
\frac B4
\right\}.
\]
By the preceding maximal bound,
\[
\mathbb P(\mathcal E_{\rm ep}\mid\mathcal F_s)
\ge
1-C\delta.
\]

\medskip
\noindent
\textbf{Step 4: contraction on \(\mathcal E_{\rm ep}\).}

We prove by induction that, on \(\mathcal E_{\rm ep}\),
\[
Q_{s+\ell}
\le
\alpha^\ell B+\frac B4
\qquad
\text{for every }0\le \ell\le L_{\rm ep}.
\]
For \(\ell=0\), this follows from \(Q_s\le B\).

Assume the claim has been proved up to time \(s+\ell-1\), where
\(1\le \ell\le L_{\rm ep}\). Then, for every \(0\le j\le \ell-1\),
\[
Q_{s+j}
\le
\alpha^jB+\frac B4
\le
B+\frac B4
<
2B.
\]
Hence \(\tau_B>s+\ell-1\), so the original increments and the stopped
increments agree on the interval \(s,\dots,s+\ell-1\). Iterating
\eqref{eq:local-epoch-stopped-recursion} gives
\[
Q_{s+\ell}
\le
\alpha^\ell Q_s
+
\sum_{j=0}^{\ell-1}
\alpha^{\ell-1-j}\Delta N_{s+j+1}
=
\alpha^\ell Q_s+\widetilde S_\ell.
\]
On \(\mathcal E_{\rm ep}\), this yields
\[
Q_{s+\ell}
\le
\alpha^\ell B+\frac B4.
\]
This closes the induction. In particular,
\[
Q_{s+\ell}<2B
\qquad
\text{for every }0\le \ell\le L_{\rm ep},
\]
and therefore
\[
\tau_B>s+L_{\rm ep}.
\]

It remains to choose \(C_{\rm ep}\) so that the endpoint is at most \(B/2\).
Choose \(C_{\rm ep}\) large enough that
\[
e^{-\lambda C_{\rm ep}}\le \frac14.
\]
Since
\[
L_{\rm ep}=\left\lceil \frac{C_{\rm ep}}{\delta}\right\rceil,
\qquad
\alpha=1-\lambda\delta,
\]
we have, for \(d\) large enough,
\[
\alpha^{L_{\rm ep}}
\le
e^{-\lambda\delta L_{\rm ep}}
\le
e^{-\lambda C_{\rm ep}}
\le
\frac14.
\]
Thus, on \(\mathcal E_{\rm ep}\),
\[
Q_{s+L_{\rm ep}}
\le
\alpha^{L_{\rm ep}}B+\frac B4
\le
\frac B4+\frac B4
=
\frac B2.
\]
Together with \(\tau_B>s+L_{\rm ep}\), this gives
\[
\mathcal E_{\rm ep}
\subseteq
\left\{
Q_{s+L_{\rm ep}}\le \frac B2
\ \text{and}\
\tau_B>s+L_{\rm ep}
\right\}.
\]
Since
\[
\mathbb P(\mathcal E_{\rm ep}\mid\mathcal F_s)
\ge
1-C\delta,
\]
the proposition follows.
\end{proof}

\begin{proposition}[Iterated local contraction]
\label{prop:iterated-local-contraction}
Assume Proposition~\ref{prop:one-local-epoch}. Let
\[
\delta=\delta_{\rm loc}=\frac{c_{\rm loc}}d
\]
satisfy the local step-size condition, and let
\[
L_{\rm ep}:=
\left\lceil \frac{C_{\rm ep}}{\delta}\right\rceil
\]
be the epoch length from Proposition~\ref{prop:one-local-epoch}. Let \(s\) be
a stopping time and suppose that, conditionally on \(\mathcal F_s\),
\[
Q_s\le B_0,
\qquad
0<B_0\le \frac{r_{\rm loc}^2}{32}.
\]
For \(\varepsilon\in(0,B_0]\), define
\[
J_\varepsilon
:=
\left\lceil \log_2\frac{B_0}{\varepsilon}\right\rceil,
\qquad
N_\varepsilon
:=
J_\varepsilon L_{\rm ep}.
\]
Then
\begin{align}
\mathbb P\left(
Q_{s+N_\varepsilon}\le \varepsilon
\ \text{and}\
Q_t\le 2B_0\ \text{for every }s\le t\le s+N_\varepsilon
\;\middle|\;
\mathcal F_s
\right)
\ge
1-C\delta J_\varepsilon .
\end{align}
In particular, since \(\delta=c_{\rm loc}/d\),
\begin{align}
N_\varepsilon
\le
C d\left(1+\log\frac{B_0}{\varepsilon}\right),
\end{align}
and
\begin{align}
\mathbb P\left(
Q_{s+N_\varepsilon}\le \varepsilon
\;\middle|\;
\mathcal F_s
\right)
\ge
1-
C\frac{\log(2B_0/\varepsilon)}d .
\end{align}
\end{proposition}

\begin{proof}
If \(\varepsilon=B_0\), then \(J_\varepsilon=0\), \(N_\varepsilon=0\), and the
claim follows from \(Q_s\le B_0=\varepsilon\). Hence assume
\(\varepsilon<B_0\), so \(J_\varepsilon\ge1\).

Set
\[
s_j:=s+jL_{\rm ep},
\qquad
B_j:=2^{-j}B_0,
\qquad
j=0,1,\dots,J_\varepsilon .
\]
Since \(s\) is a stopping time, each \(s_j\) is also a stopping time. Moreover,
\[
B_j\le B_0\le \frac{r_{\rm loc}^2}{32},
\]
so Proposition~\ref{prop:one-local-epoch} is applicable at every epoch as
soon as \(Q_{s_j}\le B_j\).

For \(j=0,\dots,J_\varepsilon-1\), define the epoch-success event
\[
\mathcal A_j
:=
\left\{
Q_{s_{j+1}}\le B_{j+1}
\ \text{and}\
Q_t\le 2B_j\ \text{for every }s_j\le t\le s_{j+1}
\right\}.
\]
Also define
\[
\mathcal G_m
:=
\bigcap_{j=0}^{m-1}\mathcal A_j,
\qquad
m=1,\dots,J_\varepsilon,
\]
with \(\mathcal G_0\) equal to the whole probability space.

We claim that, for every \(m\le J_\varepsilon\), on \(\mathcal G_m\),
\[
Q_{s_m}\le B_m
\]
and
\[
Q_t\le 2B_0
\qquad
\text{for every }s\le t\le s_m.
\]
Indeed, the first statement follows by induction from the endpoint conclusion
in each \(\mathcal A_j\). The second follows because, on the \(j\)-th epoch,
\[
Q_t\le 2B_j\le 2B_0.
\]

Now fix \(j\in\{0,\dots,J_\varepsilon-1\}\). On \(\mathcal G_j\), we have
\(Q_{s_j}\le B_j\). Applying Proposition~\ref{prop:one-local-epoch}
conditionally on \(\mathcal F_{s_j}\) gives
\[
\mathbb P\left(\mathcal A_j^c\mid\mathcal F_{s_j}\right)
\le
C\delta
\qquad
\text{on }\mathcal G_j.
\]
Therefore,
\[
\mathbb P\left(\mathcal G_j\cap\mathcal A_j^c\mid\mathcal F_s\right)
=
\mathbb E\left[
\mathbf 1_{\mathcal G_j}
\mathbb P\left(\mathcal A_j^c\mid\mathcal F_{s_j}\right)
\,\middle|\,
\mathcal F_s
\right]
\le
C\delta .
\]
Taking a union bound over the \(J_\varepsilon\) epochs yields
\[
\mathbb P\left(\mathcal G_{J_\varepsilon}^c\mid\mathcal F_s\right)
\le
C\delta J_\varepsilon .
\]
Hence
\[
\mathbb P\left(\mathcal G_{J_\varepsilon}\mid\mathcal F_s\right)
\ge
1-C\delta J_\varepsilon .
\]

On \(\mathcal G_{J_\varepsilon}\), we have
\[
Q_{s+N_\varepsilon}
=
Q_{s_{J_\varepsilon}}
\le
B_{J_\varepsilon}
=
2^{-J_\varepsilon}B_0
\le
\varepsilon,
\]
by the definition of \(J_\varepsilon\). Also, as shown above,
\[
Q_t\le 2B_0
\qquad
\text{for every }s\le t\le s+N_\varepsilon.
\]
This proves the probability estimate.

It remains only to record the sample complexity. Since
\[
L_{\rm ep}
=
\left\lceil \frac{C_{\rm ep}}{\delta}\right\rceil
\le
\frac C\delta
\]
for \(d\) sufficiently large, and
\[
J_\varepsilon
=
\left\lceil \log_2\frac{B_0}{\varepsilon}\right\rceil
\le
C\left(1+\log\frac{B_0}{\varepsilon}\right),
\]
we get
\[
N_\varepsilon
=
J_\varepsilon L_{\rm ep}
\le
C\delta^{-1}
\left(1+\log\frac{B_0}{\varepsilon}\right).
\]
With \(\delta=c_{\rm loc}/d\), this becomes
\[
N_\varepsilon
\le
C d\left(1+\log\frac{B_0}{\varepsilon}\right).
\]
Finally,
\[
\delta J_\varepsilon
\le
\frac C d\log\frac{2B_0}{\varepsilon},
\]
after increasing \(C\). This gives
\[
\mathbb P\left(
Q_{s+N_\varepsilon}\le \varepsilon
\;\middle|\;
\mathcal F_s
\right)
\ge
1-
C\frac{\log(2B_0/\varepsilon)}d .
\]
The proof is complete.
\end{proof}

\begin{theorem}[Discrete-time weak-to-strong recovery]
\label{thm:discrete-weak-to-strong-recovery}
Assume Theorem~\ref{thm:sgd-escape-mediocrity-pullup} holds with event \( G_T^{\rm weak}\), stopping time \(t_{\rm esc}\), and constants \(T_{\rm esc},c_{\rm esc},
C_{\eta,{\rm esc}}\), such that on \( G_T^{\rm weak}\),
\[
t_{\rm esc}\le T_{\rm esc},
\qquad
\min\{\rho_{t_{\rm esc}},\chi_{t_{\rm esc}}\}\ge c_{\rm esc},
\qquad
\eta_{t_{\rm esc}}
\le
C_{\eta,{\rm esc}}\frac{\log d}{\sqrt d}.
\]
Assume \ref{ass:favorable-initialization}. After \(t_{\rm esc}\), switch to the local step size $\delta_{\rm loc}=\frac{c_{\rm loc}}d,$ where \(c_{\rm loc}>0\) is chosen sufficiently small so that the assumptions of Lemma~\ref{lem:entry-window-control}, Proposition~\ref{prop:discrete-entry-local-basin}, and Proposition~\ref{prop:one-local-epoch} hold. Let
\[
Q_{\rm in}:=\frac{r_{\rm loc}^2}{64}.
\]
Then, for every fixed or polynomially small \(\nu_{\rm in}\), there exists an
event \(E_{\rm str}\) satisfying
\begin{align}
\mathbb P(E_{\rm str})
\ge
1-\mathbb P( {G_T^{\rm weak}}^c)-\nu_{\rm in},
\end{align}
on which there is a stopping time \(t_1\) such that
\begin{align}
t_1\le t_{\rm esc}+C_{\rm ent}\delta_{\rm loc}^{-1}
\le T_{\rm esc}+C_{\rm ent}\delta_{\rm loc}^{-1},
\end{align}
and
\begin{align}
Q_{t_1}\le Q_{\rm in}\le \frac{r_{\rm loc}^2}{32}.
\end{align}

Moreover, for every \(\varepsilon\in(0,Q_{\rm in}]\), define
\[
J_\varepsilon
:=
\left\lceil \log_2\frac{Q_{\rm in}}{\varepsilon}\right\rceil,
\qquad
L_{\rm ep}:=
\left\lceil \frac{C_{\rm ep}}{\delta_{\rm loc}}\right\rceil,
\qquad
N_\varepsilon:=J_\varepsilon L_{\rm ep}.
\]
Then
\begin{align}
\mathbb P\left(
Q_{t_1+N_\varepsilon}\le \varepsilon
\right)
\ge
1-\mathbb P( {G_T^{\rm weak}}^c)-\nu_{\rm in}
-
C\delta_{\rm loc}J_\varepsilon.
\end{align}
Equivalently, since \(\delta_{\rm loc}=c_{\rm loc}/d\),
\[
\mathbb P\left(
Q_{t_1+N_\varepsilon}\le \varepsilon
\right)
\ge
1-\mathbb P( {G_T^{\rm weak}}^c)-\nu_{\rm in}
-
C\frac{\log(2Q_{\rm in}/\varepsilon)}d.
\]
On the same event,
\[
1-\rho_{t_1+N_\varepsilon}\le \varepsilon,
\qquad
1-\chi_{t_1+N_\varepsilon}\le \varepsilon,
\]
and hence
\begin{align}
\min\{\rho_{t_1+N_\varepsilon},\chi_{t_1+N_\varepsilon}\}
\ge
1-\varepsilon.
\end{align}

In particular,
\[
N_\varepsilon
\le
C d\left(1+\log\frac{Q_{\rm in}}{\varepsilon}\right).
\]
With the weak phase satisfies \(T_{\rm esc}=O(d\log^2 d)\), the total number
of SGD samples needed to reach \(Q\le\varepsilon\) is
\begin{align}
O(d\log^2 d)+O\left(d\log\frac1\varepsilon\right).
\end{align}
\end{theorem}

\begin{proof}
Let
\[
t_0:=t_{\rm esc},
\qquad
\delta:=\delta_{\rm loc}.
\]
Work first on the weak-recovery event \( G_T^{\rm weak}\). On this event,
\[
\min\{\rho_{t_0},\chi_{t_0}\}\ge c_{\rm esc},
\qquad
\eta_{t_0}
\le
C_{\eta,{\rm esc}}\frac{\log d}{\sqrt d}.
\]
Together with the incoherence bound
\[
\|\theta^*\|_\infty\le C_*\sqrt{\frac{\log d}{d}},
\]
these are exactly the assumptions of Lemma~\ref{lem:entry-window-control}.

Apply Lemma~\ref{lem:entry-window-control} with \(C_{\rm in}\ge C_{\rm ent}\).
Conditionally on \(\mathcal F_{t_0}\), there is an event \(E_{\rm in}\) such
that
\[
\mathbb P(E_{\rm in}\mid\mathcal F_{t_0})
\ge
1-\nu_{\rm in}.
\]
Hence, by the tower property,
\[
\mathbb P( G_T^{\rm weak}\cap E_{\rm in})
=
\mathbb E\left[
\mathbf 1_{ G_T^{\rm weak}}
\mathbb P(E_{\rm in}\mid\mathcal F_{t_0})
\right]
\ge
(1-\nu_{\rm in})\mathbb P( G_T^{\rm weak}).
\]
Therefore
\[
\mathbb P( G_T^{\rm weak}\cap E_{\rm in})
\ge
1-\mathbb P( {G_T^{\rm weak}}^c)-\nu_{\rm in}.
\]

On \( G_T^{\rm weak}\cap E_{\rm in}\), Proposition~\ref{prop:discrete-entry-local-basin}
gives a time
\[
t_1\le t_{\rm esc}+C_{\rm ent}\delta^{-1}
\]
such that
\[
\rho_{t_1}\ge \cos(r_{\rm loc}/8),
\qquad
\chi_{t_1}\ge \cos(r_{\rm loc}/8),
\]
and consequently
\[
Q_{t_1}\le \frac{r_{\rm loc}^2}{64}=Q_{\rm in}.
\]
Choosing \(t_1\) as the stopping time constructed in
Proposition~\ref{prop:discrete-entry-local-basin}, this proves the first part.
Since \(t_{\rm esc}\le T_{\rm esc}\) on \( G_T^{\rm weak}\), we also have
\[
t_1\le T_{\rm esc}+C_{\rm ent}\delta^{-1}.
\]
Set
\[
E_{\rm str}^{(0)}:= G_T^{\rm weak}\cap E_{\rm in}.
\]
Then
\[
\mathbb P(E_{\rm str}^{(0)})
\ge
1-\mathbb P( {G_T^{\rm weak}}^c)-\nu_{\rm in}.
\]

Now fix \(\varepsilon\in(0,Q_{\rm in}]\). On \(E_{\rm str}^{(0)}\), we may
apply Proposition~\ref{prop:iterated-local-contraction} conditionally on
\(\mathcal F_{t_1}\), with
\[
s=t_1,
\qquad
B_0=Q_{\rm in}.
\]
Indeed,
\[
Q_{t_1}\le Q_{\rm in}
\le
\frac{r_{\rm loc}^2}{32}.
\]
Thus, conditionally on \(\mathcal F_{t_1}\),
\[
\mathbb P\left(
Q_{t_1+N_\varepsilon}\le \varepsilon
\ \middle|\ \mathcal F_{t_1}
\right)
\ge
1-C\delta J_\varepsilon.
\]
Using the tower property again,
\[
\mathbb P\left(
E_{\rm str}^{(0)}
\cap
\{Q_{t_1+N_\varepsilon}>\varepsilon\}
\right)
\le
C\delta J_\varepsilon.
\]
Therefore
\[
\mathbb P\left(
Q_{t_1+N_\varepsilon}\le \varepsilon
\right)
\ge
\mathbb P(E_{\rm str}^{(0)})-C\delta J_\varepsilon.
\]
Combining this with the lower bound on \(\mathbb P(E_{\rm str}^{(0)})\) gives
\[
\mathbb P\left(
Q_{t_1+N_\varepsilon}\le \varepsilon
\right)
\ge
1-\mathbb P( {G_T^{\rm weak}}^c)-\nu_{\rm in}
-
C\delta J_\varepsilon.
\]
Since
\[
\delta=\frac{c_{\rm loc}}d
\]
and
\[
J_\varepsilon
=
\left\lceil \log_2\frac{Q_{\rm in}}{\varepsilon}\right\rceil
\le
C\log\frac{2Q_{\rm in}}{\varepsilon},
\]
we obtain
\[
C\delta J_\varepsilon
\le
C\frac{\log(2Q_{\rm in}/\varepsilon)}d.
\]

It remains to convert the \(Q\)-bound into the overlap bound. If
\[
Q_{t_1+N_\varepsilon}\le \varepsilon,
\]
then
\[
\operatorname{dist}_{\mathbb S}(\theta_{t_1+N_\varepsilon},\theta^*)^2
+
\operatorname{dist}_{\mathbb S}(\beta_{t_1+N_\varepsilon},\beta^*)^2
=
2Q_{t_1+N_\varepsilon}
\le
2\varepsilon.
\]
In particular,
\[
\operatorname{dist}_{\mathbb S}(\theta_{t_1+N_\varepsilon},\theta^*)^2
\le
2\varepsilon,
\qquad
\operatorname{dist}_{\mathbb S}(\beta_{t_1+N_\varepsilon},\beta^*)^2
\le
2\varepsilon.
\]
Using \(1-\cos x\le x^2/2\), we get
\[
1-\chi_{t_1+N_\varepsilon}
=
1-\cos\!\left(
\operatorname{dist}_{\mathbb S}(\theta_{t_1+N_\varepsilon},\theta^*)
\right)
\le
\varepsilon,
\]
and similarly
\[
1-\rho_{t_1+N_\varepsilon}\le \varepsilon.
\]
Thus
\[
\min\{\rho_{t_1+N_\varepsilon},\chi_{t_1+N_\varepsilon}\}
\ge
1-\varepsilon.
\]

Finally, Proposition~\ref{prop:iterated-local-contraction} gives
\[
N_\varepsilon
=
J_\varepsilon L_{\rm ep}
\le
C d\left(1+\log\frac{Q_{\rm in}}{\varepsilon}\right).
\]
Since
\[
t_1\le t_{\rm esc}+C_{\rm ent}\delta^{-1}
=
t_{\rm esc}+O(d),
\]
the total number of samples after initialization is bounded by
\[
t_{\rm esc}
+
O(d)
+
O\left(d\log\frac{Q_{\rm in}}{\varepsilon}\right).
\]
Because \(Q_{\rm in}=r_{\rm loc}^2/64\) is a fixed constant independent of
\(d\), this is
\[
t_{\rm esc}
+
O\left(d\log\frac1\varepsilon\right).
\]
If \(T_{\rm esc}=O(d\log^2 d)\), then on \( G_T^{\rm weak}\),
\[
t_1+N_\varepsilon
\le
O(d\log^2 d)
+
O\left(d\log\frac1\varepsilon\right).
\]
The proof is complete.
\end{proof}

\section{Auxiliary Lemmas}

\subsection{Stein Bounds}\label{subsec:stein-bounds}

Here, we present the comparison bounds between the Riemannian Hessian of the true loss and that of the Gaussian model via a Stein-type approximation, for $\theta$-Hessian and $\beta$-Hessian respectively. They are used in the Lemma~\ref{lem:teacher-full-hessian-comparison}. We prove our results based on this theorem from \cite{bruna2023singleindexmodelsgaussian}:

\begin{theorem}[Stein coupling for product measures {\cite[Thm.\ D.1]{bruna2023singleindexmodelsgaussian}}]\label{thm:stein-product}
Let $X=(X_1,\dots,X_d)$ be a random vector with independent coordinates such that
$\E[X]=0$, $\E[XX^\top]=I_d$, and $\E|X_i|^3=\tau_i^3<\infty$.
Let $G\sim\mathcal N(0,I_d)$ be standard Gaussian.
If $h:\R^d\to\R$ is three-times differentiable, then
\begin{equation}\label{eq:stein-product}
\big|\E[h(X)]-\E[h(G)]\big|
\ \le\ \frac{5}{6}\sum_{i=1}^d \tau_i^3 \,\big\|\partial_{x_i}^3 h\big\|_\infty .
\end{equation}
\end{theorem}

\paragraph{Stein Bound for \texorpdfstring{$\theta$}{theta}-Hessian}

\begin{corollary}[Block Stein coupling with moment remainder]
\label{cor:stein-block}
Let $(X_i,X_i^*)_{i=1}^d$ be independent $\R^2$-valued random vectors with
\[
\E[X_i]=\E[X_i^*]=0,\qquad
\E[X_i^2]=\E[(X_i^*)^2]=1,\qquad
\E[X_i X_i^*]=m,
\]
and $\tau:=\max_i(\E|X_i|^5+\E|X_i^*|^5)<\infty$.
Let $(G_i,G_i^*)_{i=1}^d$ be iid $\mathcal N(0,\Sigma_m)$ with
$\Sigma_m=\bigl(\begin{smallmatrix}1&m\\m&1\end{smallmatrix}\bigr)$.
If $h:\R^d\times\R^d\to\R$ is three-times differentiable in each block $(x_i,x_i^*)$,
then
\begin{equation}\label{eq:stein-block}
\big|\E[h(X,X^*)]-\E[h(G,G^*)]\big|
\le
C(\tau)\sum_{i=1}^d
\sum_{\substack{a,b\ge 0\\a+b=3}}
\sup_{j\le d}\;
\E\bigg[\Big|\frac{\partial^{a+b}h}{\partial x_i^a\,\partial(x_i^*)^b}
(W^{(j)},W^{*(j)})\Big|\bigg],
\end{equation}
where $(W^{(j)},W^{*(j)})$ is the hybrid that agrees with $(G,G^*)$ on blocks
$1,\ldots,j$ and with $(X,X^*)$ on blocks $j{+}1,\ldots,d$.
\end{corollary}
 
\begin{proof}
We have $(W^{(0)},W^{*(0)})=(X,X^*)$ and $(W^{(d)},W^{*(d)})=(G,G^*)$, so
\begin{equation}\label{eq:block-telescope}
\E[h(X,X^*)]-\E[h(G,G^*)]
=
\sum_{j=1}^d
\Bigl(\E\bigl[h(W^{(j-1)},W^{*(j-1)})\bigr]
-\E\bigl[h(W^{(j)},W^{*(j)})\bigr]\Bigr).
\end{equation}
Each summand differs only in block~$j$. With all other blocks frozen, define
$\varphi_j(s,t):=h\bigl(W^{(j)}\big|_{(s,t)\text{ in block }j}\bigr)$,
so the $j$-th summand equals
$\E[\varphi_j(X_j,X_j^*)-\varphi_j(G_j,G_j^*)]$.
Taylor-expand $\varphi_j$ around $(0,0)$ to second order with integral remainder:
\[
\varphi_j(s,t)
=
\sum_{\substack{a,b\ge 0\\a+b\le 2}}
\frac{1}{a!\,b!}
\frac{\partial^{a+b}\varphi_j}{\partial s^a\partial t^b}\bigg|_{(0,0)}
s^a t^b
\;+\;R_j(s,t),
\]
where the integral form of the remainder gives
\[
|R_j(s,t)|
\le
C_0\,(|s|+|t|)^3
\sum_{\substack{a,b\ge 0\\a+b=3}}
\sup_{|\xi|\le|s|,\,|\zeta|\le|t|}
\Big|\frac{\partial^{a+b}\varphi_j}{\partial s^a\partial t^b}(\xi,\zeta)\Big|.
\]
Taking $\E[\varphi_j(X_j,X_j^*)]-\E[\varphi_j(G_j,G_j^*)]$:
the zeroth-order terms cancel trivially; the first-order terms cancel since
$\E[X_j]=\E[G_j]=0$ and $\E[X_j^*]=\E[G_j^*]=0$; the second-order terms
cancel since
\[
\E[X_j^2]=\E[G_j^2]=1,\quad
\E[(X_j^*)^2]=\E[(G_j^*)^2]=1,\quad
\E[X_jX_j^*]=\E[G_jG_j^*]=m.
\]
Hence only the remainder contributes:
\[
\bigl|\E\bigl[\varphi_j(X_j,X_j^*)-\varphi_j(G_j,G_j^*)\bigr]\bigr|
\le
C_0\bigl(\E(|X_j|{+}|X_j^*|)^3+\E(|G_j|{+}|G_j^*|)^3\bigr)
\sum_{\substack{a,b\ge 0\\a+b=3}}
\E\Big[\Big|\frac{\partial^{a+b}\varphi_j}{\partial s^a\partial t^b}\Big|\Big],
\]
where the moment prefactors are bounded by $C'(\tau)$, and the expectations of the
third-order partials use H\"older to separate the $(s,t)$-dependence from the
frozen blocks. Summing over $j$ and noting
$\partial^{a+b}\varphi_j/\partial s^a\partial t^b
=\partial^{a+b}h/\partial x_j^a\partial(x_j^*)^b$ evaluated at the hybrid
yields~\eqref{eq:stein-block}.
\end{proof}

\begin{corollary}[Stein estimate for a weighted quadratic statistic]
\label{cor:stein-linear-quadratic}
Let \(X_1,\ldots,X_d\) be independent, centered, unit-variance random
variables with \(\sup_i \E|X_i|^8<\infty\), and let
\(\Gamma\sim N(0,I_d)\). Let \(a,u\in \R^d\) satisfy
\(\|a\|_2=\|u\|_2=1\). If \(F\in C^3(\R)\) and
\(\|F^{(k)}\|_\infty<\infty\) for \(1\le k\le 3\), then
\[
\left|
\E\left[F(a\cdot X)(u\cdot X)^2\right]
-
\E\left[F(a\cdot \Gamma)(u\cdot \Gamma)^2\right]
\right|
\le
C_F
\left(
\sum_{i=1}^d |a_i|u_i^2
+
\sum_{i=1}^d a_i^2|u_i|
+
\sum_{i=1}^d |a_i|^3
\right),
\]
where \(C_F\) depends only on \(F\) and the law of \(X_1\), but not on
\(d,a,u\). In particular,
\[
\left|
\E\left[F(a\cdot X)(u\cdot X)^2\right]
-
\E\left[F(a\cdot \Gamma)(u\cdot \Gamma)^2\right]
\right|
\le
C_F\|a\|_\infty .
\]
\end{corollary}

\begin{proof}
Apply the scalar-block moment version of the Lindeberg--Stein comparison,
as in Corollary~\ref{cor:stein-block}, to
\[
h(x):=F(a\cdot x)(u\cdot x)^2.
\]
For each coordinate \(i\),
\[
\partial_i^3 h(x)
=
a_i^3 F^{(3)}(a\cdot x)(u\cdot x)^2
+
6a_i^2u_iF''(a\cdot x)(u\cdot x)
+
6a_iu_i^2F'(a\cdot x).
\]
At every hybrid vector appearing in the Lindeberg interpolation,
\[
\E|u\cdot W|^2\le C\|u\|_2^2=C,
\qquad
\E|u\cdot W|\le C,
\]
because the coordinates have uniformly bounded second moments. Hence
\[
\E|\partial_i^3h(W)|
\le
C_F\left(|a_i|^3+a_i^2|u_i|+|a_i|u_i^2\right).
\]
Summing over \(i\) gives the first estimate.

For the final bound, use \(\|a\|_2=\|u\|_2=1\):
\[
\sum_i |a_i|u_i^2\le \|a\|_\infty,
\qquad
\sum_i a_i^2|u_i|
\le
\left(\sum_i a_i^4\right)^{1/2}\|u\|_2
\le
\|a\|_\infty,
\]
and
\[
\sum_i |a_i|^3\le \|a\|_\infty\sum_i a_i^2=\|a\|_\infty.
\]
This proves the claim.
\end{proof}

\begin{lemma}[Teacher-level \(\theta\)-Hessian comparison]
\label{lem:teacher-theta-hessian-comparison}
Let
\[
A_\psi:=\E[\psi'(G)^2],
\qquad G\sim N(0,1).
\]
Assume \(\E[\phi(G)]=0\), \(\E[\phi(G)^2]=1\), \(\E|\phi(G)|^8<\infty\), and
\(\psi\in C^4(\R)\) with bounded derivatives up to order \(4\). Then there
exists \(C_\theta>0\), independent of \(d\), such that for every
\(u\in T_{\theta^*}\mathbb S^{d-1}\),
\[
\left|
\operatorname{Hess}_{\mathbb S,\theta}L(y^*)[u,u]
-
2A_\psi\|u\|^2
\right|
\le
C_\theta\|\theta^*\|_\infty\|u\|^2.
\]
Equivalently,
\[
\left|
\operatorname{Hess}_{\mathbb S,\theta}L(y^*)[u,u]
-
\operatorname{Hess}_{\mathbb S,\theta}L_G(y^*)[u,u]
\right|
\le
C_\theta\|\theta^*\|_\infty\|u\|^2.
\]
\end{lemma}

\begin{proof}
By quadratic scaling, it is enough to prove the result for \(\|u\|=1\).
Set
\[
a:=\theta^*,
\qquad
X_i:=\phi(G_i),
\qquad
S:=a\cdot X=\sum_{i=1}^d a_iX_i,
\qquad
A_u:=u\cdot X=\sum_{i=1}^d u_iX_i,
\]
where \(G_i\stackrel{\mathrm{iid}}{\sim}N(0,1)\). Also define
\[
F:=\psi'^2.
\]
Since \(u\in T_{\theta^*}\mathbb S^{d-1}\), we have \(a\cdot u=0\).

At the teacher, the residual vanishes pointwise. Therefore, along a
spherical geodesic in the \(\theta\)-variable with initial velocity \(u\),
\[
\operatorname{Hess}_{\mathbb S,\theta}L(y^*)[u,u]
=
2\E\left[F(S)A_u^2\right].
\]
Let \(\Gamma\sim N(0,I_d)\). By Corollary~\ref{cor:stein-linear-quadratic},
\[
\left|
\E\left[F(a\cdot X)(u\cdot X)^2\right]
-
\E\left[F(a\cdot \Gamma)(u\cdot \Gamma)^2\right]
\right|
\le
C_F\|a\|_\infty
=
C_F\|\theta^*\|_\infty .
\]
Since \(a\cdot u=0\) and \(\|a\|=\|u\|=1\), the Gaussian random variables
\(a\cdot\Gamma\) and \(u\cdot\Gamma\) are independent standard normals. Hence
\[
\E\left[F(a\cdot \Gamma)(u\cdot \Gamma)^2\right]
=
\E[F(G)]\E[G^2]
=
A_\psi.
\]
Thus
\[
\left|
\operatorname{Hess}_{\mathbb S,\theta}L(y^*)[u,u]
-
2A_\psi
\right|
\le
C_\theta\|\theta^*\|_\infty
\]
for every unit \(u\in T_{\theta^*}\mathbb S^{d-1}\). Homogeneity gives the
claim for arbitrary \(u\).

Finally, the Gaussian teacher Hessian satisfies
\[
\operatorname{Hess}_{\mathbb S,\theta}L_G(y^*)[u,u]
=
2A_\psi\|u\|^2,
\]
because in the Gaussian model the same computation gives
\(2\E[F(a\cdot\Gamma)(u\cdot\Gamma)^2]\), and \(a\cdot\Gamma\) is independent
of \(u\cdot\Gamma\). This proves the comparison with \(L_G\).
\end{proof}

\paragraph{Stein Bound for \texorpdfstring{$\beta$}{beta}-Hessian}

To give a Stein bound for $\beta$-Hessian, we start by deriving this corollary from Theorem~\ref{thm:stein-product}.

\begin{corollary}[One-dimensional linear statistic]\label{cor:stein-1d-linear}
Assume the setting of Theorem~\ref{thm:stein-product}. Let $a\in\R^d$ and
$f\in C^3(\R)$ with $\|f^{(3)}\|_\infty<\infty$.
Define $h(x):=f(\langle a,x\rangle)$. Then
\[
\big|\E[f(\langle a,X\rangle)]-\E[f(\langle a,G\rangle)]\big|
\ \le\ \frac{5}{6}\|f^{(3)}\|_\infty\sum_{i=1}^d \tau_i^3 |a_i|^3.
\]
In particular, if $\sup_i\tau_i\le \tau$ and $\|a\|_2=1$, then
\begin{align}\label{eq:beta-stein-in-particular}
    \big|\E[f(\langle a,X\rangle)]-\E[f(\langle a,G\rangle)]\big|
\ \le\ C(\tau,f)\,\|a\|_\infty,
\end{align}
since $\sum_i |a_i|^3\le \|a\|_\infty\sum_i a_i^2=\|a\|_\infty$.
\end{corollary}
\begin{proof}
    Define $h(x) := f(\langle a, x\rangle)$. By the chain rule, all partial derivatives act only through $\langle a, x\rangle$:
$$\frac{\partial^3 h}{\partial x_i^3}(x) = a_i^3 f'''(\langle a, x\rangle).$$
Therefore $$\left\|\frac{\partial^3 h}{\partial x_i^3}\right\|_\infty = |a_i|^3\|f^{(3)}\|_\infty.$$
Applying Theorem~\ref{thm:stein-product} directly gives $$\big|\mathbb{E}[f(\langle a, X\rangle)] - \mathbb{E}[f(\langle a, G\rangle)]\big| \leq \frac{5}{6}\sum_{i=1}^d \tau_i^3 \left\|\partial_{x_i}^3 h\right\|_\infty = \frac{5}{6}\|f^{(3)}\|_\infty \sum_{i=1}^d \tau_i^3 |a_i|^3.$$
If $\sup_i \tau_i \leq \tau$ and $\|a\|_2 = 1$, then
$$\sum_{i=1}^d \tau_i^3 |a_i|^3 \leq \tau^3 \sum_{i=1}^d |a_i|^3 \leq \tau^3 \|a\|_\infty \sum_{i=1}^d a_i^2 = \tau^3 \|a\|_\infty,$$
giving $C(\tau, f) = \frac{5}{6}\tau^3 |f^{(3)}|_\infty$, so we obtained \eqref{eq:beta-stein-in-particular}. 
\end{proof}

\begin{lemma}[Teacher-level \(\beta\)-Hessian comparison]
\label{lem:teacher-beta-hessian-comparison}
Let
\[
A_\psi:=\mathbb E[\psi'(G)^2],
\qquad
B_\phi:=\mathbb E[\phi'(G)^2]=\kappa'(1),
\qquad
G\sim N(0,1).
\]
Assume \ref{ass:regularity-and-moments}. Then there exists a constant \(C_\beta>0\), independent of \(d\), such that
for every \(v\in T_{\beta^*}\mathbb S^{d-1}\),
\[
\left|
H_{L,\beta\beta}(y^*)[v,v]
-
2A_\psi B_\phi\|v\|^2
\right|
\le
C_\beta\|\theta^*\|_\infty\|v\|^2.
\]
Equivalently,
\[
\left|
H_{L,\beta\beta}(y^*)[v,v]
-
H_{L_G,\beta\beta}(y^*)[v,v]
\right|
\le
C_\beta\|\theta^*\|_\infty\|v\|^2.
\]
\end{lemma}

\begin{proof}
By quadratic scaling, it is enough to consider \(\|v\|=1\). Set
\[
a_i:=\theta_i^*,
\qquad
X_i:=\phi(G_i),
\qquad
q_i:=\phi'(G_i)^2,
\qquad
S:=\sum_{i=1}^d a_iX_i,
\qquad
S_{-i}:=S-a_iX_i,
\]
and define
\[
F(s):=\psi'(s)^2.
\]
By Lemma~\ref{lem:teacher-hessian-formula},
\[
\frac12 H_{L,\beta\beta}(y^*)[v,v]
=
\sum_{i=1}^d a_i^2
\mathbb E[F(S)q_i].
\]

We first decouple the \(i\)-th coordinate. Since \(S_{-i}\) is independent of
\((X_i,q_i)\),
\[
\mathbb E[F(S_{-i})q_i]
=
\mathbb E[F(S_{-i})]\mathbb E[q_i]
=
B_\phi\,\mathbb E[F(S_{-i})].
\]
Moreover, by the boundedness of \(F'\) and the moment assumption,
\[
\left|
\mathbb E[F(S)q_i]-\mathbb E[F(S_{-i})q_i]
\right|
\le
\|F'\|_\infty |a_i|
\mathbb E[|X_i|q_i]
\le
C|a_i|.
\]
Multiplying by \(a_i^2\) and summing gives
\[
\sum_{i=1}^d a_i^2
\mathbb E[F(S)q_i]
=
B_\phi
\sum_{i=1}^d a_i^2\mathbb E[F(S_{-i})]
+
O\left(\sum_{i=1}^d |a_i|^3\right).
\]
Since \(\sum_i a_i^2=1\) and
\[
\left|
\mathbb E[F(S)]-\mathbb E[F(S_{-i})]
\right|
\le
\|F'\|_\infty |a_i|\mathbb E|X_i|,
\]
we also have
\[
\sum_{i=1}^d a_i^2\mathbb E[F(S_{-i})]
=
\mathbb E[F(S)]
+
O\left(\sum_{i=1}^d |a_i|^3\right).
\]
Therefore,
\[
\sum_{i=1}^d a_i^2
\mathbb E[F(S)q_i]
=
B_\phi\,\mathbb E[F(S)]
+
O\left(\sum_{i=1}^d |a_i|^3\right).
\]

Now apply the one-dimensional Stein comparison in Corollary~\ref{cor:stein-1d-linear} with
\[
X_i:=\phi(G_i),\qquad a_i:=\theta_i^*,\qquad f:=F=\psi'^2.
\]
By the assumptions on \(\phi\),
\[
\E[X_i]=0,\qquad \E[X_i^2]=1,\qquad
\tau_\phi^3:=\E|X_i|^3<\infty,
\]
and by the regularity assumptions on \(\psi\), \(F\in C^3(\mathbb R)\) with
\(\|F^{(3)}\|_\infty<\infty\). Moreover \(\|a\|_2=\|\theta^*\|_2=1\).
Let \(\Gamma=(\Gamma_1,\ldots,\Gamma_d)\sim N(0,I_d)\). Then
\(\langle a,\Gamma\rangle\sim N(0,1)\). Hence Corollary~\ref{cor:stein-1d-linear}
gives
\[
\left|
\E F\left(\sum_{i=1}^d a_iX_i\right)
-
\E F\left(\langle a,\Gamma\rangle\right)
\right|
\le
\frac56 \|F^{(3)}\|_\infty
\sum_{i=1}^d \tau_\phi^3 |a_i|^3.
\]
Since \(\langle a,\Gamma\rangle\sim G\sim N(0,1)\), this becomes
\[
\left|
\E[F(S)]-\E[F(G)]
\right|
\le
C_F\sum_{i=1}^d |a_i|^3,
\]
where \(C_F:=\frac56\tau_\phi^3\|F^{(3)}\|_\infty\). Finally,
\[
\sum_{i=1}^d |a_i|^3
\le
\|a\|_\infty\sum_{i=1}^d a_i^2
=
\|\theta^*\|_\infty.
\]
Therefore
\[
\left|
\E[F(S)]-\E[F(G)]
\right|
\le
C_F\|\theta^*\|_\infty.
\]
Combining this with the preceding leave-one-out estimate yields
\[
\sum_{i=1}^d a_i^2
\E[F(S)q_i]
=
B_\phi\,\E[F(G)]
+
O(\|\theta^*\|_\infty)
=
B_\phi A_\psi
+
O(\|\theta^*\|_\infty).
\]
Multiplying by \(2\|v\|^2\) proves the true \(\beta\beta\)-block estimate.

For the Gaussian loss, since we have \eqref{eq:L_G} in Lemma~\ref{lem:gaussian-local-strong-convexity},
\[
L_G(\theta,\beta)
=
2\sum_{k\ge1}\alpha_k^2
\left(1-(\kappa(\rho)\chi)^k\right)
\],
gives, at \(y^*\),
\[
H_{L_G,\beta\beta}(y^*)[v,v]
=
2\kappa'(1)
\left(\sum_{k\ge1}k\alpha_k^2\right)
\|v\|^2.
\]
But
\[
\sum_{k\ge1}k\alpha_k^2
=
\mathbb E[\psi'(G)^2]
=
A_\psi,
\qquad
\kappa'(1)=B_\phi.
\]
Thus
\[
H_{L_G,\beta\beta}(y^*)[v,v]
=
2A_\psi B_\phi\|v\|^2,
\]
and the comparison follows.
\end{proof}

\subsection{Gradient Bounds}\label{subsec:gradient-bounds}

In the lemma below, we provide bounds for gradients $g^{(\theta)}, g^{(\beta)}$ to help us to control the evolution of $Z_t$ and $\eta_t$ in weak recovery, and that of $Q_t$ in strong recovery.

\begin{lemma}[Technical gradient bounds for blockwise spherical SGD]
\label{lem:sgd-tech-bounds}

Assume \ref{ass:regularity-and-moments}. Then there exists a constant
\(K<\infty\), depending only on \(L_\phi,L_\psi,\phi(0)\), such that if we draw a fresh sample $x_t$ then with
\[
g^{(\beta)}(\beta,\theta;x_t):=\nabla_\beta^{\mathcal S}\ell(\beta,\theta;x_t),
\qquad
g^{(\theta)}(\beta,\theta;x_t):=\nabla_\theta^{\mathcal S}\ell(\beta,\theta;x_t),
\]
we have the moment bounds
\begin{align*}
\sup_{\beta,\theta}\ \E\big[\|g^{(\beta)}(\beta,\theta;x_t)\|^2\big] &\le K d,\\
\sup_{\beta,\theta}\ \E\big[\|g^{(\theta)}(\beta,\theta;x_t)\|^2\big] &\le K d,\\
\sup_{\beta,\theta}\ \E\big[\|g^{(\beta)}(\beta,\theta;x_t)\|^4\big] &\le K d^2,\\
\sup_{\beta,\theta}\ \E\big[\|g^{(\theta)}(\beta,\theta;x_t)\|^4\big] &\le K d^2.
\end{align*}
Moreover, for the centered gradient-noise fields defined as
\[
\begin{gathered}
\Delta g^{(\beta)}(\beta,\theta;x_t)
:= g^{(\beta)}(\beta,\theta;x_t)-\E[g^{(\beta)}(\beta,\theta;x_t)],\\
\Delta g^{(\theta)}(\beta,\theta;x_t)
:= g^{(\theta)}(\beta,\theta;x_t)-\E[g^{(\theta)}(\beta,\theta;x_t)],
\end{gathered}
\]
we have the directional variance bounds
\begin{align*}
\sup_{\beta,\theta}\ \sup_{\|u\|=1}\ \E\big[(\Delta g^{(\beta)}(\beta,\theta;x_t)\cdot u)^2\big] &\le K,\\
\sup_{\beta,\theta}\ \sup_{\|u\|=1}\ \E\big[(\Delta g^{(\theta)}(\beta,\theta;x_t)\cdot u)^2\big] &\le K.
\end{align*}
Consequently, along the SGD trajectory $(\beta_t,\theta_t)$, since $x_t$ is independent of $\mathcal F_t$,
the same bounds hold conditionally on $\mathcal F_t$:
\[
\E\big[\|g_t^{(\beta)}\|^2\mid\mathcal F_t\big]\le K d,\quad
\E\big[\|g_t^{(\beta)}\|^4\mid\mathcal F_t\big]\le K d^2,\quad
\sup_{\|u\|=1}\E\big[((\Delta g_t^{(\beta)})\cdot u)^2\mid\mathcal F_t\big]\le K,
\]
and similarly for the $\theta$-block.
\end{lemma}
\begin{proof}

Since spherical gradient is the orthogonal projection of the Euclidean gradient onto the tangent space via $\nabla^{\mathcal{S}}F(x)=\nabla F(x)-(\nabla F(x) \cdot x) x,$  we have $\|\nabla^{\mathcal{S}}F(x)\|\leq \|\nabla F(x)\|$ and $|u\cdot\nabla^{\mathcal{S}}F(x)|\leq |u\cdot \nabla F(x)| \leq \|\nabla F(x)\|$ for $|u|=1$, it suffices to bound the Euclidean gradients.  

Based on the model definition in Section~\ref{sec:setup} and define the per-sample loss as 
\begin{align}
    l=(f-f^*)^2,
\end{align}
we have the per-sample Euclidean gradients w.r.t. each layer as:
\begin{align}
    \nabla_{\theta}l &=2(f-f^*)\cdot \psi'(s)\cdot z \\
    \nabla_{\beta}l &= 2(f-f^*)\cdot \psi'(s)\cdot  \sum_{i=1}^d\theta_i\phi'(u_i)x^{(i)}.
\end{align}
We simplify the expression by defining 
\begin{align}\label{eq:persamplegradient}
A(x;\beta,\theta):=2(f-f^*)\cdot \psi'(s)\in \mathbb{R}.
\end{align}

\paragraph{$\theta$ bounds}
\begin{align}
    \mathbb{E}[\|g^{(\theta)}\|^2] 
    &\leq \mathbb{E}[\|\nabla_{\theta}l\|^2] \\ 
    &= \mathbb{E}[|A|^2\|z\|^2] \\
    &\leq (\mathbb{E}|A|^4) ^{1/2}\cdot(\mathbb{E}\|z\|^4) ^{1/2} \text{ by Cauchy-Schwarz}\label{eq:secondmomentcs}
\end{align}
For the first term of \eqref{eq:secondmomentcs}, with the bounded $\psi,\psi'$ assumption, we have 
\begin{align} \label{eq:Abound}
    |A|=2|\psi(s)-\psi(s^*)||\psi'(s)|\leq 2(|\psi(s)|+|\psi(s^*)|)|\psi'(s)|\leq 4C_{\psi}C_{\psi'}.
\end{align}
For the second term, given $z$ has i.i.d. entries, we have
\begin{align}
    \mathbb{E}[\|z\|^4]
&=\mathbb{E}\big[\sum_{i=1}^dz_i^2\big]^2 \\
&=\mathbb{E}\big[\big(\sum_{i=1}^dz_i^2\big)\big(\sum_{i=1}^dz_i^2\big)]\\
&=\mathbb{E}\big[\sum_{i=1}^dz_i^4+\sum_{i\neq j}z_i^2z_j^2 \big]\\
&=\sum_{i=1}^d\mathbb{E}[z_i^4]+\sum_{i\neq j}\mathbb{E}[z_i^2]\mathbb{E}[z_j^2]\\
&= d \mu_4 + d(d-1)\mu_2^2 \\
&= \Theta(d^2).
\end{align}
Therefore, there exists a unifying constant $K$ such that
\begin{align}
    \mathbb{E}[\|g^{(\theta)}\|^2] |\leq \big[4C_{\psi}C_{\psi'}\big]^{2} \big[\Theta(d^2)\big]^{1/2}= Kd.
\end{align}
Similarly, to bound the forth moment, using the Minkowski inequality in $L^4$ and the i.i.d. of $\{z_i\},$ we have $$\|\sum_{i=1}^d z_i^2\|_{L^4}=\big(\mathbb{E}\big[\sum_{i=1}^dz_i^2\big]^4\big)^{1/4}\leq \sum_{i=1}^d \big(\mathbb{E}[z_i^8] \big)^{1/4}=d(\mathbb{E}[z_i^8])^{1/4}$$
Raise both sides to the 4th power, we have 
$$\mathbb{E}[\|z\|^8]=\mathbb{E}\big(\sum_{i=1}^d z_i^2 \big)^4 \leq d^4 \mathbb{E}[z_1^8]=O(d^4).$$
Thus, with the same, finite $A,$ there exists a unifying constant $K$ such that
\begin{align}
    \mathbb{E}[\|g^{(\theta)}\|^4]\leq \mathbb{E}[|A|^4\|z\|^4]\leq \mathbb{E}[|A|^8]^{1/2}\mathbb{E}[\|z\|^8]^{1/2} \leq Kd^2
\end{align}
\medskip
\noindent
\emph{\(\theta\)-block directional variance.}
Fix \((\beta,\theta)\in\mathbb S^{d-1}\times\mathbb S^{d-1}\), and condition on
\(\mathcal F_t\), so that \((\beta,\theta)\) are deterministic. For a fresh
sample \(x_t\), write
\[
u_i:=x_t^{(i)}\cdot\beta,
\qquad
z_i:=\phi(u_i),
\qquad
z:=(z_1,\ldots,z_d),
\qquad
s:=z\cdot\theta.
\]
Also write
\[
A_t:=2(f_t-f_t^*)\psi'(s).
\]
By the boundedness of \(\psi\) and \(\psi'\),
\[
|A_t|\le C_A.
\]
The Euclidean \(\theta\)-gradient is \(A_t z\), hence the spherical
\(\theta\)-gradient is
\[
g_t^{(\theta)}
=
P_{\theta^\perp}(A_t z)
=
A_tP_{\theta^\perp}z,
\qquad
P_{\theta^\perp}:=I-\theta\theta^\top .
\]

Let \(v\in\mathbb R^d\) be any unit vector. Since \(P_{\theta^\perp}\) is
self-adjoint,
\[
\langle g_t^{(\theta)},v\rangle
=
A_t\langle z,P_{\theta^\perp}v\rangle .
\]
Set
\[
\widetilde v:=P_{\theta^\perp}v.
\]
Then \(\|\widetilde v\|_2\le1\). Since \(\mathbb E[\phi(G)]=0\) and
\(\mathbb E[\phi(G)^2]<\infty\), the coordinates \(z_i=\phi(u_i)\) are iid
centered variables with bounded second moment. Therefore
\[
\mathbb E\!\left[
\langle z,\widetilde v\rangle^2
\,\middle|\,\mathcal F_t
\right]
=
\mathbb E[\phi(G)^2]\|\widetilde v\|_2^2
\le C.
\]
Using \(|A_t|\le C_A\),
\[
\mathbb E\!\left[
\langle g_t^{(\theta)},v\rangle^2
\,\middle|\,\mathcal F_t
\right]
\le C.
\]
Hence
\[
\mathbb E\!\left[
\bigl(
\langle g_t^{(\theta)},v\rangle
-
\mathbb E[\langle g_t^{(\theta)},v\rangle\mid\mathcal F_t]
\bigr)^2
\,\middle|\,\mathcal F_t
\right]
\le
\mathbb E\!\left[
\langle g_t^{(\theta)},v\rangle^2
\,\middle|\,\mathcal F_t
\right]
\le C.
\]
Equivalently,
\[
\sup_{\|v\|_2=1}
\mathbb E\!\left[
\bigl((\Delta g_t^{(\theta)})\cdot v\bigr)^2
\,\middle|\,\mathcal F_t
\right]
\le C.
\]

\paragraph{$\beta$ bounds}
Recall
\[
\nabla_\beta l
=2(f-f^*)\,\psi'(s)\sum_{i=1}^d \theta_i \phi'(u_i)\,x^{(i)},
\qquad
u_i:=x^{(i)}\cdot \beta,
\qquad
s:=z\cdot\theta,\ z_i=\phi(u_i),
\]
and define the scalar
\[
A(x;\beta,\theta):=2(f-f^*)\psi'(s).
\]
By boundedness of $\psi,\psi'$, we have the uniform bound
\[
|A|\le 2\big(|\psi(s)|+|\psi(s^*)|\big)\,|\psi'(s)|
\le 4C_\psi C_{\psi'}=:C_A.
\]

Let $P_\beta:=I-\beta\beta^\top$ be the orthogonal projector onto $\beta^\perp$.
The $\beta$-spherical gradient is
\[
g^{(\beta)}(\beta,\theta;x)
=\nabla_\beta^{\mathcal S}l(\beta,\theta;x)
=P_\beta \nabla_\beta l
=A\sum_{i=1}^d \theta_i \phi'(u_i)\,P_\beta x^{(i)}.
\]
Now exploit the Gaussian block structure: each block $x^{(i)}\sim\mathcal N(0,I_d)$
independently. Decompose
\[
x^{(i)} = u_i \beta + v_i,
\qquad
u_i:=x^{(i)}\cdot\beta\sim\mathcal N(0,1),
\qquad
v_i:=P_\beta x^{(i)}\sim\mathcal N(0,P_\beta),
\]
and for each $i$, $u_i$ is independent of $v_i$ (and the pairs $(u_i,v_i)$ are i.i.d. over $i$).
Hence
\[
g^{(\beta)} = A\sum_{i=1}^d \theta_i \phi'(u_i)\,v_i.
\]
Conditioned on $(u_1,\dots,u_d)$, the vector
\[
S:=\sum_{i=1}^d \theta_i \phi'(u_i)\,v_i
\]
is Gaussian with mean $0$ and covariance
\[
\mathrm{Cov}(S\mid u)=\Big(\sum_{i=1}^d \theta_i^2\phi'(u_i)^2\Big)P_\beta.
\]
Write
\[
\sigma^2(u):=\sum_{i=1}^d \theta_i^2\phi'(u_i)^2.
\]
Since $\mathrm{rank}(P_\beta)=d-1$, standard Gaussian moment formulas give
\[
\mathbb E\big[\|S\|^2\mid u\big]=(d-1)\sigma^2(u),
\qquad
\mathbb E\big[\|S\|^4\mid u\big]=(d-1)(d+1)\sigma^4(u).
\]

At this point we use the regularity of $\phi$.
For instance, if $\phi$ is $L_\phi$-Lipschitz (as in our regularity assumptions), then
$|\phi'|\le L_\phi$ a.e., so
\[
\sigma^2(u)\le L_\phi^2\sum_{i=1}^d \theta_i^2 = L_\phi^2,
\qquad
\sigma^4(u)\le L_\phi^4.
\]
Therefore,
\[
\mathbb E\|g^{(\beta)}\|^2
=\mathbb E\big[|A|^2\|S\|^2\big]
\le C_A^2\,\mathbb E\|S\|^2
\le C_A^2 (d-1) L_\phi^2
\le K d,
\]
and similarly
\[
\mathbb E\|g^{(\beta)}\|^4
=\mathbb E\big[|A|^4\|S\|^4\big]
\le C_A^4\,\mathbb E\|S\|^4
\le C_A^4 (d-1)(d+1) L_\phi^4
\le K d^2.
\]

Finally, we bound the directional variance of the centered noise field.
Fix any unit vector $q\in\mathbb R^d$.
Conditioned on $u$, $(q\cdot S)$ is Gaussian with variance
\[
\mathrm{Var}(q\cdot S\mid u)= q^\top \mathrm{Cov}(S\mid u)\,q
= \sigma^2(u)\,q^\top P_\beta q
\le \sigma^2(u).
\]
Thus,
\[
\mathbb E[(q\cdot g^{(\beta)})^2]
=\mathbb E\big[|A|^2 (q\cdot S)^2\big]
\le C_A^2\,\mathbb E[\sigma^2(u)]
\le C_A^2 L_\phi^2.
\]
Since $\mathrm{Var}(X)\le \mathbb E[X^2]$, we get
\[
\sup_{\|q\|=1}\ \mathbb E\big[((\Delta g^{(\beta)})\cdot q)^2\big]
\le \sup_{\|q\|=1}\ \mathbb E\big[(g^{(\beta)}\cdot q)^2\big]
\le C_A^2 L_\phi^2
\le K,
\]
which proves the claimed directional variance bound (absorbing constants into $K$).

\paragraph{Along the SGD trajectory} $(\beta_t,\theta_t)$ is $\mathcal F_t$-measurable and $x_t$ is independent of
$\mathcal F_t$, hence the above bounds apply conditionally on $\mathcal F_t$ by evaluation at
$(\beta,\theta)=(\beta_t,\theta_t)$.
\end{proof}

\subsection{Summation by Parts}\label{subsec:summation-by-parts}

The following estimates are deterministic and pathwise; no martingale property is used.
They are stated on subintervals because the SGD proof applies them after stopping times.

\begin{lemma}[Summation by parts for monotone weights]\label{lem:sbp-monotone}
Let \((N_t)_{0\le t\le T}\) be a real-valued sequence and set
\[
\varepsilon_t:=N_{t+1}-N_t .
\]
Fix integers \(0\le r<t\le T\), and let \((b_k)_{k=r}^{t-1}\) be monotone with
\(0\le b_k\le 1\). If
\[
\sup_{0\le u\le T}|N_u|\le M,
\]
then
\[
\left|\sum_{k=r}^{t-1}b_k\varepsilon_k\right|\le 2M .
\]
\end{lemma}

\begin{proof}
Abel summation gives
\[
\sum_{k=r}^{t-1}b_k(N_{k+1}-N_k)
=
b_{t-1}N_t-b_rN_r
-
\sum_{k=r}^{t-2}N_{k+1}(b_{k+1}-b_k).
\]
Hence
\[
\left|\sum_{k=r}^{t-1}b_k\varepsilon_k\right|
\le
M\left(
b_{t-1}+b_r+\sum_{k=r}^{t-2}|b_{k+1}-b_k|
\right).
\]
If \((b_k)\) is increasing, then
\[
\sum_{k=r}^{t-2}|b_{k+1}-b_k|=b_{t-1}-b_r,
\]
so the coefficient is \(2b_{t-1}\le 2\). If \((b_k)\) is decreasing, then
\[
\sum_{k=r}^{t-2}|b_{k+1}-b_k|=b_r-b_{t-1},
\]
so the coefficient is \(2b_r\le 2\). This proves the claim.
\end{proof}

\begin{corollary}[Summation by parts for decreasing geometric weights]\label{lem:sbp}
Let \((N_t)_{0\le t\le T}\) be a real-valued sequence and set
\[
\varepsilon_t:=N_{t+1}-N_t .
\]
Let \(\gamma>1\). If
\[
\sup_{0\le u\le T}|N_u|\le M,
\]
then for every \(0\le r<T\) and every \(s\in\{1,\dots,T-r\}\),
\[
\left|
\sum_{k=0}^{s-1}\gamma^{-(k+1)}\varepsilon_{r+k}
\right|
\le 2M .
\]
In particular, if \(M=2\lambda\), the bound is \(4\lambda\). If moreover
\(r=0\) and \(N_0=0\), then the bound improves to \(M\).
\end{corollary}

\begin{proof}
Apply Lemma~\ref{lem:sbp-monotone} with
\[
b_{r+k}:=\gamma^{-(k+1)},\qquad 0\le k\le s-1.
\]
These weights are decreasing and lie in \([0,1]\), so the first bound follows.

If \(r=0\) and \(N_0=0\), then the initial boundary term in Abel summation vanishes:
\[
\sum_{k=0}^{s-1}\gamma^{-(k+1)}(N_{k+1}-N_k)
=
\gamma^{-s}N_s
-
\sum_{k=0}^{s-2}N_{k+1}
\left(\gamma^{-(k+2)}-\gamma^{-(k+1)}\right).
\]
Therefore
\[
\left|
\sum_{k=0}^{s-1}\gamma^{-(k+1)}\varepsilon_k
\right|
\le
M\left(
\gamma^{-s}
+
\sum_{k=0}^{s-2}
\left(\gamma^{-(k+1)}-\gamma^{-(k+2)}\right)
\right)
=
M\gamma^{-1}
\le M .
\]
\end{proof}

\begin{corollary}[Summation by parts for increasing geometric weights]\label{lem:sbp-increasing}
Let \((N_t)_{0\le t\le T}\) be a real-valued sequence and set
\[
\varepsilon_t:=N_{t+1}-N_t .
\]
Let \(\alpha\in(0,1)\). If
\[
\sup_{0\le u\le T}|N_u|\le M,
\]
then for every \(0\le r<t\le T\),
\[
\left|
\sum_{k=r}^{t-1}\alpha^{t-1-k}\varepsilon_k
\right|
\le 2M .
\]
In particular, if \(M=\lambda\), the bound is \(2\lambda\), and if \(M=2\lambda\),
the bound is \(4\lambda\).
\end{corollary}

\begin{proof}
Apply Lemma~\ref{lem:sbp-monotone} with
\[
b_k:=\alpha^{t-1-k},\qquad r\le k\le t-1.
\]
Since \(\alpha\in(0,1)\), these weights are increasing in \(k\) and lie in
\([0,1]\). The claim follows.
\end{proof}

\newpage
\input{checklist.tex}

\end{document}

%% file: neurips/assumptions.tex
\section{Formalized Assumptions}
\label{sec:assumptions}

Let \(G\sim N(0,1)\), and write the Hermite expansions
\(\phi=\sum_{k\ge0}a_kh_k\) and \(\psi=\sum_{k\ge0}\alpha_kh_k\) in the normalized
Hermite basis.  We assume:
\begin{enumerate}[label=(A\arabic*)]
    \item \textbf{Normalization.}\label{ass:normalization}
    \(\mathbb E\phi(G)=0\) and \(\mathbb E\phi(G)^2=1\).

    \item \textbf{Non-degeneracy.} \label{ass:non-degeneracy}
    The first Hermite coefficients are nonzero:
    \[
        a_1=\mathbb E[G\phi(G)]=\mathbb E[\phi'(G)]\ne0,
        \qquad
        \alpha_1=\mathbb E[G\psi(G)]=\mathbb E[\psi'(G)]\ne0.
    \]
    \item \textbf{ Regularity of activation function.} \label{ass:regularity-and-moments} The inner activation \(\phi\in C^3(\mathbb R)\) and outer activation \(\psi\in C_b^4(\mathbb R)\) have finite regularity constants
\[
L_\phi:=\|\phi'\|_\infty+\|\phi''\|_\infty+\|\phi'''\|_\infty,
\qquad
L_\psi:=\max_{0\le j\le4}\|\psi^{(j)}\|_\infty.
\]
Here \(C_b^4(\mathbb R)\) requires \(\psi\) and its derivatives through order four to be bounded; \(\phi\) itself need not be bounded.
    \item \textbf{Initialization.}\label{ass:favorable-initialization} The initial second layer student parameter and the teacher parameter are uniformly sampled from the unit sphere: $\theta_0,\theta^* \sim \text{Unif}(\mathbb{S}^{d-1})$ and set the initial overlap to be of the order $d^{-1/2}$
    The initialization distribution of the overlap satisfies 
    \[\mathcal E_{\rm ov}:=\left\{\rho_0,\chi_0\in[c_{\rm init}d^{-1/2},C_{\rm init}d^{-1/2}]\right\}.
    \]
\end{enumerate}

Below, we made two remarks, for \ref{ass:regularity-and-moments} and for \ref{ass:favorable-initialization} respectively:

\begin{remark}
\label{rem:activation-moment-consequences}
Assumption~\ref{ass:regularity-and-moments} implies all moment conditions
used below. Indeed, since \(\|\phi'\|_\infty<\infty\),
\[
    |\phi(x)|
    \le
    |\phi(0)|+\|\phi'\|_\infty |x|.
\]
Hence \(\phi(G)\) has finite Gaussian moments of every order. Moreover, for
\(1\le j\le3\), the derivatives \(\phi^{(j)}\) are bounded. Therefore, for
every finite \(p\ge1\),
\[
    \max_{0\le j\le3}
    \mathbb E\left[
        (1+|G|)^p |\phi^{(j)}(G)|^p
    \right]
    <\infty,
    \qquad G\sim N(0,1),
\]
where \(\phi^{(0)}=\phi\). In particular,
\[
    \mathbb E|\phi(G)|^8<\infty,
    \qquad
    \mathbb E|\phi'(G)|^8<\infty,
\]
and
\[
    \mathbb E\!\left[(1+|\phi(G)|^2)|\phi'(G)|^2\right]<\infty,
    \qquad
    \mathbb E\!\left[|\phi(G)|\,|\phi'(G)|^2\right]<\infty.
\]
Finally, if \(Y\) is sub-Gaussian, then \(\phi(Y)\) is sub-Gaussian, with
\[
    \|\phi(Y)\|_{\psi_2}
    \le
    C\bigl(|\phi(0)|+L_\phi\|Y\|_{\psi_2}\bigr),
\]
for a universal constant \(C\). This follows from the same Lipschitz bound.
\end{remark}

\begin{remark}\label{eeee}
By standard concentration of the uniform measure (e.g. \cite{bietti2023learning}), the event $\mathcal E_{\rm ov}$ in \ref{ass:favorable-initialization} will happen with probability at least \(p_0>0\) independent of \(d\) for the uniform initialization. 

Fix \(K>0\). We work on the event
\[
\mathcal E_0(K)
:=
\mathcal E_{\rm ov}
\cap
\mathcal E_{\rm deloc}(K),
\]
where
\[
\mathcal E_{\rm deloc}(K)
:=
\left\{
\|\theta_0\|_\infty+\|\theta^*\|_\infty
\le
C_K\sqrt{\frac{\log d}{d}},
\quad
\|\theta_0\|_3^3+\|\theta^*\|_3^3
\le
C_Kd^{-1/2}
\right\}.
\]
By Lemma~\ref{lem:random-spherical-delocalization}, given $\theta_0,\theta^* \sim \text{Unif}(\mathbb{S}^{d-1})$, \(\mathcal E_{\rm deloc}(K)\) holds with probability at least \(1-d^{-K}\).
\end{remark}

%% file: checklist.tex
\section*{NeurIPS Paper Checklist}

\begin{enumerate}

\item {\bf Claims}
    \item[] Question: Do the main claims made in the abstract and introduction accurately reflect the paper's contributions and scope?
    \item[] Answer: \answerYes{} 
    \item[] Justification: We provide proofs of the main claims and clearly distinguish between claims and open questions. 

\item {\bf Limitations}
    \item[] Question: Does the paper discuss the limitations of the work performed by the authors?
    \item[] Answer: \answerYes{} 
    \item[] Justification: In the final conclusion section

\item {\bf Theory assumptions and proofs}
    \item[] Question: For each theoretical result, does the paper provide the full set of assumptions and a complete (and correct) proof?
    \item[] Answer: \answerYes{} 
    \item[] Justification: N/A

    \item {\bf Experimental result reproducibility}
    \item[] Question: Does the paper fully disclose all the information needed to reproduce the main experimental results of the paper to the extent that it affects the main claims and/or conclusions of the paper (regardless of whether the code and data are provided or not)?
    \item[] Answer: \answerYes{} 
    \item[] Justification: We include an appendix with experimental details

\item {\bf Open access to data and code}
    \item[] Question: Does the paper provide open access to the data and code, with sufficient instructions to faithfully reproduce the main experimental results, as described in supplemental material?
    \item[] Answer: \answerYes{} 
    \item[] Justification: We provide the code used to run the experiments. 

\item {\bf Experimental setting/details}
    \item[] Question: Does the paper specify all the training and test details (e.g., data splits, hyperparameters, how they were chosen, type of optimizer) necessary to understand the results?
    \item[] Answer: \answerYes{} 
    \item[] Justification: 

\item {\bf Experiment statistical significance}
    \item[] Question: Does the paper report error bars suitably and correctly defined or other appropriate information about the statistical significance of the experiments?
    \item[] Answer: \answerYes{} 
    \item[] Justification: 

\item {\bf Experiments compute resources}
    \item[] Question: For each experiment, does the paper provide sufficient information on the computer resources (type of compute workers, memory, time of execution) needed to reproduce the experiments?
    \item[] Answer: \answerYes{} 
    \item[] Justification:
    
\item {\bf Code of ethics}
    \item[] Question: Does the research conducted in the paper conform, in every respect, with the NeurIPS Code of Ethics \url{https://neurips.cc/public/EthicsGuidelines}?
    \item[] Answer: \answerYes{} 
    \item[] Justification: 

\item {\bf Broader impacts}
    \item[] Question: Does the paper discuss both potential positive societal impacts and negative societal impacts of the work performed?
    \item[] Answer: \answerNA{} 
    \item[] Justification:
    
\item {\bf Safeguards}
    \item[] Question: Does the paper describe safeguards that have been put in place for responsible release of data or models that have a high risk for misuse (e.g., pre-trained language models, image generators, or scraped datasets)?
    \item[] Answer: \answerNA{} 
    \item[] Justification: 

\item {\bf Licenses for existing assets}
    \item[] Question: Are the creators or original owners of assets (e.g., code, data, models), used in the paper, properly credited and are the license and terms of use explicitly mentioned and properly respected?
    \item[] Answer: \answerNA{} 
    \item[] Justification: 

\item {\bf New assets}
    \item[] Question: Are new assets introduced in the paper well documented and is the documentation provided alongside the assets?
    \item[] Answer: \answerNA{} 
    \item[] Justification: 

\item {\bf Crowdsourcing and research with human subjects}
    \item[] Question: For crowdsourcing experiments and research with human subjects, does the paper include the full text of instructions given to participants and screenshots, if applicable, as well as details about compensation (if any)? 
    \item[] Answer: \answerNA{} 
    \item[] Justification: 

\item {\bf Institutional review board (IRB) approvals or equivalent for research with human subjects}
    \item[] Question: Does the paper describe potential risks incurred by study participants, whether such risks were disclosed to the subjects, and whether Institutional Review Board (IRB) approvals (or an equivalent approval/review based on the requirements of your country or institution) were obtained?
    \item[] Answer: \answerNA{} 
    \item[] Justification: 

\item {\bf Declaration of LLM usage}
    \item[] Question: Does the paper describe the usage of LLMs if it is an important, original, or non-standard component of the core methods in this research? Note that if the LLM is used only for writing, editing, or formatting purposes and does \emph{not} impact the core methodology, scientific rigor, or originality of the research, declaration is not required.
    \item[] Answer: \answerNA{} 
    \item[] Justification: 

\end{enumerate}